\documentclass[lettersize,journal]{IEEEtran}
\usepackage{amsmath,amsfonts}
\usepackage{algorithmicx}
\usepackage{algorithm}
\usepackage{array}
\usepackage[caption=false,font=normalsize,labelfont=sf,textfont=sf]{subfig}
\usepackage{textcomp}
\usepackage{stfloats}
\usepackage{url}
\usepackage{verbatim}
\usepackage{graphicx}
\usepackage{cite}
\usepackage{bm}
\usepackage{xcolor}
\usepackage{booktabs}

\usepackage{amssymb}
\usepackage{tabularx}
\usepackage{multirow}
\usepackage{balance}
\usepackage{algpseudocode}

\begin{document}

\title{Micro-Structures Graph-Based Point Cloud Registration for Balancing Efficiency and Accuracy}

\author{Rongling Zhang, Li Yan, Pengcheng Wei, Hong Xie, Pinzhuo Wang, Binbing Wang
\thanks{Rongling Zhang, Li Yan, Pengcheng Wei, Hong Xie, Pinzhuo Wang and Binbing Wang are with School of Geodesy and Geomatics,  Hubei Luojia Laboratory, Wuhan University, Wuhan, 430079, China. E-mail: {whuzrl}@whu.edu.cn, {lyan}@sgg.whu.edu.cn, {wei.pc}@whu.edu.cn, {hxie}@sgg.whu.edu.cn, {wangpz}@whu.edu.cn, {whu\_wangbb}@whu.edu.cn.}
\thanks{Corresponding Authors: Hong Xie}
\thanks{Rongling Zhang and  Li Yan are co-first authors of the article}
}
\markboth{Journal of \LaTeX\ Class Files,~Vol.~14, No.~8, August~2024}%
{Shell \MakeLowercase{\textit{et al.}}: A Sample Article Using IEEEtran.cls for IEEE Journals}


\maketitle
\begin{abstract}

Point Cloud Registration (PCR) is a fundamental and significant issue in photogrammetry and remote sensing, aiming to seek the optimal rigid transformation between sets of points. Achieving efficient and precise PCR poses a considerable challenge. We propose a novel micro-structures graph-based global point cloud registration method. The overall method is comprised of two stages. 1) Coarse registration (CR): We develop a graph incorporating micro-structures, employing an efficient graph-based hierarchical strategy to remove outliers for obtaining the maximal consensus set. We propose a robust GNC-Welsch estimator for optimization derived from a robust estimator to the outlier process in the Lie algebra space, achieving fast and robust alignment.
2) Fine registration (FR): To refine local alignment further, we use the octree approach to adaptive search plane features in the micro-structures. By minimizing the distance from the point-to-plane, we can obtain a more precise local alignment, and the process will also be addressed effectively by being treated as a planar adjustment algorithm combined with Anderson accelerated optimization (PA-AA). After extensive experiments on real data, our proposed method performs well on the 3DMatch and ETH datasets compared to the most advanced methods, achieving higher accuracy metrics and reducing the time cost by at least one-third.
\end{abstract}

\begin{IEEEkeywords}
Point cloud registration, correspondence graph, robust estimator, planar adjustment, Anderson acceleration
\end{IEEEkeywords}    
\section{Introduction}
\label{sec:intro}

\IEEEPARstart{P}{oint} cloud registration is a fundamental problem in remote sensing and geometric processing \cite{yu2019advanced}.
Since an individual laser scan typically cannot capture all the scene data, combining multiple scans from different perspectives is essential to acquire a complete point cloud. Point cloud registration aligns these scans to the same coordinate system and determines the optimal rigid transformation between the sets of points. It has a wide range of applications, including scene reconstruction \cite{wang2020robust},  3D object recognition \cite{guo20143d,tao2020pipeline} and forest inventory survey \cite{kelbe2016marker}. Existing point cloud registration techniques commonly employ a coarse-to-fine registration strategy. Coarse registration methods typically estimate transformation based on the correspondence sets and establish an initial pose for subsequent fine registration. After obtaining the sets, the focus optimization avenue of coarse registration methods is acquiring the maximal consensus set, which involves various strategies \cite{ge2017automatic,cai2019practical} to enhance the correspondence set's consistency. For instance, the Gore \cite{bustos2017guaranteed} method effectively rejects outliers by seeking lower and upper bounds on consensus size, and the Max \cite{zhang20233d} method searches for maximal cliques in the graph. Notably, these strategies come with higher time consumption. As the scale of the correspondence set increases, the time cost significantly rises.
Moreover, it is practically challenging to completely eliminate outliers, which poses difficulties for accurate transformation estimation. Alternatively, the corresponding features can be learned using deep learning-based methods \cite{bai2021pointdsc,ao2023buffer,liu2024deep} to obtain the maximum consensus point set. While many learning methods \cite{guo2020deep} have achieved state-of-the-art performance on dataset metrics, they require substantial data for training, and their generalization across different datasets is not always promising. Therefore, this paper primarily focuses on non-learning-based methods.

Iterative Closest Point (ICP) is a popular method for fine registration. It relies on providing a good initial transformation from coarse registration but tends to converge more slowly \cite{rabbani2007integrated}. Numerous improvements to ICP have since emerged. Leveraging the rich geometric properties of point clouds has become a crucial area of improvement. For instance, point-to-plane ICP \cite{censi2008icp} was introduced to expedite convergence. To tackle the issue of iterations getting stuck in a local minimum, NICP \cite{serafin2015nicp} added local feature constraints during the query phase, and the work \cite{zhou2016fast,barron2019general,li2020robust} utilized robust geometric metrics.
Nevertheless, advanced geometric features usually pay a high computational cost, especially in the context of large-scale applications, making these methods challenging to apply in real-time or resource-constrained environments. Planar (bundle) adjustment \cite{zhou2020efficient,zhou2021lidar} provides an effective solution for this challenge. This approach jointly optimizes plane parameters and poses rather than estimating plane parameters from local sets prior to optimization, thereby enhancing the efficiency and accuracy of the registration process. For example, the work in \cite{ferrer2019eigen,liu2021balm,zhou2023efficient,liu2023efficient} significantly lowered the optimization dimensions by deriving a closed-form solution to eliminate the plane parameters. However, they typically spend considerable time searching for corresponding features.

The above analysis indicates that existing coarse and fine registration methods struggle to balance the trade-offs between computational efficiency and registration accuracy, and a limited number of approaches deal with both processes simultaneously. However, precise and effective registration typically requires optimization of both stages. In this work, we introduce a micro-structures graph as the cornerstone of a novel registration and derive the GNC-Welsch estimator in the Lie algebra space and PA-AA optimization to achieve coarse-to-fine point cloud registration, distinguished by its high accuracy and efficiency. First, we employ an octree-based approach to construct a micro-structures graph. Following our previous work \cite{yan2022new}, we implement an efficient hierarchical strategy to remove outliers in the CR, which involves the reliability of graph nodes and edges. Then, we derive the GNC-Welsch estimator based on the outlier process to obtain initial rigid transformation, making the CR more robust. With a solid initial transformation in place, followed by the FR, we establish explicit correspondences of the micro-structures through the graph. Here, we focus solely on partially reliable micro-structures, a crucial strategy for improving the method's efficiency. By leveraging the octree method to detect plane features within the micro-structures, we can quickly obtain feature points associated with the same plane. PA-AA joint optimization is applied to refine the pose and plane parameters, and Anderson acceleration is introduced to enhance this process further. To summarize, our contributions are as follows:

\begin{itemize}

\item{We propose a micro-structures graph-based global point cloud registration method that thoroughly exploits the information within the micro-structures graph. Our method balances efficiency and accuracy well.}

\item{We derive an enhanced GNC-Welsch estimator optimized from a robust estimation to the outlier process approach executed within the Lie algebra space. The estimator reduces sensitivity to outliers and initial values while converging towards heightened precision.}

\item{We propose PA-AA joint optimization to refine the micro-structures alignment, making our fine registration highly efficient and also achieving higher accuracy.}

\end{itemize}

\section{Related Work}
\label{sec:related_work}
\subsection{Coarse Registration}

The primary objective of coarse registration methods is to establish approximate alignment between two sets of point clouds, typically estimated based on the correspondence set. This process generally involves two key steps: acquiring the correspondence set and obtaining the maximal consensus set.

A mainstream strategy for acquiring correspondence sets is to identify key points \cite{zhong2009intrinsic,sipiran2011harris,liu2022rethinking} as substitutes for the original point cloud, as it can improve efficiency and reduce dependency on the scene. Then, the feature point matching will be performed based on the feature descriptor. Widely adopted feature descriptors include Fast Point Feature Histogram (FPFH) \cite{rusu2009fast}, 3D Scale Invariant Feature Transform (3D-SIFT) \cite{jiao2019point}, and learning-based 3D features like Fully Convolutional Geometric Features (FCGF) \cite{choy2019fully} and 3DFeat-Net \cite{yew20183dfeat}. However, these methods often lead to a high outlier rate in the correspondence set due to noise, uneven point cloud density, occlusion, and repetitive structures in the scene. 
\begin{figure}[t]
  \centering
   \includegraphics[width=\linewidth]{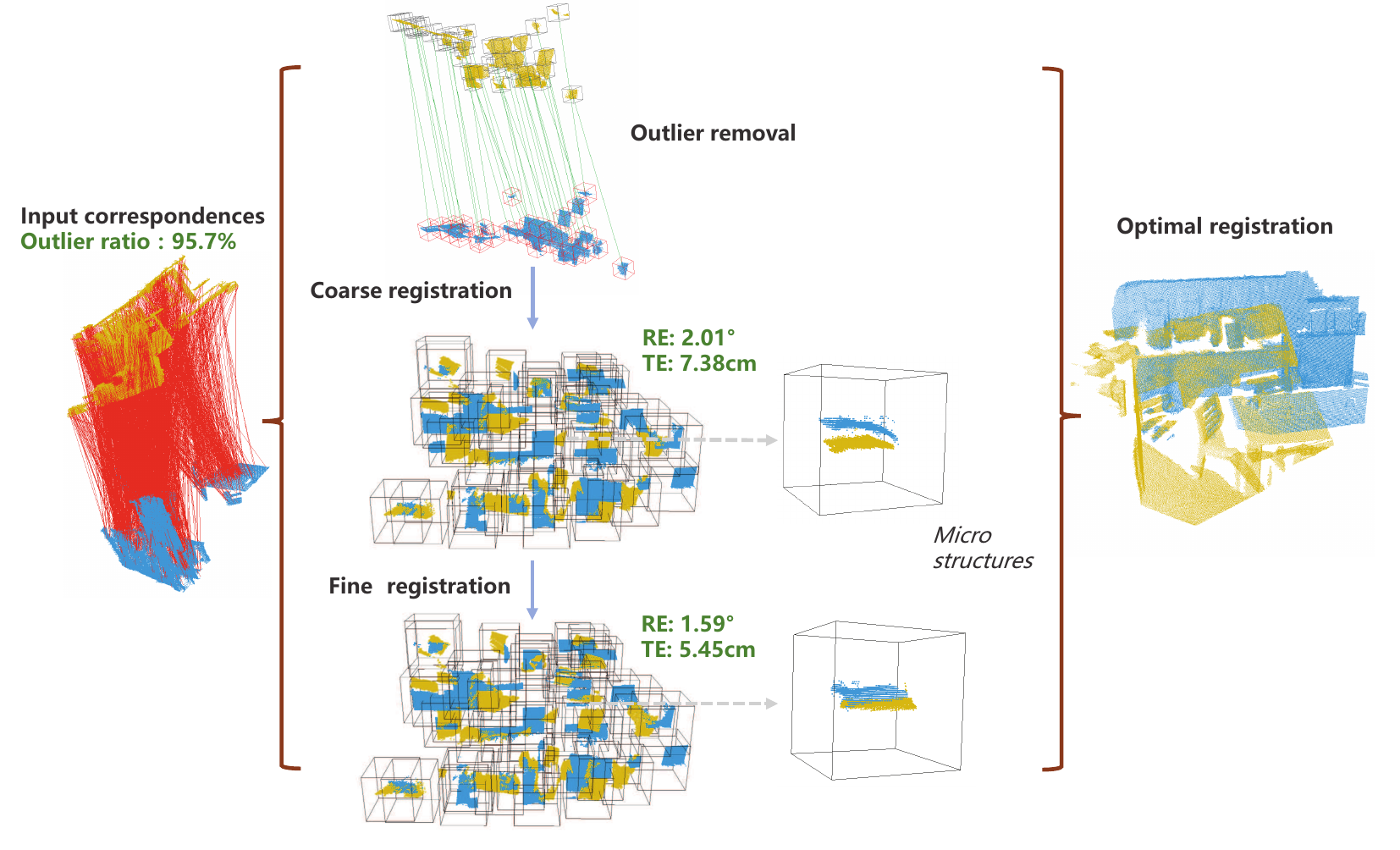}
   \caption{An example of the key output steps of our method: Starting with a correspondence set with a 95.7\% outlier rate, coarse registration achieves initial alignment, followed by fine registration of micro-structures, significantly improving accuracy.}
   \label{fig:teaser}
\end{figure}
Obtaining the maximal consensus set is essential, as direct estimation without outlier removal usually fails to yield satisfactory results. The random sampling consensus (RANSAC) \cite{fischler1981random} is a classic method applied for correspondence matching, essentially seeking the maximal consensus set. It iteratively samples the correspondence set until a satisfactory solution is found. Nonetheless, the reliance on extensive hypotheses and trials becomes its clear drawback. Currently, leveraging graph properties is a popular approach to obtaining the maximal consensus set, attributed to its superior efficiency in data representation. Numerous efficient algorithms based on graph theory have been proposed. For example, GC-RANSAC \cite{Barath_2018_CVPR} runs the graph-cut algorithm into RANSAC to solve the local optimization (LO), enhancing the efficiency of the LO step. In addition to point-to-point constraints, second-order constraints \cite{livi2013graph,yan2022new,han2016enhanced} like edge-to-edge similarity are commonly used during the sampling process. Furthermore, higher-order constraints, such as triangle or polygon similarity, are also employed in specific scenarios. For instance, Yang \textit{et al.}\cite{yang2021sac} proposed sorting and sampling compatible triangles formed by ternary cycles to generate the maximal consensus set, reinforcing consistency using these higher-order constraints. However, graph optimization problems are NP-hard, generally leading to lower efficiency when dealing with large-scale point sets. Methods like Max, proposed by Zhang \textit{et al.} \cite{zhang20233d}, construct large-scale second-order compatible graphs, paying considerable time cost. There are also many attempts \cite{yan2022new,duchenne2011tensor} to deal with complexity. Zhou \textit{et al.} \cite{zhou2015factorized} utilized factor graph matching to decompose large pairwise affinity matrices into smaller ones, although it is prone to the local minimum. TEASER++ \cite{yang2020teaser} uses the maximal clique algorithm to find consistent matching pairs, avoiding solving semi-definite programming problems. It performs with high efficiency on sparse compatibility graphs but requires substantial memory space for dense ones. Our proposed hierarchical removal strategy, based on the micro-structures graph, utilizes both first-order and second-order constraints and performs at a high efficiency level, even when facing large-scale sets.

\begin{figure*}[ht]
\centering
\includegraphics[width=\textwidth]{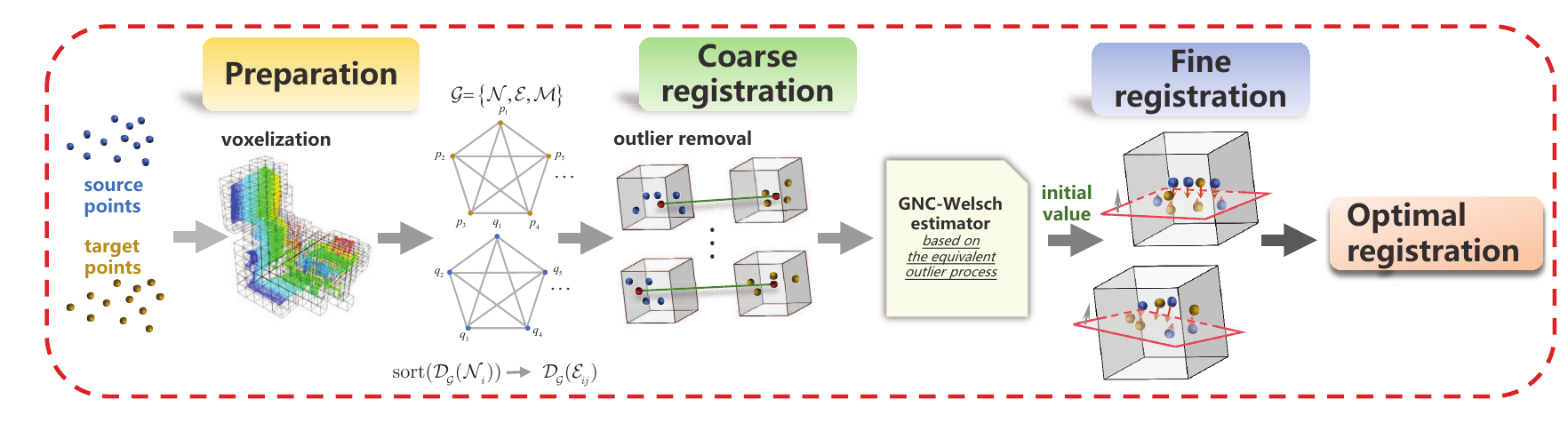}
\caption{{\bf Pipeline of our method.} 1. Voxelize the input point clouds and construct the graph $\mathsf{\mathcal{G}}$; 2. Remove outliers using a graph-based hierarchical strategy and estimate the initial value based on the GNC-Welsch estimator; 3. Align the micro-structures by PA-AA algorithm.}
\label{fig:framework}
\end{figure*}

\subsection{Fine Registration}

Fine registration methods rely on the initial alignment provided by coarse registration, typically involving more detailed local adjustments and optimal methods. Here, we focus on the classic ICP family and the planar (bundle) adjustment algorithm. The latter is mainly used for multi-view registration and provides an efficient and accurate optimization framework for fine registration.

One of the classic representatives of fine registration is the ICP \cite{besl1992method} algorithm and its variants. The ICP algorithm is characterized by its simplicity and efficiency. It alternates between the query for the nearest point in the target set and the minimization of the distance between corresponding points to obtain the optimal rigid registration matrix. It has given rise to many ICP variants \cite{rusinkiewicz2001efficient,rusinkiewicz2019symmetric,segal2009generalized,li2022robust} due to its sensitivity to noise and outliers, as well as its heavy reliance on the initial pose. Yang \textit{et al.} \cite{yang2013go} introduced Go-ICP, utilizing an octree data structure and branch-and-bound techniques to address local minimum issues. Nevertheless, its performance still relies on the quality of the initialization. Although the introduction of point-to-plane metrics \cite{ramalingam2013theory,censi2008icp} enhances robustness, it also increases the complexity, prompting the introduction of efficiency-improving methods. Sparse-ICP \cite{bouaziz2013sparse} represents the registration problem as a sparse ${{\ell }_{q}}$ optimization to reduce computational load but introduces significant performance degradation. To address this issue, Mavridis \textit{et al.}\cite{mavridis2015efficient} proposed a solution combining simulated annealing search with standard sparse ICP in a mixed optimization system. AA-ICP \cite{pavlov2018aa} treats the registration problem as a fixed-point iteration and incorporates Anderson acceleration into the iterative process. However, the results of Anderson acceleration do not consistently converge as expected.

Planar adjustment (PA) is considered a more formal term for the lidar bundle adjustment\cite{zhou2020efficient}, and they both work to jointly optimize sensor pose and plane parameters for minimizing reprojection errors. PA was designed to leverage geometric features for registration issues in large-scale or multi-view scenes. There typically needs to be enough overlap between views, and it is also categorized as fine registration. In recent years, many outstanding works have been proposed; for example, Eigen-factor (EF)\cite{ferrer2019eigen} minimizes the feature point-to-plane distance and uses the feature factor to implicitly represent the plane parameters, thus reducing the optimization scale. However, employing the gradient descent method leads to slow convergence; BALM\cite{liu2021balm} derives the Hessian matrix and uses LM (Levenberg-Marquardt)\cite{ranganathan2004levenberg} optimization to accelerate convergence, along with adaptive voxelization for searching corresponding feature relations. BALM still requires enumerating every feature point, resulting in high complexity. BALM improvement work \cite{liu2023efficient} encodes for feature points through point clusters, significantly reducing complexity. There is also another work\cite{zhou2023efficient} with similar ideas. It derives a closed-form solution for the Hessian matrix and the gradient vector, which makes the optimization independent of the number of feature points. These efforts enhance the advantages of PA optimization in large-scale scenarios. However, they all require searching for shared plane features before optimization, which can be time-consuming when aiming for accurate plane feature identification. Our fine registration inherits this optimization efficiency while allowing for the rapid acquisition of corresponding plane features.

\section{Approach}
\label{sec:algorithm}

\subsection{Problem Formulation}
Consider $\mathsf{\mathcal{X}}=\left\{ {{\mathbf{x}}_{i}}\mid {{\mathbf{x}}_{i}}\in {{\mathbb{R}}^{3}},i\in 1,2, \cdots, N \right\}$ and $\mathsf{\mathcal{Y}}=\left\{ {{\mathbf{y}}_{i}}\mid {{\mathbf{y}}_{i}}\in {{\mathbb{R}}^{3}},i\in 1,2, \cdots, M\right\}$ two point sets, which are to be aligned in 3D space.
Assuming that the $\mathsf{\mathcal{K}}$ sets of  the correspondence $\left\{{{\mathbf{x}}_{i}},{{\mathbf{y}}_{i}} \right\}_{1}^{\mathsf{\mathcal{K}}}$ are completely ideal, the following model will be generated ${{\mathbf{y}}_{i}}=\mathbf{R}{{\mathbf{x}}_{i}}+\mathbf{t},i\in 1,2, \cdots, \mathsf{\mathcal{K}}$.
\begin{equation}
\underset{{}}{\mathop{(\mathbf{\hat{R}},\mathbf{\hat{t}})=\underset{(\mathbf{R},\mathbf{t})}{\mathop{\arg \min }}\,}}\,(\sum\limits_{1}^{\mathcal{K}}{{{\left\| {{\mathbf{y}}_{i}}-(\mathbf{R}{{\mathbf{x}}_{i}}+\mathbf{t}) \right\|}^{\text{2}}}}),\mathbf{R}\in \mathcal{S}\mathcal{O}(3),\text{t}\in {{\mathbb{R}}^{3}}
\label{eq:1}
\end{equation}
$\left\|\cdot \right\|$ consistently represents the second-order norm in this paper. Apparently, this is a minimization problem on the maximal consensus set to obtain optimal transformation parameter $\mathbf{\hat{R}},\mathbf{\hat{t}}$. The correspondence sets are usually obtained through 3D keypoint detection and descriptor-based matching.
\begin{table}[tb]
\centering
\caption{Nomenclature}
\renewcommand{\arraystretch}{1.2}
\begin{tabularx}{\linewidth}{rX}
\toprule 
\multicolumn{1}{c}{\textbf{Notation}} & \multicolumn{1}{c}{\textbf{Explanation}} \\
\midrule
${{\mathsf{\mathcal{C}}}_{1}},{{\mathsf{\mathcal{C}}}_{2}},{{\mathsf{\mathcal{C}}}_{3}}$ & The correspondence sets with sizes ${\mathsf{\mathcal{K}}_{1}}, {\mathsf{\mathcal{K}}_{2}}, {\mathsf{\mathcal{K}}_{3}}$, respectively. \\
$\ell $ & The downsampling resolution. \\
$\varepsilon$ & The threshold of determining inlier: $\varepsilon=2\ell$. \\
$w_{n}$ & The weight of node: $w_{n}(x)=\exp(-\frac{{{x}^{2}}}{0.6*\ell}).$ \\
${{K}_{opt}}$ & The number of  reliable nodes, set by ${\mathsf{\mathcal{K}}_{1}}$ . \\
$\mu$ &Annealing rate of the shape parameter $\sigma$. \\
$(\cdot)^{*}$ & The homogeneous coordinate of a point. \\
${\mathbf{p}}^{*}_{f},{\mathbf{q}}^{*}_{f}$ & Plane feature points from point cloud $\mathsf{\mathcal{P}}$ and point cloud $\mathsf{\mathcal{Q}}$ respectively. \\
${\mathbf{p}}^{*}_{Mf}$ & All feature points of the same plane. \\
${\mathbf{p}}^{*}_{M}$ & The feature point coordinate in the frame of reference. \\
$h$ & The number of historical information in Anderson acceleration, $h=5$. \\
${{{\mathbf{\hat{T}}}}_{c}}$ & The transformation matrix estimated with CR. \\
${{{\mathbf{\hat{T}}}}_{AA}}$ & The transformation matrix estimated with Anderson acceleration. \\
${{{\mathbf{\hat{T}}}}_{f}}$ & The transformation matrix estimated with FR. \\
\bottomrule
\end{tabularx}
\end{table}

\subsection{Construct Micro-Structures Graph}
Downsampling the point cloud \cite{ervan2023histogram} is an effective step in data processing, especially when dealing with large-scale point clouds.  To enable rapid search and indexing, we downsample the original point clouds $\mathsf{\mathcal{X}}$ and $\mathsf{\mathcal{Y}}$ using octree-based sampling, with a resolution of $\ell$. The resulting representative point sets $\mathsf{\mathcal{P}}$ and $\mathsf{\mathcal{Q}}$ are then used to establish correspondence set ${{\mathsf{\mathcal{C}}}_{1}}=\left\{ {{\mathbf{p}}_{i}},{{\mathbf{q}}_{i}} \right\}_{1}^{\mathsf{\mathcal{K}}_{1}}$ by extracting key points \cite{zhong2009intrinsic}, calculating descriptors, and performing feature matching \cite{rusu2009fast}.

We build an undirected graph based on the correspondence set ${{\mathsf{\mathcal{C}}}_{1}}$. Here, $\mathsf{\mathcal{G}}=\left\{ \mathsf{\mathcal{N}},\mathsf{\mathcal{E}},\mathsf{\mathcal{M}} \right\}$
 represents a connected graph, where $\mathsf{\mathcal{N}}$ represents the graph nodes, $\mathsf{\mathcal{E}}\subseteq\mathsf{\mathcal{N}}\times \mathsf{\mathcal{N}}$
 denotes the edges connecting the nodes, and $\mathsf{\mathcal{M}}$ denotes the micro-structures of the nodes. The voxels of the corresponding nodes are regarded as micro-structures as shown in figure \ref{fig:teaser}. Micro-structure is defined as representing not only smaller scale but also tightly confined spatial region. In this way, we have constructed point cloud graphs $\mathsf{\mathcal{G}}{{\mathsf{}}^{\mathsf{}}}^{\mathsf{\mathcal{P}}}\text{=}\left\{ {{\mathsf{\mathcal{N}}}^{\mathsf{\mathcal{P}}}},{{\mathsf{\mathcal{E}}}^{\mathsf{\mathcal{P}}}},{{\mathsf{\mathcal{M}}}^{\mathsf{\mathcal{P}}}} \right\}$ and $\mathsf{\mathcal{G}}{{\mathsf{}}^{\mathsf{}}}^{\mathsf{\mathcal{Q}}}\text{=}\left\{ {{\mathsf{\mathcal{N}}}^{\mathsf{\mathcal{Q}}}},{{\mathsf{\mathcal{E}}}^{\mathsf{\mathcal{Q}}}},\mathsf{\mathcal{M}}{{\mathsf{}}^{\mathsf{\mathcal{Q}}}} \right\}$. Our method is developed based on micro-structures graphs.

\subsection{Coarse Registration (CR)}
\begin{figure}[tb]
\centering
\includegraphics[scale=0.60]{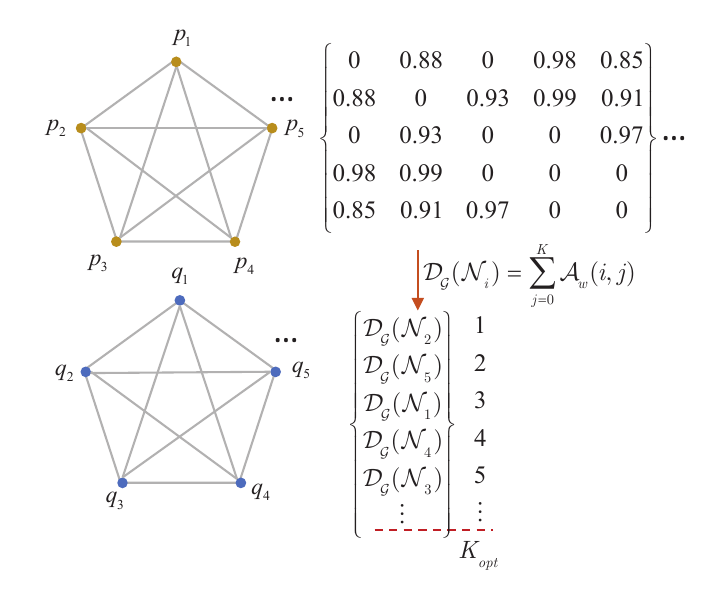}
\caption{An example for computing the node weights reliability}
\label{fig:outlier_removal}
\end{figure}
\subsubsection{Graph-Based Hierarchical Outlier Removal\label{sec:Graph Edges}}
Due to point cloud outliers, noises, and quality, the correspondence set obtained through descriptor-based matching often has a high outlier rate, sometimes including extreme cases, (e.g., 99\% of the correspondences are outliers \cite{yang2020teaser}). Advanced optimization algorithms may also fail to guarantee accurate transformation under a high outlier rate by following (\ref{eq:1}). To ensure the precision and efficiency of alignment, we perform a graph-based hierarchical outlier removal strategy for preprocessing. This hierarchical strategy allows us to refine the initial correspondence set ${{\mathsf{\mathcal{C}}}_{1}}$ into the maximal consensus set ${{\mathsf{\mathcal{C}}}_{2}}$.

\begin{itemize}

\item {\bf Reliability-Based Removal of Graph Nodes (RGN).\label{sec:Graph Nodes}} We propose the reliability of graph-based node weights, which enhances the expressive power of nodes compared to methods relying solely on voting \cite{yan2022new}. We use a weighted adjacency matrix $\mathsf{\mathcal{A}}\in {{\mathbb{Z}}^{{\mathsf{\mathcal{K}}_{1}}\times {\mathsf{\mathcal{K}}_{1}}}}$ from graph theory to record the relationships and define the reliability of nodes as ${{\mathsf{\mathcal{D}}}_{\mathsf{\mathcal{G}}}}({{\mathsf{\mathcal{N}}}_{i}})=\sum\limits_{j=0}^{K}{{{\mathsf{\mathcal{A}}}_{w}}(i,j)}$.
Moreover, the relative positions of points within a rigid point cloud remain invariant under rigid transformations. In graph $\mathsf{\mathcal{G}}{{\mathsf{}}^{\mathsf{}}}^{\mathsf{\mathcal{P}}}$, for nodes ${{\mathsf{\mathcal{N}}}_{i}}^{\mathsf{\mathcal{P}}}$ and ${{\mathsf{\mathcal{N}}}_{j}}^{\mathsf{\mathcal{P}}}$ connected by edge ${\vec{\mathcal{E}}}_{ij}^{\mathsf{\mathcal{P}}}$, if the corresponding correct nodes in graph $\mathsf{\mathcal{G}}{{\mathsf{}}^{\mathsf{}}}^{\mathsf{\mathcal{Q}}}$ are ${{\mathsf{\mathcal{N}}}_{i}}^{\mathsf{\mathcal{Q}}}$ and ${{\mathsf{\mathcal{N}}}_{j}}^{\mathsf{\mathcal{Q}}}$, then the Euclidean distances between two corresponding edges should ideally be equal without noise. We assume that the inlier noise is $\varepsilon$ and obtain the following constraints.
\begin{equation}
\Delta d_{ij} = \mid \lVert \vec{\mathcal{E}}_{ij}^{\mathcal{P}} \rVert - \lVert \vec{\mathcal{E}}_{ij}^{\mathcal{Q}}  \rVert \mid \in \left\{
    \begin{array}{ll}
        [0, \varepsilon] & \text{inlier} \\
        \text{else} & \text{outlier}
    \end{array}
\right.
\label{eq:2}
\end{equation}
The adjacency matrix $\mathsf{\mathcal{A}}$ is recorded as follows: for nodes $\left\{ \begin{matrix}
   {{\mathsf{\mathcal{N}}}_{i}}^{\mathsf{\mathcal{P}}}, {{\mathsf{\mathcal{N}}}_{i}}^{\mathsf{\mathcal{Q}}}  \\
\end{matrix} \right\}$ and $\left\{ \begin{matrix}
   {{\mathsf{\mathcal{N}}}_{j}}^{\mathsf{\mathcal{P}}}, {{\mathsf{\mathcal{N}}}_{j}}^{Q}  \\
\end{matrix} \right\}$
, if considered inliers, compute ${{\mathsf{\mathcal{A}}}_{w}}(i,j)=w_{n}(\Delta {{d}_{ij}})$ where $w_{n}$ is the exponential function; if considered outliers, the value is set to zero. This matrix is symmetric. By calculating the weight of each node, we obtain the reliability ${{\mathsf{\mathcal{D}}}_{\mathsf{\mathcal{G}}}}({{\mathsf{\mathcal{N}}}_{i}})$. The nodes can then be sorted by reliability from highest to lowest, selecting the top ${{K}_{opt}}$ nodes. While this criterion is necessary but not sufficient, we can eliminate most outliers through this strategy, thereby enhancing efficiency and allowing for further, more strict removal. The complexity of this strategy is $O(N)$ in figure \ref{fig:outlier_removal}.

\item {\bf Reliability-Based Removal of Graph Edges (RGE).}
Our previous work \cite{yan2022new} introduced a constraint function $\mathsf{\mathcal{F}}$ to evaluate the reliability of identically named edges. It starts with an initially zero-valued affinity matrix $\mathsf{\mathcal{B}}\in {{\mathbb{Z}}^{({{K}_{opt}}-2)\times 1}}$ in graph theory, setting elements to 1 if $\mathsf{\mathcal{F}}<0$, and defines the reliability of edges as ${{\mathsf{\mathcal{D}}}_{\mathsf{\mathcal{G}}}}({{\mathsf{\mathcal{E}}}_{ij}})=\sum\limits_{j=1}^{{{K}_{opt}}-2}{\mathsf{\mathcal{B}}(i,1)}$. We obtain the maximal consensus set by calculating ${{{\mathsf{\mathcal{D}}}_{\mathsf{\mathcal{G}}}}(\mathsf{\mathcal{E}}_{ij}^{\mathsf{\mathcal{P}\mathcal{Q}}})|}_{{\mathsf{\mathcal{F}}}_{1}}$ and ${{{\mathsf{\mathcal{D}}}_{\mathsf{\mathcal{G}}}}(\mathsf{\mathcal{E}}_{ij}^{\mathsf{\mathcal{P}\mathcal{Q}}})|}_{{\mathsf{\mathcal{F}}}_{2}}$ in turn, as done in the work\cite{yan2022new}.

{\emph{Loose function:}}
\begin{equation}
\label{eq:4}
\centering
{{\mathsf{\mathcal{F}}}_{1}}(\mathsf{\mathcal{E}}_{ij}^{\mathsf{\mathcal{P}\mathcal{Q}}}, \mathsf{\mathcal{N}}_{k}^{\mathsf{\mathcal{P}\mathcal{Q}}}) = \mid \lvert Prj_{{\vec{\mathcal{E}}}_{ij}^{\mathsf{\mathcal{P}}}} {\vec{\mathcal{E}}}_{ik}^{\mathsf{\mathcal{P}}} \rvert - \lvert Prj_{{\vec{\mathcal{E}}}_{ij}^{\mathsf{\mathcal{Q}}}} {\vec{\mathcal{E}}}_{ik}^{\mathsf{\mathcal{Q}}} \rvert \mid - \varepsilon
\end{equation}

{\emph {Tight function:}}
\begin{equation}
\centering
{{\mathsf{\mathcal{F}}}_{2}}(\mathsf{\mathcal{E}}_{ij}^{\mathsf{\mathcal{P}\mathcal{Q}}},\mathsf{\mathcal{N}}_{k}^{\mathsf{\mathcal{P}\mathcal{Q}}})=\left\| \mathsf{\mathcal{N}}_{k}^{\mathsf{\mathcal{P}}}-R(\theta ,{\vec{\mathcal{E}}}_{ij}^{\mathsf{\mathcal{P}\mathcal{Q}}})\mathsf{\mathcal{N}}_{k}^{\mathsf{\mathcal{Q}}} \right\|-\varepsilon 
\label{eq:5}
\end{equation}
$\mathsf{\mathcal{E}}_{ij}^{\mathsf{\mathcal{P}\mathcal{Q}}}$ denotes the pair edge $(\mathsf{\mathcal{E}}_{ij}^{\mathsf{\mathcal{P}}} , \mathsf{\mathcal{E}}_{ij}^{\mathsf{\mathcal{Q}}})$.  $Pr{{j}_{{{{{\vec{\mathcal{E}}}}}_{1}}}}{{{\vec{\mathcal{E}}}}_{2}}$ represents the projection from ${{{\vec{\mathcal{E}}}}_{2}}$ to ${{{\vec{\mathcal{E}}}}_{\text{1}}}$. $R(\theta ,{\vec{\mathcal{E}}}_{ij}^{\mathsf{\mathcal{P}\mathcal{Q}}})$ represents the rotation matrix obtained through $\theta $, with ${\vec{\mathcal{E}}}_{ij}^{{\mathcal{P}\mathcal{Q}}}$ as the rotation axis. The optimal $\theta $ is obtained by searching for the maximum consensus set of $\theta $. The complexity is denoted by $O(N\log N)$.

\end{itemize}

\noindent We ultimately refine the initial correspondence set ${{\mathsf{\mathcal{C}}}_{1}}$ into the maximal consensus set ${{\mathsf{\mathcal{C}}}_{2}}\text{= }\{{{\bf{p}}_{i}},{{\bf{q}}_{i}}\}_{1}^{{{\mathsf{\mathcal{K}}}_{2}}}$. Although the edge-based strategy is more strict, the scale of the correspondence set has been reduced by the node-based strategy, resulting in less time consumption overall. Thus, the hierarchical strategy is theoretically highly efficient. The contribution of node-based and edge-based strategies in removing outliers will also be demonstrated in \ref{sec:Analysis_Experiments}.

\subsubsection{GNC-Welsch Estimator Based on an Equivalent Outlier Process\label{sec:GNC-Welsch_Estimator}} 
Through the hierarchical strategy, most outliers have already been removed. We can proceed by (\ref{eq:1}) estimating the registration parameters. However, ${\mathsf{\mathcal{C}}}_{2}$ still contains some outliers. Compared to direct SVD estimation, a robust estimator yields a more accurate result, as it imposes greater penalties on larger residuals while remaining sensitive to smaller residuals. There have been many studies on robust estimators \cite{mactavish2015all}, including the Tukey, Huber, Geman-McClure, Cauchy, and Welsch functions, which all play similar roles in the optimization process but have different sensitivities to outliers.
The Tukey loss function rejects outliers beyond the threshold, providing strong robustness against outliers but challenging threshold adjustment. Its non-smooth nature near the threshold can cause optimization to get stuck in local minima. The Huber loss function, due to its piecewise nature, also encounters similar issues. Geman-McClure and Cauchy loss functions gradually stabilize with increasing residuals, offering some resistance to outliers but less effectively than the Welsch function. The Welsch function rapidly stabilizes for large residuals, significantly reducing sensitivity to outliers. Thus, we propose an enhanced GNC-Welsch estimator by combining the Welsch function with the Graduated Non-Convexity (GNC) framework. This approach transitions from convex to non-convex optimization, avoiding local minima and reducing sensitivity to initial values, ensuring quick convergence to a good solution. This leads to the following objective function, replacing (\ref{eq:1}).
\begin{equation}
\label{eq:6}
\underset{(\mathbf{R},\mathbf{t})}{\mathop{\arg \min }}\,(\sum\limits_{({{\mathbf{p}}_{i}},{{\mathbf{q}}_{i}})\in {{\mathcal{C}}_{2}}}^{{}}{\rho (\left\| {{\mathbf{p}}_{i}}-(\mathbf{R}{{\mathbf{q}}_{i}}+\mathbf{t}) \right\|}))
\end{equation}


\begin{figure}[t]
	\centering
	\includegraphics[scale=0.5]{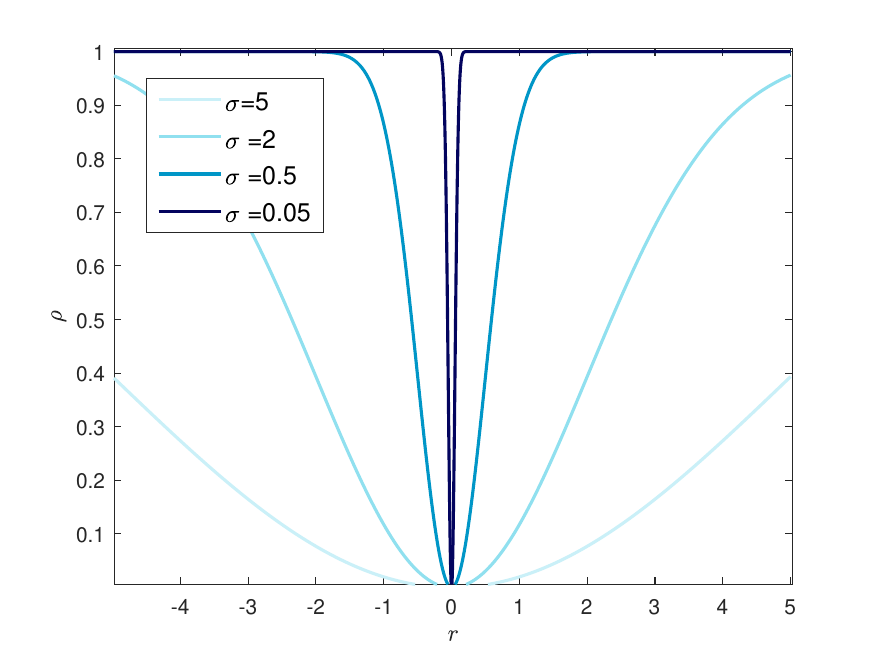}
	\caption{Welsch function under different $\sigma$ values}
	\label{fig:welsch-function}
\end{figure}

The function $\rho (r)=1-\exp (-\frac{{{r}^{2}}}{2{{\sigma }^{2}}})$ defines the GNC-Welsch metric, where $r=\left\| {{\mathbf{p}}_{i}}-(\mathbf{R}{{\mathbf{q}}_{i}}+\mathbf{t}) \right\|$ represents residuals. The parameter $\sigma (\sigma>0)$ controls the shape of the function, as illustrated in figure  \ref{fig:welsch-function}. Its initial value is determined based on the mean distance of the initial set ${{\mathsf{\mathcal{C}}}_{2}}$. The fundamental idea behind GNC\cite{blake1987visual} is to control $\sigma \leftarrow \frac{\sigma }{\mu}$ to anneal. In addition, we transform $\{\mathbf{R}, \mathbf{t}\}$ into homogeneous coordinates to represent the transformation as $\mathbf{T}$ for simpler notation. The same applies for the conversion ${{\mathbf{p}}_{i}}\to \mathbf{p}_{i}^{*}$, ${{\mathbf{q}}_{i}}\to \mathbf{q}_{i}^{*}$.

However, (\ref{eq:6}) is difficult to solve directly. To improve the computational efficiency of robust estimation and leverage existing optimization algorithms, following the Black-Rangarajan duality \cite{black1996unification,eriksson2018rotation}, we equivalently transform the robust cost function $\rho (r(\mathbf{T}))$ into a simpler Welsch outlier process, defining $\phi (\omega )=\rho (\sigma \sqrt{2\omega })$, the outlier process $z\equiv {\phi }'(\omega )=\exp(-\omega)$, $(0<z\le 1)$. $\omega$ is a scaling variable with respect to $r$ introduced to simplify the expression.
\begin{equation}
\label{eq:7}
E(\mathbf{T},z)=\frac{{{r(\mathbf{T})}^{2}}}{2{{\sigma }^{2}}}z+\Psi (z)
\end{equation} 

In order to achieve the equivalence in minimizing $\rho (r(\mathbf{T}))$ and $E(\mathbf{T},z)$, it is necessary to find an appropriate function $\Psi (z)$. Black and Rangarajan provided a unified procedure from robust function estimation to outlier processes.
\begin{equation}
\label{eq:8}
\Psi (z)=\phi ({{(\phi ')}^{-1}}(z))-z{{(\phi ')}^{-1}}(z)=z\log z-z+1
\end{equation}

By substituting (\ref{eq:8}) back into (\ref{eq:7}), we can employ alternate minimization to solve it. For instance, in the $t$-th iteration, by first fixing $z_{i}^{(t-1)}$, (\ref{eq:9}) becomes a minimization problem solely with respect to $\mathbf{T}$. Then, by fixing the estimated result ${{\mathbf{\hat{T}}}^{(t)}}$, (\ref{eq:10}) transforms into a straightforward minimization problem with a closed-form solution for $z$. Thus, solving the objective function (\ref{eq:6}) reduces to a simple alternating optimization between (\ref{eq:9}) and (\ref{eq:10}).

\begin{equation} 
\label{eq:9}
\begin{aligned}
\hat{\mathbf{T}}^{(t)} = \underset{\mathbf{T}}{\arg\min} \Bigg(
  \sum_{\substack{r_{i}= \left\|\mathbf{p}_{i}^{*}-\mathbf{T}^{(t-1)}\mathbf{q}_{i}^{*}\right\|  \\ \left\{ {{\mathbf{p}}_{i}^{*}},{{\mathbf{q}}_{i}^{*}} \right\} \in \mathcal{C}_{2}}}   \frac{r_{i}^{2}}{2\sigma^{2}}z_{i}^{(t-1)}  &+  z_{i}^{(t-1)}\log z_{i}^{(t-1)}  \\ 
&-z_{i}^{(t-1)}+1  \Bigg)
\end{aligned} 
\end{equation}

\begin{equation}
\label{eq:10}
\begin{aligned}
{{z}^{(t)}}=\underset{z}{\mathop{\arg \min }}\,\sum\limits_{{{z}_{i}}\in (0,1]}{(\frac{{{r}_{i}}^{2}}{2{{\sigma }^{2}}}{{z}_{i}}+{{z}_{i}}\log {{z}_{i}}-{{z}_{i}}+1)}
\end{aligned} 
\end{equation}

\noindent{\bf Optimization in Lie Algebra Spaces.} 
Lie algebra provides a continuous and stable rotation representation, avoiding the gimbal lock issue associated with Euler angles. Using the Gauss-Newton algorithm, we treat (\ref{eq:9}) as an optimization problem in Lie algebra space and update the state with the left perturbation approach. Convergence is reached when the changes are minimal, resulting in a reliable initial transformation matrix ${{{\mathbf{\hat{T}}}}_{c}}=\{{{{{\mathbf{\hat{R}}}}_{c}},{{{\mathbf{\hat{t}}}}_{c}}}\}$.

\begin{algorithm}[tb]
\caption{GNC-Welsch estimator}
\begin{algorithmic}[1]
\Require ${{\mathsf{\mathcal{C}}}_{2}}=\left\{ {{\mathbf{p}}_{i}^{*}},{{\mathbf{q}}_{i}^{*}} \right\}_{1}^{\mathsf{\mathcal{K}}_{2}}$, $\ell$;
\Ensure  ${{{\mathbf{\hat{T}}}}_{c}}$;
\State Set $\sigma_{min}=0.5*\ell$; $iteration\_nums=100$;
\State  $\sigma \gets 10*Mean\_Distance({{\mathsf{\mathcal{C}}}_{2}})$;
\State  $\mu \gets \sigma /20$;
\For{$it = 0$ to $iteration\_nums$}
    \If{$\sigma > \sigma_{min}$ AND $it\%2==0$}
        \State $\sigma \gets \sigma /\mu$;
    \EndIf
    \State Set $\mathbf{J}_{{{r}}}$ = 0, $\mathbf{r}$ = 0;
    \For{$each \left\{ {{\mathbf{p}}_{i}^{*}},{{\mathbf{q}}_{i}^{*}} \right\} \in {{\mathsf{\mathcal{C}}}_{2}}$}
    \State Compute $z_{i}$ by $(\ref{eq:10})$;
    \State $r_{i} \gets \left\|{{\mathbf{p}}_{i}^{*}}-{{\mathbf{q}}_{i}^{*}}\right\| $;
    \State Compute the matrix $\mathbf{J}_{r_{i}}$;
    \State Update  $\mathbf{J}_{{{r}}}$ and $\mathbf{r}$;
    \EndFor
    \State $\Delta{\bm \xi} \gets -(\mathbf{J}_{{{r}}}^{T}{{\mathbf{J}}_{{{r}}}})^{-1}\mathbf{J}^{T}_{{{r}}}\mathbf{r}$;
     \State Update  ${{\mathbf{T}}^{it}} \gets \exp (\Delta {\bm\xi}^{\wedge }){{\mathbf{T}}^{it-1}}$;
    \State $trans \gets \exp (\Delta {\bm\xi}^{\wedge })$;
    \If{$\left\|last\_trans -trans \right\|<1e-8$}
        \State $break$;
    \EndIf
    \State $last\_trans \gets trans$;
\EndFor

\end{algorithmic}
\end{algorithm}

\subsection{Fine Registration (FR)}
\subsubsection{Detecting Planar Features in Corresponding Micro-structures \label{sec:search-plane}}
After coarse registration, we further optimize the initially aligned point cloud to achieve better alignment on a finer scale. Our key idea is that neighborhood points of corresponding inliers should possess consistent geometric features. This stems from the inherent geometric continuity of object surfaces and the uniformity among neighboring points within a scene. After the point clouds are registered, the neighborhoods of corresponding inliers are expected to share the same features. Moreover, the correspondence set originates from key points, such as Intrinsic Shape Signatures (ISS) key points.   They are typically points with high curvature, such as corners, often found at geometric transitions or areas with significant surface variation.
The micro-structures represent a smaller scale than the nodes, allowing us to preserve the finer local structures around the nodes through the micro-structure graph.

Before beginning, we refine the initial correspondence set ${{\mathsf{\mathcal{C}}}_{1}}$ to obtain the inlier set ${{\mathsf{\mathcal{C}}}_{3}}=\left\{ {{\mathbf{p}}_{i}^{*}},{{\mathbf{q}}_{i}^{*}} \right\}_{1}^{{{\mathsf{\mathcal{K}}}_{3}}}$ using the initial transformation matrix ${{{\mathbf{\hat{T}}}}_{c}}$. Points for which $\lVert{{\mathbf{p}}_{i}^{*}}-{{\mathbf{\hat{T}}}_{c}}{{\mathbf{q}}_{i}^{*}}\rVert<2\varepsilon$ are identified  as inliers. We can also obtain micro-structures $\left\{ \mathsf{\mathcal{M}}_{i}^{\mathsf{\mathcal{P}}},\mathsf{\mathcal{M}}_{i}^{\mathsf{\mathcal{Q}}} \right\}_{1}^{{{\mathsf{\mathcal{K}}}_{3}}}$, where the correspondence is already included. Therefore, we can directly search for shared plane features. We use only a subset of micro-structures rather than all of them, which is also the key to improving efficiency. Typically, we also consider the micro-structures within $L$ distance of ${\mathcal{M}}$ to appropriately expand the ${\mathcal{M}}$.

\begin{figure}[bp]
\centering
\includegraphics[scale=0.20]{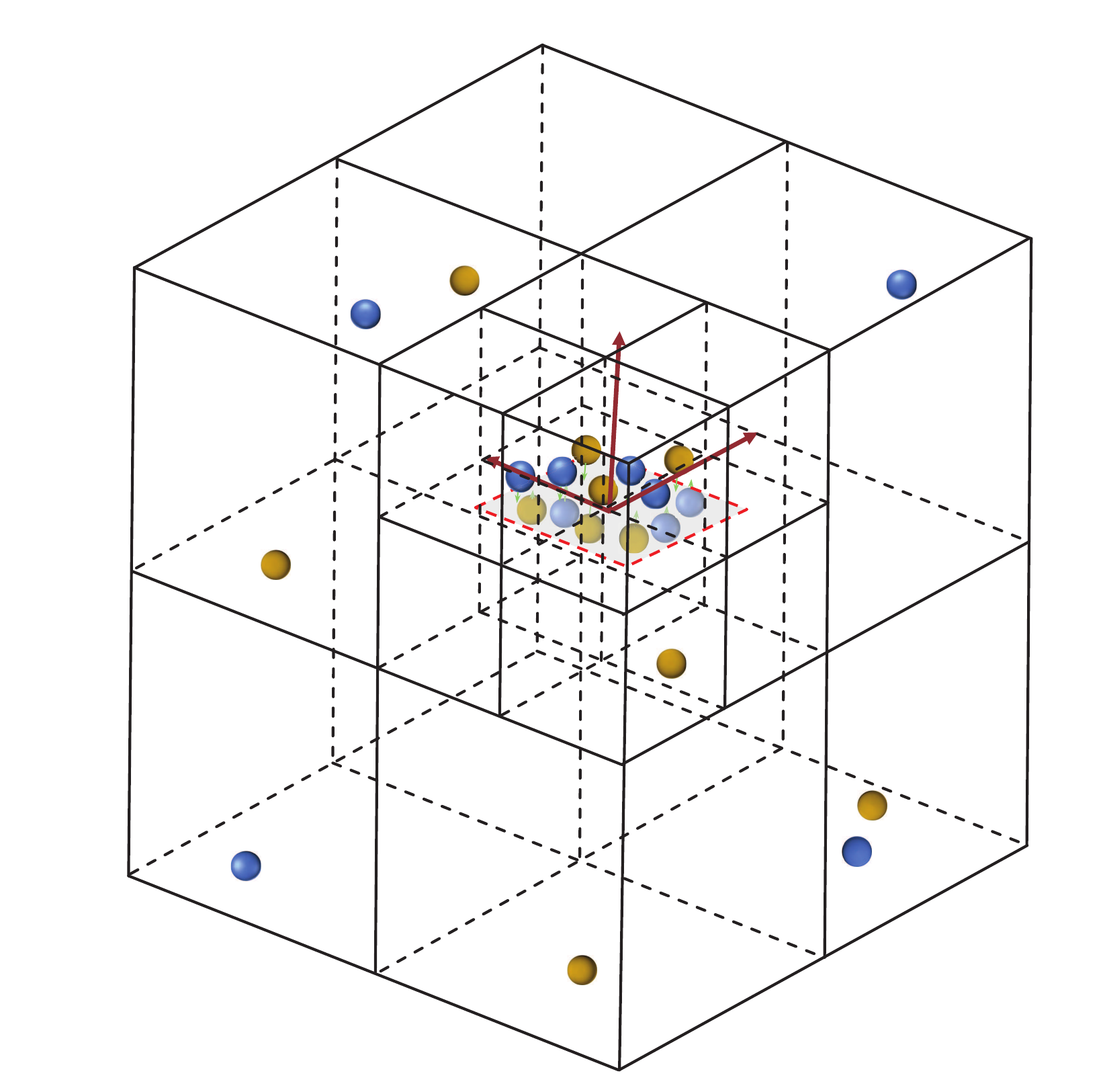}
\caption{Adaptive search for plane features in the micro-structure}
\label{fig:cube}
\end{figure}

Each pair of micro-structures $\left\{ \mathsf{\mathcal{M}}_{i}^{\mathsf{\mathcal{P}}},\mathsf{\mathcal{M}}_{i}^{\mathsf{\mathcal{Q}}} \right\}$ serves as neighborhoods for inliers. After the initial transformation, planar features are detected, and points with similar features are associated with the same plane. As shown in figure \ref{fig:cube}, we employ the octree approach to locate as many plane feature points as possible. This involves computing the eigenvalue ratio of the covariance matrix \cite{mishra2017multivariate} and residuals within a cube to assess whether it represents a plane. If not, the cube is further subdivided until the point count within the field is insufficient. If it does, the points are considered plane feature points associated with the same plane, denoted as  ${{\mathbf{p}}_{f_{i}}^{*}}$ or ${{\mathbf{q}}_{f_{i}}^{*}}$,  representing points from $\mathsf{\mathcal{M}}^{\mathsf{\mathcal{P}}}$ or  $\mathsf{\mathcal{M}}^{\mathsf{\mathcal{Q}}}$. We assume that ${{{\mathsf{\mathcal{K}}}_{f}}}$ planes are detected, each associated with ${N}_{f_{K}}$ feature points, where ${a}_{K}$ are from $\mathsf{\mathcal{M}}^{\mathsf{\mathcal{P}}}$. For the $K$-th detected plane, the associated feature points are denoted as $\mathbf{p}_{{{Mf}_{K}}}^{*}={{[\mathbf{p}_{{{f}_{1}}}^{*},\cdots ,\mathbf{p}_{{{f}_{{{a}_{K}}}}}^{*},\mathbf{q}_{{{f}_{1}}}^{*},\cdots \mathbf{q}_{{{f}_{{{N}_{f_{K}}-{a}_{K}}}}}^{*}]}_{{{N}_{f_{K}}}}}$.  

\subsubsection{Planar Adjustment Combined with Anderson Accelerated (PA-AA) \label{sec:p2pl}}
After detecting geometric plane features, a more efficient approach involves jointly  optimizing the plane parameters and pose, rather than first calculating the normal vectors of points and then minimizing the point-to-plane distance to obtain the optimal registration parameters. Therefore, we propose the PA-AA optimization process. In the derivation, we eliminate the plane parameters to reduce the optimization dimensions and incorporate Anderson acceleration to further minimize the objective function value, achieving higher precision. Initially, we translate the minimization of point-to-plane distances into the joint optimization of the orientation and pose of plane features. With the pose of $\mathsf{\mathcal{M}}^{\mathsf{\mathcal{P}}}$ as the reference (identity matrix),  the objective is to optimize the pose of $\mathsf{\mathcal{M}}^{\mathsf{\mathcal{Q}}}$ relative to $\mathsf{\mathcal{M}}^{\mathsf{\mathcal{P}}}$. The feature points in the reference frame are as follows.
\begin{equation}
\begin{aligned}
\label{eq:133}
&\mathbf{p}_{{{M}_{K_{i}}}}^{*}=\mathbf{T}^{g}_{j}\mathbf{p}_{{{Mf}_{K_{i}}}}^{*},\\
i\in[1,\cdots,{N}_{f_{K}}],\,&j=\left\{
    \begin{array}{ll}
        1,& i\leq a_{K} \\
        2,& i>a_{K}
    \end{array}
\right., \mathbf{T}^{g}=[\mathbf I_{4},{{\mathbf{\hat{T}}}_{c}}]
\end{aligned}
\end{equation}
The cost term is the mean distance from each plane feature point $\mathbf{p}_{{{M}_{K_{i}}}}^{*}$ to its corresponding plane.
\begin{equation}
\label{eq:13}
\underset{\mathbf{T},\mathbf{n}}{\mathop{\arg \min }}\,\sum\limits_{K=1}^{{{\mathsf{\mathcal{K}}}_{f}}}{\frac{1}{{{N}_{f_{K}}}}\sum\limits_{i=1}^{{{N}_{f_{K}}}}{{{\left\| {{\mathbf{n}}_{K}}^{T}(\mathbf{p}_{{{M}_{K_{i}}}}^{*}-{{{\overline{\mathbf{p}}}^{*}_{{{M}_{K}}}}}) \right\|}^{2}}}}
\end{equation}
Where ${\mathbf{n}_{K}}$ represents the unit normal vector of the plane, and ${{{\overline{\mathbf{p}}}^{*}_{{{M}_{K}}}}}$ denotes the point on the plane.
When ${{{\overline{\mathbf{p}}}^{*}_{{{M}_{K}}}}}=\frac{1}{{{N}_{f_{K}}}}\sum\limits_{i=1}^{{{N}_{f_{K}}}}{{{\mathbf{p}}^{*}_{{{M}_{K_{i}}}}}}$, the cost function can be reformulated as (\ref{eq:14}), defining $\mathcal{C}\mathcal{O}\mathcal{V}\triangleq \sum\limits_{i=1}^{{{N}_{f_{K}}}}{\frac{1}{{{N}_{f_{K}}}}(\mathbf{p}_{{{M}_{{{K}_{i}}}}}^{*}-\overline{\mathbf{p}}_{{{M}_{K}}}^{*}){{(\mathbf{p}_{{{M}_{{{K}_{i}}}}}^{*}-\overline{\mathbf{p}}_{{{M}_{K}}}^{*})}^{T}}}$. The quadratic form ${{\mathbf{n}}_{K}}^{T}\mathcal{C}\mathcal{O}\mathcal{V}{{\mathbf{n}}_{K}}$ is a standard eigenvalue problem.  It can be ascertained that the inner minimum solution of (\ref{eq:14}) is the eigenvector ${\mathbf{u}}_{\min }$ corresponding to the minimum eigenvalue $\lambda_{\min}$ of the covariance matrix $\mathsf{\mathcal{C}\mathcal{O}\mathcal{V}}$.
\begin{equation}
\label{eq:14}
\underset{\mathbf{T}}{\mathop{\arg \min }}\,\sum\limits_{K=1}^{{{\mathsf{\mathcal{K}}}_{f}}}{\underbrace{\underset{\mathbf{n}}{\mathop{\min }}\,({{\mathbf{n}}_{K}}^{T}\mathsf{\mathcal{C}\mathcal{O}\mathcal{V}}{{\mathbf{n}}_{K}}\text{)}}_{{{{\mathbf{\hat{n}}}}_{K}}={{\mathbf{u}}_{\min }}}}
\end{equation}

The objective function (\ref{eq:13}) can also be reformulated as (\ref{eq:15}). By eliminating the plane parameter via a closed-form solution, the optimization is reduced to a function solely dependent on $\mathbf{T}$, thereby lowering the dimension of the problem.
\begin{equation}
\label{eq:15}
\underset{\mathbf{T}}{\mathop{\mathbf{\hat{T}}=\arg \min }}\,\sum\limits_{K=1}^{{{\mathsf{\mathcal{K}}}_{4}}}{{{{\lambda}}_{{{\min }}}}(\mathbf{p}_{{{{M}_{K}}}}^{*}(\mathbf{T}))}
\end{equation}

The LM algorithm can be directly applied to solve (\ref{eq:15}) in a relatively straightforward manner. Additionally, the problem can be transformed into the Lie algebra space for optimization, as described in \ref{sec:GNC-Welsch_Estimator}. Given that the pose of $\mathsf{\mathcal{M}}^{\mathsf{\mathcal{P}}}$ is treated as the identity matrix, we only need to enumerate ${{\mathbf{q}}_{f_{i}}^{*}}$ when calculating the Jacobian matrix and second-order derivatives \cite{zhou2023efficient}.
\begin{equation}
\begin{aligned}
\label{eq:16}
{\mathbf J_{K_{i}}}&=\frac{\partial {{{\lambda}}_{{{\min_{K}}}}}}{\partial \Delta \bm{\xi} }\\
&=\left\{    
\begin{array}{ll}
        0&,i\leq a_{K} \\
        \frac{\partial {{{\lambda}}_{{{\min_{K} }}}}}{\partial ({\mathbf{p}_{M_{K_{i}}}^{*}}(\bm{\xi} ))}\cdot \frac{\partial (\exp ({{\bm{\xi} }^{\wedge }}){{\mathbf{p}}^{*}_{{{Mf}_{K_{i}}}}})}{\partial \Delta \bm{\xi}}&,i>{a_{K}}
\end{array}
\right.
\end{aligned}
\end{equation}

\begin{equation}
\label{eq:17}
\frac{\partial {{{\lambda}}_{{{\min }_{K}}}}}{\partial (\mathbf{p}^{*}_{M_{K_{i}}}(\bm{\xi }))}=\frac{2}{{N}_{f_{K}}}(\exp ({{\bm{\xi} }^{\wedge }})\mathbf{p}^{*}_{M_{K_{i}}}-{{\mathbf{\bar{p}}}^{*}_{M_{K}}}){{\mathbf{u}}_{{{\min }_{K}}}}{\mathbf{u}}_{{{\min }_{K}}}^{T}
\end{equation}

We will incorporate Anderson acceleration into the LM algorithm to optimize each iteration. This optimization problem, represented by (\ref{eq:13}), can be viewed as an iterative procedure to find a fixed point \cite{walker2011anderson}. (\ref{eq:15}) can also be rewritten as ${{\mathbf{T}}^{k+1}}=G({{\mathbf {T}}^{k}})$, where $H(\mathbf {T})=G(\mathbf {T})-\mathbf {T}$. A fixed point ${{\mathbf {T}}^{*}}$ exists such that when $\mathbf {T}$ converges to ${{\mathbf {T}}^{*}}$, $H(\mathbf {T}^{*})=0$. This formulation enables the use of Anderson acceleration. By leveraging the current $k$-th iteration and the previous $h$ iterations, Anderson acceleration improves the results of the LM algorithm, facilitating faster convergence for the $(k+1)$ iteration.
\begin{equation}
\label{eq:18}
\bm{\xi}_{AA}^{(k+1)}=G({\bm{\xi }^{(k)}})-\sum\limits_{j=1}^{h}{\theta _{j}^{*}}(G({{\bm{\xi} }^{(k-j+1)}})-G({{\bm{\xi} }^{(k-j)}}))
\end{equation}
\begin{equation}
\label{eq:19}
\begin{aligned}
{{\bm{\Theta} }^{*}}=\arg \min {{\left\| {{H}^{(k)}}-\sum\limits_{j=1}^{h}{{{\theta }_{j}}({{H}^{(k-j+1)}}-{{H}^{(k-j)}})} \right\|}^{2}}
\end{aligned}
\end{equation}
Where $\sum\limits_{j=1}^{h}{{{\theta }_{j}}=1}$ and  ${{\bm{\Theta} }^{*}}=\left\{ \theta _{1}^{*},\ldots ,\theta _{h}^{*} \right\}$ are obtained through least squares estimation. The result is updated by $\mathbf{\hat{T}}_{AA}^{(k+1)}=\exp (\bm{\xi} {{_{AA}^{(k+1)}}^{\wedge}})$.
To preserve performance, we only accept the accelerated value as a new iteration if it results in a lower cost than the current one, as defined by (\ref{eq:13}). Finally, we obtain the optimized result ${{{\mathbf{\hat{T}}}}_{f}}$. 

\section{Experiments}
\label{sec:experiment}

\subsection{Experimental Setup}
\noindent{\bf Datasets.} We consider two well-known real-world scene datasets: the ETH PRS TLS Registration (ETH) dataset \cite{theiler2015globally} and the 3DMatch dataset \cite{zeng20173dmatch}. The ETH dataset comprises large-scale data from four outdoor scenes and one indoor scene: arch, courtyard, facade, office, and trees, including 77 matching pairs. It represents real-world scanning scenes. The 3DMatch dataset, on the other hand, contains data from 8 indoor scenes used for evaluation, including 1623 (FPFH) matching pairs and 1546 (FCGF) matching pairs. It features diverse local geometries at various scales and reconstructs from RGB-D data. These datasets pose various registration challenges, such as partial overlaps, noise, and outliers. Our experiments aim to validate the effectiveness and versatility of our approach across these diverse indoor and outdoor scene datasets. The details of each data set are shown in table \ref{tab:data_detail}.

\noindent{\bf Evaluation Criteria.}
We will employ four evaluation criteria to assess the registration results of different methods: rotational error ($RE$), translational error ($TE$), success rate ($RR$), and computational time.
\begin{equation}
RE=\arccos \left( \frac{tr(\mathbf{\tilde{R}}{{{\mathbf{\hat{R}}}}^{T}})-1}{2} \right) ,
TE=\left\| \mathbf{\hat{t}}-\mathbf{\tilde{t}} \right\|
\label{eq:metric}
\end{equation}
Here, $\mathbf{\hat{R}}$ represents the estimated rotation matrix, and $\mathbf{\tilde{R}}$ is the ground truth rotation matrix. $tr(\cdot )$ denotes the trace of a matrix. $\mathbf{\hat{t}}$ is the estimated translation matrix, and $\mathbf{\tilde{t}}$ is the ground truth translation matrix. For the 3DMatch dataset, registration is considered successful when $RE\le {{15}^{{}^\circ }}$ and $TE\le 30\text{cm}$ \cite{zhang20233d}; similarly, for the ETH dataset, successful registration is defined as $RE\le {{5}^{{}^\circ }}$ and $TE\le 50\text{cm}$\cite{theiler2015globally}. The success rate is defined as the ratio of successful registrations to the total number of pairs, and computational time is measured from the input of initial correspondence points to the final successful estimation of the transformation matrix.

\noindent{\bf Implementation Details.} Our method is implemented in C++ based on the Point Cloud Library (PCL) and the Eigen and Sophus libraries. We employ ISS key points detection and FPFH \cite{rusu2009fast} descriptors for the ETH dataset to generate initial correspondence sets, and both FPFH descriptors and FCGF \cite{choy2019fully} learned descriptors for the 3DMatch dataset. All experiments were conducted on a system with a 12th Gen Intel(R) Core(TM) i9-12900KF and 32 GB RAM.
\begin{table*}[tbp]
\centering
\caption{detailed settings of the compared algorithms}
\label{tab:detailed settings of the experiment}
\renewcommand{\arraystretch}{1.5}
\begin{tabularx}{\linewidth}{lXX}
\toprule 
\textbf{Method}                 & \textbf{Parameters}                                                                       & \textbf{Implementation}           \\ 
\midrule
RANSAC-M             & Maximum number of iterations: M;                                 inlier threshold: 2$\ell$; \newline confidence: 0.99. & C++ code; single thread      \newline  \url{https://github.com/PointCloudLibrary}       
\\ 
Tri-FGR             & Annealing rate: 1.4;                                 maximum correspondence distance: $\ell$; \newline maximum number of iterations: 100. & C++ code; single thread      \newline  \url{https://github.com/intel-isl/FastGlobalRegistration}   
\\
TEASER++              & Noise bound: 0.05 (3DMatch FPFH), 0.1 (ETH), \newline0.0075 (3DMatch FCGF); \newline rotation max iterations: 100; \newline rotation gnc factor: 1.4; rotation cost threshold: 0.005.                                  & C++ code; single thread      \newline  \url{https://github.com/MIT-SPARK/TEASER-plusplus}       
\\ 
MAC               & $t_{cmp}=0.99$; $d_{cmp}=10\ell$; inlier threshold: 2$\ell$. & C++ code; single thread    \newline  \url{https://github.com/zhangxy0517/3D-Registration-with-Maximal-Cliques}              
\\ 
GROR                  & 3DMatch: K = 0.7${\mathsf{\mathcal{K}}_{1}}$; $\rho=\ell$; \newline  ETH: K = 800; $\rho=\ell$. & C++ code; single thread  \newline \url{https://github.com/WPC-WHU/GROR} \\ 
pl-ICP               & Maximum number of iterations: 100; \newline maximum correspondence distance: 2$\ell$; \newline minimum transformationEpsilon: 1e-8; \newline compute normal of RadiusSearch: 4$\ell$   & C++ code; single thread   \newline \url{https://github.com/PointCloudLibrary}             
\\ 
GICP                          &  Maximum number of iterations: 100; \newline maximum correspondence distance: 2$\ell$; \newline minimum transformationEpsilon: 1e-8.
                                           & C++ code; single thread      \newline \url{https://github.com/PointCloudLibrary}    
\\ 
BALM                       & Voxel\_size: 8$\ell$. & C++  code; single thread    \newline \url{https://github.com/hku-mars/BALM}    
\\
Proposed                       & 3DMatch: ${{K}_{opt}}$=0.7${\mathsf{\mathcal{K}}_{1}}$; $\ell=0.05$; $L=5\ell$ \newline  ETH: ${{K}_{opt}}=800$; $\ell=0.1$; $L=2\ell$. & C++  code; single thread    \newline \url{https://github.com/Rolin-zrl/} (will be available)
\\ 
\bottomrule
\end{tabularx}
\end{table*}
\begin{table}[tb]
\centering
\caption{more information about each dataset}
\label{tab:data_detail}
\begin{tabularx}{\linewidth}{lllXXX}
\toprule
\textbf{Dataset} & \textbf{scenes}& \textbf{pairs}  & \textbf{Average number of points (10\textsuperscript{6})}  & \textbf{Average number of correspondence } & \textbf{Outlier ratio} \\ 
\midrule
3DMatch (FPFH)  &8 & 1623  & 0.34  & 4709 & 92.85\% \\
3DMatch (FCGF)  &8 & 1546  & 0.34 & 4656 & 81.26\% \\
ETH-Arch &1 & 8  & 27.64  & 14317 & 98.74\% \\
ETH-Courtyard  &1 & 28  & 13.54  & 17820 & 96.27\% \\
ETH-Facade &1 & 21  & 19.80  & 2047 & 96.84\% \\
ETH-Office &1 & 10  & 10.72  & 6090 & 97.79\% \\
ETH-Trees &1 & 10  & 20.17  & 21885 & 99.64\% \\

\bottomrule
\end{tabularx}
\end{table}

\begin{table*}[htbp]
\centering
\caption{ETH dataset registration results.}
\label{tab:ETH}
\resizebox{\linewidth}{!}{
\begin{tabular}{@{}lcccccccccccccccc@{}}
\toprule
\multirow{2}{*}{Method} & \multicolumn{3}{c}{Arch} & \multicolumn{3}{c}{Courtyard} & \multicolumn{3}{c}{Facade} & \multicolumn{3}{c}{Office} & \multicolumn{3}{c}{Trees} \\
\cmidrule(lr){2-4} \cmidrule(lr){5-7} \cmidrule(lr){8-10} \cmidrule(lr){11-13} \cmidrule(lr){14-16}
  & RR(\%) & RE($^\circ$) & TE(cm) & RR(\%) & RE($^\circ$) & TE(cm) & RR(\%) & RE($^\circ$) & TE(cm) & RR(\%) & RE($^\circ$) & TE(cm) & RR(\%) & RE($^\circ$) & TE(cm) \\
\midrule
RANSAC-100K & 50.00 & 1.50 & 37.94 & 96.43 & 0.35 & 12.40 & 100.00 & 0.42 & 8.19 & 100.00 & 2.24 & 20.00 & 50.00 & 1.37 &14.76 \\
Tri-FGR & 12.50 & 0.19 & 16.36 & 96.43 & 0.25 & 14.04 & 85.70 & 0.51 & 11.07 & 30.00 & 0.54 & 9.59 & 50.00 & 0.43 & 14.76 \\

TEASER++ & 87.50 & 0.26 & 7.27 & 100.00 & 0.055 & 4.00 & 90.48 & 0.23 & 4.90 & 90.00 & {\bf0.27} & 3.77 & 100.00 & 0.32 & 4.02 \\

MAC & 62.50 & {\bf0.17} &{\bf6.18} & 100.00 & {0.044} & {3.58} & 71.43 & 0.11 & {\bf2.14} & 90.00 & 0.38 & 3.12 & 100.00 & 0.72 & 5.74 \\
GROR & 100.00 & 0.40 & 9.81 & 100.00 & 0.055 & 3.67 & 100.00 & 0.17 & 3.60 & 100.00 & 0.42 & 3.67 & 100.00 & 0.18 & 3.67 \\
Proposed (CR) & {\bf100.00} & 0.27 & {6.35} & {\bf100.00} & {\bf0.044} & {\bf3.52} & {\bf100.00} & {\bf0.098} & 2.24 & {\bf100.00} & 0.32 & {\bf2.29} & {\bf100.00} & {\bf0.17} & {\bf2.70} \\ \\
RANSAC-100K+pl-ICP & {50.00} & {0.036} & {1.15} & {96.43} & {\bf0.031} & {3.45} & {100.00} & {0.036} & {0.97} & {100.00} & {0.093} & {\bf0.77} & {50.00} & {\bf0.045} & {\bf0.39} \\ 
MAC+GICP & {62.50} & {\bf0.018} & {\bf0.55} & {100.00} & {0.033} & {3.54} & {76.19} & {0.037} & {0.82} & {90.00} & {0.083} & {0.78} & {100.00} & {0.077} & {1.20} \\
GROR+BALM & {100.00} & {0.27} & {3.90} & {100.00} & {0.034} & {3.77} & {100.00} & {0.047} & {0.96} & {100.00} & {0.11} & {1.29} & {100.00} & {0.059} & {0.89}\\
CR+BALM & {100.00} & {0.12} & {3.52} & {100.00} & {0.034} & {3.56} & {100.00} & {0.039} & {0.89} & {100.00} & {0.097} & {1.17} & {100.00} & {0.063} & {0.77} \\
Proposed (CR+FR) & {\bf100.00} & 0.096 & {2.59} & {\bf100.00} & 0.034 & {\bf3.43} & {\bf100.00} & {\bf0.036} & {\bf0.81} & {\bf100.00} & {\bf0.082} & {0.78} & {\bf100.00} & {0.053} & {0.69} \\
\bottomrule
\end{tabular}
}
\end{table*}

\begin{table}[htbp]
\centering
\caption{3DMatch dataset registration results.}
\label{tab:3DMatch}
\resizebox{\linewidth}{!}{
\begin{tabular}{@{}lcccccc@{}}
\toprule
\multirow{2}{*}{Method} & \multicolumn{3}{c}{FPFH } & \multicolumn{3}{c}{FCGF} \\ 
\cmidrule(lr){2-4} \cmidrule(lr){5-7} 
& RR(\%) & RE($^\circ$) & TE(cm) & RR(\%) & RE($^\circ$) & TE(cm) \\ \midrule
RANSAC-50K & 77.14 & 3.80 & 10.87 & 90.88 & 3.38 & 10.40 \\
Tri-FGR & 66.73 & 2.37 & 6.68 & 88.78 & 2.21 & 6.74 \\
TEASER++ & 78.68 & 2.29 & 7.06 & 86.08 & 2.50 & 8.15 \\
MAC & {\bf83.80} & 2.11 & 6.78 & {\bf93.72} & 2.03 & 6.55 \\
GROR & 82.38 & 2.60 & 7.60 & 92.54 & 2.39 & 7.17 \\
Proposed (CR) & 83.43 & {\bf2.05} & {\bf6.50} & 93.53 & {\bf2.00} & {\bf6.37} \\ \\
RANSAC-50K+pl-ICP & {83.67} & 1.68 & 6.35 & 93.53 & 1.76 & 6.55 \\
MAC+GICP & {\bf84.47} & 1.66 & 6.09 & {\bf93.72} & 1.65 & 6.60 \\
GROR+BALM & 83.05 & 1.90 & 6.63 & 92.85 & 1.97 & 6.70 \\
CR+BALM & {83.98} & 1.70 & 6.06 & 93.16 & 1.76 & 6.25 \\
Proposed (CR+FR) & {84.16} & {\bf1.65} & {\bf5.75} & 93.53 & {\bf1.65} & {\bf5.88} \\  

\bottomrule
\end{tabular}
}
\end{table}

\subsection{Results on 3DMatch and ETH Datasets}
\begin{figure}[t]
\centering
\includegraphics[width=\linewidth]{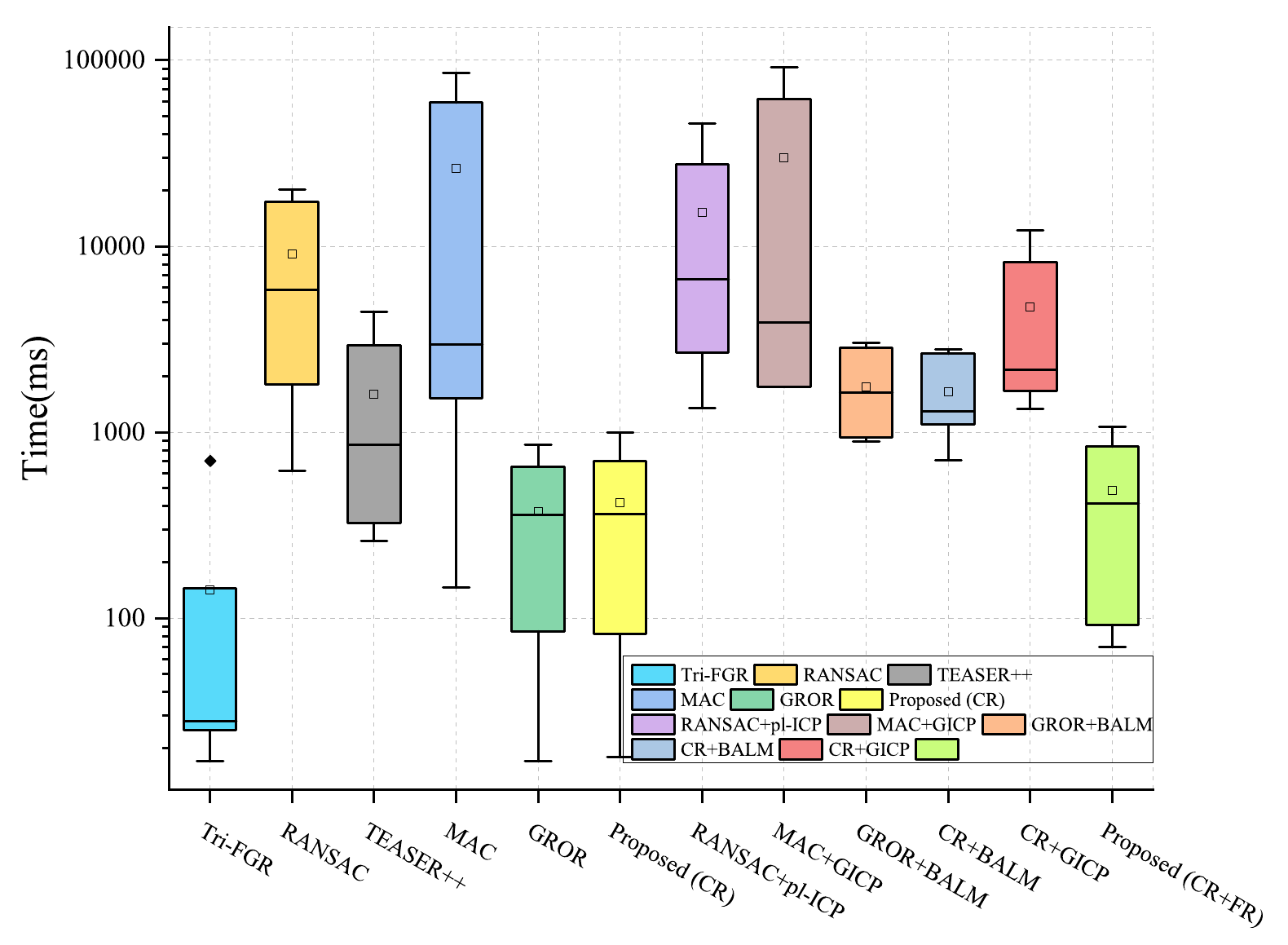}
\caption{Box-plot of time-consuming on 3DMatch and ETH datasets.}
\label{fig:time_box}
\end{figure}

\begin{figure*}[t]
\centering
\includegraphics[width=\textwidth]{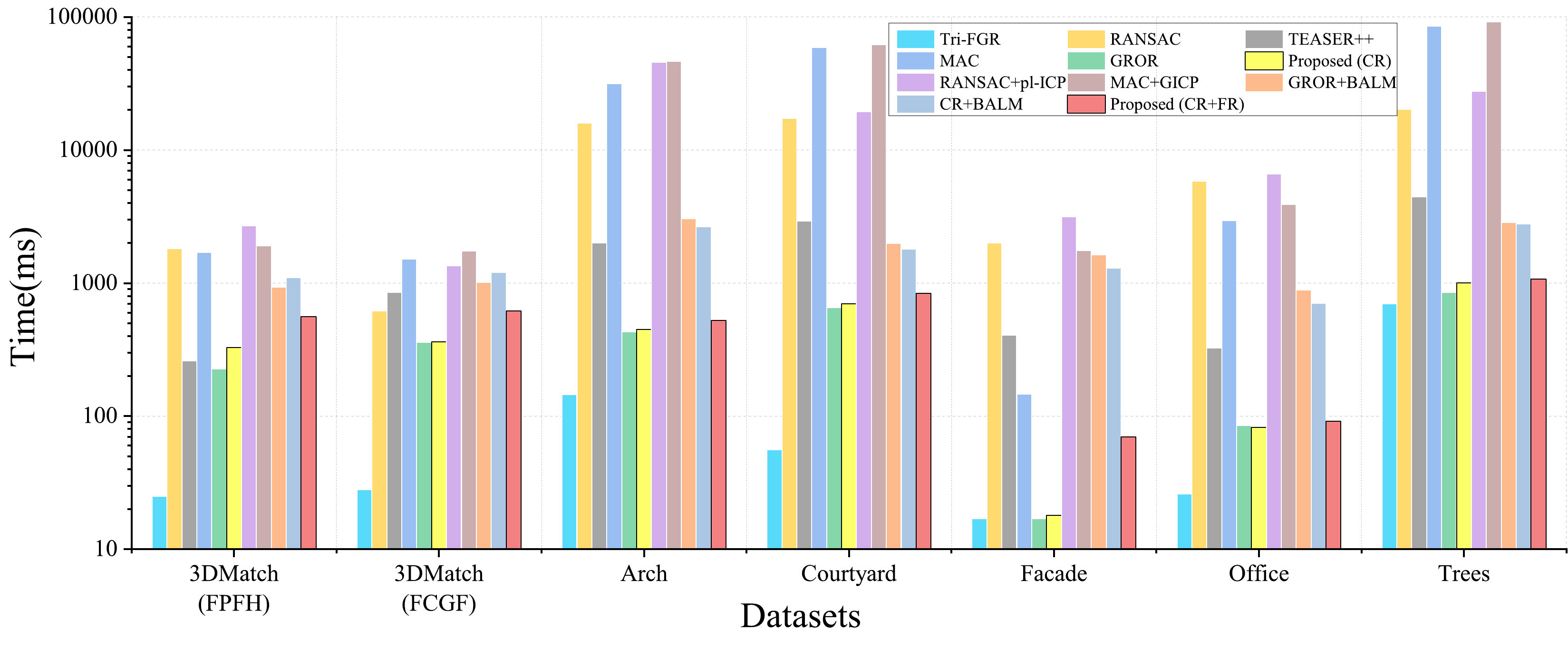}
\caption{Time consumption comparison of different methods on 3DMatch and ETH datasets.}
\label{fig:time}
\end{figure*}

We conducted comparative tests with some advanced baselines: RANSAC \cite{fischler1981random}, GoICP-Trimming \cite{yang2013go}+FGR \cite{zhou2016fast}(Tri-FGR), TEASER++ \cite{yang2020teaser}, MAC \cite{zhang20233d}, GROR \cite{yan2022new}, pl-ICP \cite{censi2008icp}, GICP \cite{segal2009generalized}, and BALM \cite{liu2023efficient}. The last three are fine registration methods, whereas the rest are coarse registration. We also combined some advanced coarse-to-fine registration methods, such as RANSAC+pl-ICP, MAC+GICP, and GROR+BALM. Before the fine registration, we downsample the point cloud with a resolution of $\ell$. We assessed the average $RE$, $TE$, and time of all matching pairs in the datasets. The detailed settings of the compared methods in the experiment are shown in table \ref{tab:detailed settings of the experiment}. The results are as follows: table \ref{tab:ETH}, table \ref{tab:3DMatch}, figure \ref{fig:time_box}, and figure \ref{fig:time}.

We report the outcomes of both coarse and fine registration within the methods and compare these results against the baseline in the same category. The following conclusions can be made: 1) our method outperforms all comparison methods of the same category on both 3DMatch and ETH datasets, regardless of coarse and fine registration; 2) our method achieves accurate alignment of all scenes on the ETH dataset of real scans. It shows significant efficiency advantages when registering large-scale outdoor scenes.

The upper sections of table \ref{tab:3DMatch} and table \ref{tab:ETH} represent the coarse registration methods, while the lower sections correspond to the coarse-to-fine registration methods. Combined with the time-consuming results in figure \ref{fig:time}, we will conduct a comprehensive analysis. First, let us take a look at the results of the coarse registration. Tri-FGR is the most efficient for coarse registration methods but demonstrates limited robustness with correspondence sets containing a high ratio of outliers. The efficiency of RANSAC and MAC is sensitive to the size of the correspondence set.   In the arch, courtyard, and trees scenes, when the input size exceeds 10,000, RANSAC requires more sampling iterations as the number of pairs increases. Meanwhile, MAC's algorithm is significantly burdened by graph construction tasks, resulting in processing times exceeding 15 seconds. In the ETH dataset's trees scene, TEASER++ also experiences a notable efficiency drop when handling dense compatibility graphs, with the average processing time rising to 4.5 seconds. However, our coarse registration (CR) employs a graph-based hierarchical removal strategy, effectively handling large-scale correspondence sets. As the size of the set increases, our method's time advantage becomes even more pronounced. Compared to the second highest-accuracy MAC method in the 3DMatch dataset, our CR reduces the average processing time to 1/3 of MAC's when the correspondence set is around 5,000. Additionally, our CR reduces $RE$ and $TE$ by 2.8\% and 4.1\%, respectively, compared to MAC, while maintaining nearly the same peak success rate ($RR$) across the 3DMatch dataset with over a thousand pairs.
For GROR, which operates at a comparable efficiency level, our CR gains a distinct advantage by using the GNC-Welsch estimator, enhancing accuracy and success rates, with $RE$ and $TE$ reduced by 21.2\% and 14.5\%, respectively.
On the ETH dataset, our CR has achieved a 100\% success rate and offers superior accuracy compared to other methods with the same success rate. The limited capability of MAC to search for maximal cliques in large and dense graphs compromises its ability to generate correct hypotheses. Although MAC has the lowest $RE$ and $TE$ in arch scenarios, it fails to register three pairs. In contrast, Our CR achieves lower metrics on its successfully registered pairs: $RE=0.11^\circ$ and $TE=4.75cm$; similarly, in the facade scenario, our CR outperforms MAC on its successfully registered pairs,  achieving $RE=0.087^\circ$ and $TE=1.99cm$.

Next, we turn our attention to the results of the coarse-to-fine registration method. While fine registration significantly improves accuracy compared to initial coarse registration, it also increases processing time. pl-ICP requires substantial time to compute the normal for each point, and GICP employs a distribution-to-distribution correspondence model that depends on costly nearest-neighbor searches. Both pl-ICP and GICP are fine registration methods whose efficiency is largely influenced by the scale of the point cloud. BALM, on the other hand, encodes all raw points associated with the same feature, making its efficiency less affected by the point cloud scale compared to the previous methods. However, segmenting corresponding voxels and identifying planes still incur some time costs. A satisfactory registration method should balance efficiency and accuracy. Due to our local micro-structures approach, the additional time cost is minimal after incorporating fine registration (FR) into our CR. The entire coarse-to-fine registration method operates at the millisecond level.
Meanwhile, the $RR$ remains high, and our method achieves the highest accuracy. On the 3DMatch benchmark, our method reduces $TE$ by 5.9\% compared to the second-highest accuracy method. Similarly, on the ETH dataset, our method maintains the highest accuracy while sustaining the same success rate. Compared to CR+BALM and GROR+BALM, improving coarse registration accuracy enhances overall coarse-to-fine registration accuracy. Our coarse registration achieves higher accuracy, contributing to the improved performance of our coarse-to-fine registration method.

\begin{figure*}[!b]
    \centering
    \vspace{2em}
    \begin{minipage}{.160\linewidth}
        \centering
        GT
    \end{minipage}
    \begin{minipage}{.160\linewidth}
        \centering
        MAC
    \end{minipage}
    \begin{minipage}{.160\linewidth}
        \centering
        GROR
    \end{minipage}
    \begin{minipage}{.160\linewidth}
        \centering
        Ours(CR)
    \end{minipage}
    \begin{minipage}{.160\linewidth}
        \centering
         MAC+GICP
    \end{minipage}
    \begin{minipage}{.160\linewidth}
        \centering
        Ours(CR+FR)
    \end{minipage}

    \subfloat{\includegraphics[trim={0pt 0pt 0pt 200pt},clip, width=0.16\linewidth]{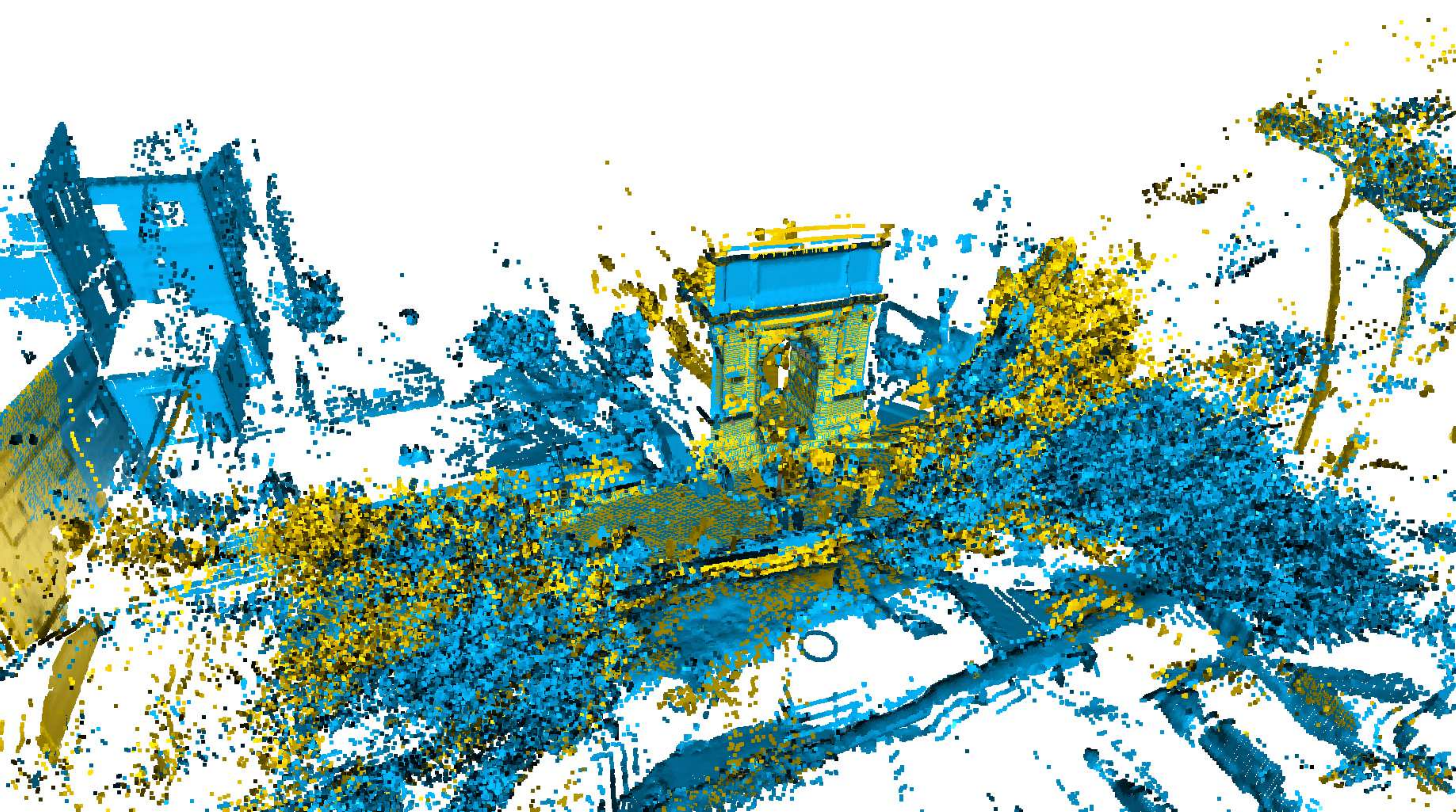}}\hfill
    \subfloat{\includegraphics[trim={0pt 0pt 0pt 200pt}, clip, width=0.16\linewidth]{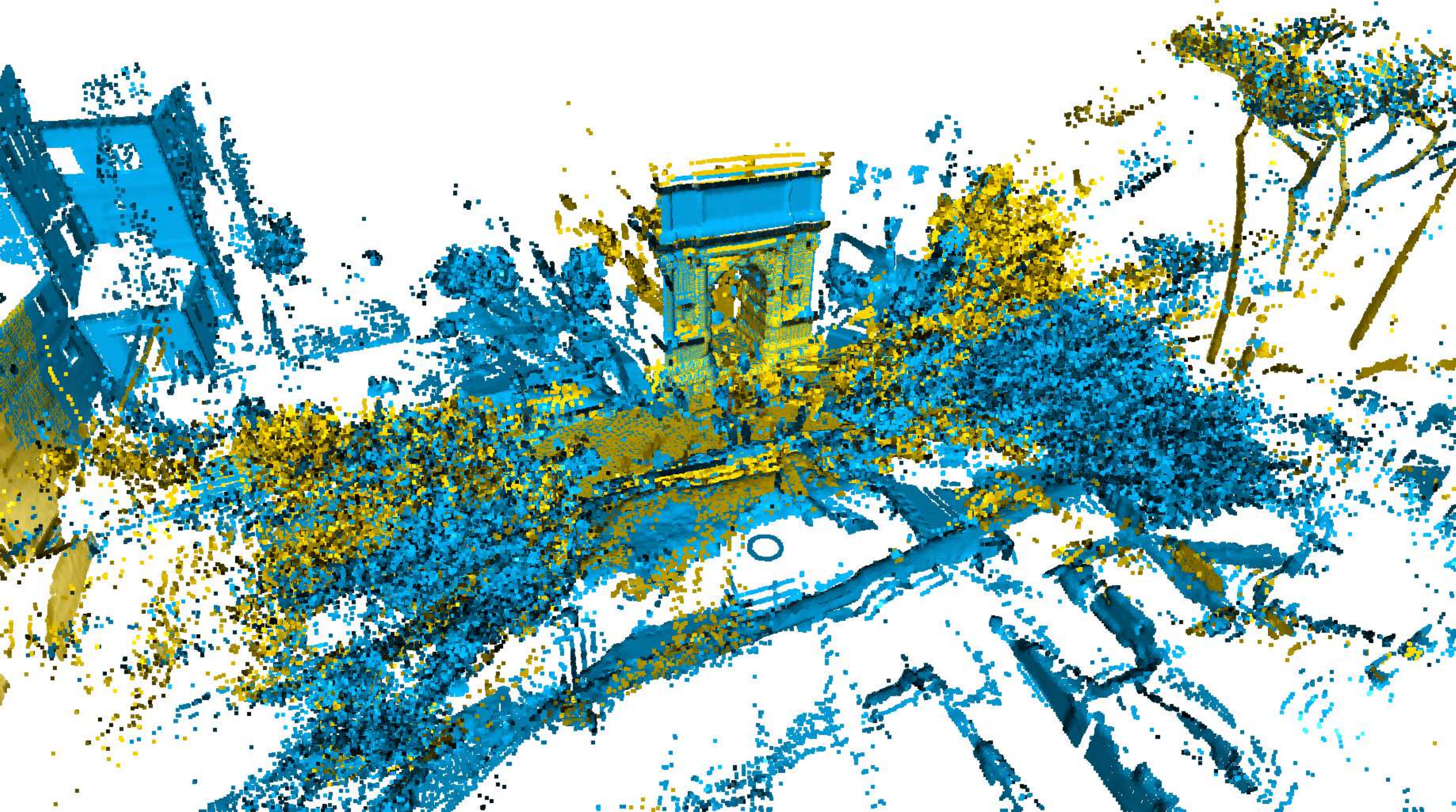}}\hfill
    \subfloat{\includegraphics[trim={0pt 0pt 0pt 200pt}, clip, width=0.16\linewidth]{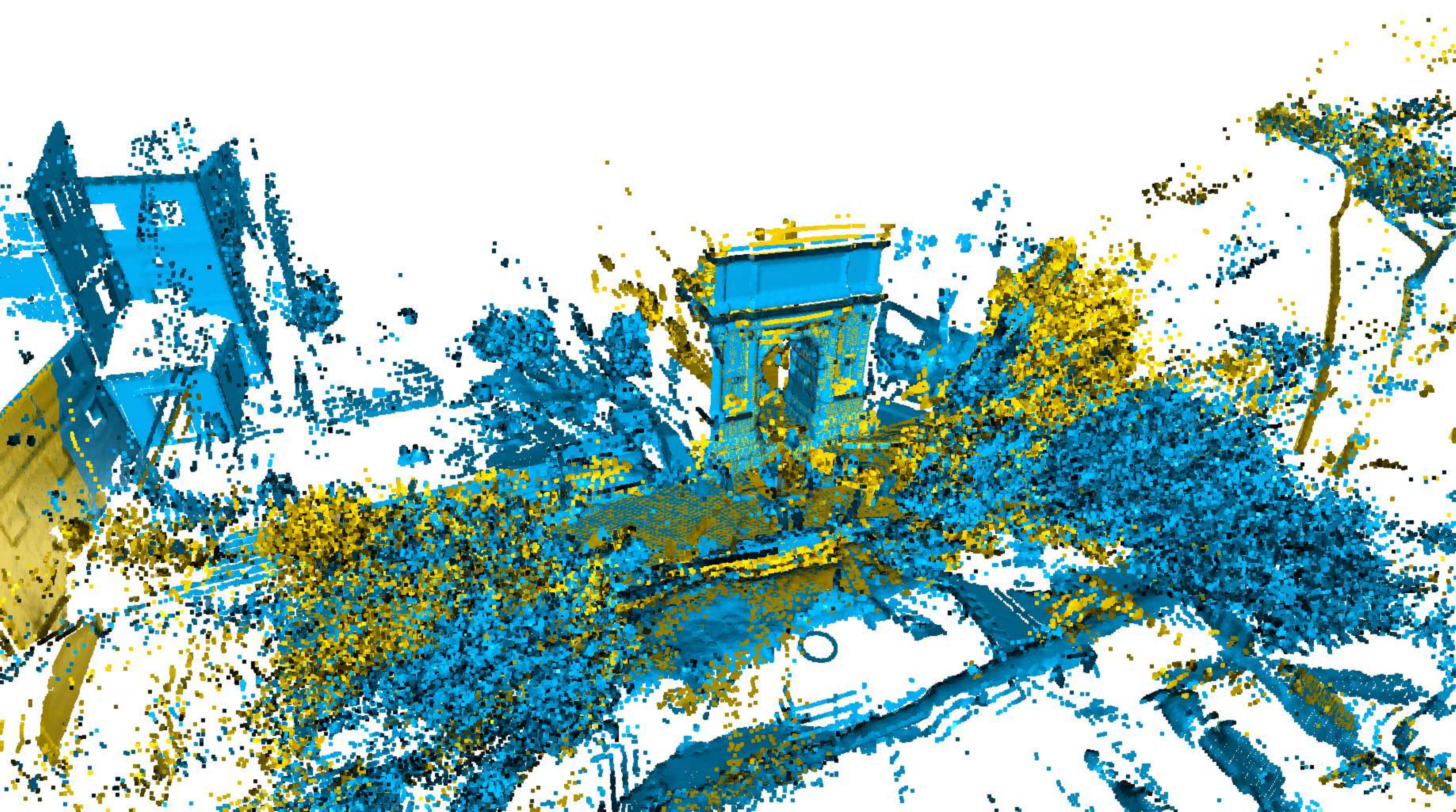}}\hfill
    \subfloat{\includegraphics[trim={0pt 0pt 0pt 200pt}, clip, width=0.16\linewidth]{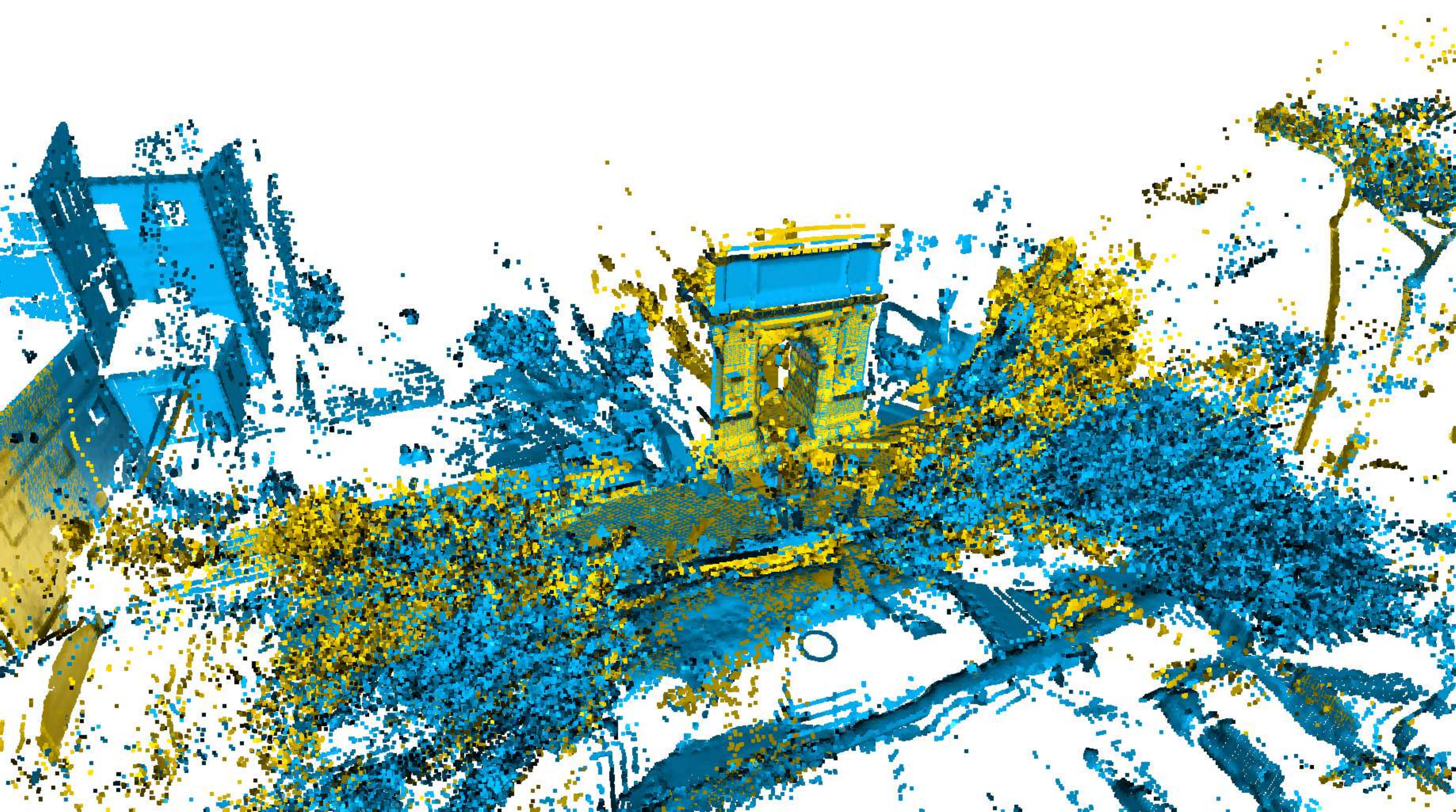}}\hfill
    \subfloat{\includegraphics[trim={0pt 0pt 0pt 200pt}, clip, width=0.16\linewidth]{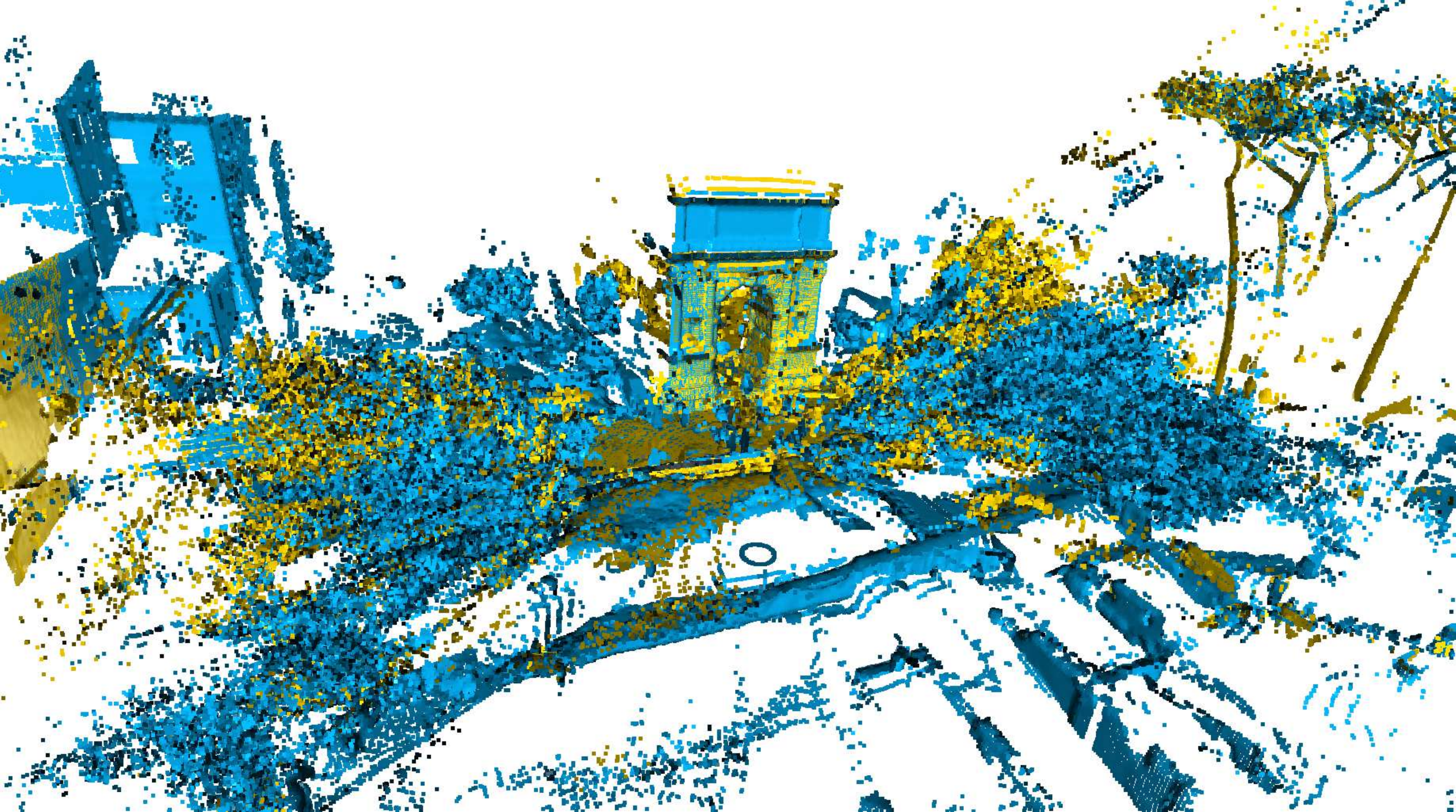}}\hfill
    \subfloat{\includegraphics[trim={0pt 0pt 0pt 200pt}, clip, width=0.16\linewidth]{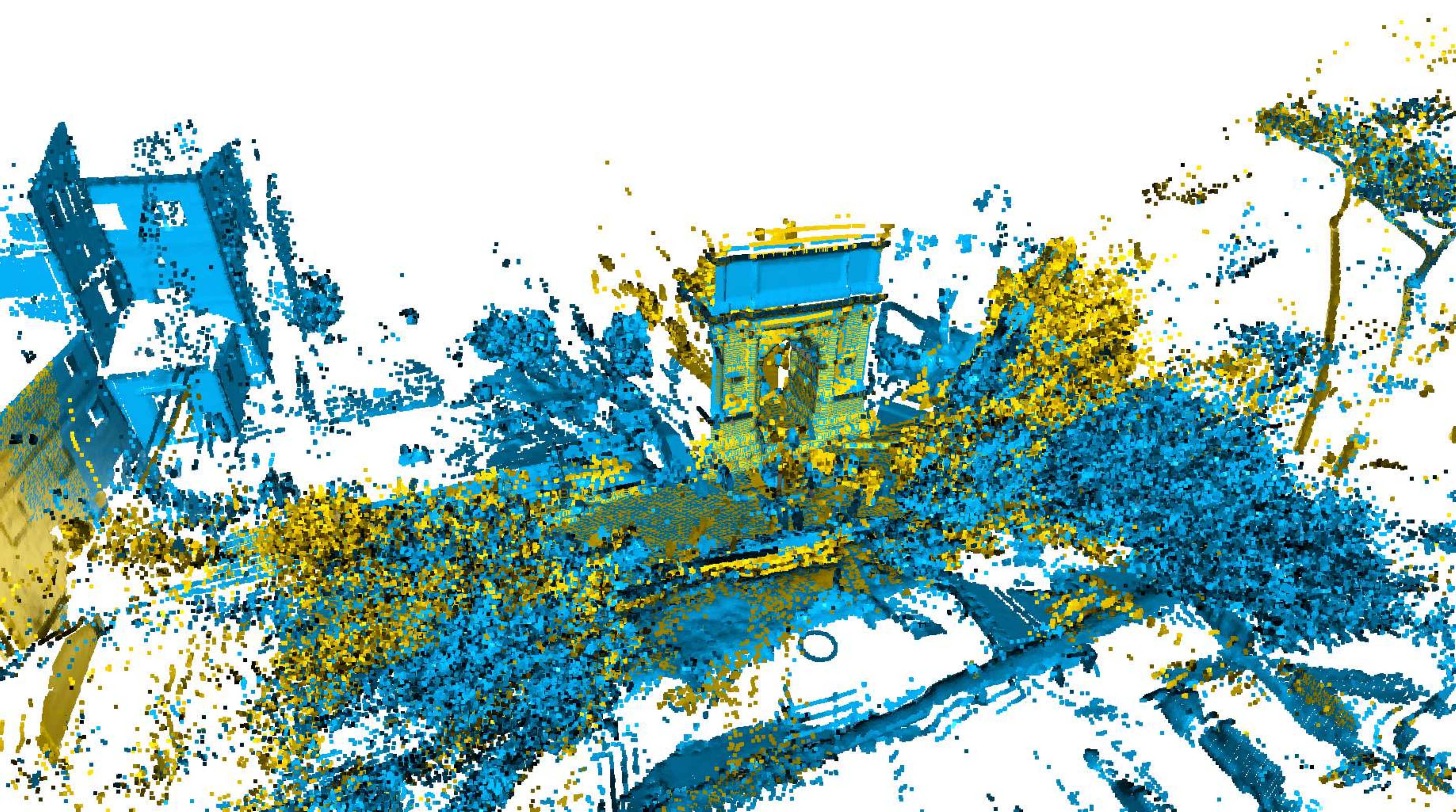}}\hfill
    \\
    \subfloat{\includegraphics[trim={200pt 0pt 200pt 0pt}, clip, width=0.16\linewidth]{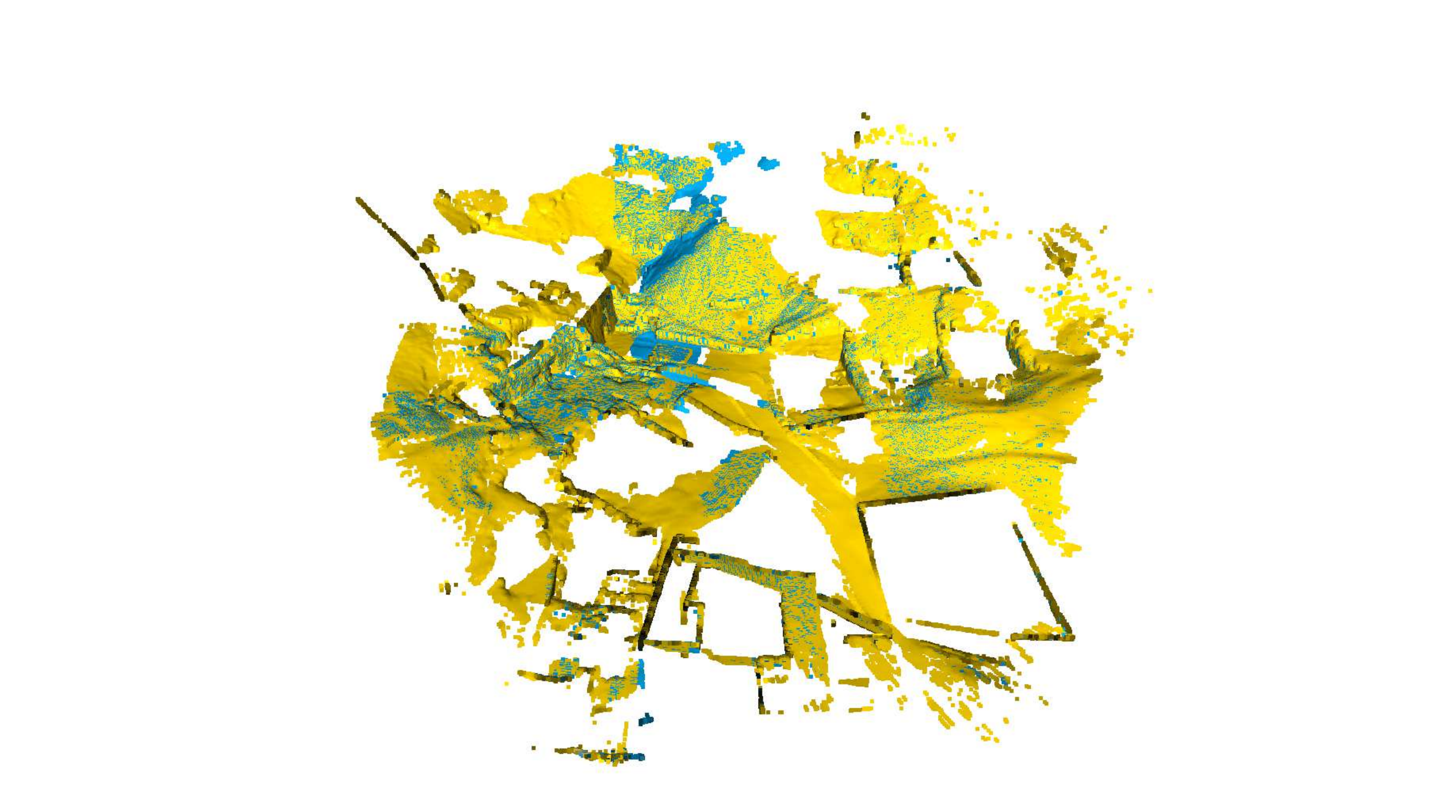}}\hfill
    \subfloat{\includegraphics[trim={200pt 0pt 200pt 0pt}, clip, width=0.16\linewidth]{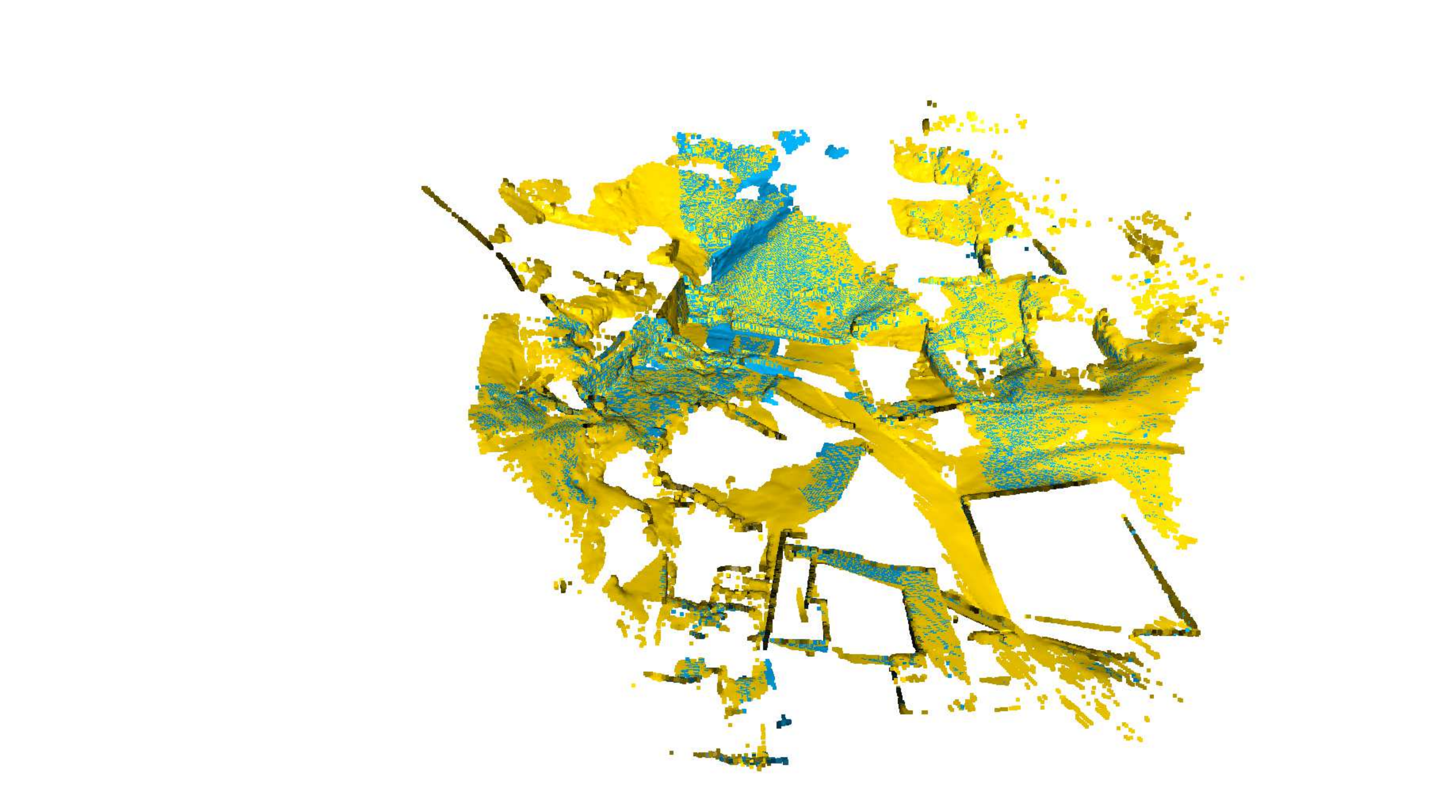}}\hfill
    \subfloat{\includegraphics[trim={200pt 0pt 200pt 0pt}, clip, width=0.16\linewidth]{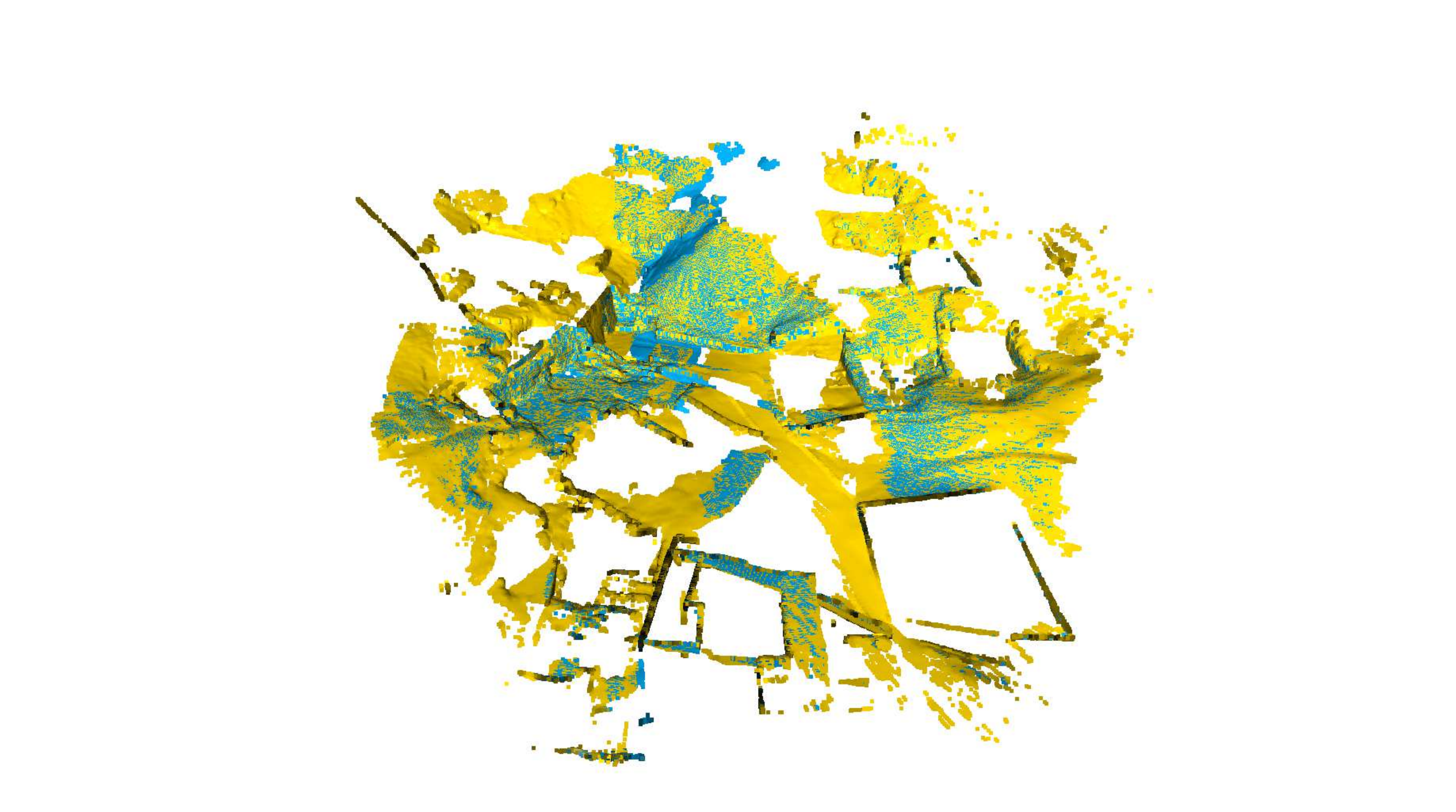}}\hfill
    \subfloat{\includegraphics[trim={200pt 0pt 200pt 0pt}, clip, width=0.16\linewidth]{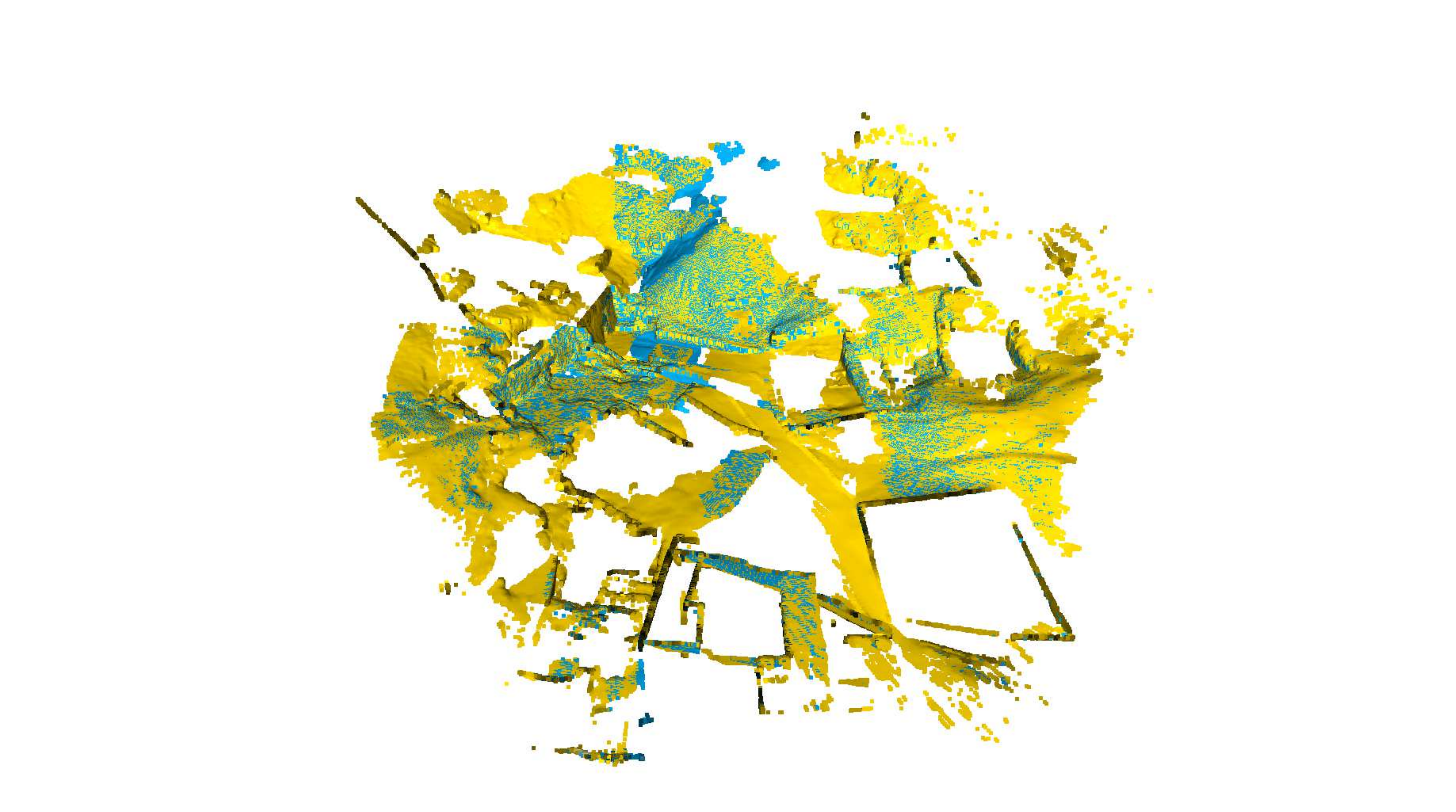}}\hfill
    \subfloat{\includegraphics[trim={200pt 0pt 200pt 0pt}, clip, width=0.16\linewidth]{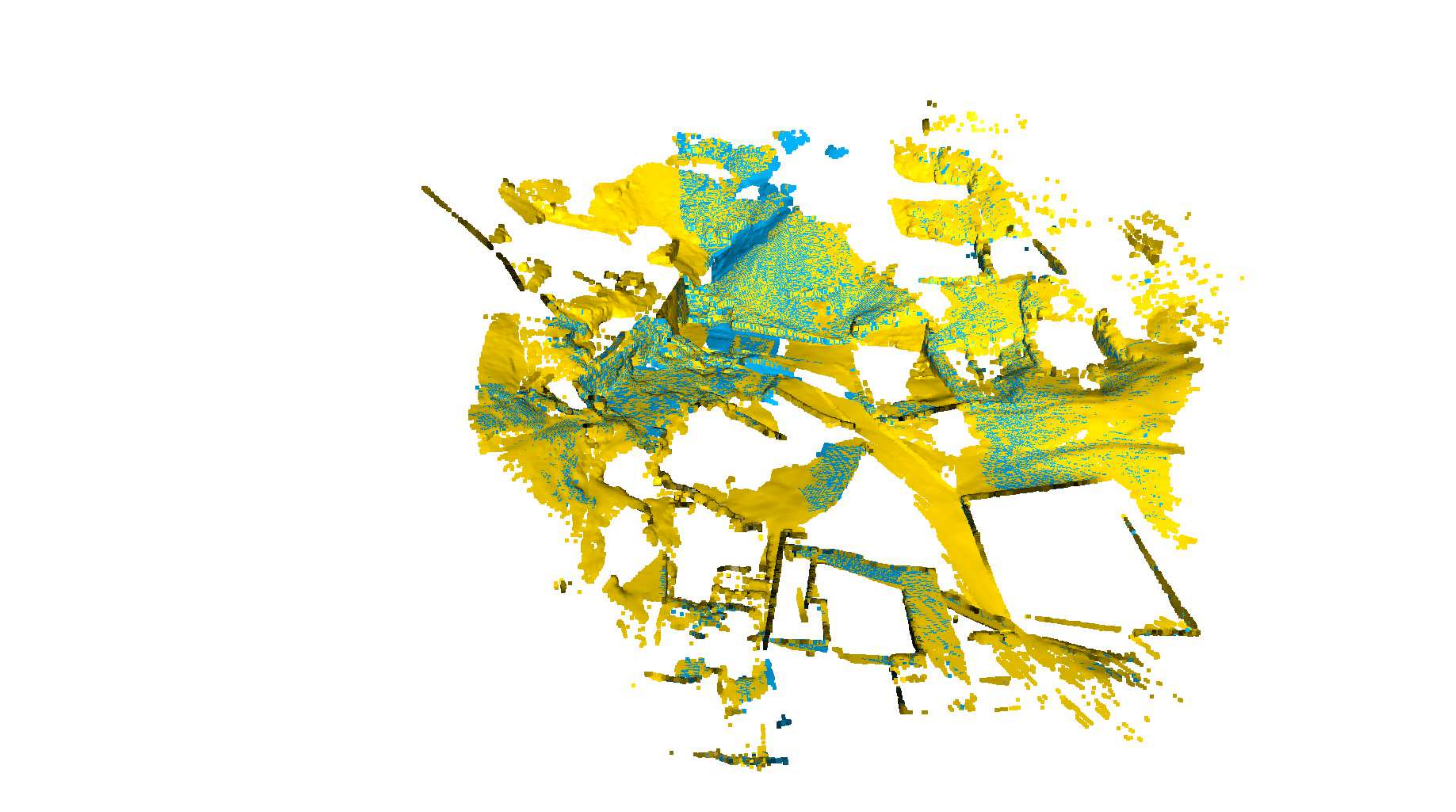}}\hfill
    \subfloat{\includegraphics[trim={200pt 0pt 200pt 0pt}, clip, width=0.16\linewidth]{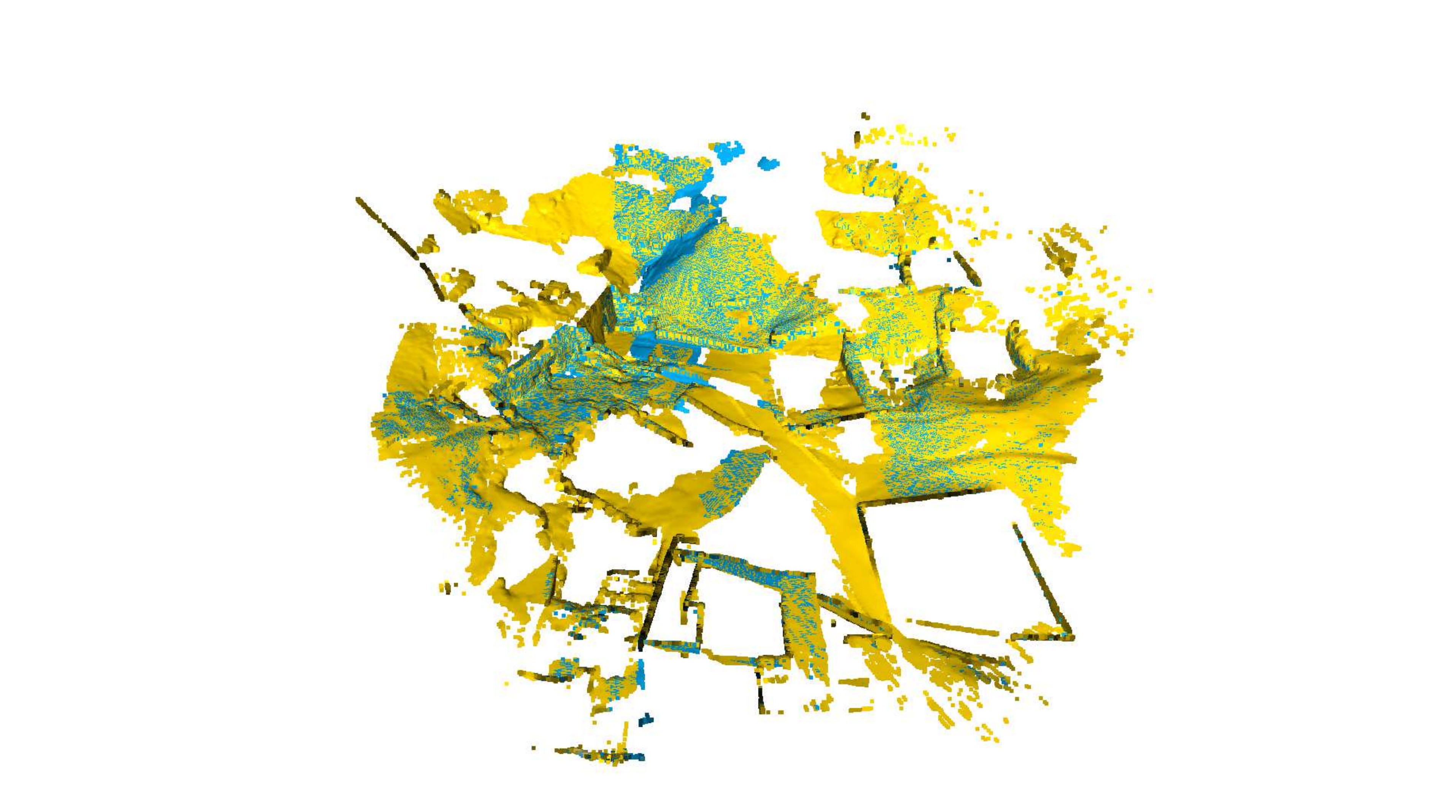}}\hfill
    \\
    \subfloat{\includegraphics[trim={500pt 200pt 300pt 300pt}, clip, width=0.16\linewidth]{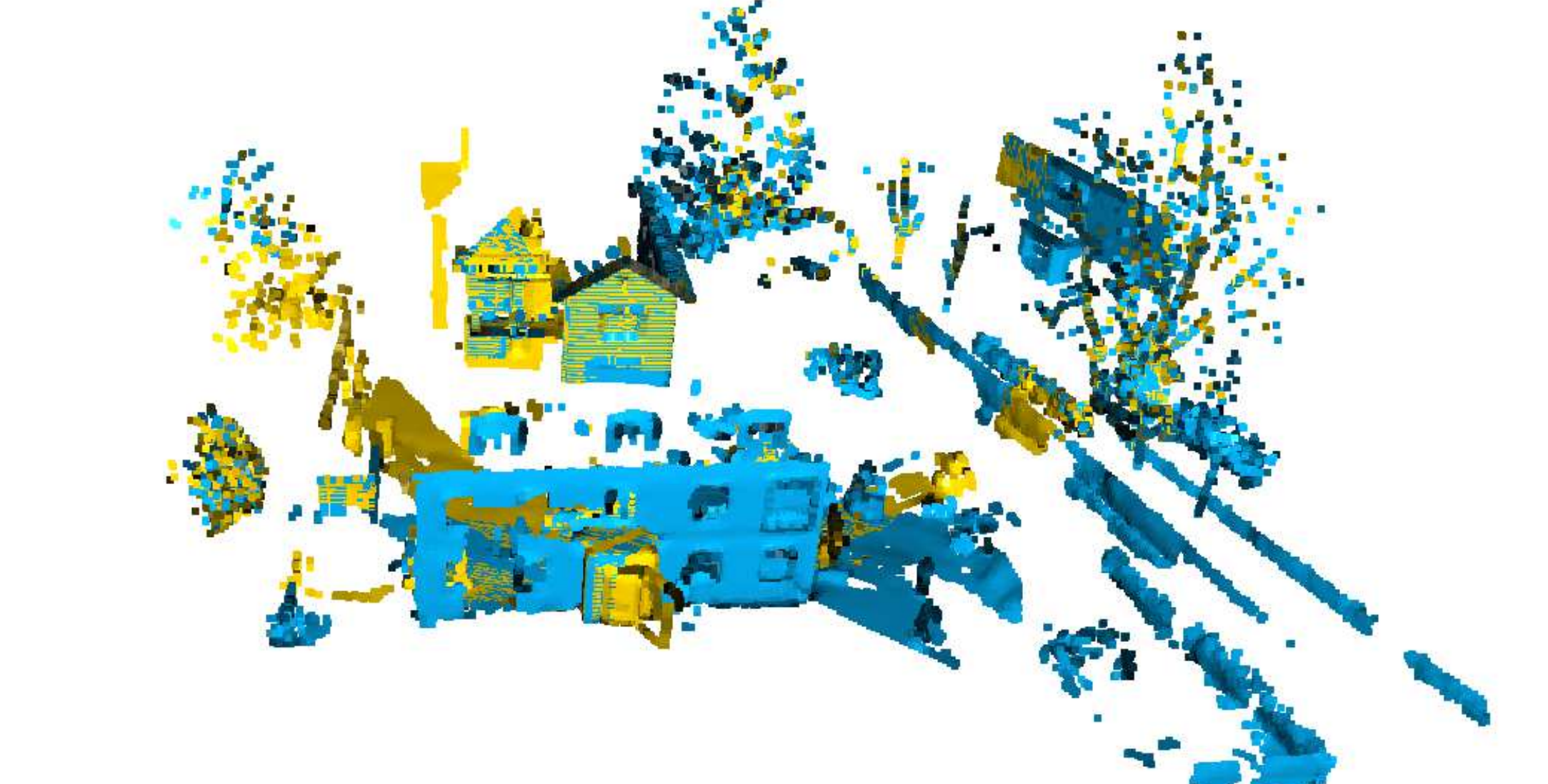}}\hfill
    \subfloat{\includegraphics[trim={500pt 200pt 300pt 300pt}, clip, width=0.16\linewidth]{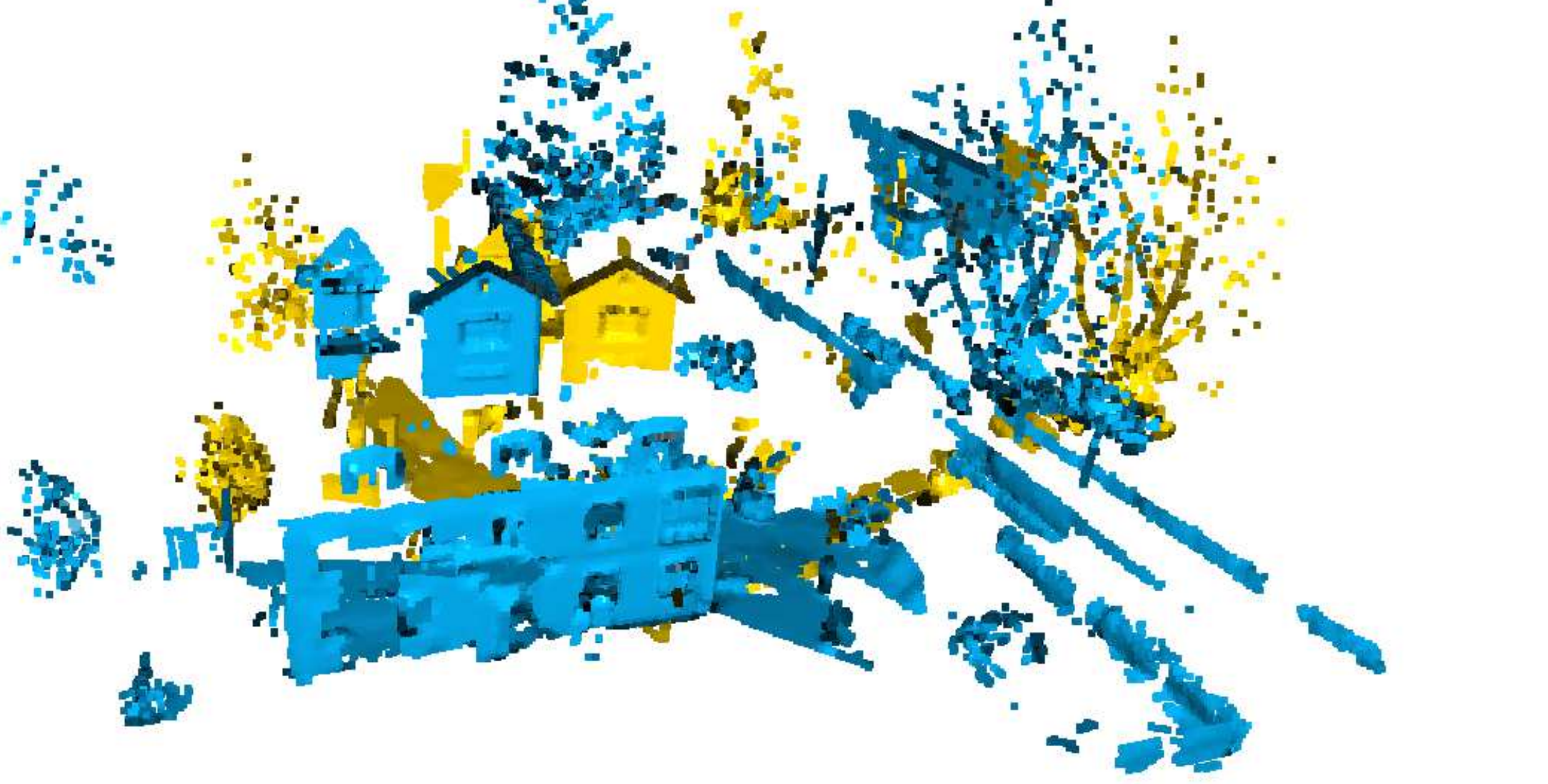}}\hfill
    \subfloat{\includegraphics[trim={500pt 200pt 300pt 300pt}, clip, width=0.16\linewidth]{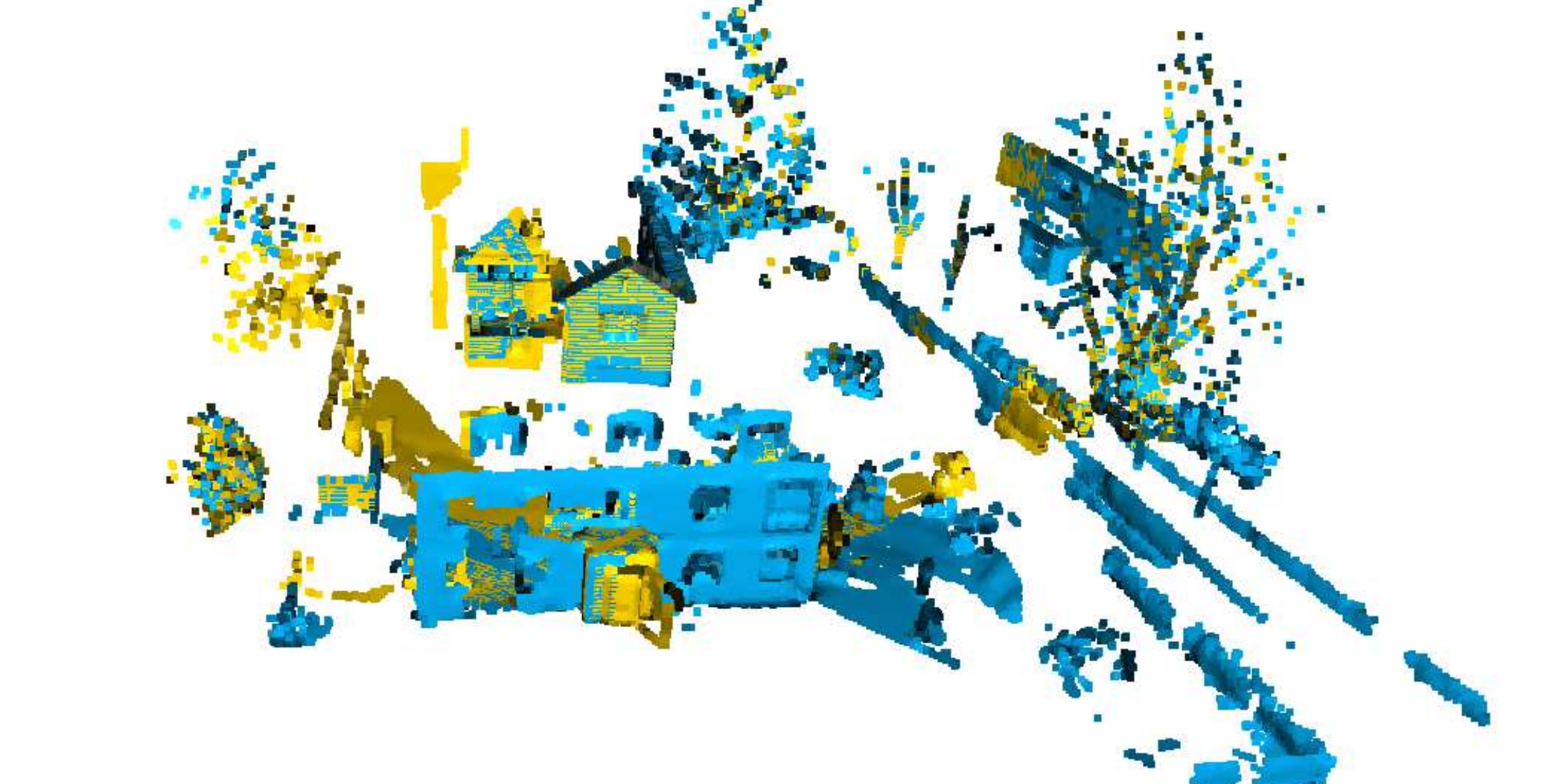}}\hfill
    \subfloat{\includegraphics[trim={500pt 200pt 300pt 300pt}, clip, width=0.16\linewidth]{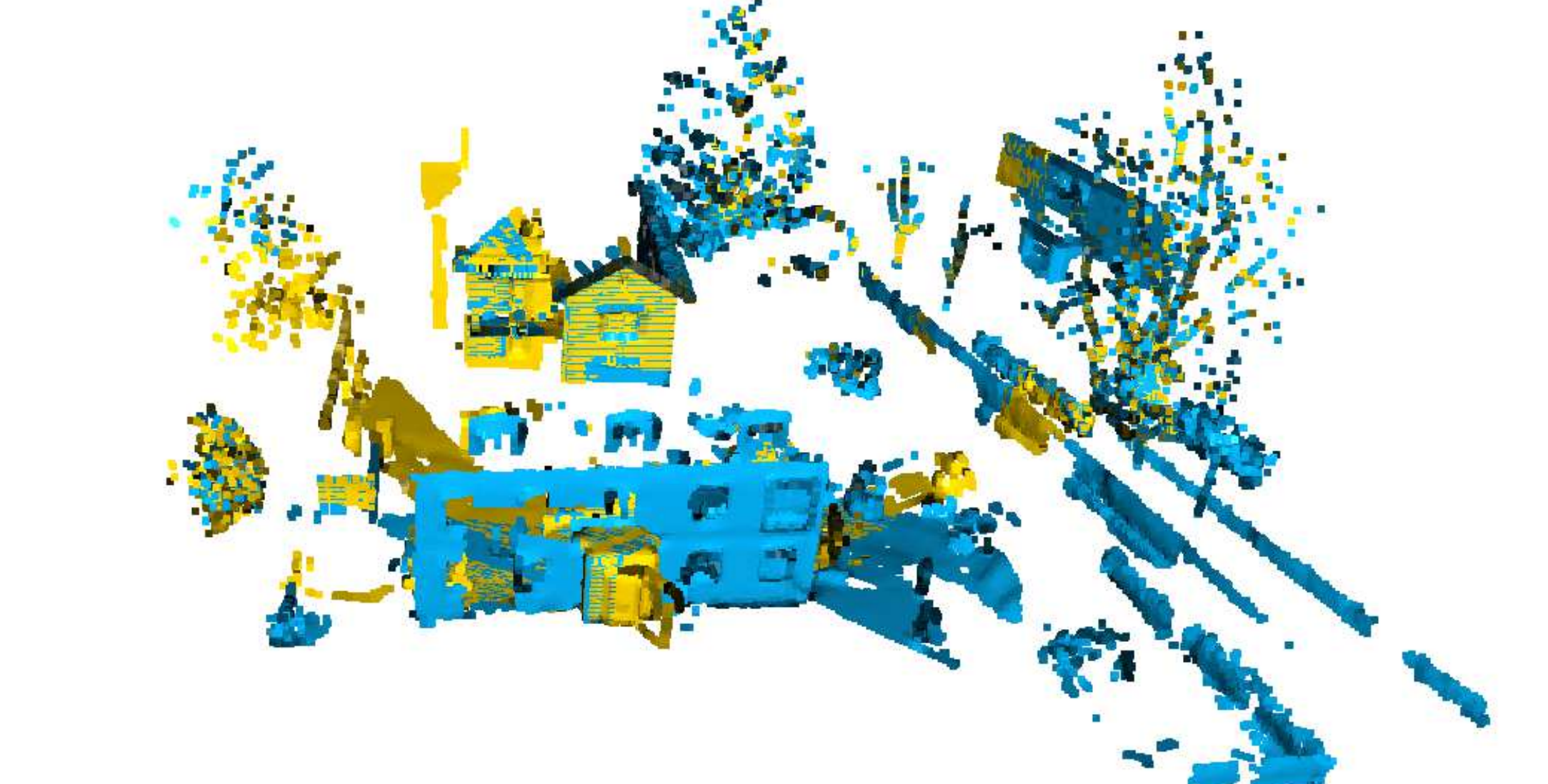}}\hfill
    \subfloat{\includegraphics[trim={500pt 200pt 300pt 300pt}, clip, width=0.16\linewidth]{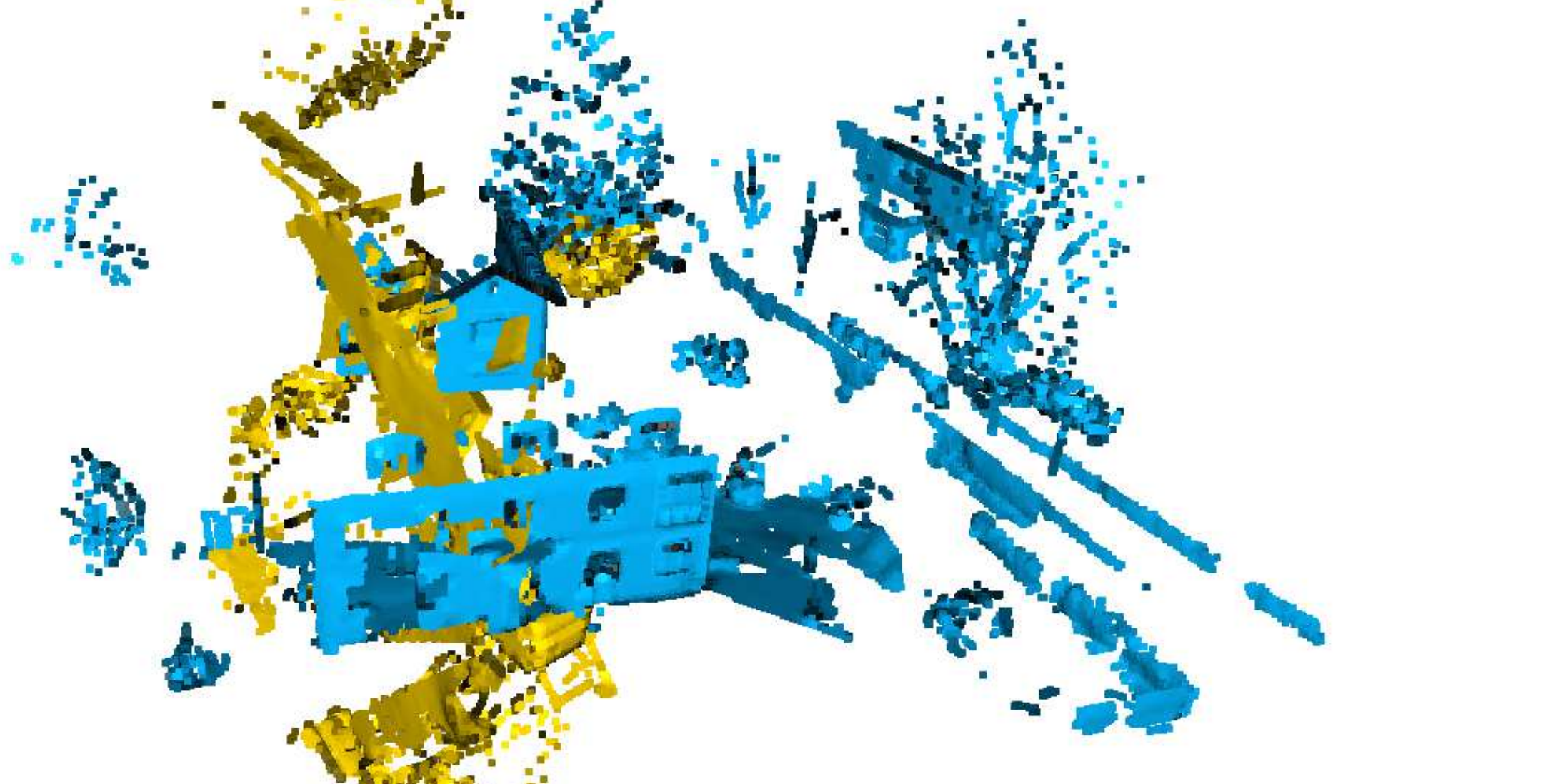}}\hfill
    \subfloat{\includegraphics[trim={500pt 200pt 300pt 300pt}, clip, width=0.16\linewidth]{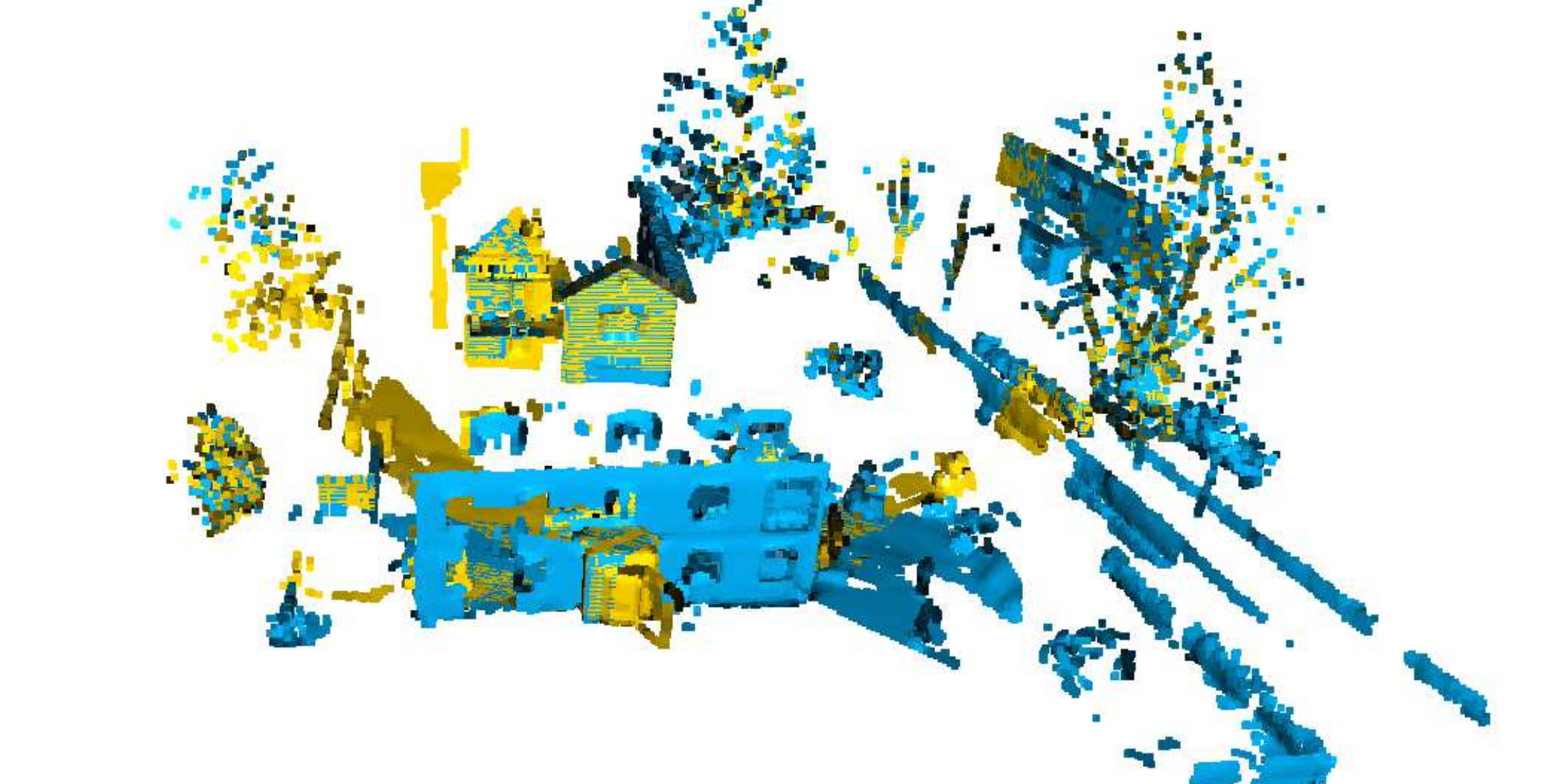}}\hfill
    \\
    \subfloat{\includegraphics[trim={500pt 250pt 300pt 300pt}, clip, width=0.160\linewidth]{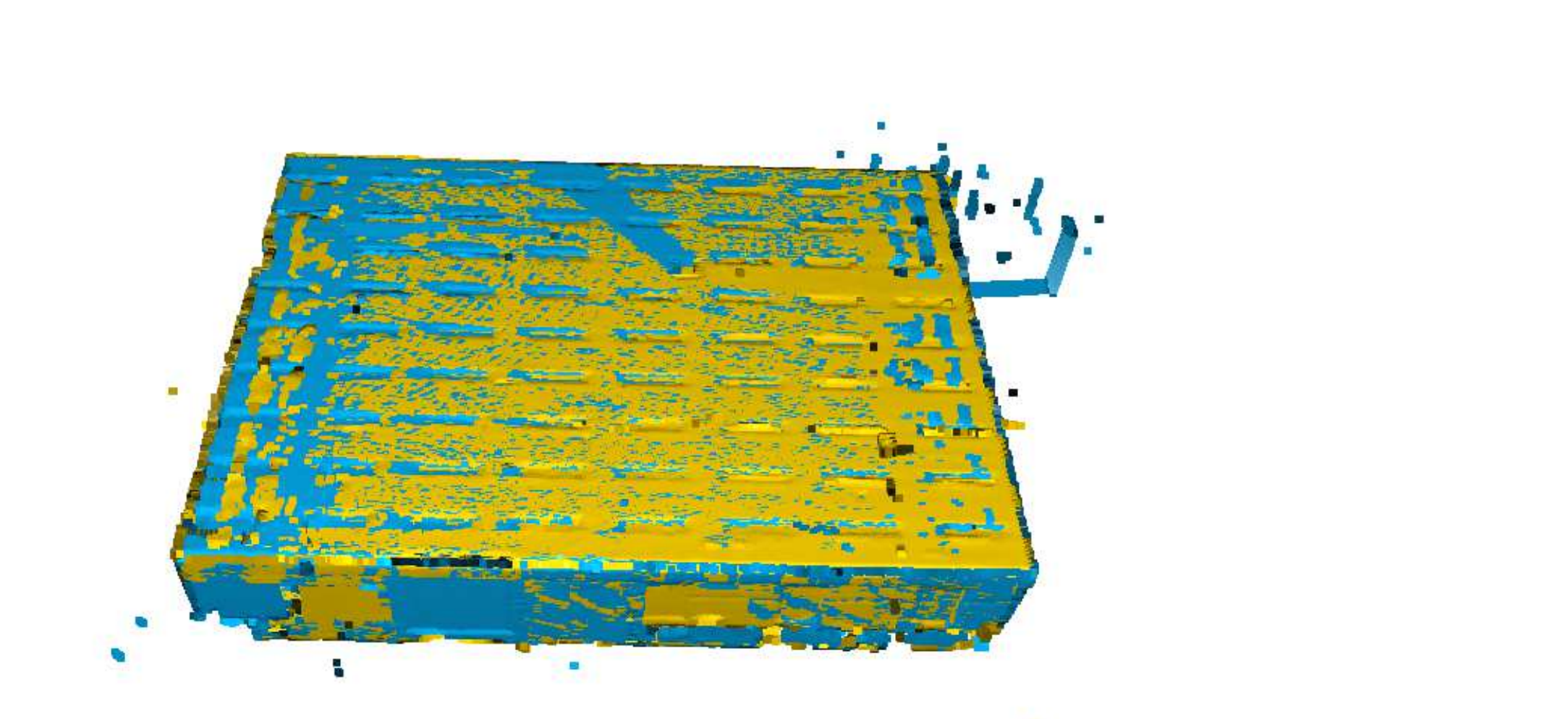}}\hfill
    \subfloat{\includegraphics[trim={500pt 250pt 300pt 300pt}, clip, width=0.160\linewidth]{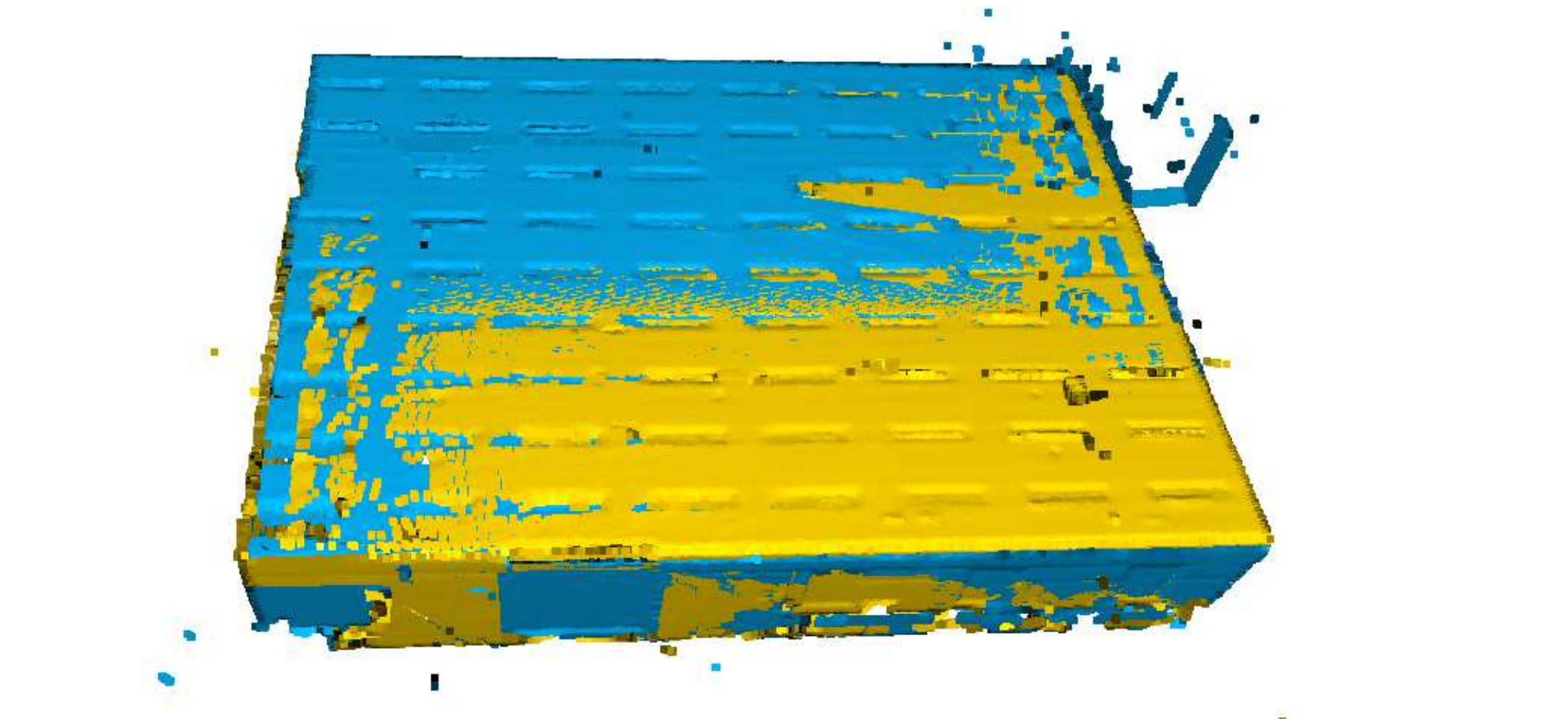}}\hfill
    \subfloat{\includegraphics[trim={500pt 250pt 300pt 300pt}, clip, width=0.160\linewidth]{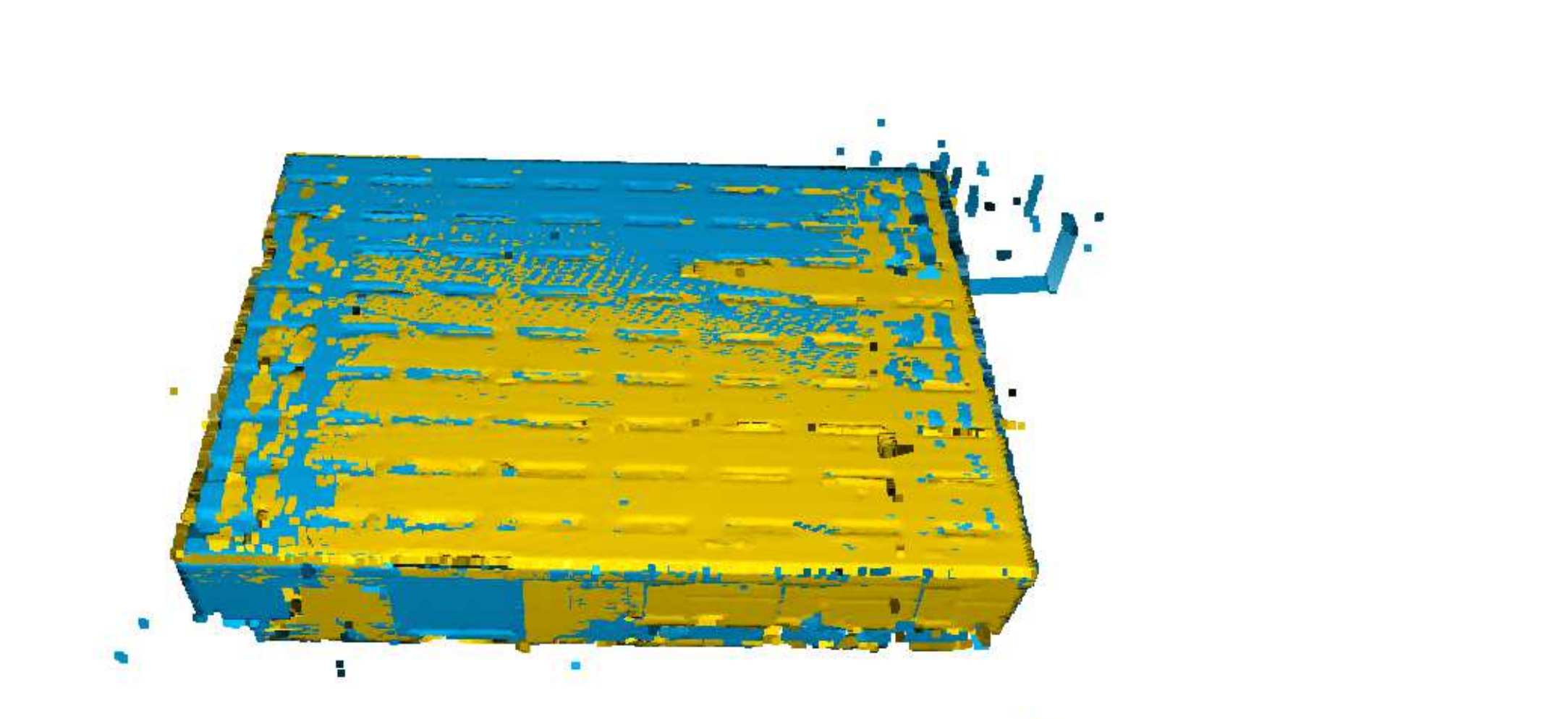}}\hfill
    \subfloat{\includegraphics[trim={500pt 250pt 300pt 300pt}, clip, width=0.160\linewidth]{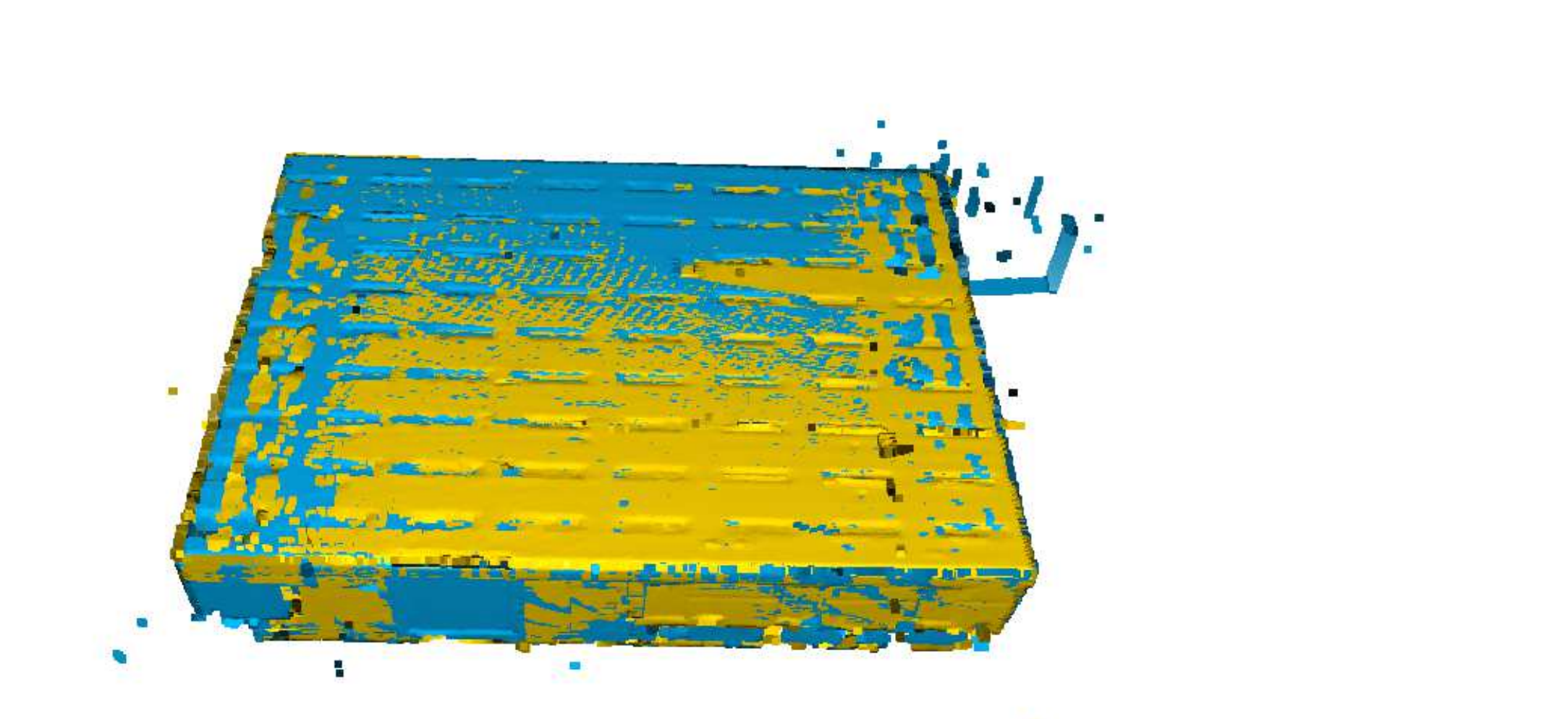}}\hfill
    \subfloat{\includegraphics[trim={500pt 250pt 300pt 300pt}, clip, width=0.160\linewidth]{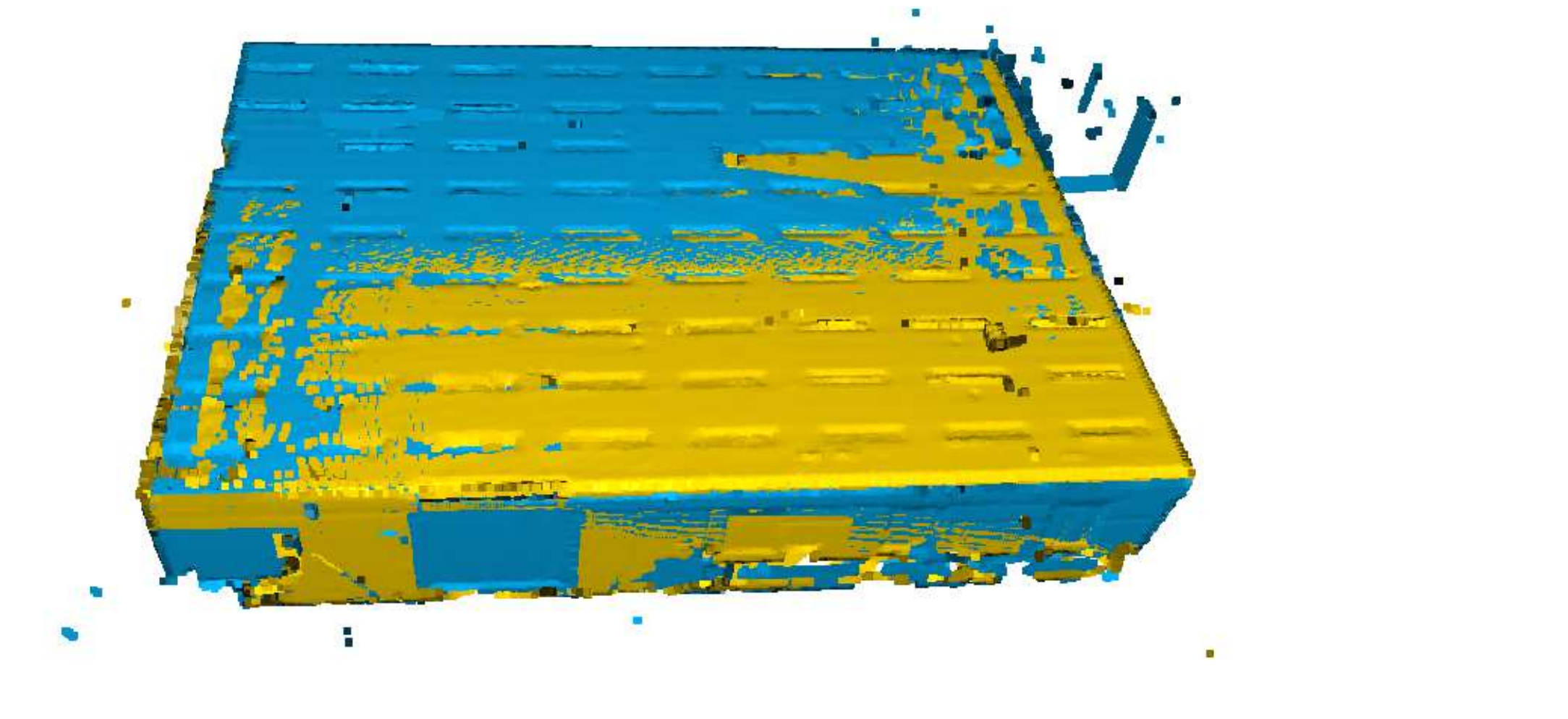}}\hfill
    \subfloat{\includegraphics[trim={500pt 250pt 300pt 300pt}, clip, width=0.160\linewidth]{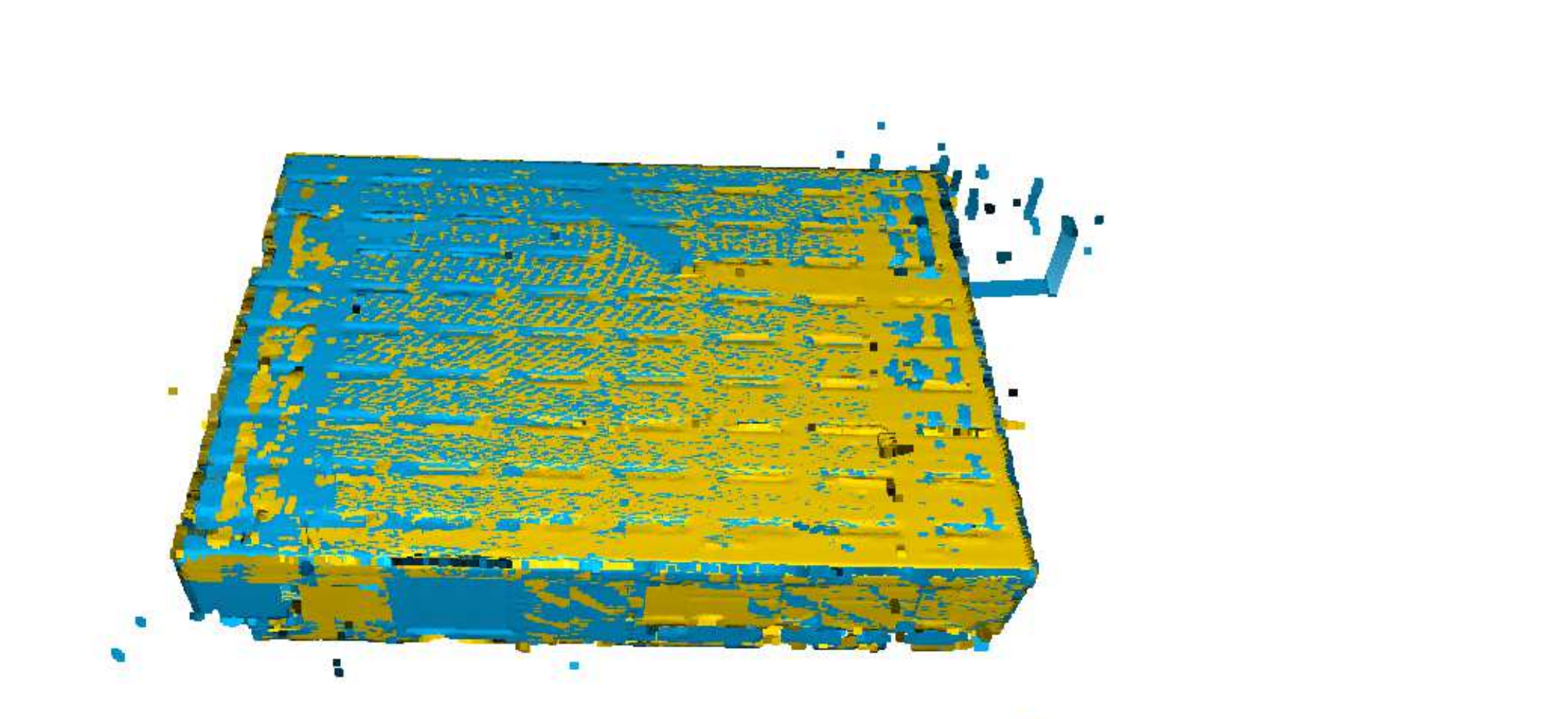}}\hfill
    \\
    \subfloat{\includegraphics[trim={0pt 0pt 0pt 0pt}, clip, width=0.16\linewidth]{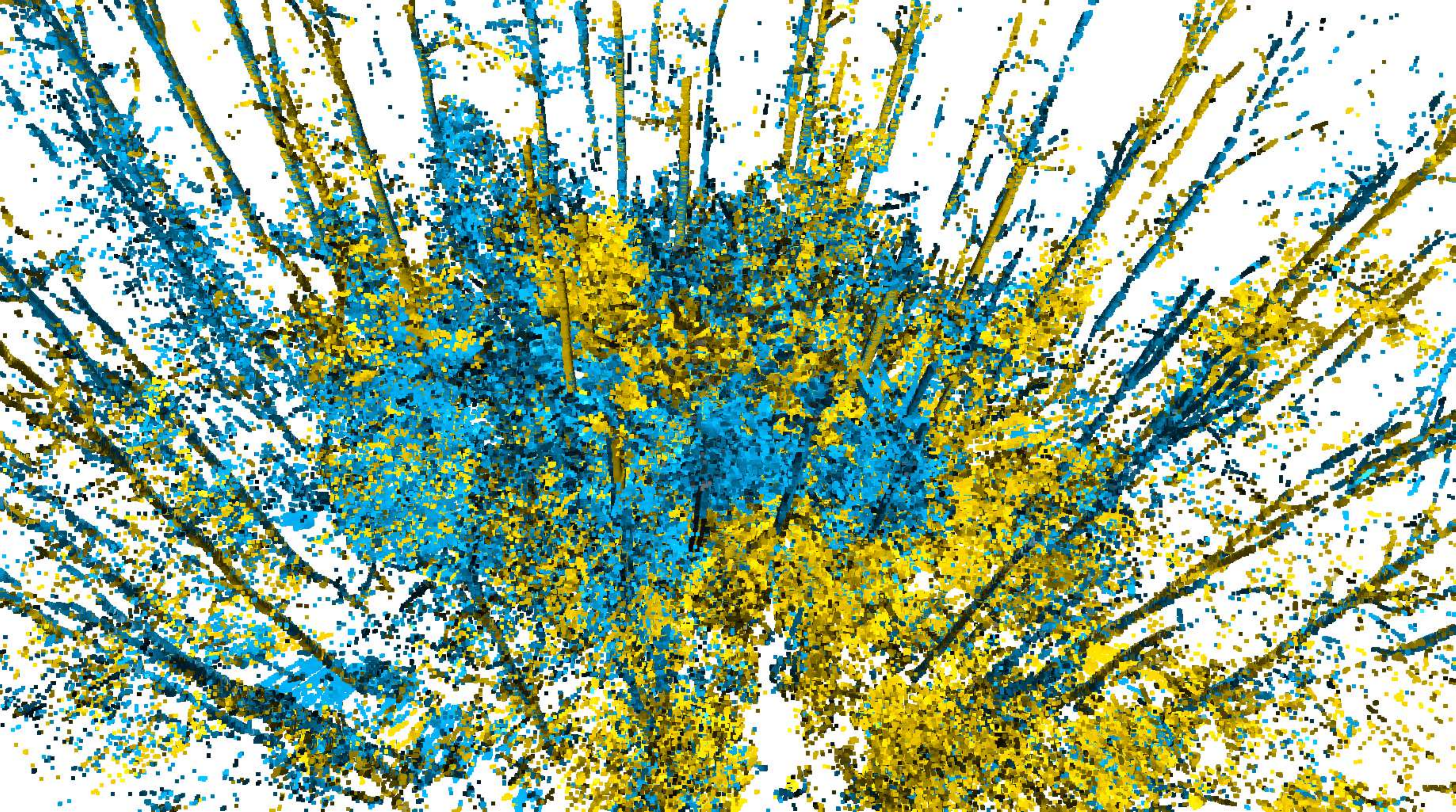}}\hfill
    \subfloat{\includegraphics[trim={0pt 0pt 0pt 0pt}, clip, width=0.16\linewidth]{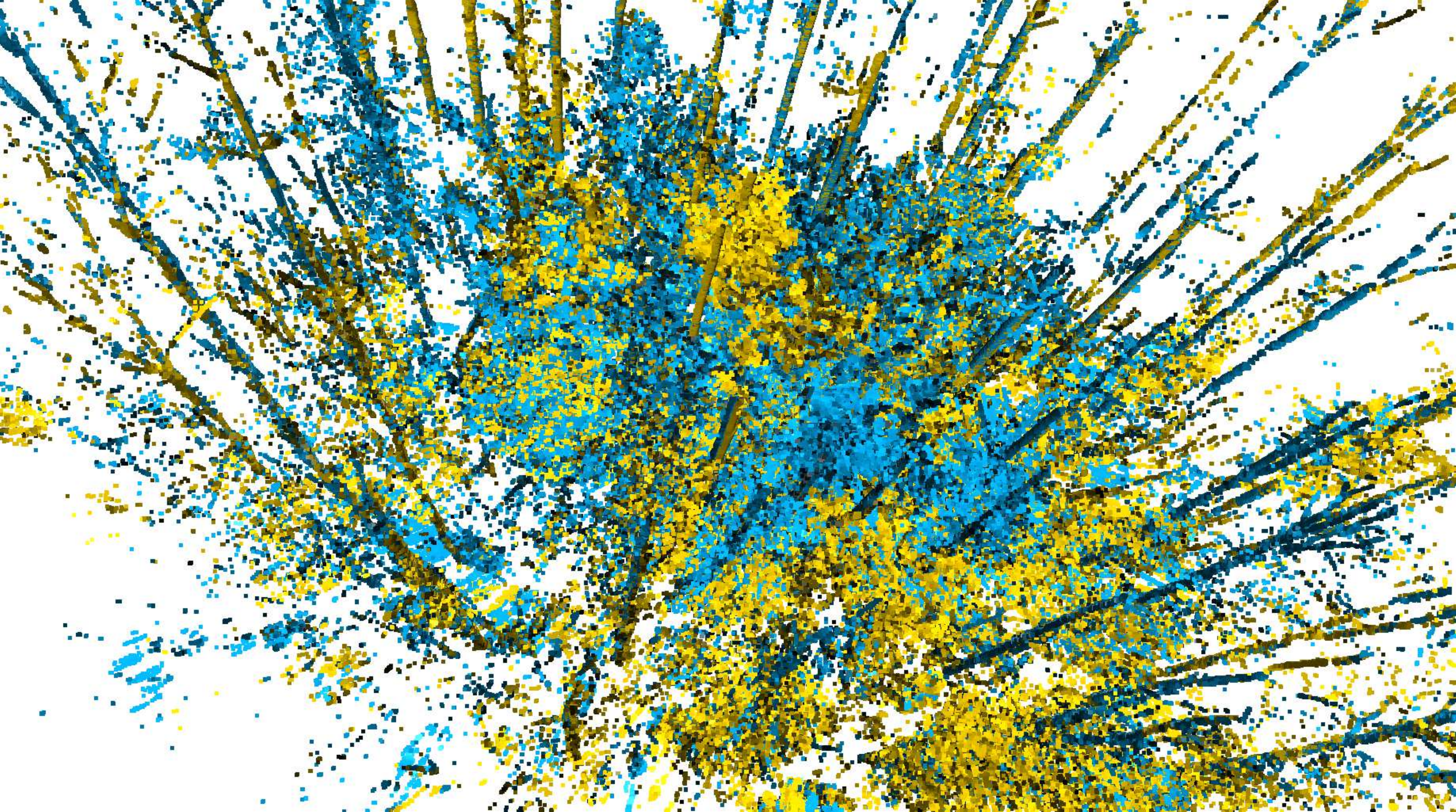}}\hfill
    \subfloat{\includegraphics[trim={0pt 0pt 0pt 0pt}, clip, width=0.16\linewidth]{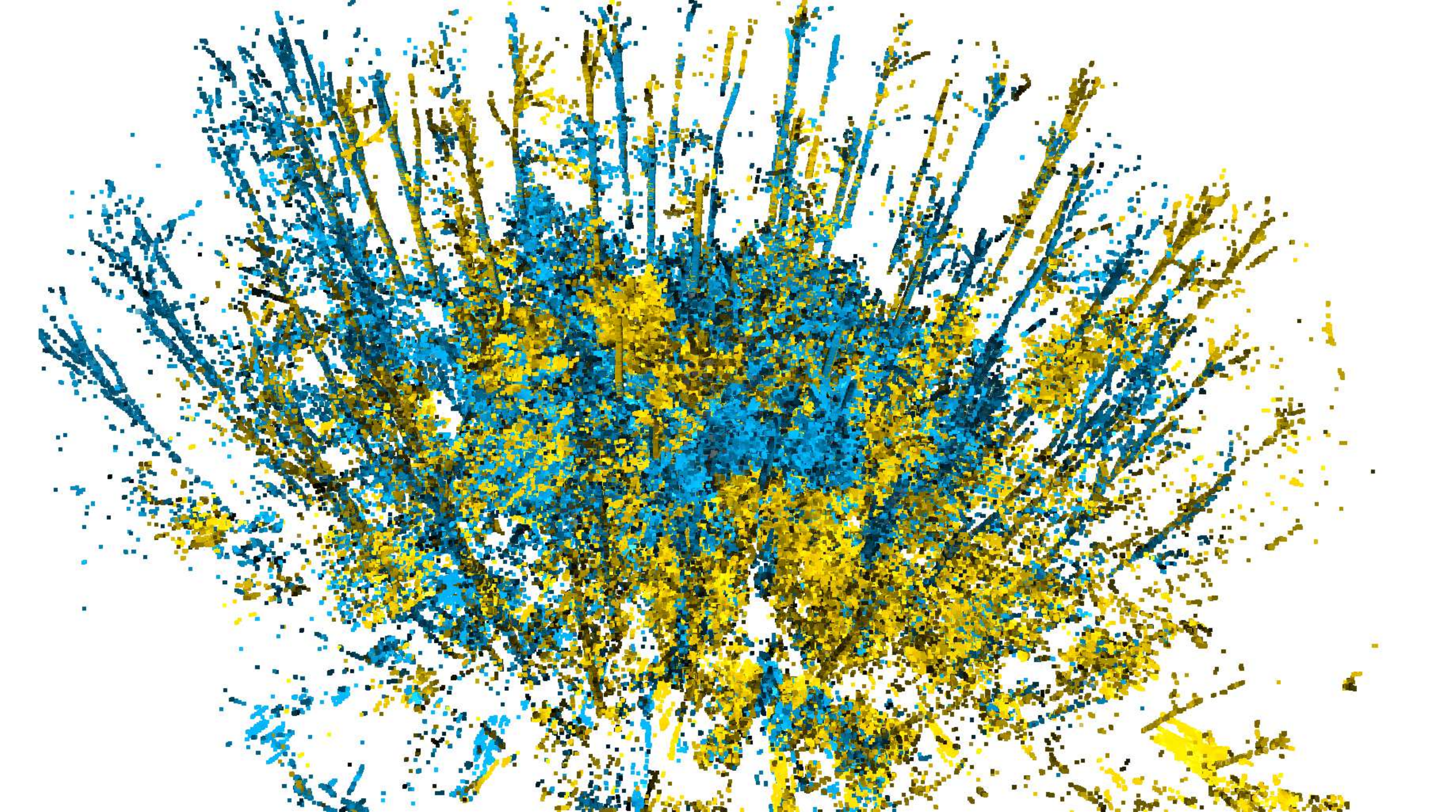}}\hfill
    \subfloat{\includegraphics[trim={0pt 0pt 0pt 0pt}, clip, width=0.16\linewidth]{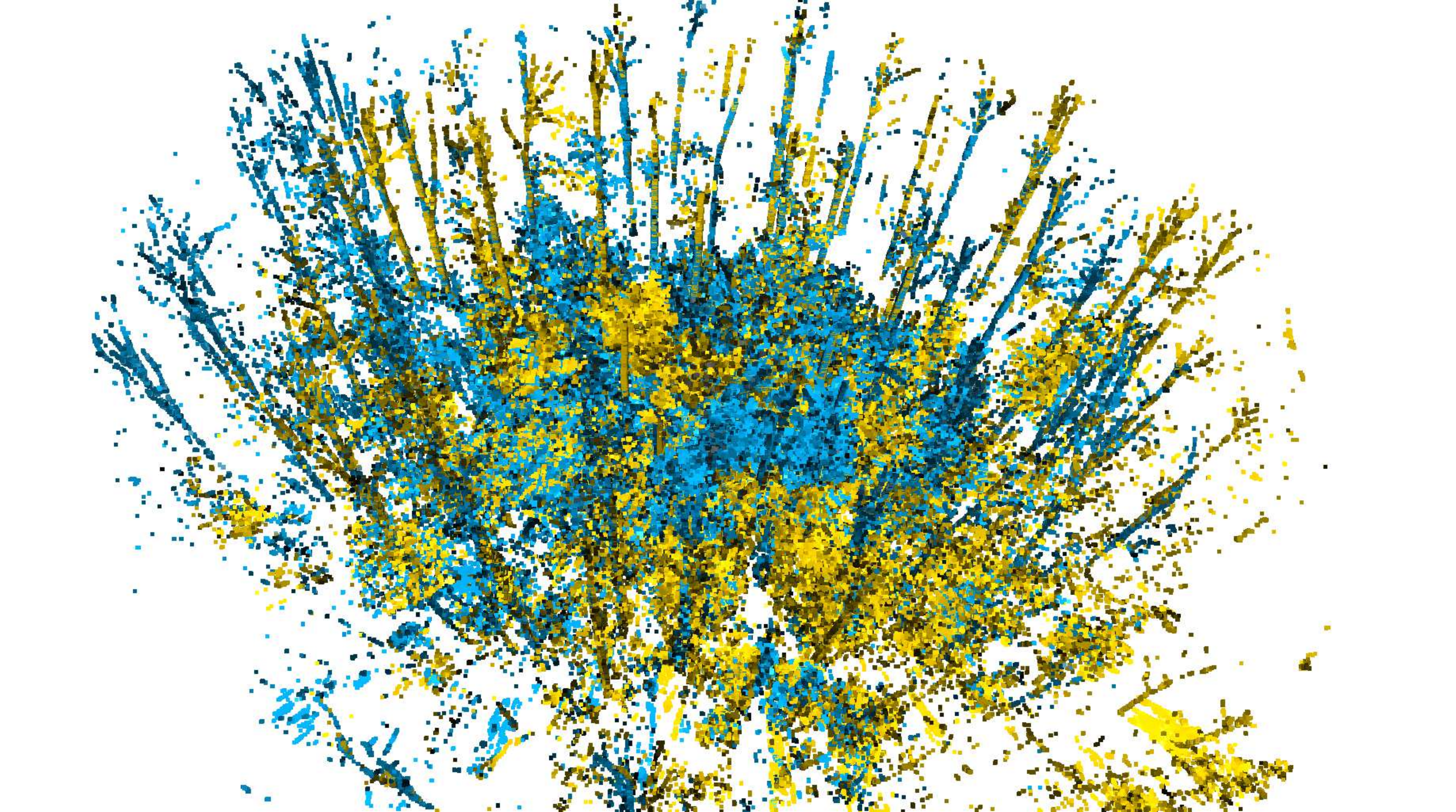}}\hfill
    \subfloat{\includegraphics[trim={0pt 0pt 0pt 0pt}, clip, width=0.16\linewidth]{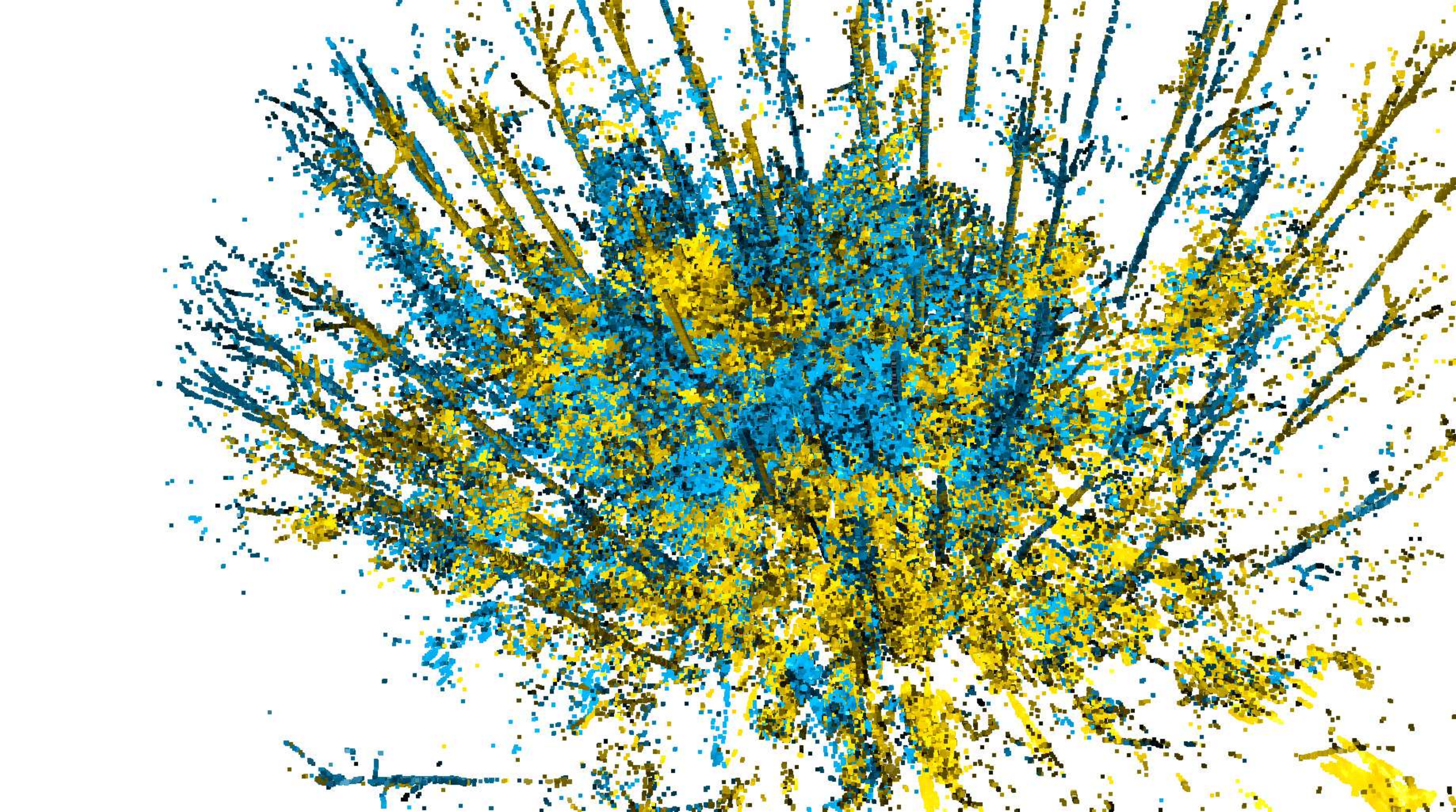}}\hfill
    \subfloat{\includegraphics[trim={0pt 0pt 0pt 0pt}, clip, width=0.16\linewidth]{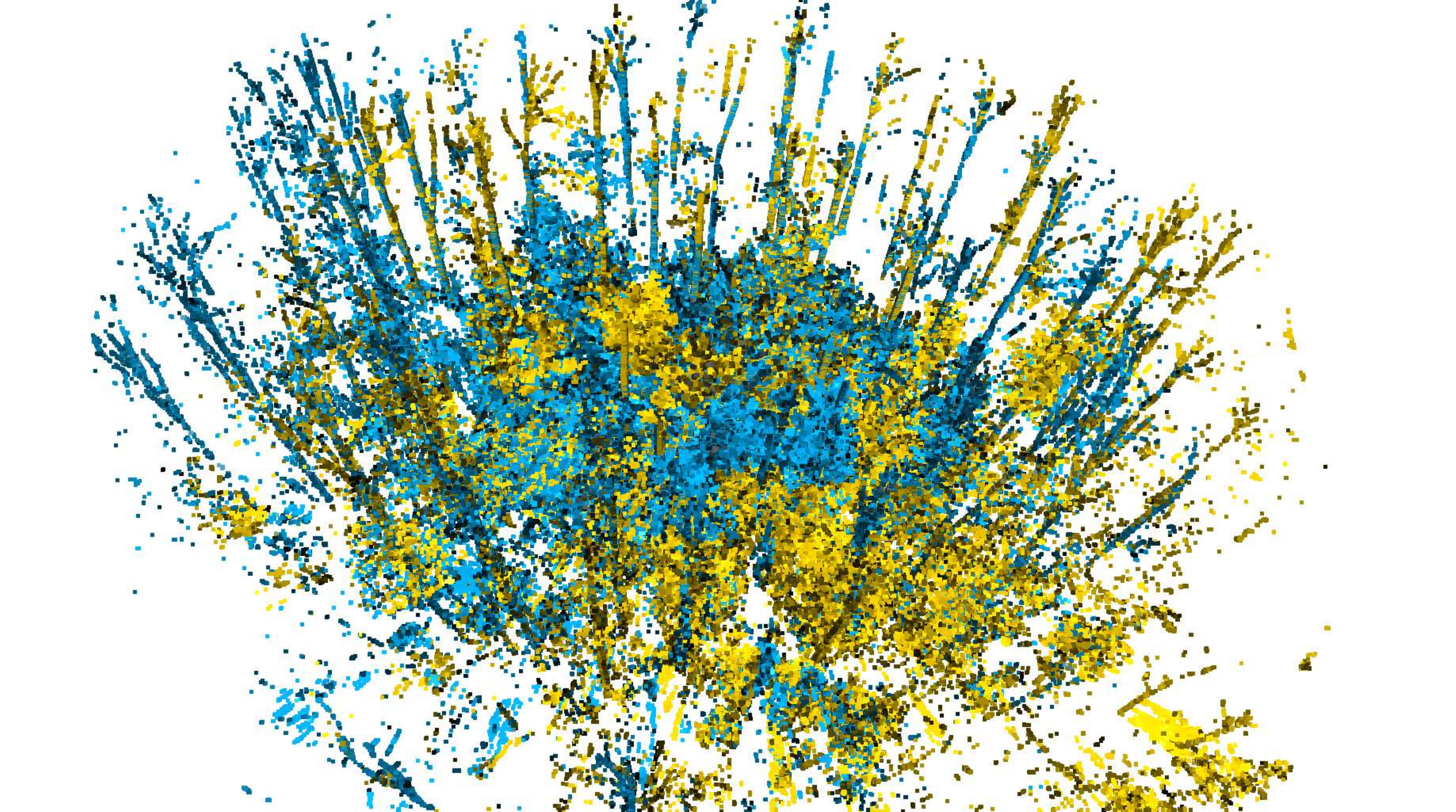}}\hfill

    \caption{Qualitative comparison on ETH. The scenes are listed in order from top to bottom as follows: arch, courtyard, facade, office, and trees.}
    \label{fig:vis_ETH}
\end{figure*}

\begin{figure*}[!h]
    \centering
    \vspace{2em}
    \begin{minipage}{.160\linewidth}
        \centering
        GT
    \end{minipage}
    \begin{minipage}{.160\linewidth}
        \centering
        MAC
    \end{minipage}
    \begin{minipage}{.160\linewidth}
        \centering
        GROR
    \end{minipage}
    \begin{minipage}{.160\linewidth}
        \centering
        Ours(CR)
    \end{minipage}
    \begin{minipage}{.160\linewidth}
        \centering
         MAC+GICP
    \end{minipage}
    \begin{minipage}{.160\linewidth}
        \centering
        Ours(CR+FR)
    \end{minipage}

    \subfloat{\includegraphics[trim={450pt 0pt 200pt 100pt},clip, width=0.16\linewidth]{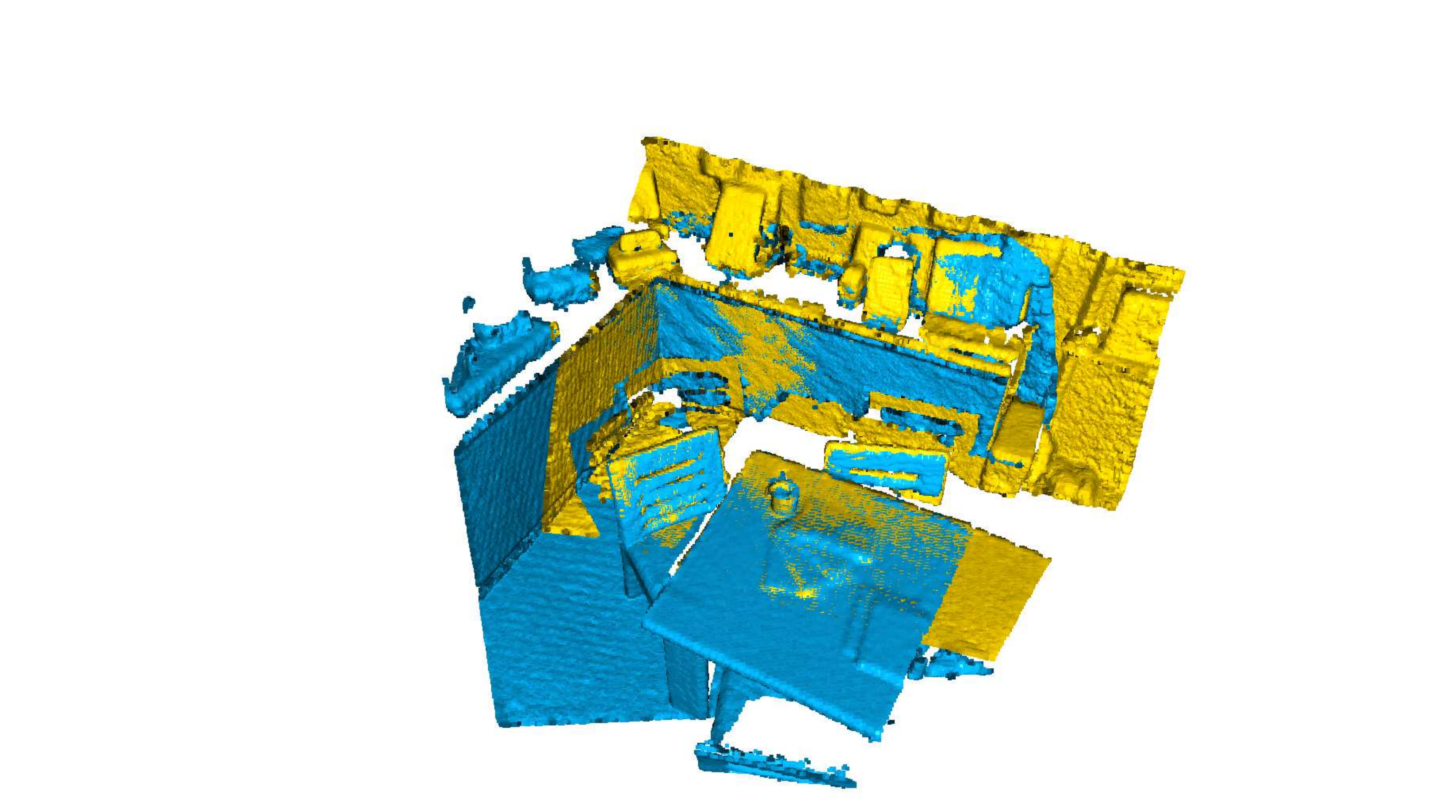}}\hfill
    \subfloat{\includegraphics[trim={450pt 0pt 200pt 100pt}, clip, width=0.16\linewidth]{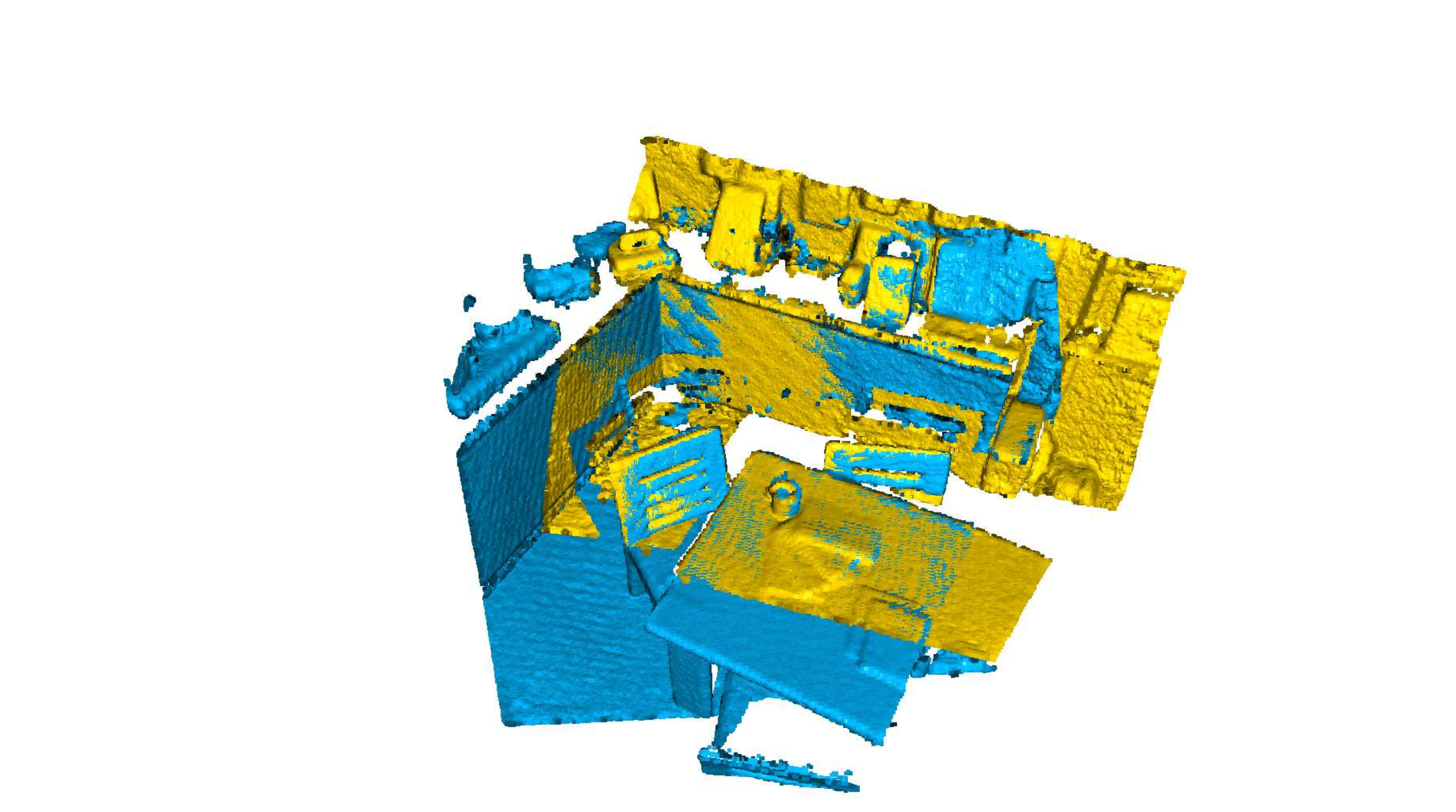}}\hfill
    \subfloat{\includegraphics[trim={450pt 0pt 200pt 100pt}, clip, width=0.16\linewidth]{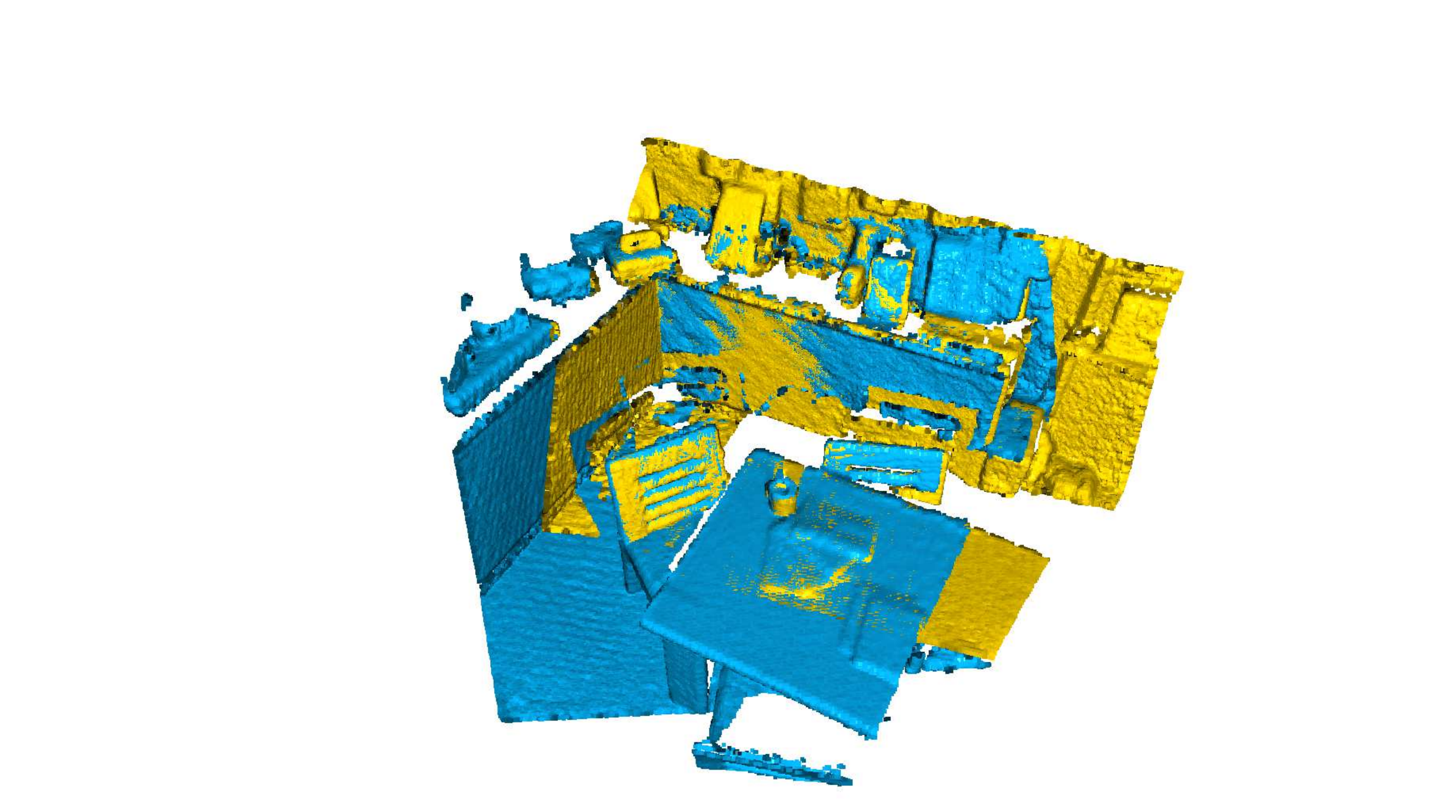}}\hfill
    \subfloat{\includegraphics[trim={450pt 0pt 200pt 100pt}, clip, width=0.16\linewidth]{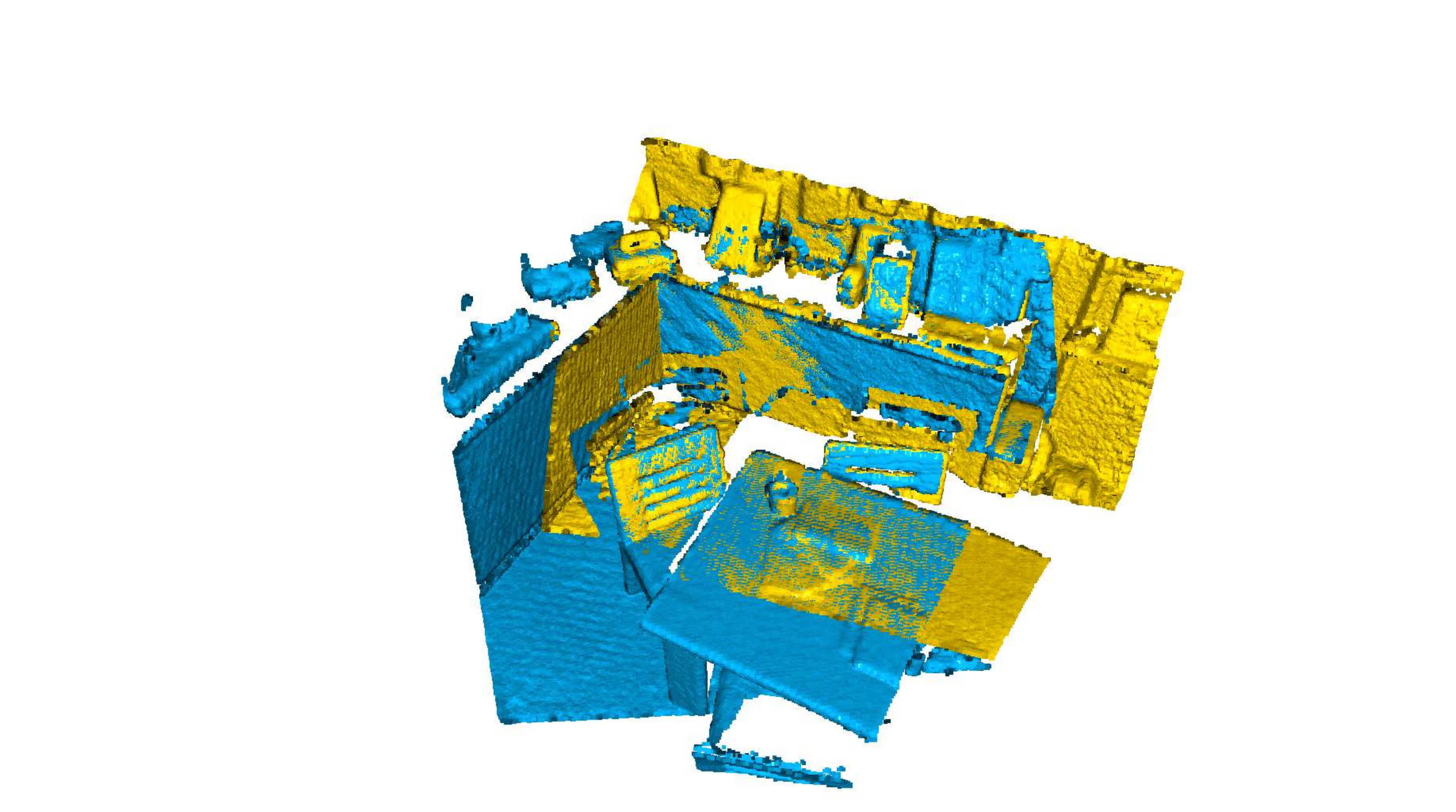}}\hfill
    \subfloat{\includegraphics[trim={450pt 0pt 200pt 100pt}, clip, width=0.16\linewidth]{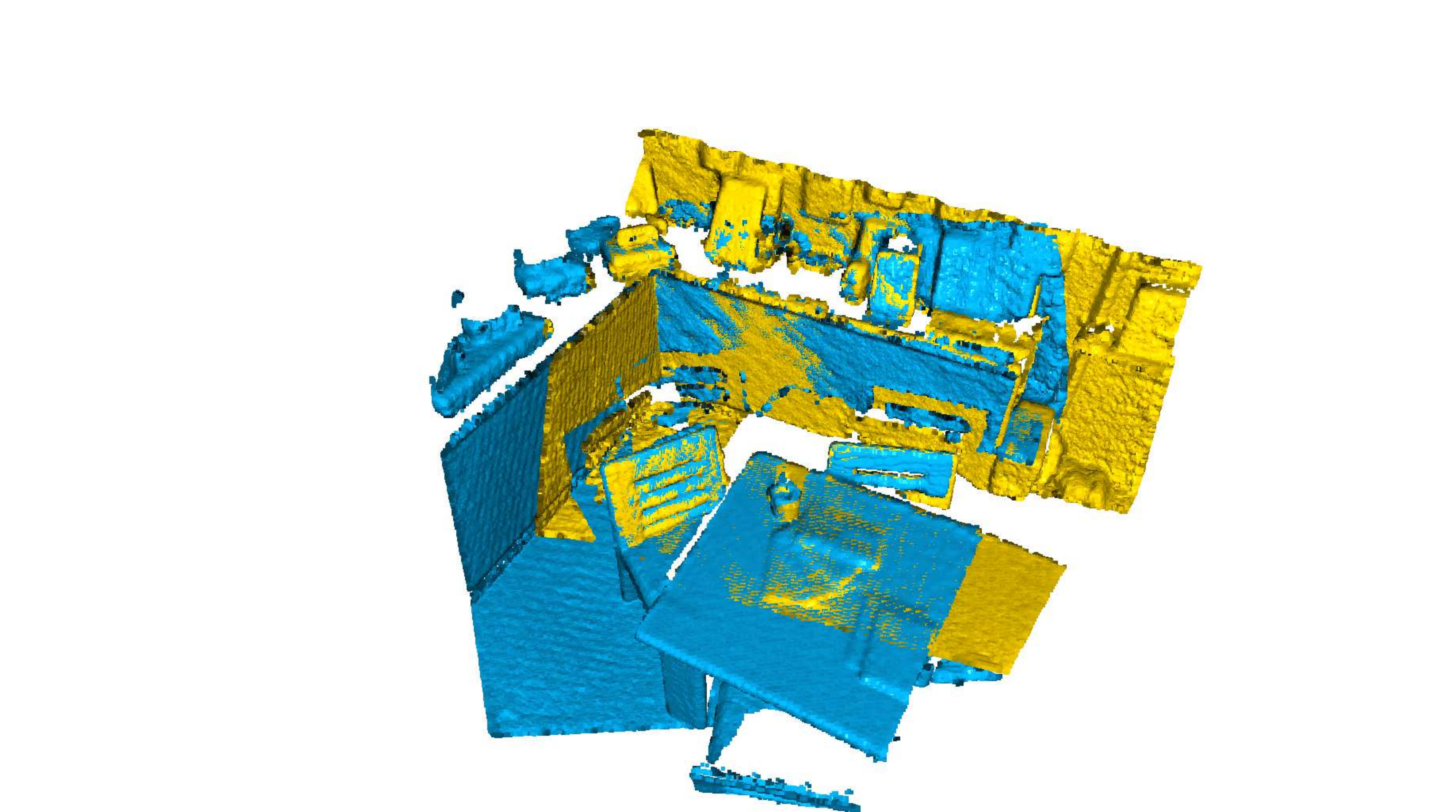}}\hfill
    \subfloat{\includegraphics[trim={450pt 0pt 200pt 100pt}, clip, width=0.16\linewidth]{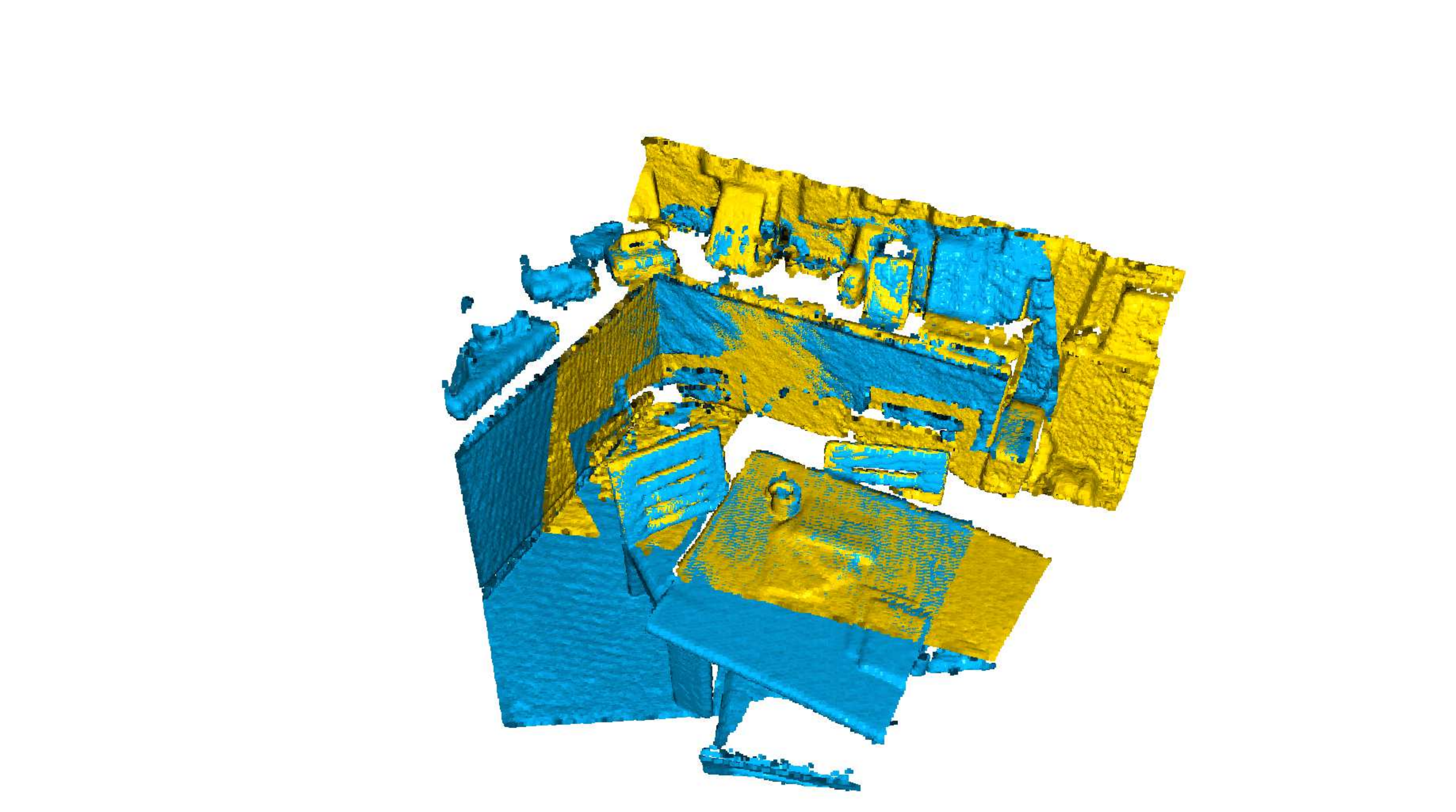}}\hfill
    \\
    \subfloat{\includegraphics[trim={500pt 300pt 500pt 100pt}, clip, width=0.16\linewidth]{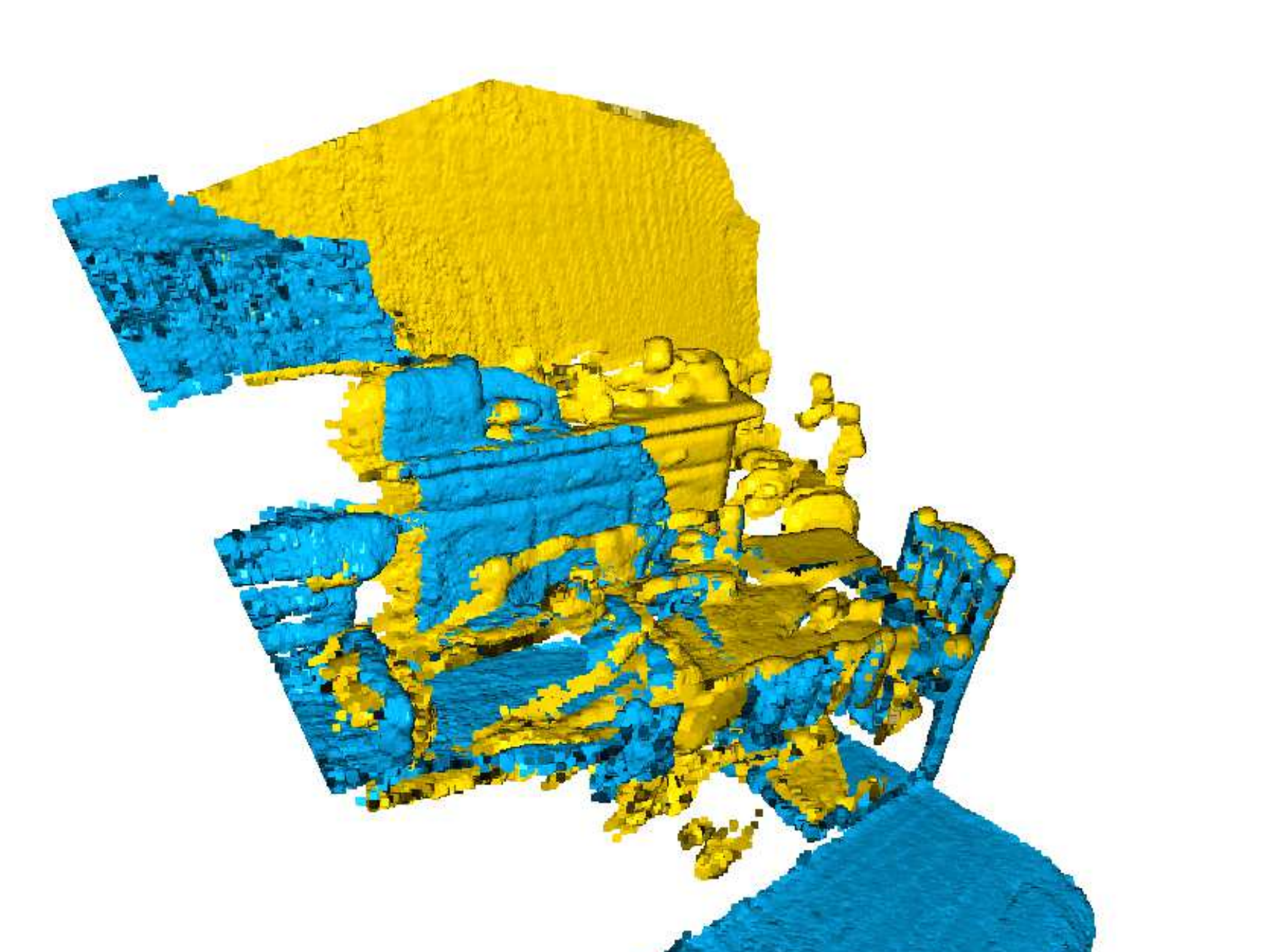}}\hfill
    \subfloat{\includegraphics[trim={500pt 300pt 500pt 100pt}, clip,width=0.16\linewidth]{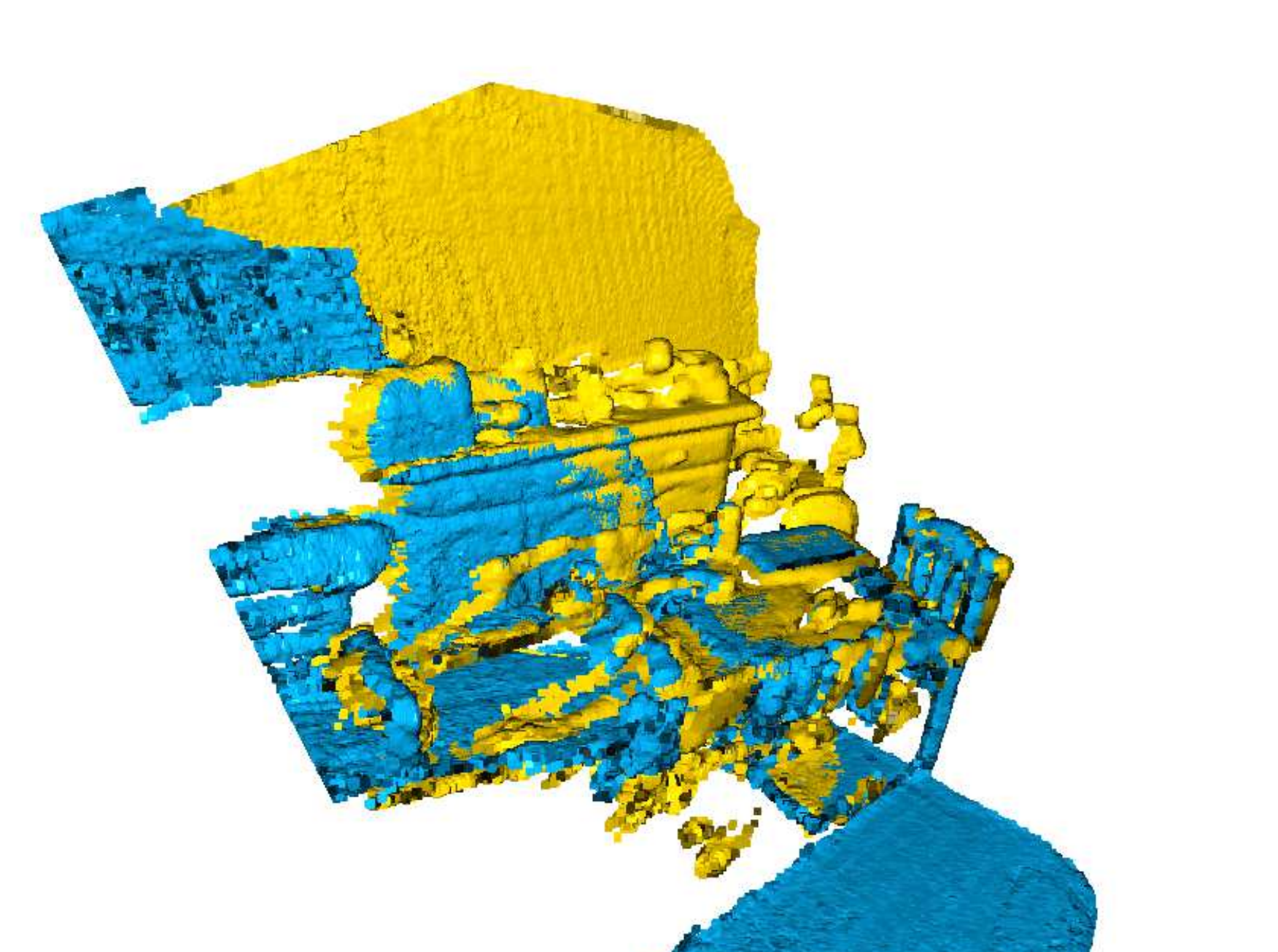}}\hfill
    \subfloat{\includegraphics[trim={500pt 300pt 500pt 100pt}, clip, width=0.16\linewidth]{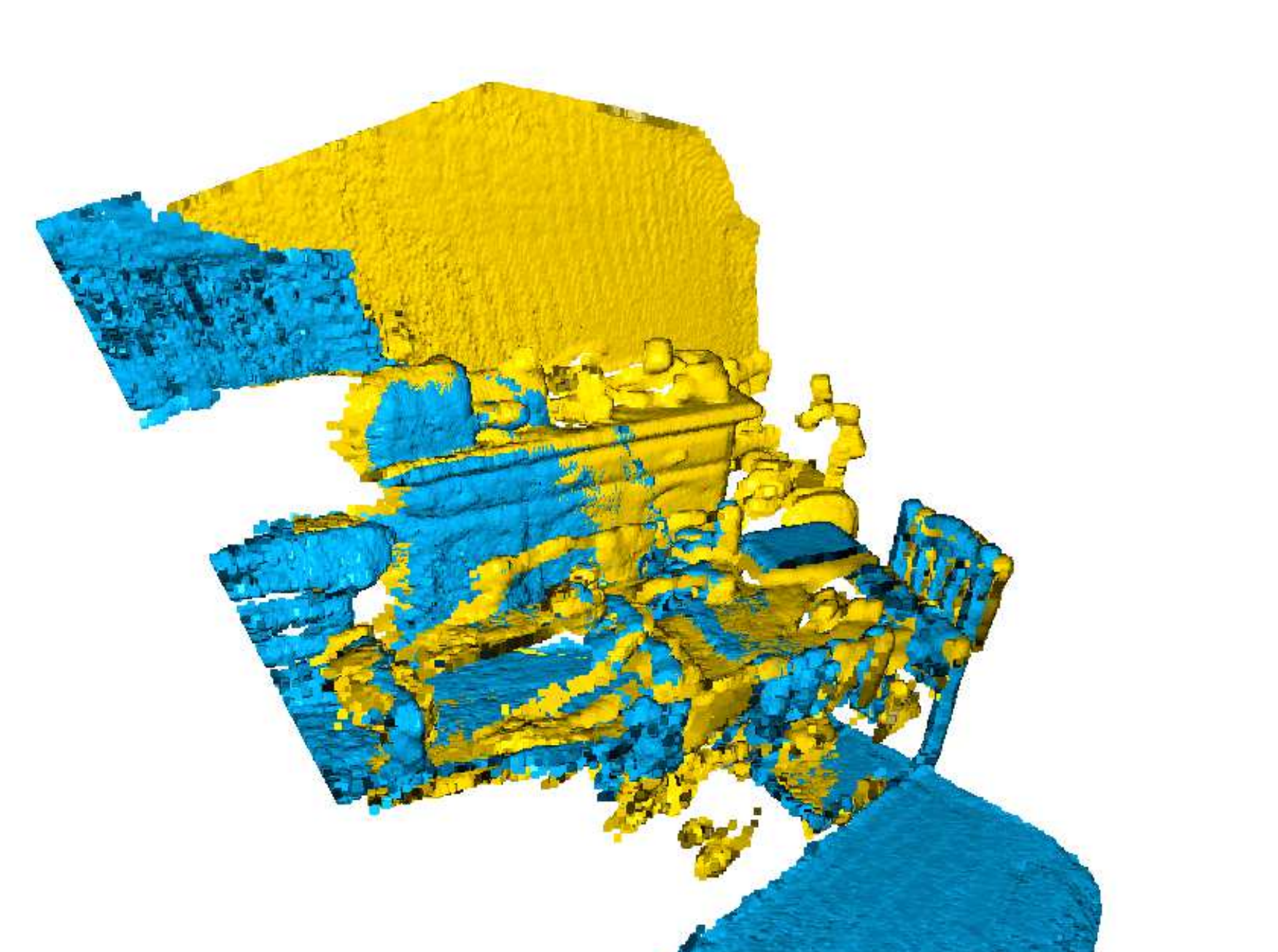}}\hfill
    \subfloat{\includegraphics[trim={500pt 300pt 500pt 100pt}, clip, width=0.16\linewidth]{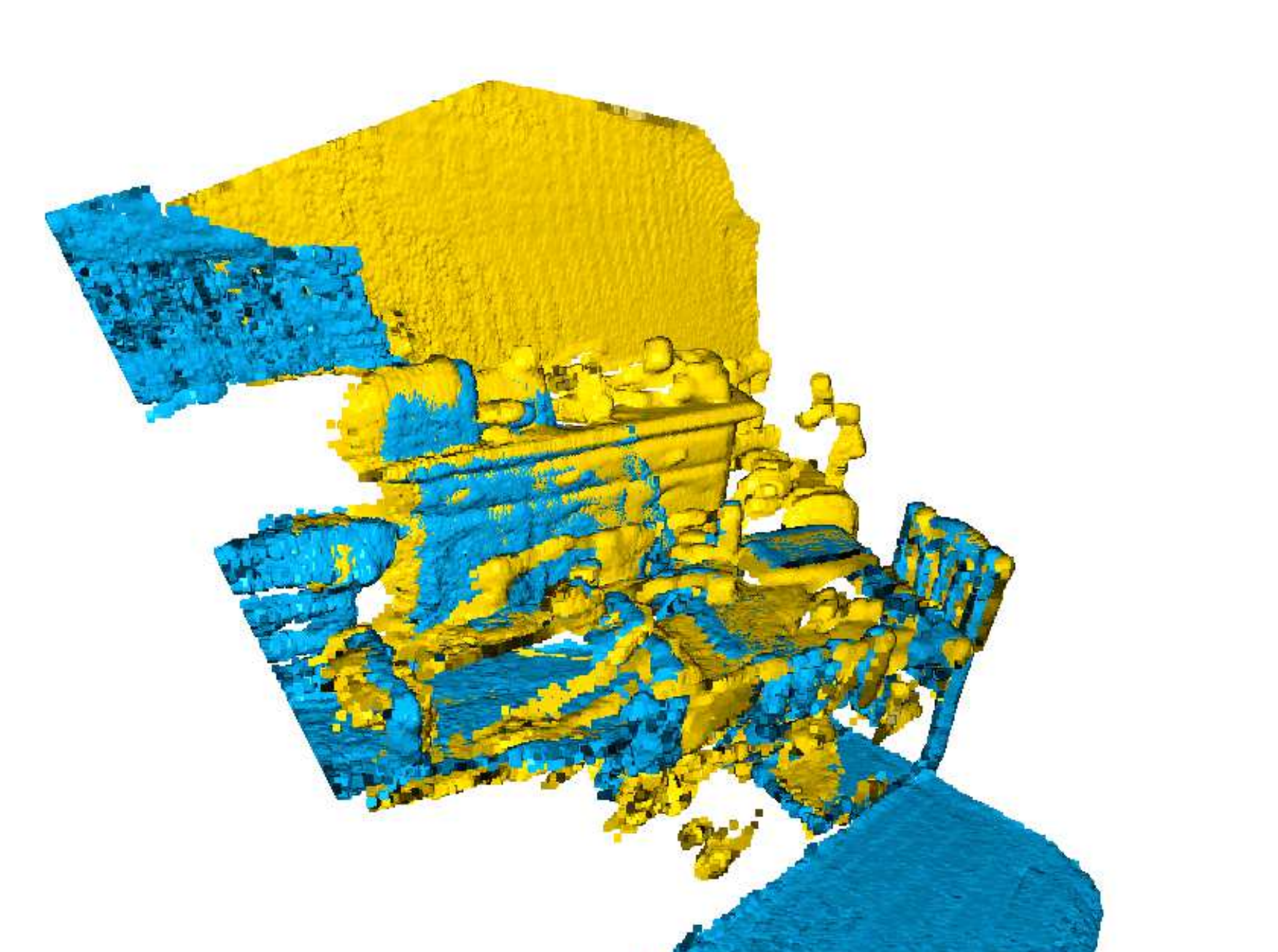}}\hfill
    \subfloat{\includegraphics[trim={500pt 300pt 500pt 100pt}, clip, width=0.16\linewidth]{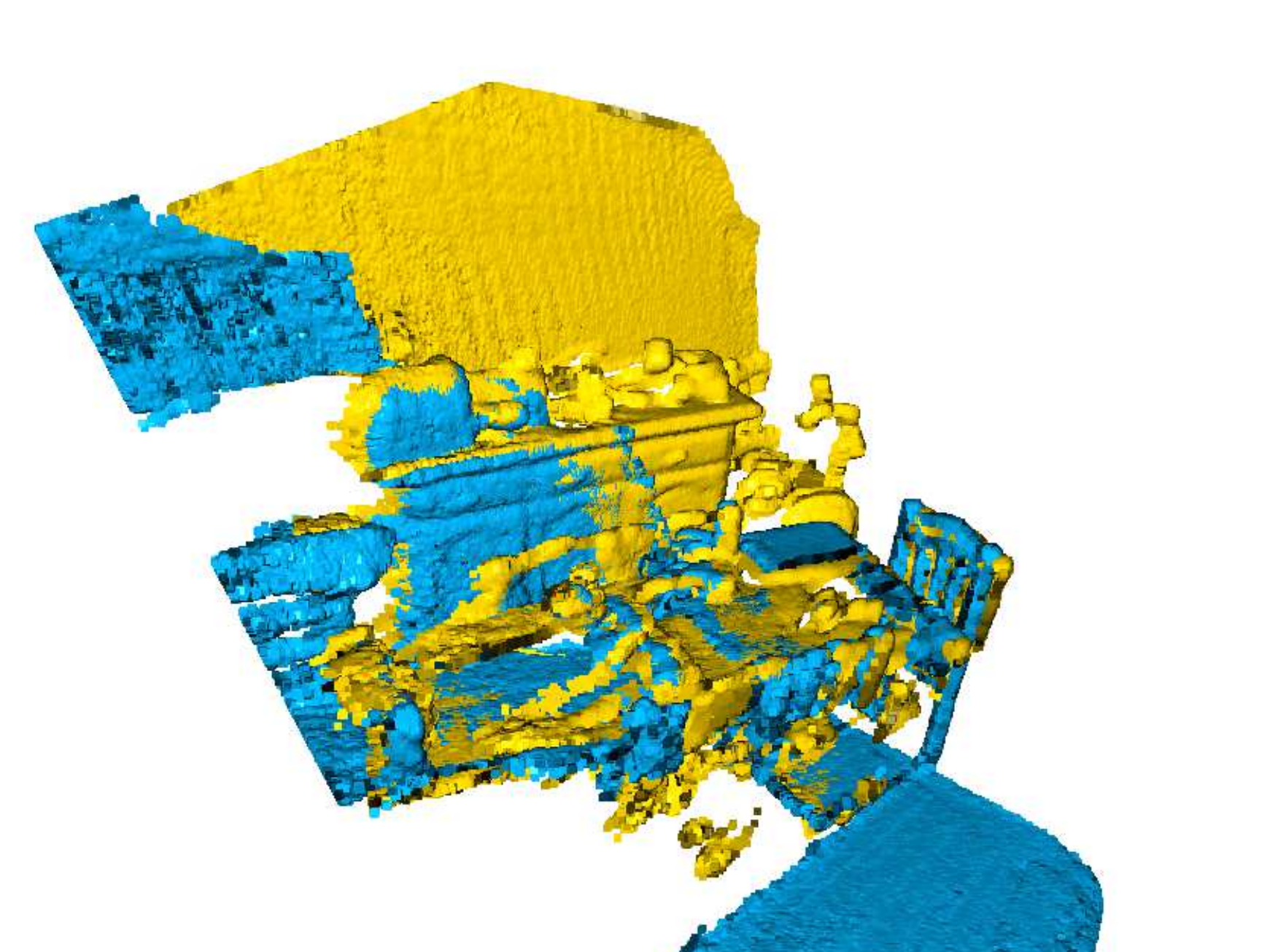}}\hfill
    \subfloat{\includegraphics[trim={500pt 300pt 500pt 100pt}, clip, width=0.16\linewidth]{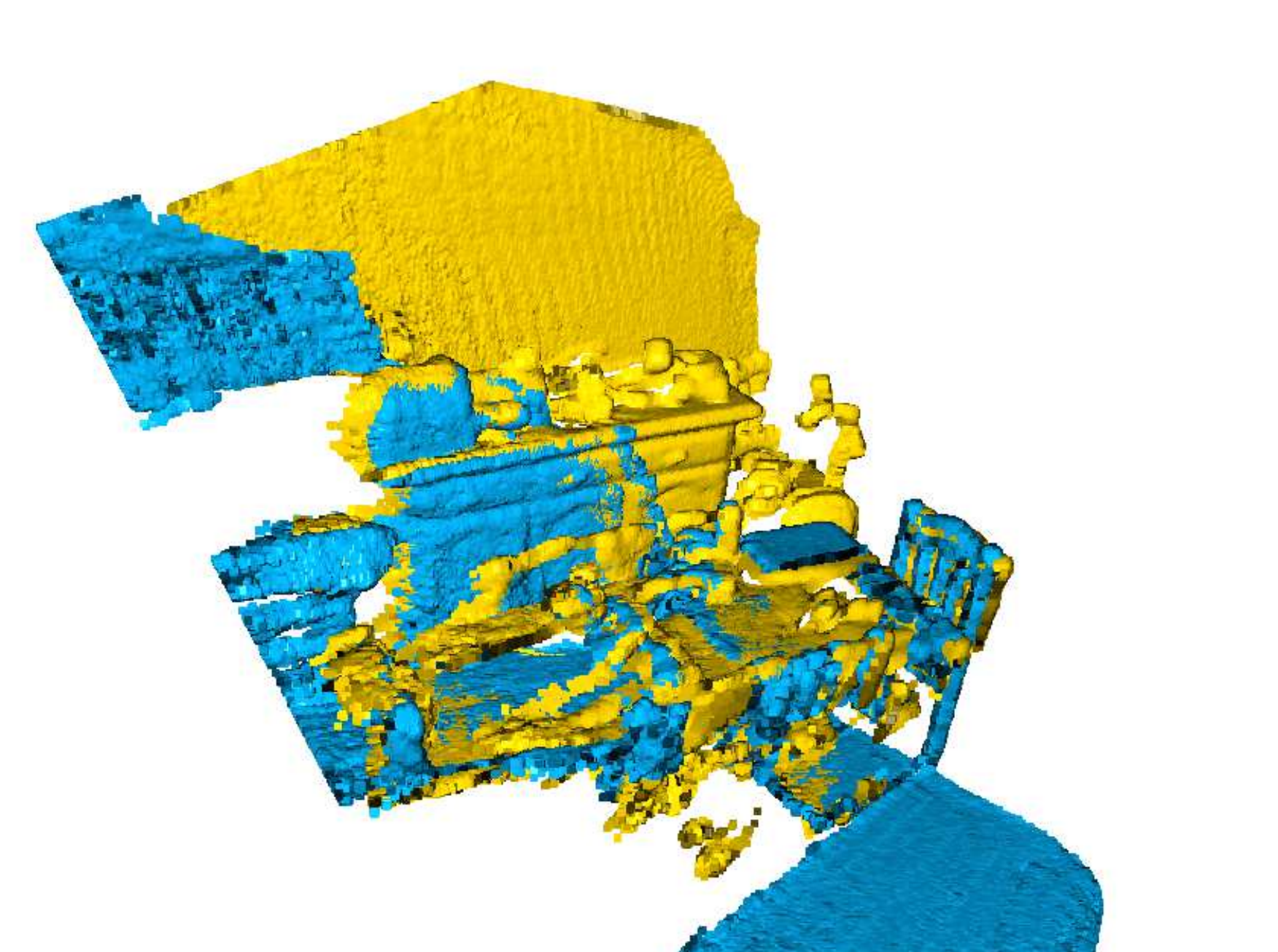}}\hfill
    \\
    \subfloat{\includegraphics[trim={600pt 200pt 450pt 100pt}, clip, scale=0.09]{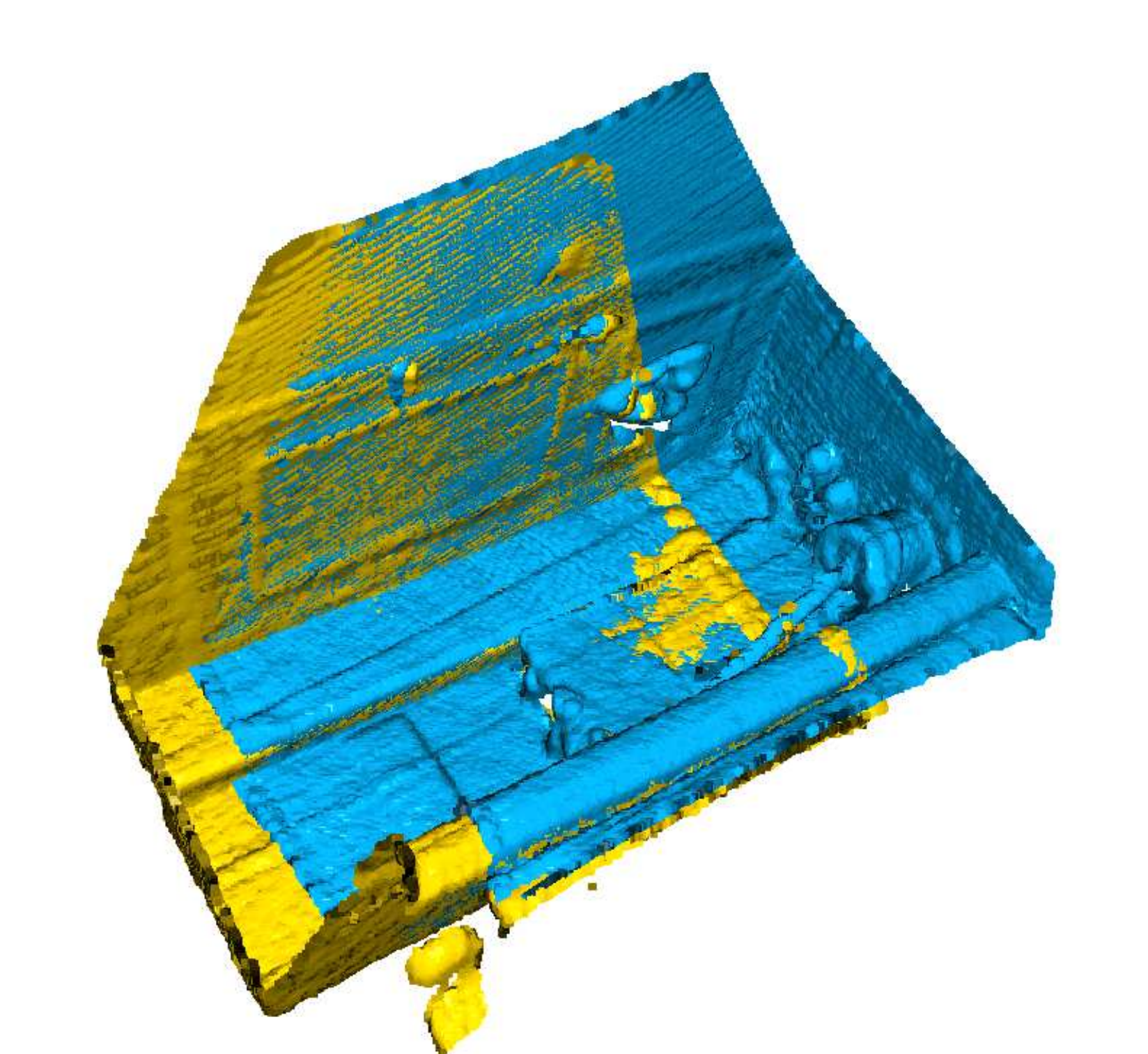}}\hfill
    \subfloat{\includegraphics[trim={600pt 200pt 450pt 100pt}, clip,  scale=0.09]{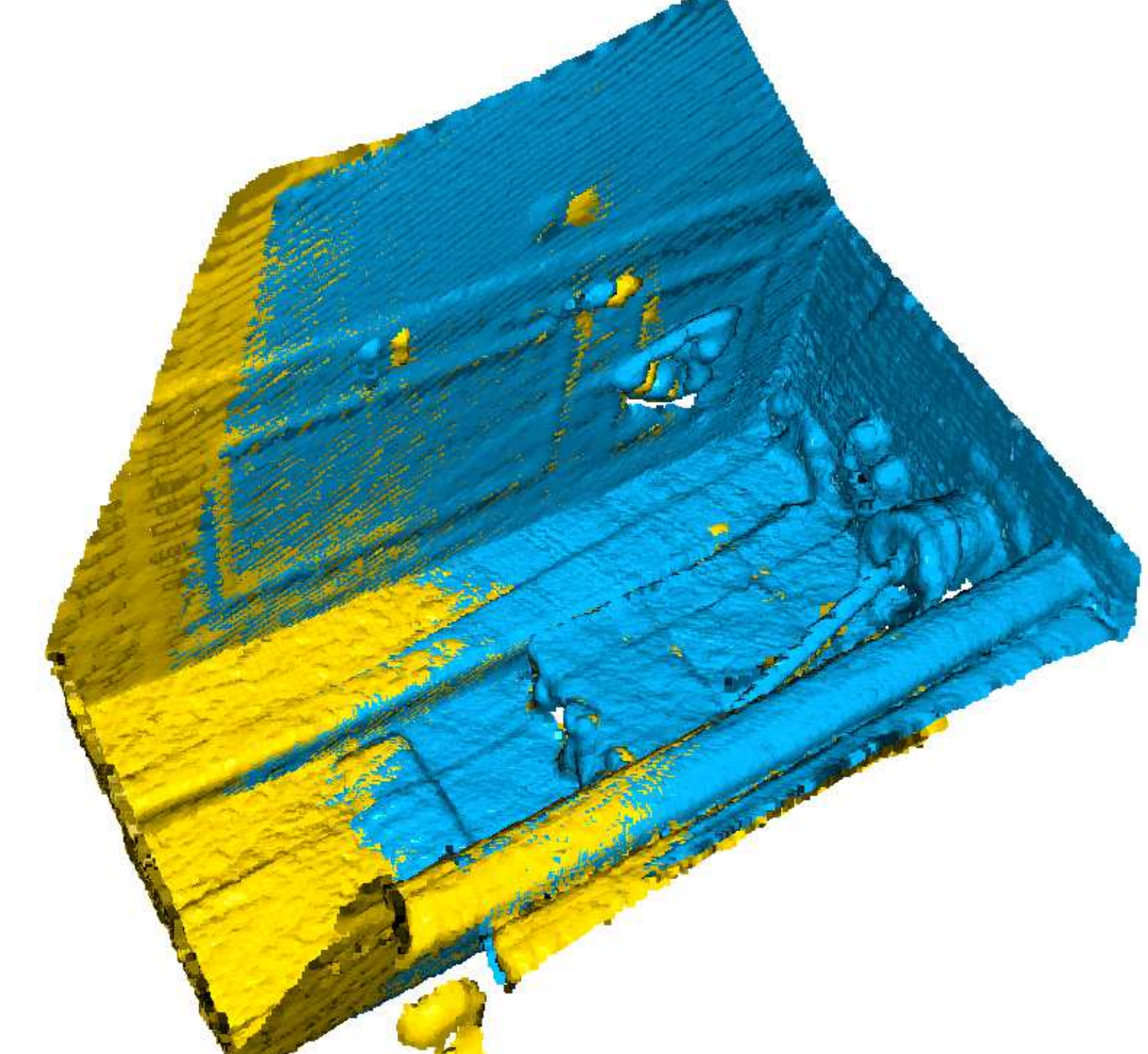}}\hfill
    \subfloat{\includegraphics[trim={600pt 200pt 450pt 100pt}, clip, scale=0.09]{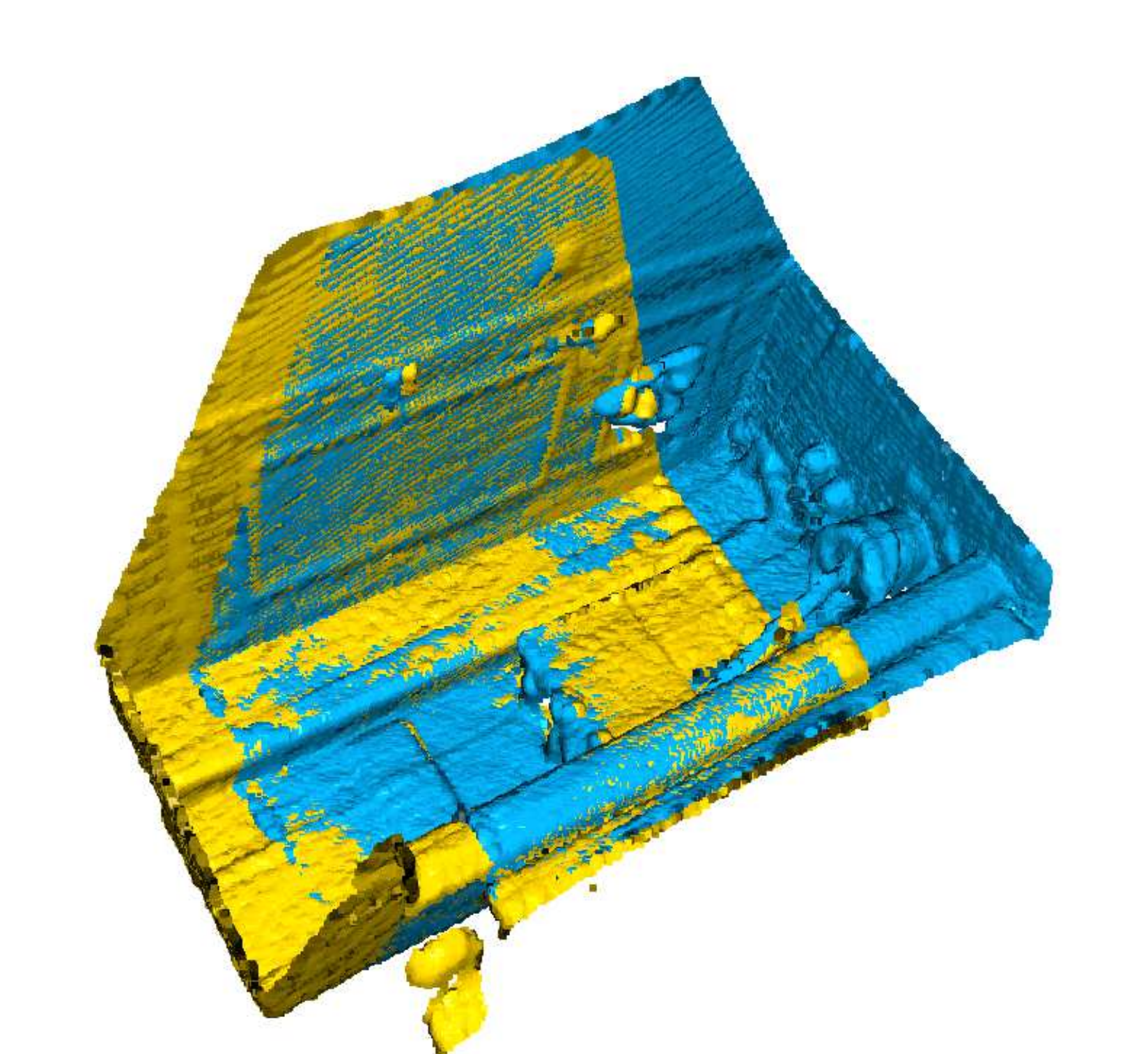}}\hfill
    \subfloat{\includegraphics[trim={600pt 200pt 450pt 100pt}, clip, scale=0.09]{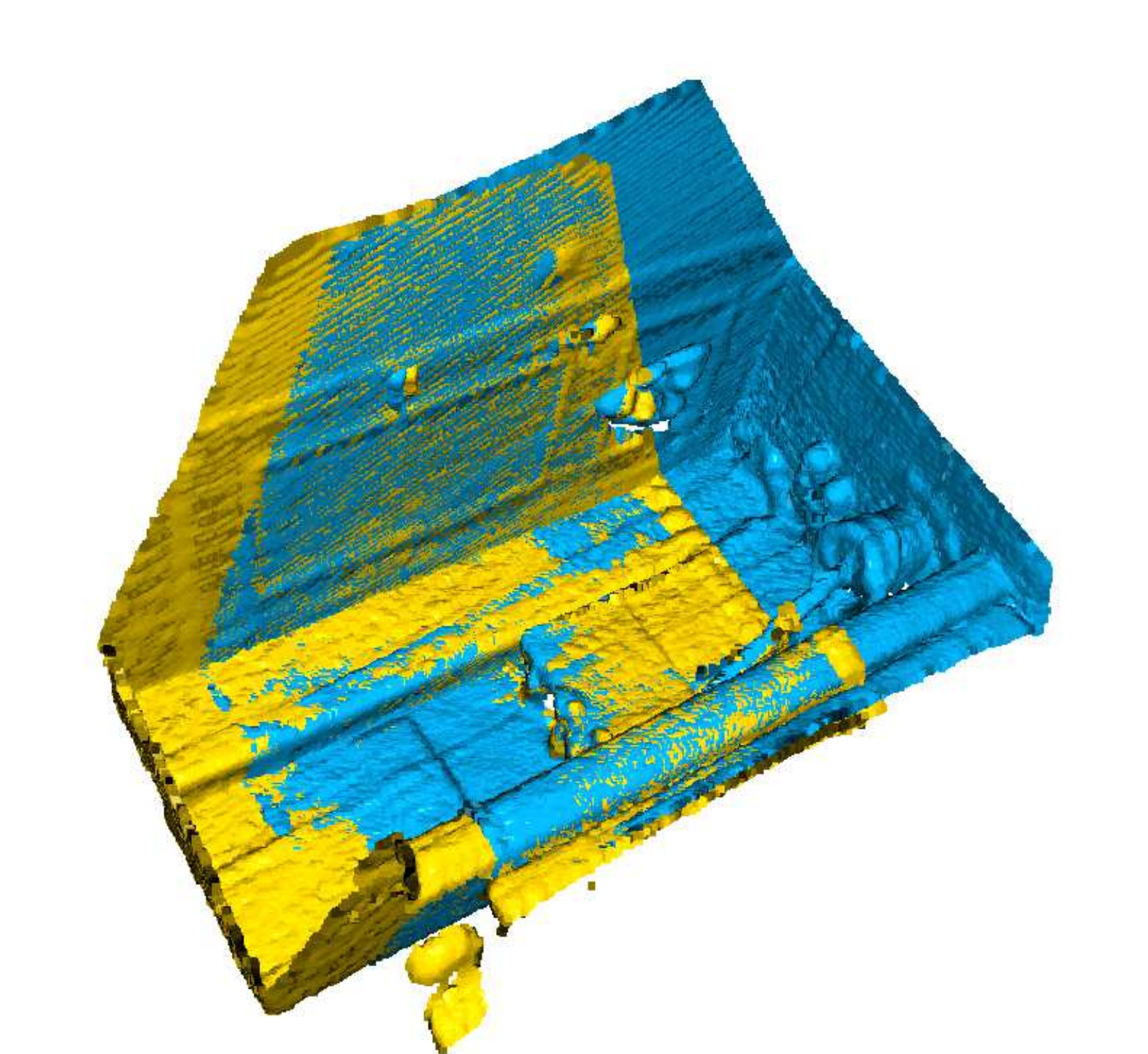}}\hfill
    \subfloat{\includegraphics[trim={600pt 200pt 450pt 100pt}, clip, scale=0.09]{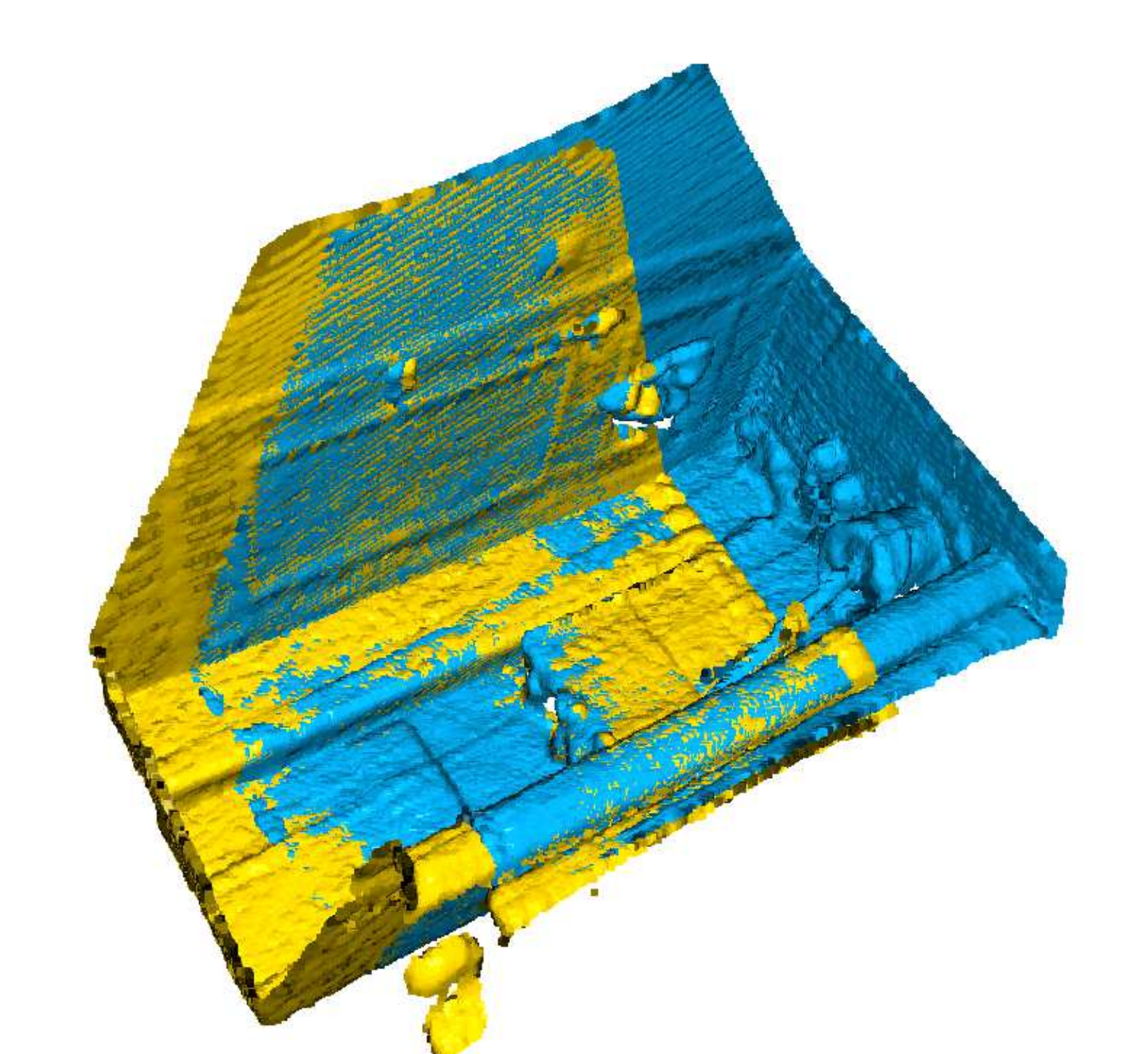}}\hfill
    \subfloat{\includegraphics[trim={600pt 200pt 450pt 100pt}, clip, scale=0.09]{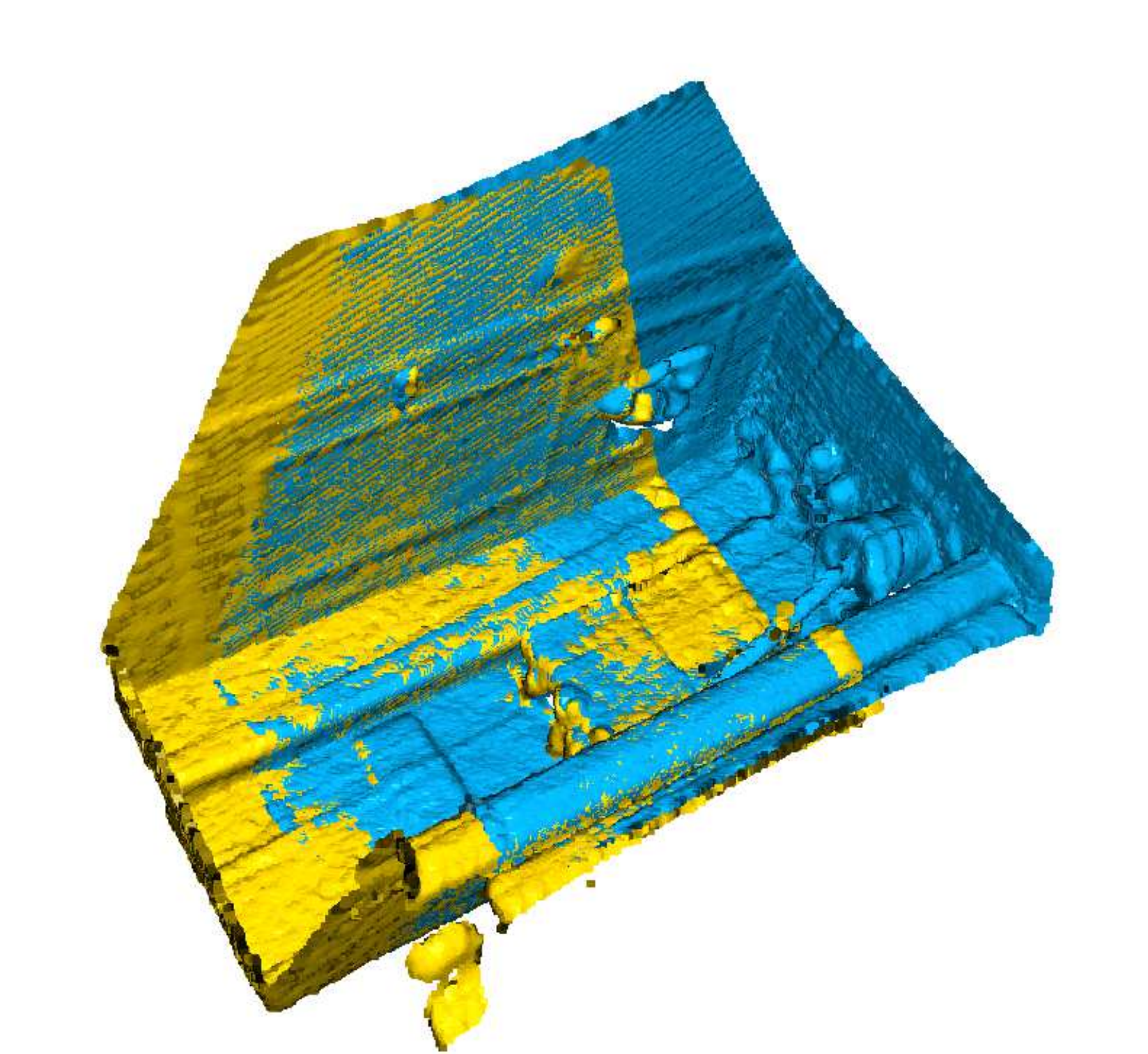}}\hfill
    \\
    \subfloat{\includegraphics[trim={450pt 150pt 450pt 200pt}, clip, width=0.160\linewidth]{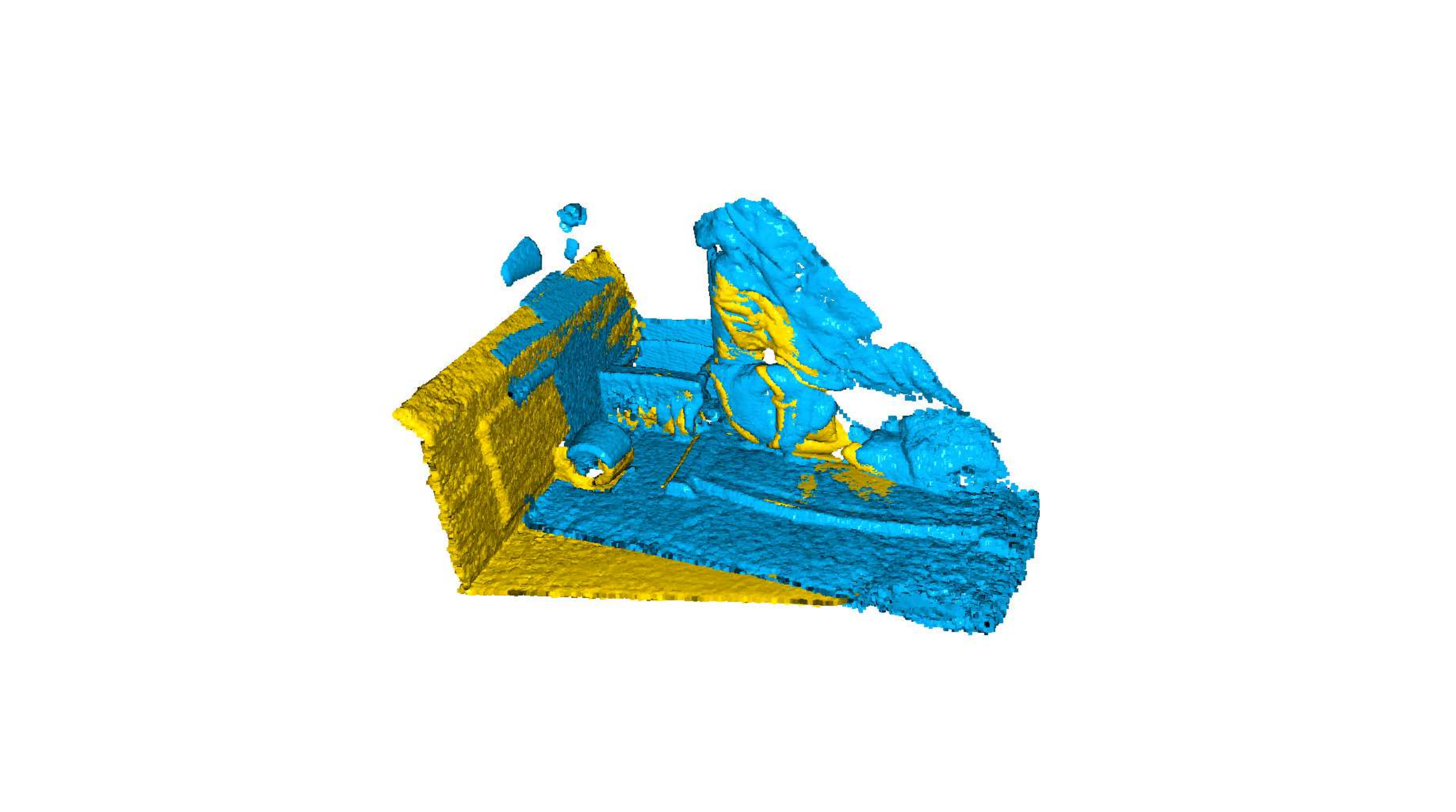}}\hfill
    \subfloat{\includegraphics[trim={450pt 150pt 450pt 200pt}, clip, width=0.160\linewidth]{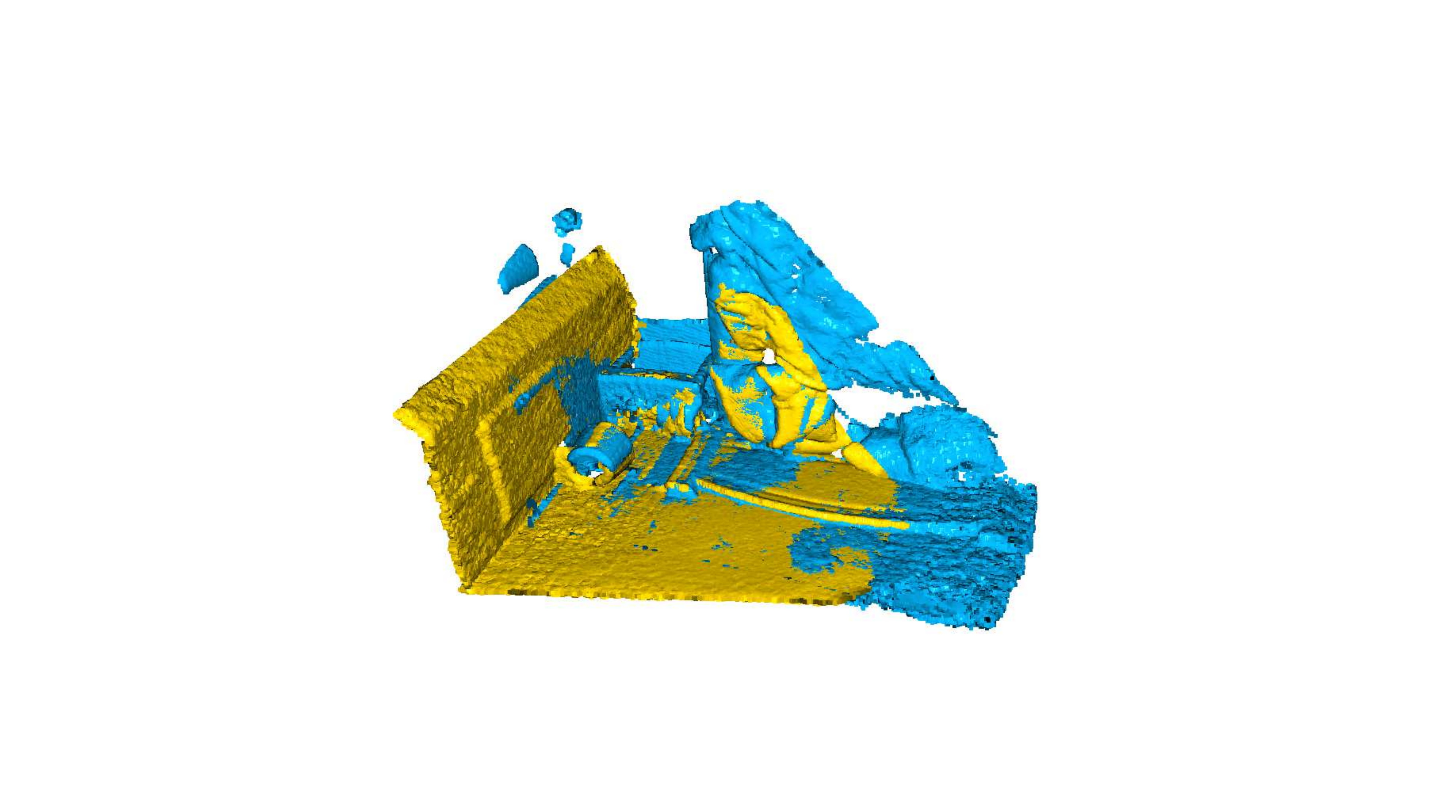}}\hfill
    \subfloat{\includegraphics[trim={450pt 150pt 450pt 200pt}, clip, width=0.160\linewidth]{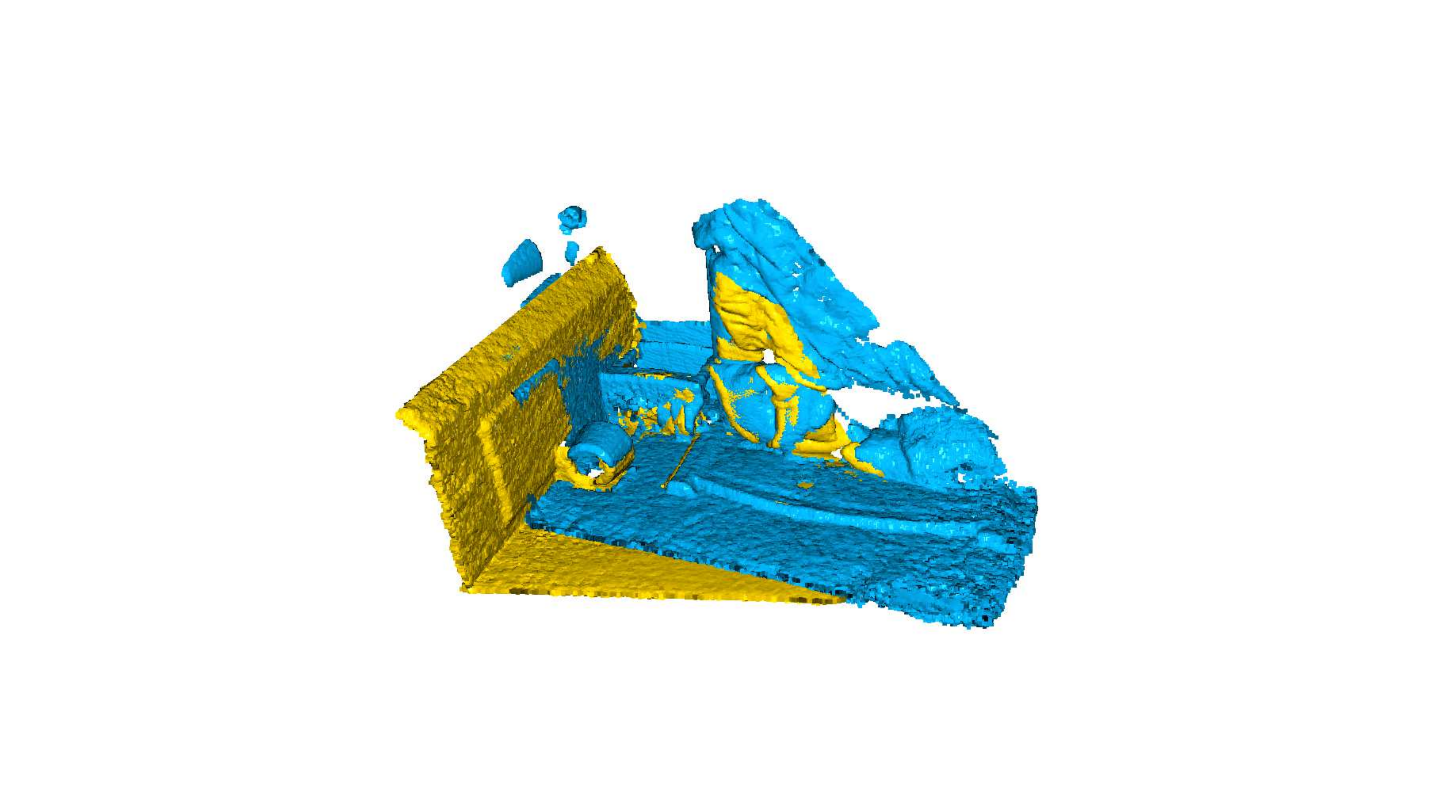}}\hfill
    \subfloat{\includegraphics[trim={450pt 150pt 450pt 200pt}, clip, width=0.160\linewidth]{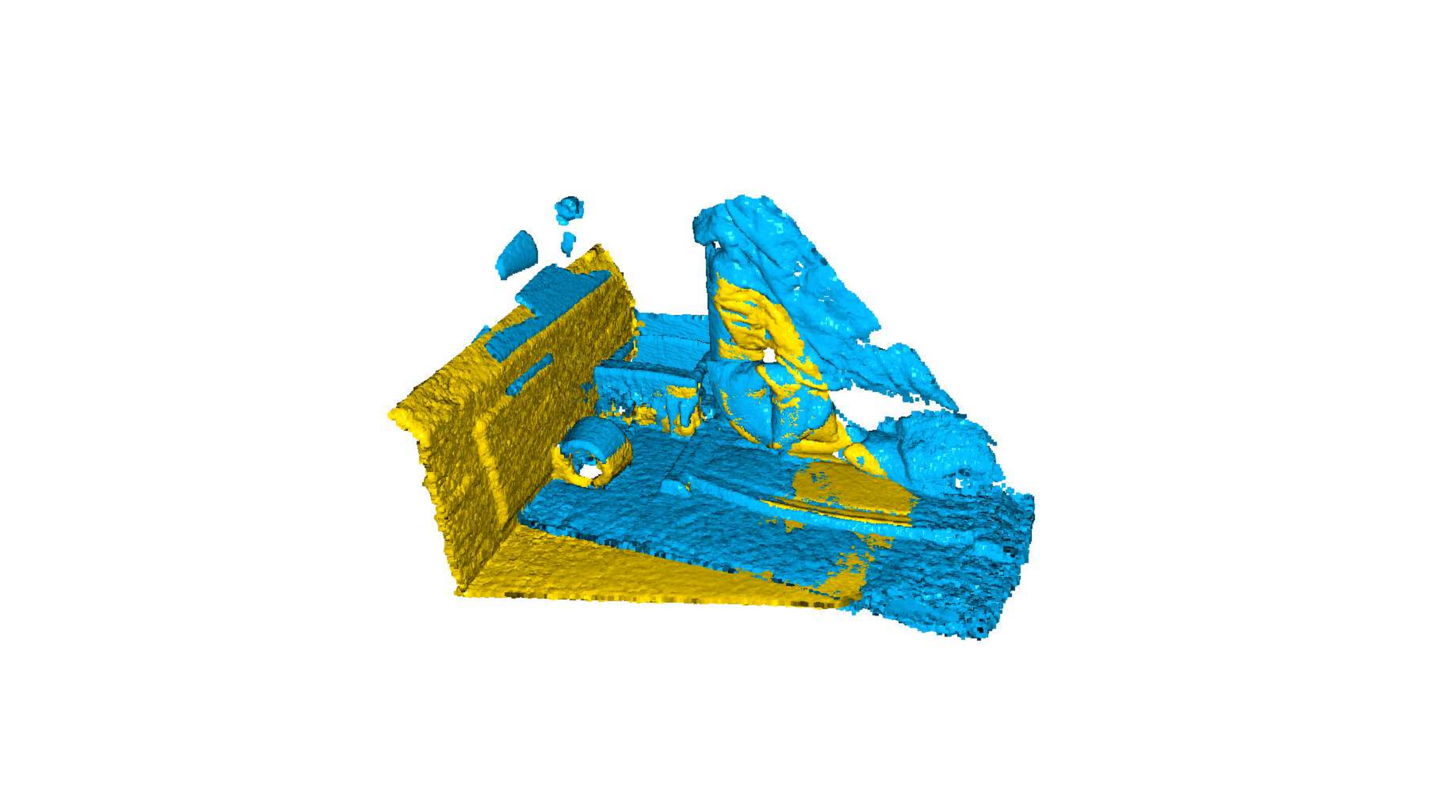}}\hfill
    \subfloat{\includegraphics[trim={450pt 150pt 450pt 200pt}, clip, width=0.160\linewidth]{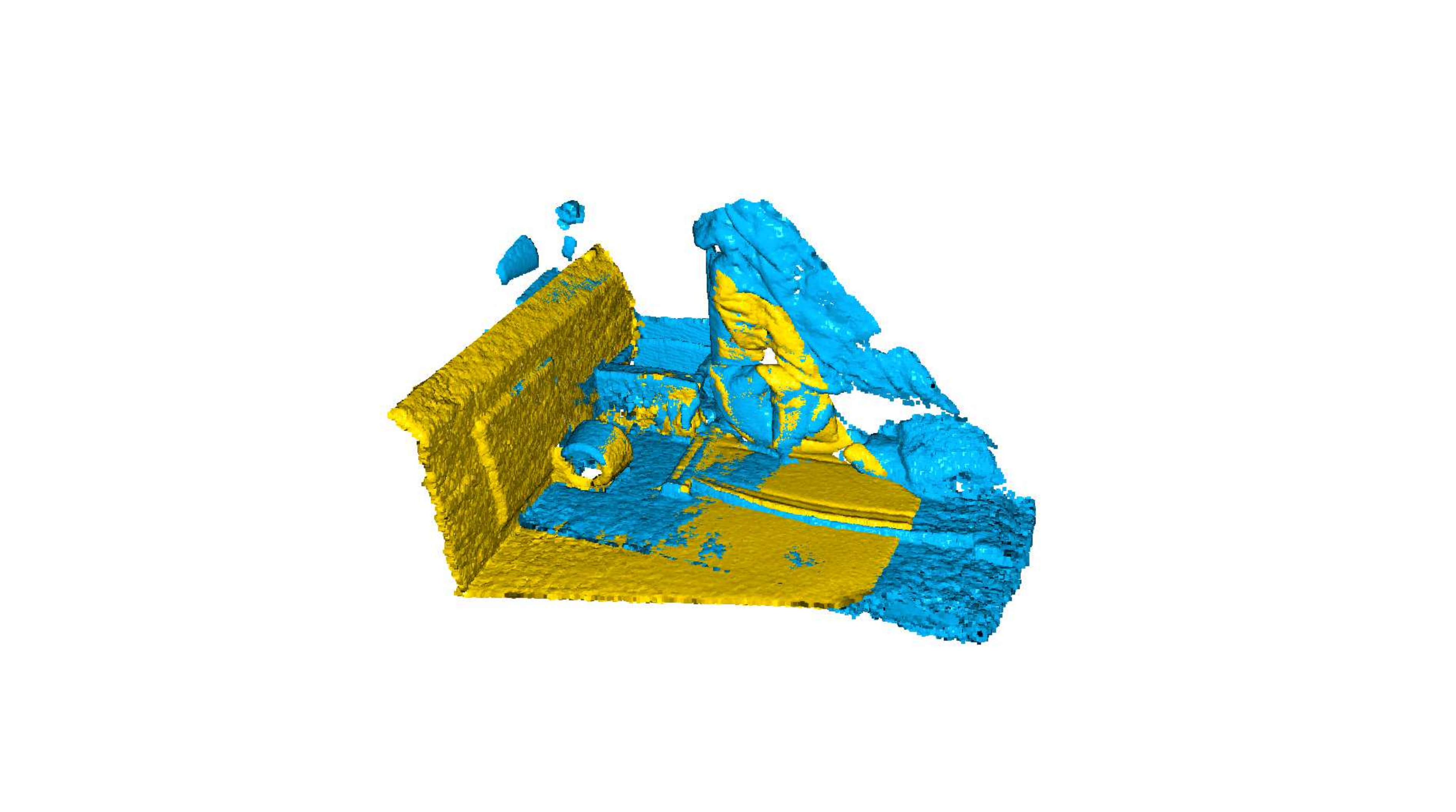}}\hfill
    \subfloat{\includegraphics[trim={450pt 150pt 450pt 200pt}, clip, width=0.160\linewidth]{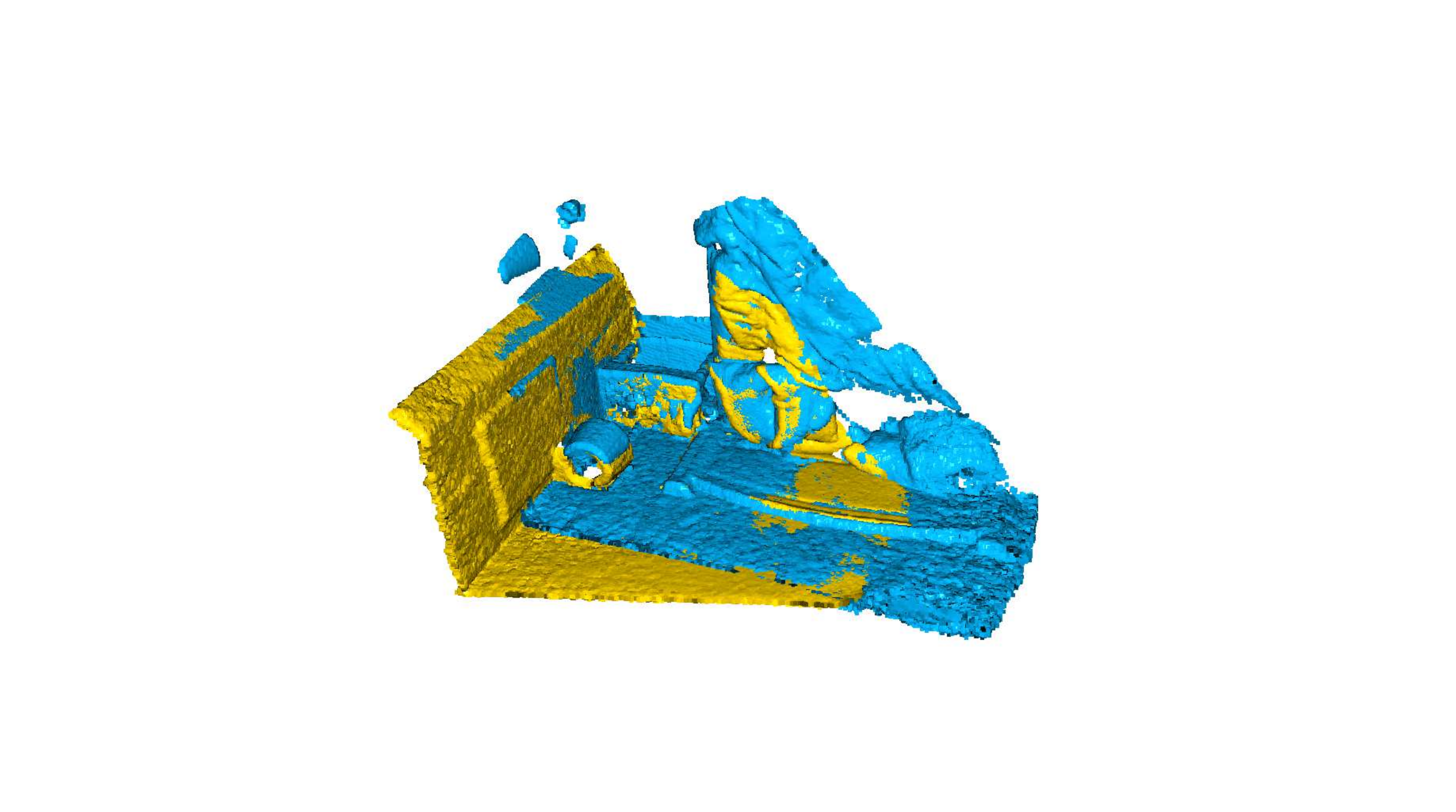}}\hfill
    \\
    \subfloat{\includegraphics[trim={550pt 200pt 600pt 100pt}, clip, scale=0.09]{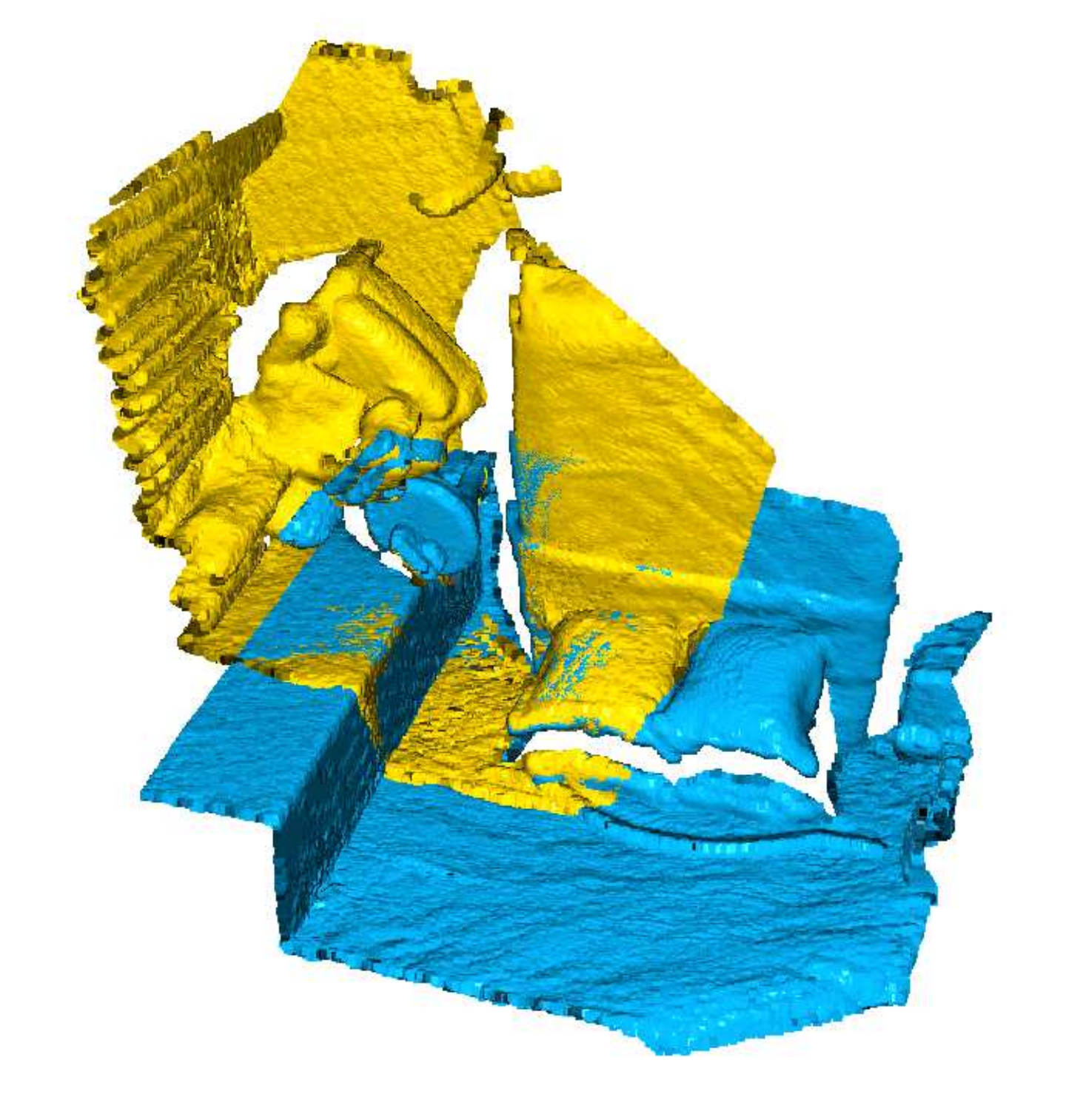}}\hfill
    \subfloat{\includegraphics[trim={550pt 200pt 600pt 100pt}, clip, scale=0.09]{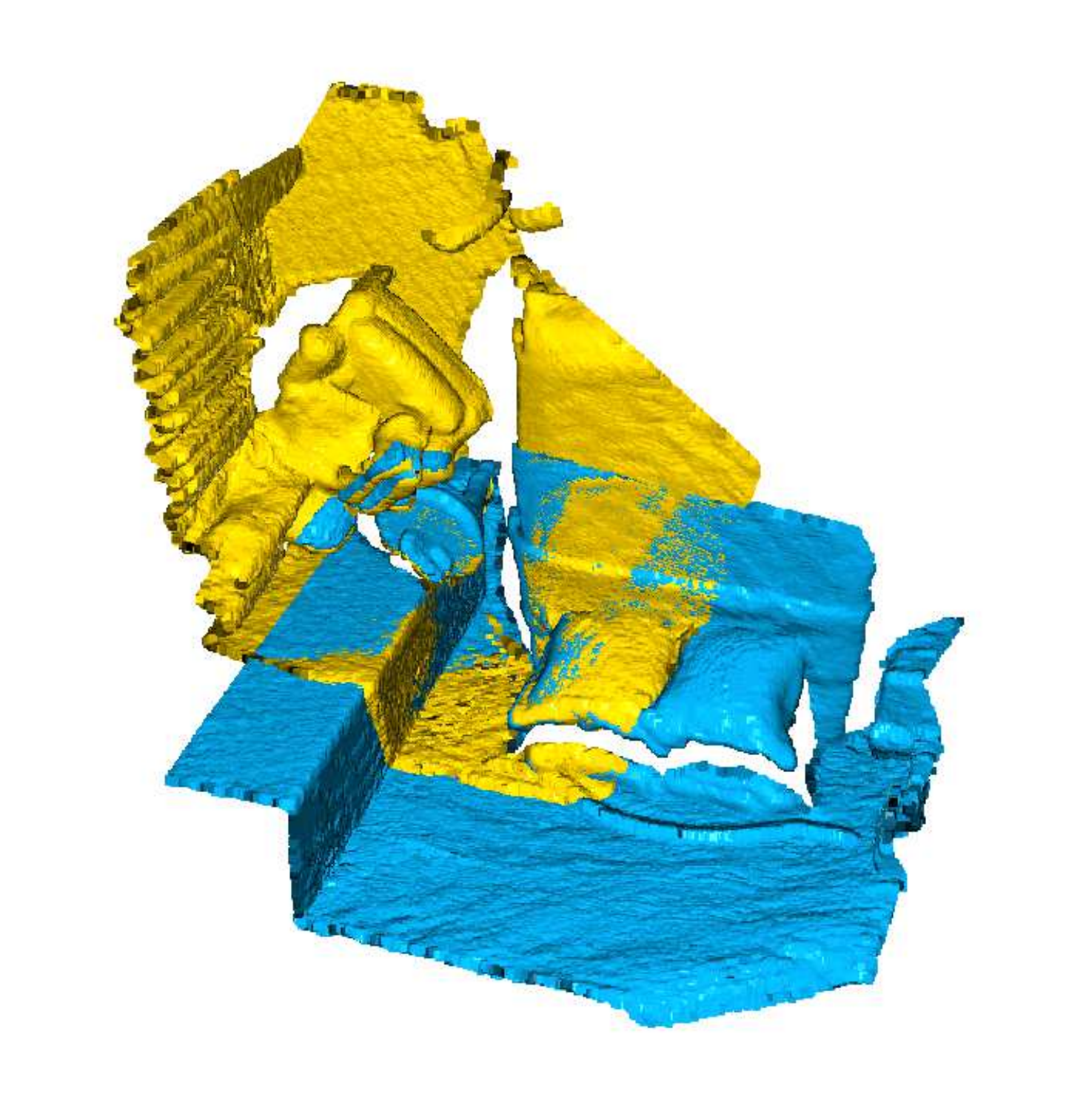}}\hfill
    \subfloat{\includegraphics[trim={550pt 200pt 600pt 100pt}, clip,scale=0.09]{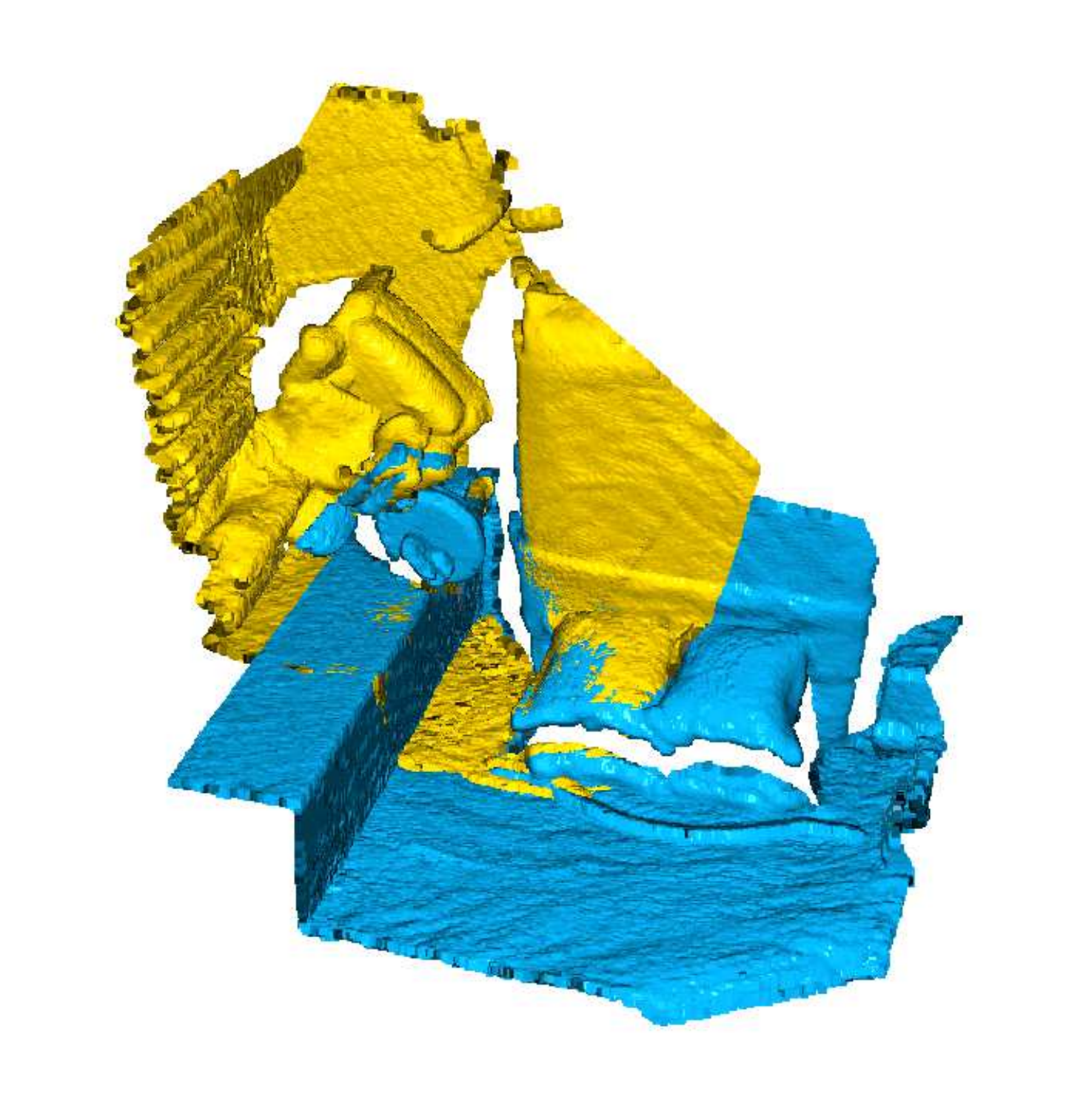}}\hfill
    \subfloat{\includegraphics[trim={550pt 200pt 600pt 100pt}, clip, scale=0.09]{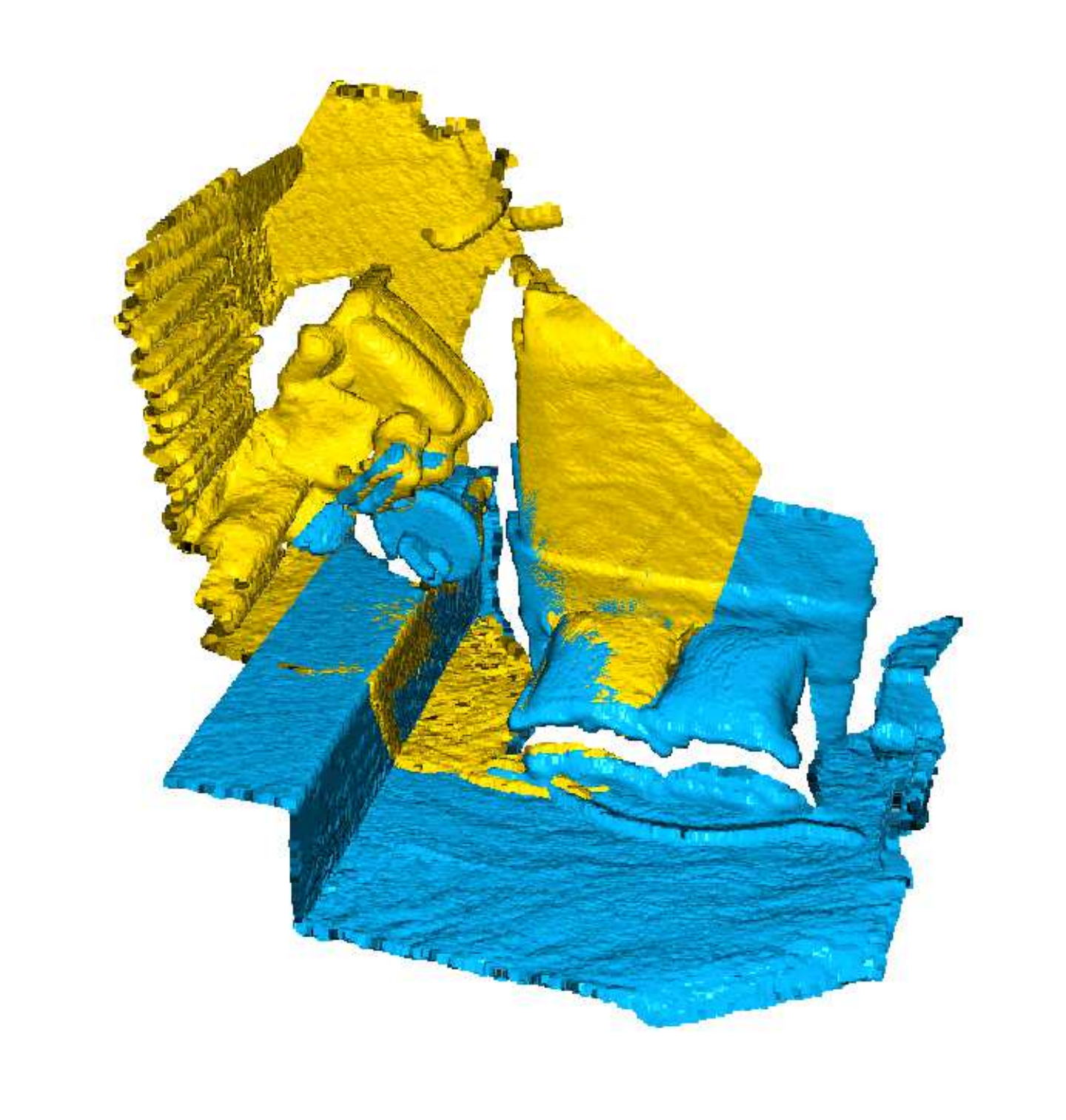}}\hfill
    \subfloat{\includegraphics[trim={550pt 200pt 600pt 100pt}, clip,scale=0.09]{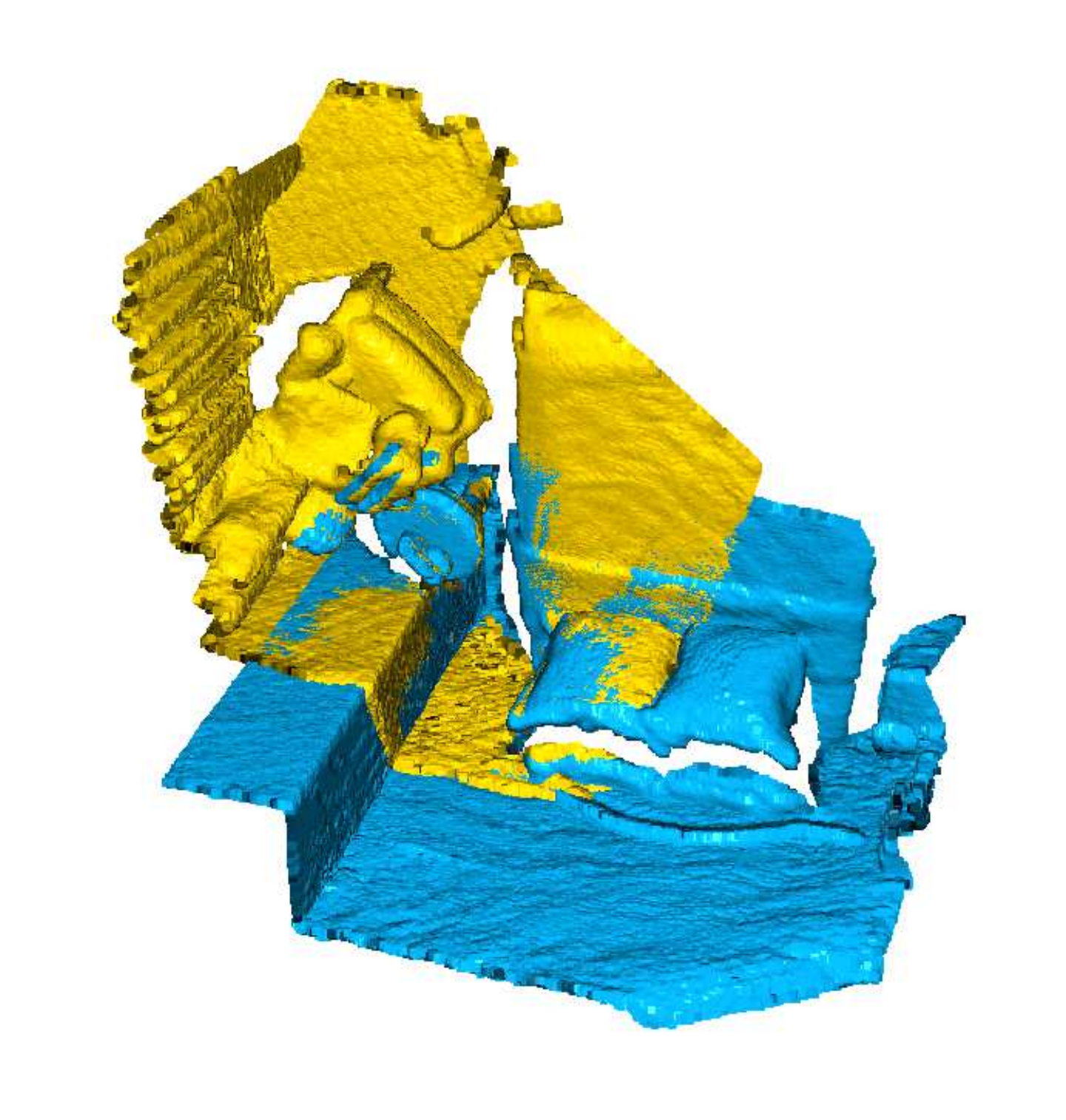}}\hfill
    \subfloat{\includegraphics[trim={550pt 200pt 600pt 100pt}, clip, scale=0.09]{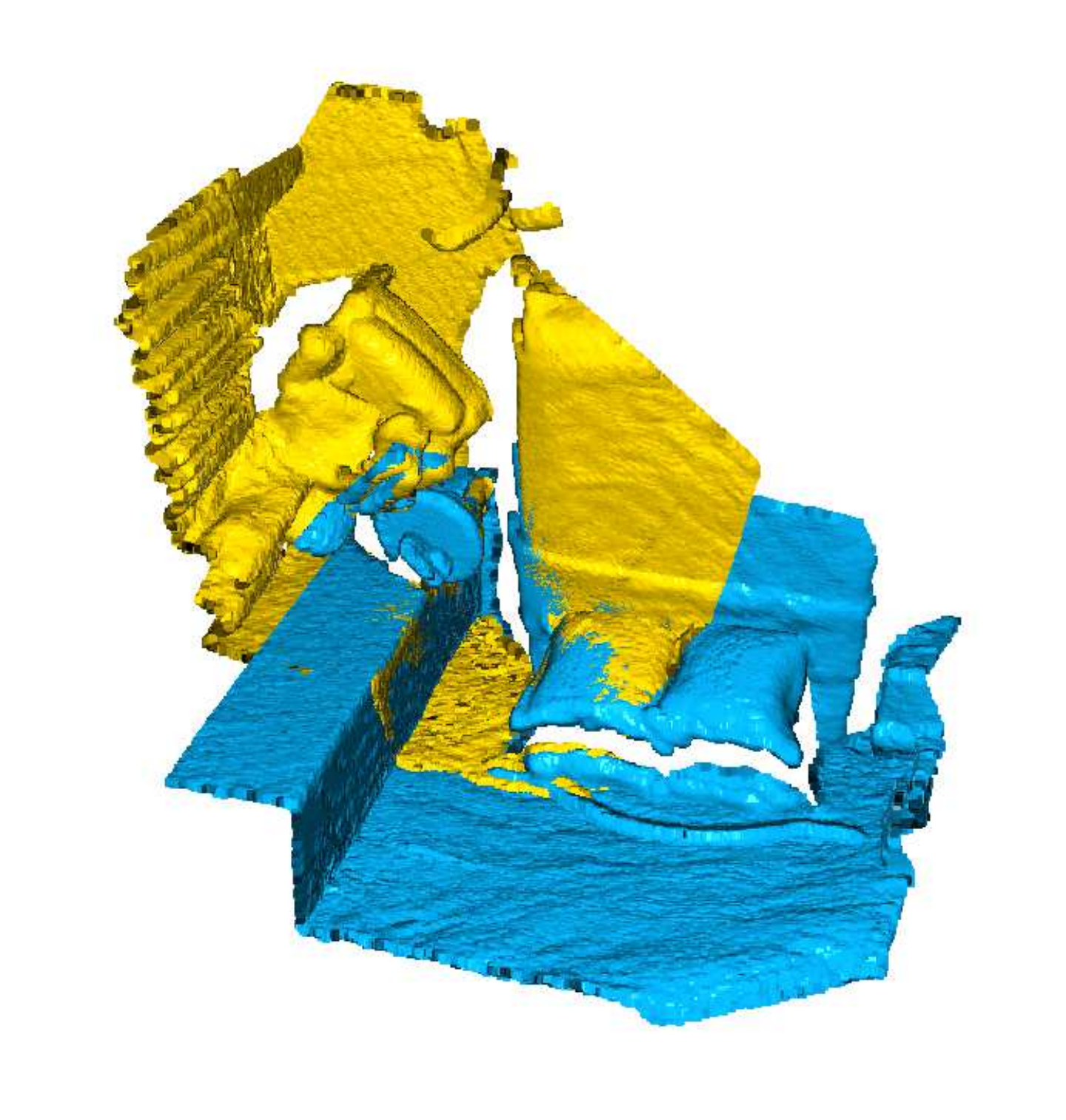}}\hfill

  \subfloat{\includegraphics[trim={550pt 200pt 600pt 300pt}, clip, width=0.16\linewidth]{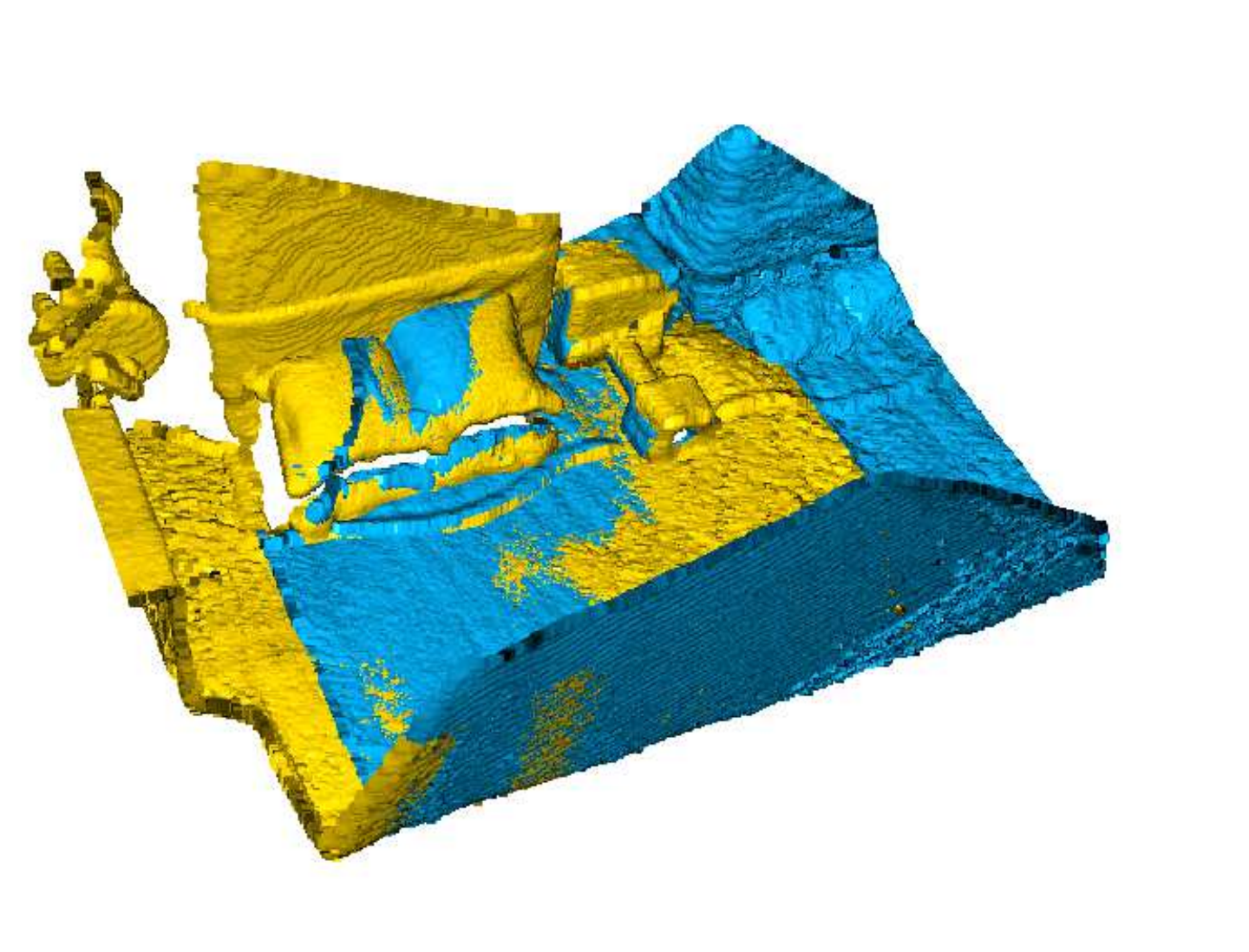}}\hfill
    \subfloat{\includegraphics[trim={550pt 200pt 600pt 300pt}, clip, width=0.16\linewidth]{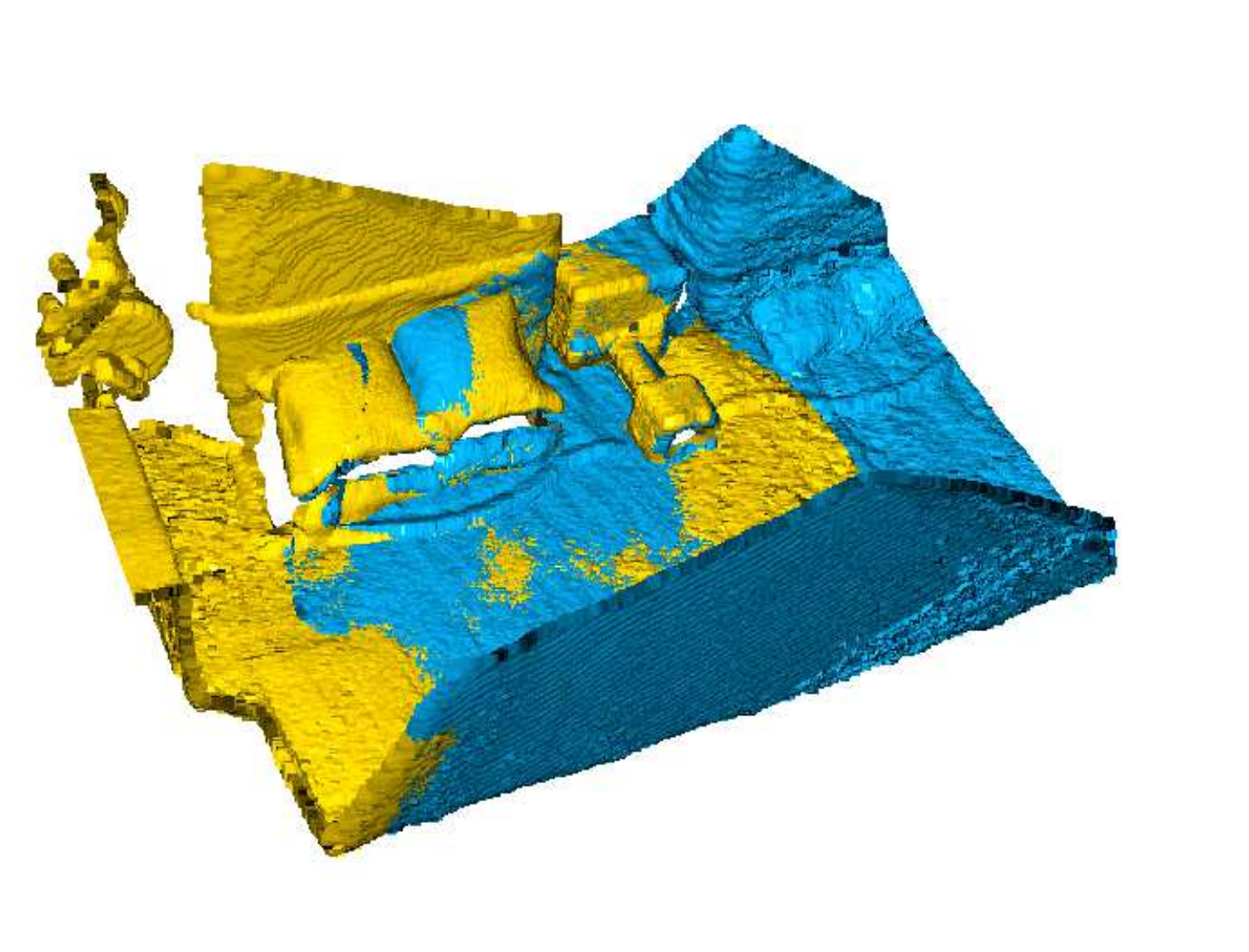}}\hfill
    \subfloat{\includegraphics[trim={550pt 200pt 600pt 300pt}, clip, width=0.16\linewidth]{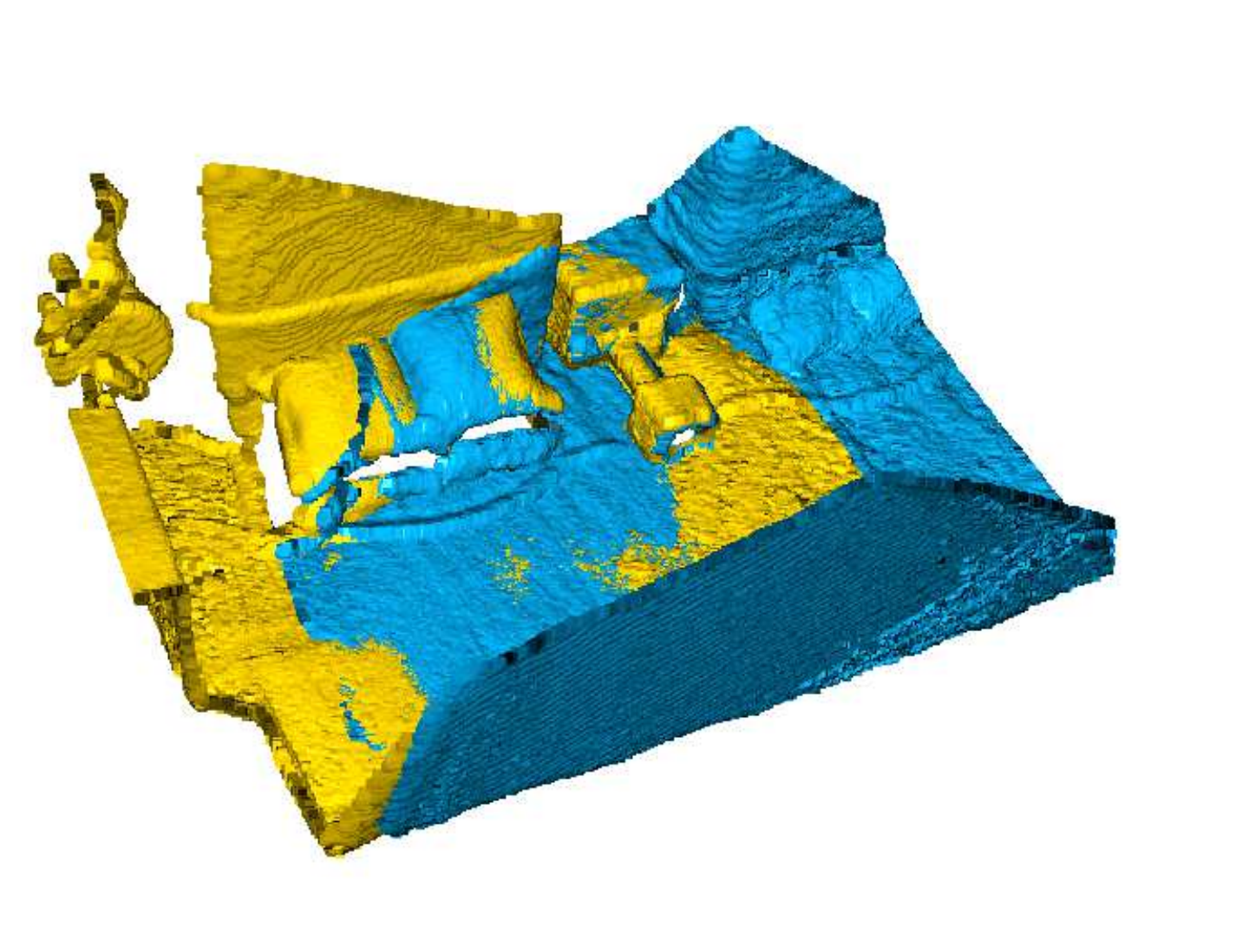}}\hfill
    \subfloat{\includegraphics[trim={550pt 200pt 600pt 300pt}, clip, width=0.16\linewidth]{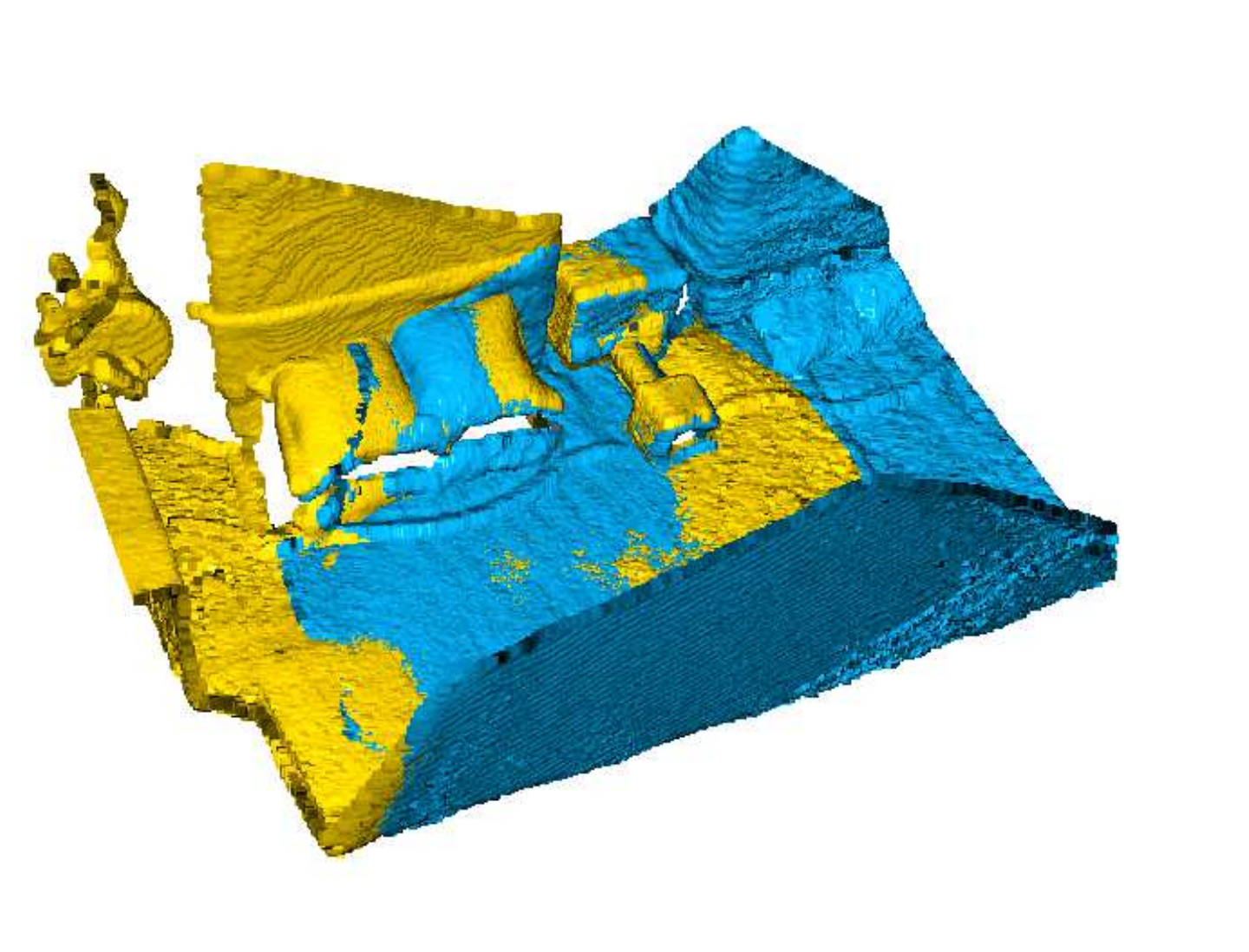}}\hfill
    \subfloat{\includegraphics[trim={550pt 200pt 600pt 300pt}, clip, width=0.16\linewidth]{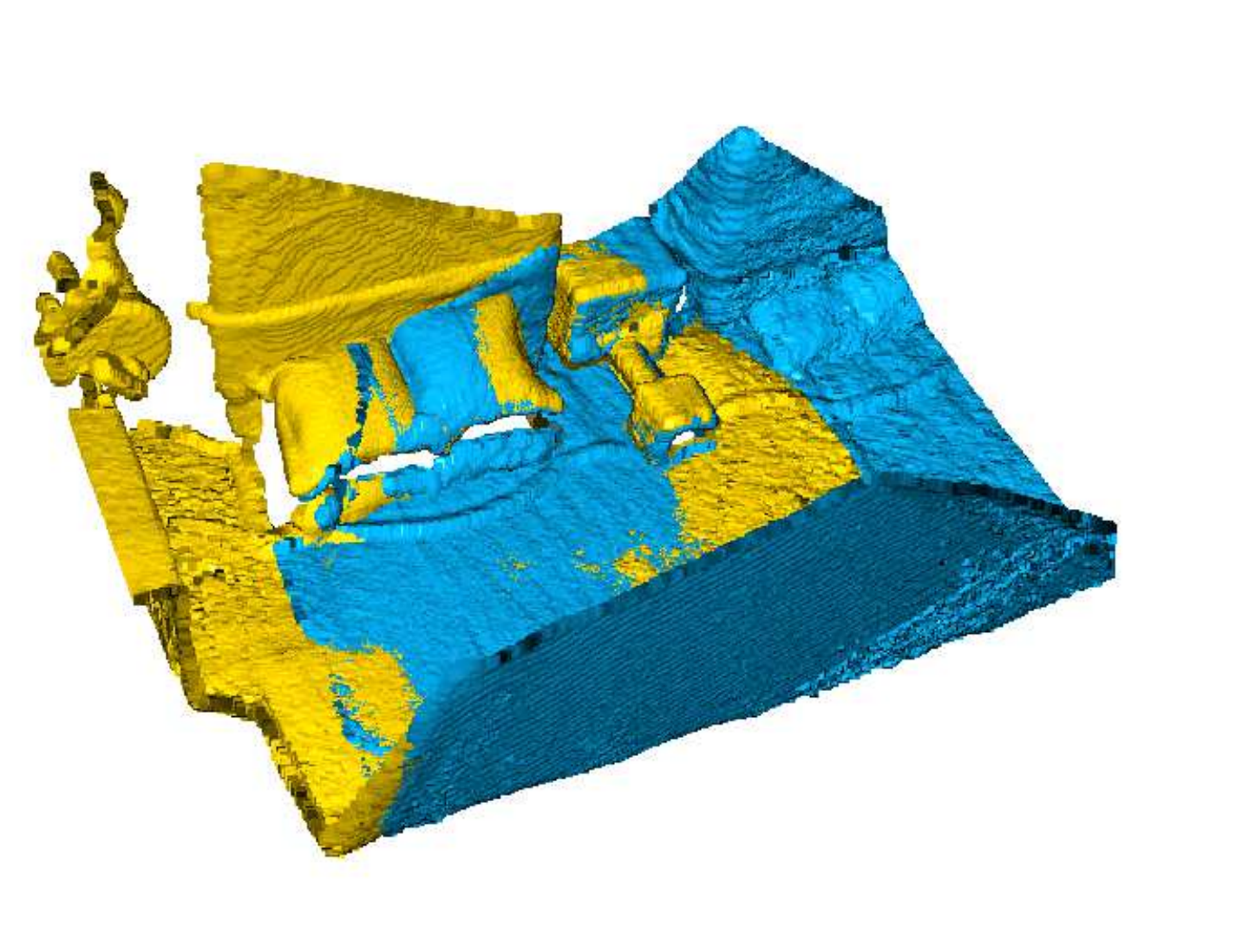}}\hfill
    \subfloat{\includegraphics[trim={550pt 200pt 600pt 300pt}, clip, width=0.16\linewidth]{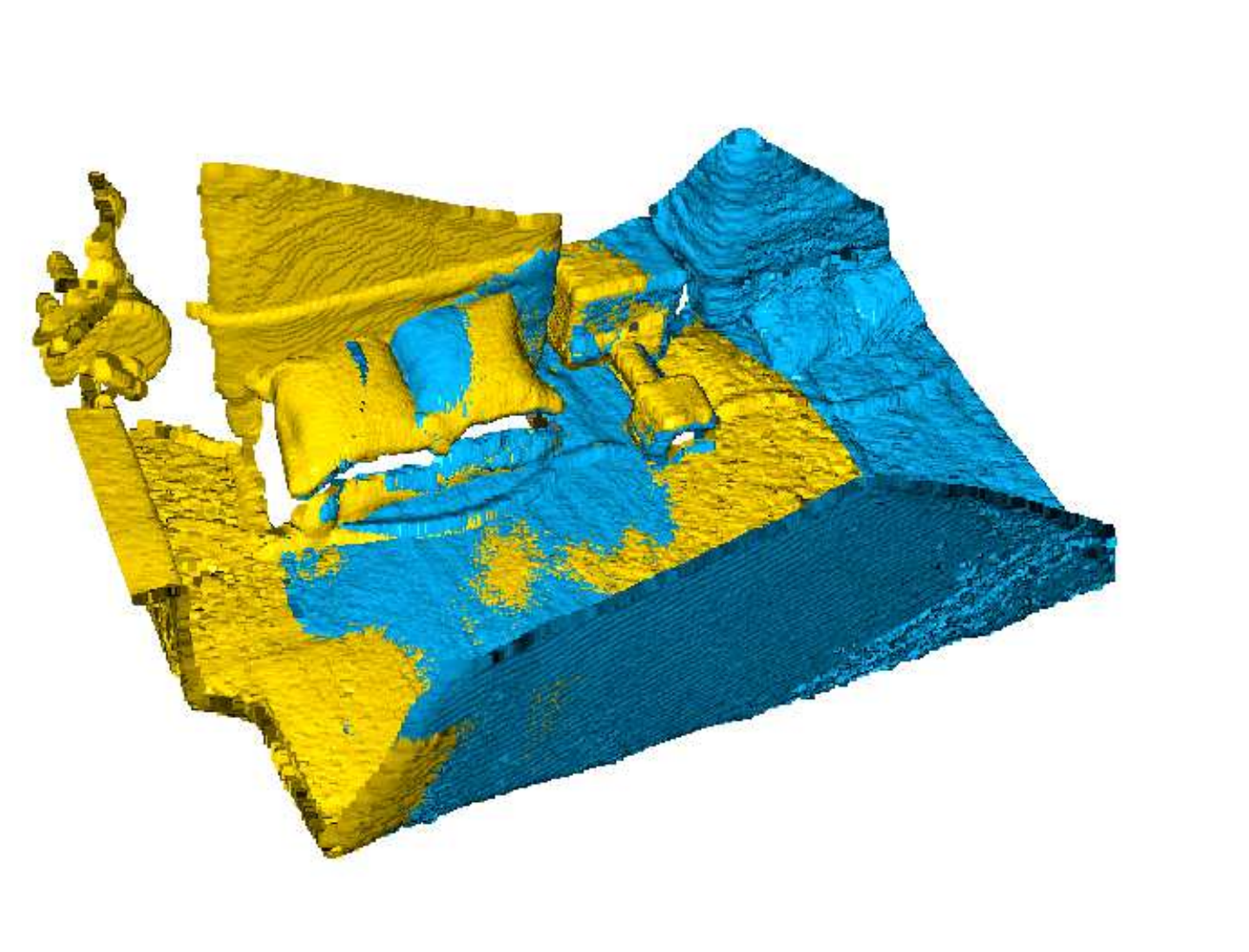}}\hfill

  \subfloat{\includegraphics[trim={500pt 200pt 400pt 250pt}, clip, width=0.16\linewidth]{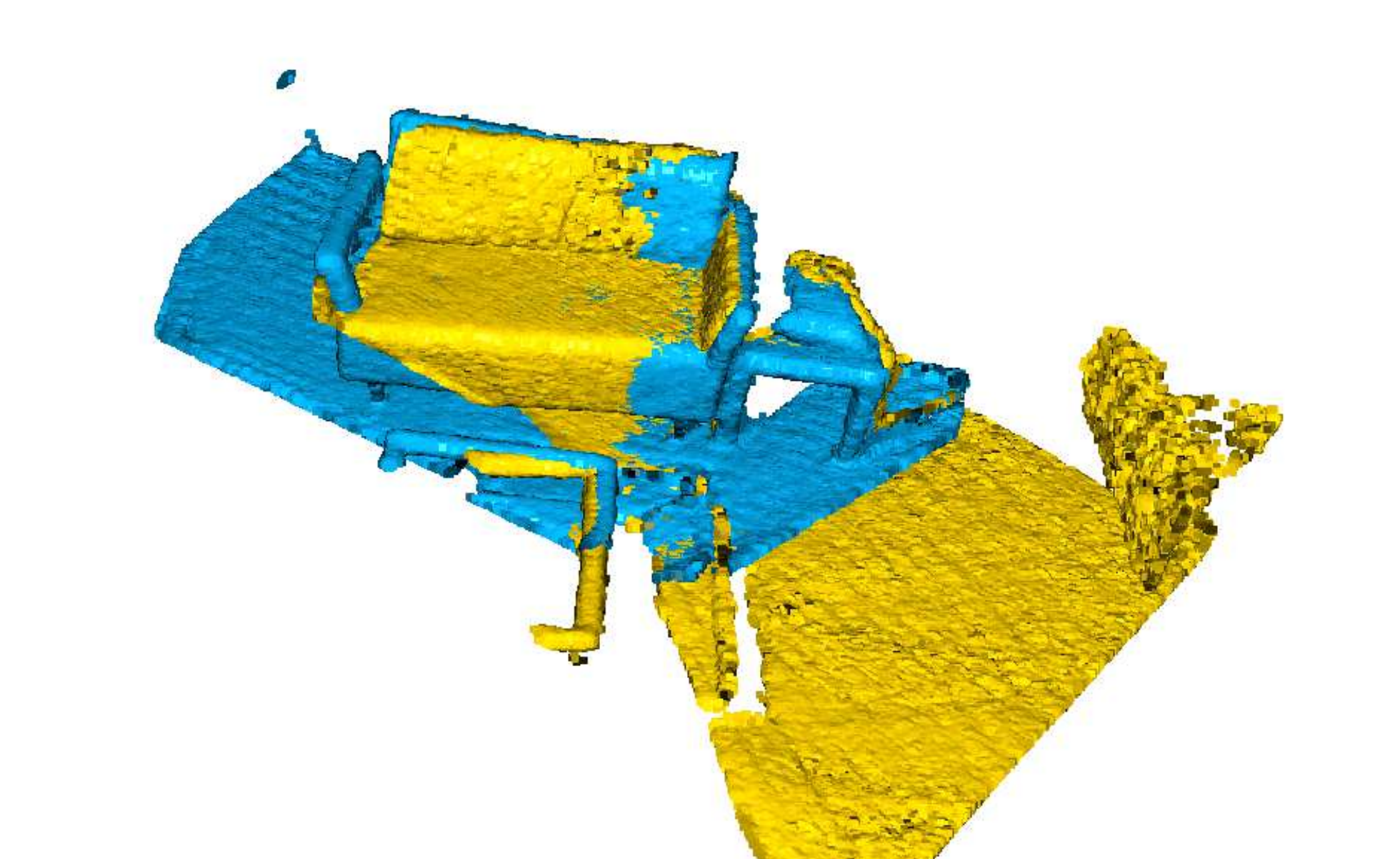}}\hfill
    \subfloat{\includegraphics[trim={500pt 200pt 400pt 250pt}, clip, width=0.16\linewidth]{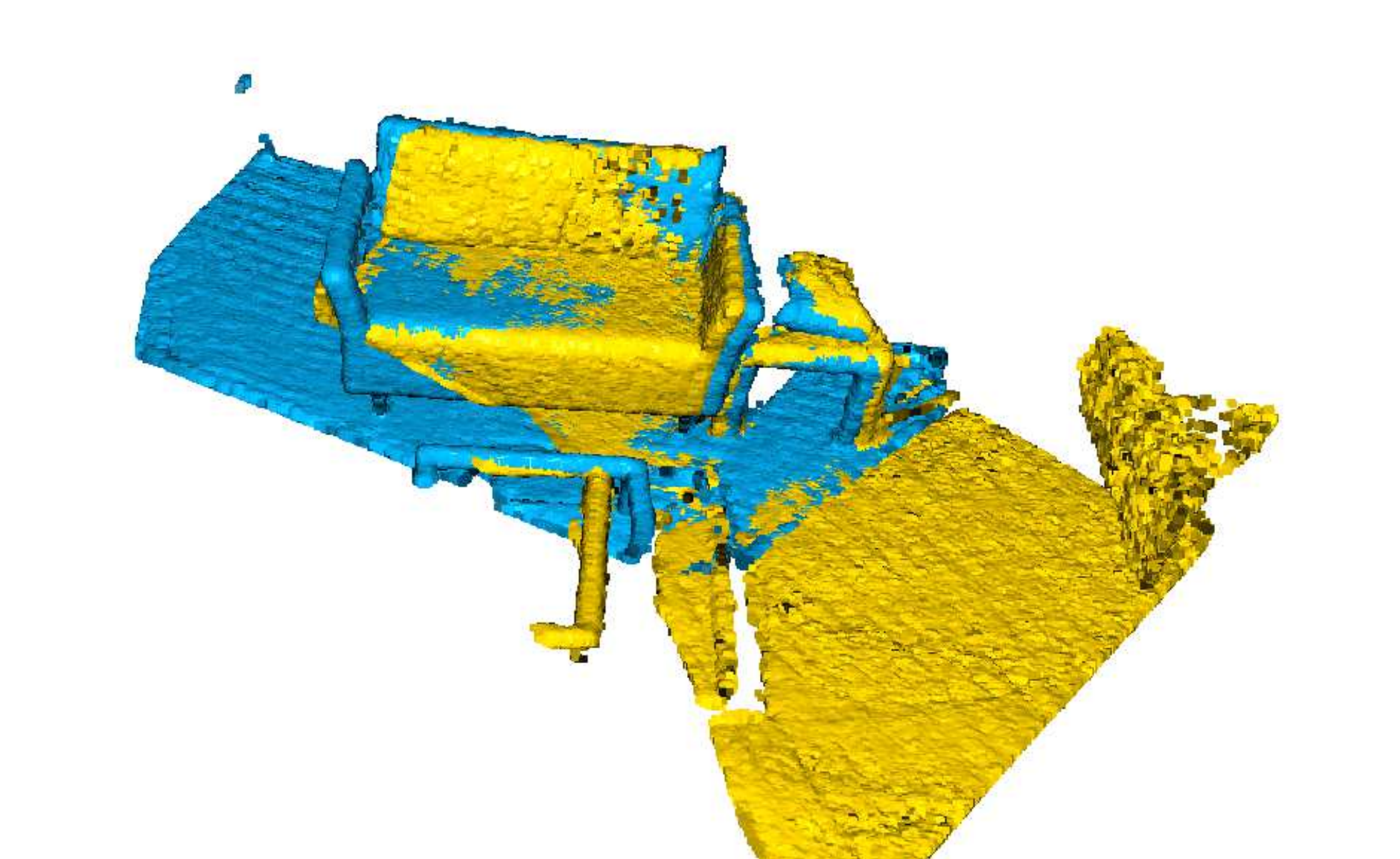}}\hfill
    \subfloat{\includegraphics[trim={500pt 200pt 400pt 250pt}, clip, width=0.16\linewidth]{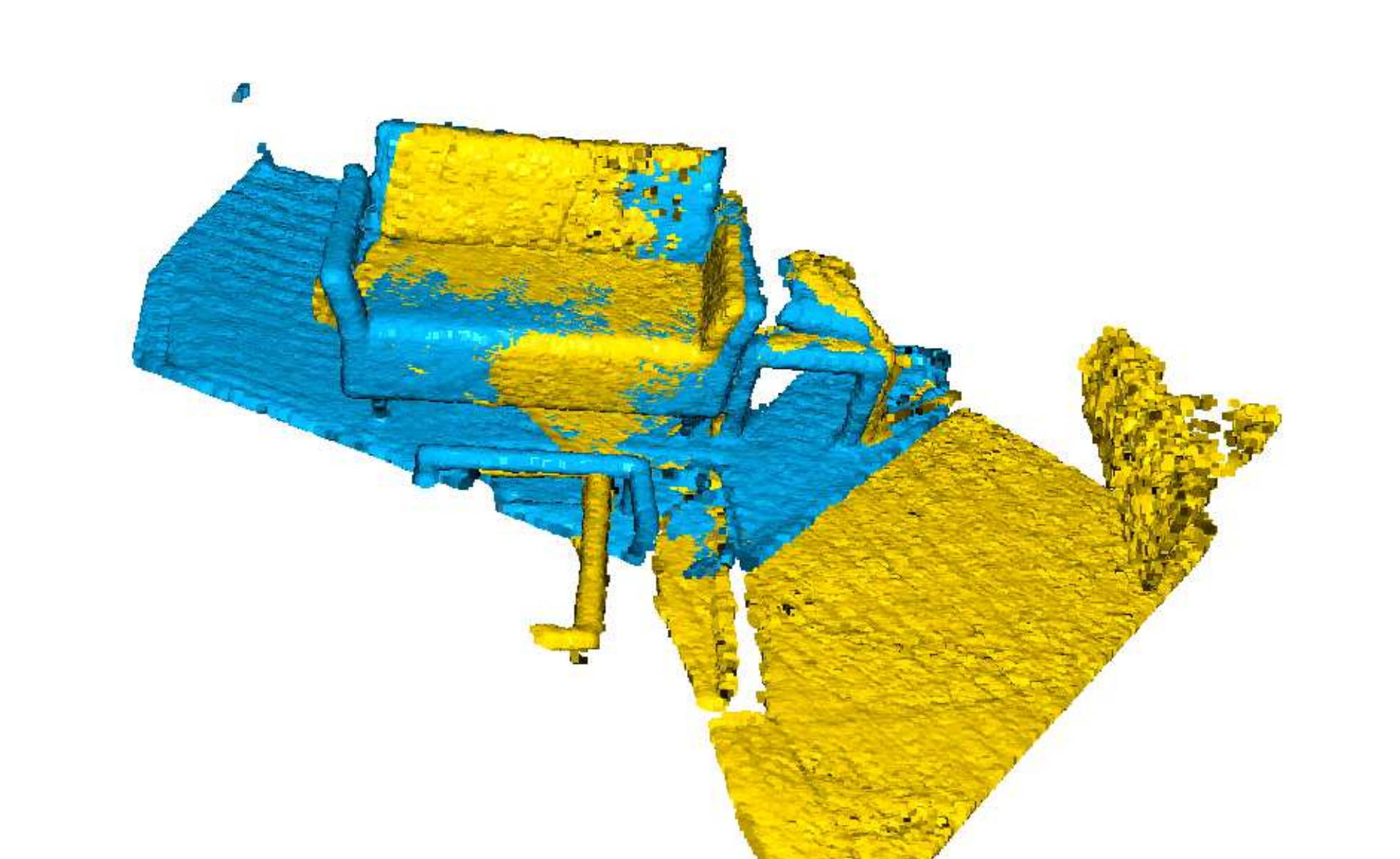}}\hfill
    \subfloat{\includegraphics[trim={500pt 200pt 400pt 250pt}, clip, width=0.16\linewidth]{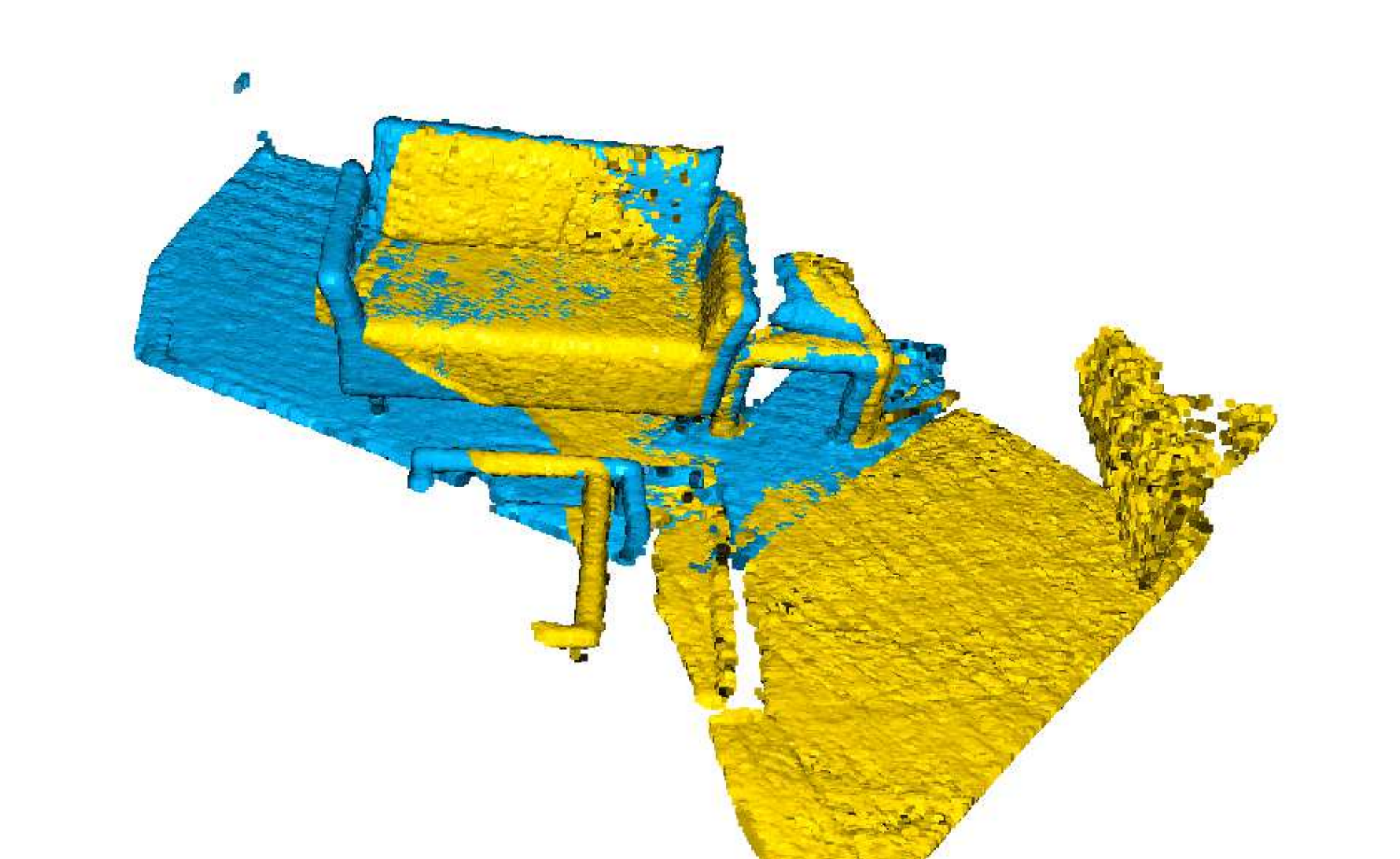}}\hfill
    \subfloat{\includegraphics[trim={500pt 200pt 400pt 250pt}, clip, width=0.16\linewidth]{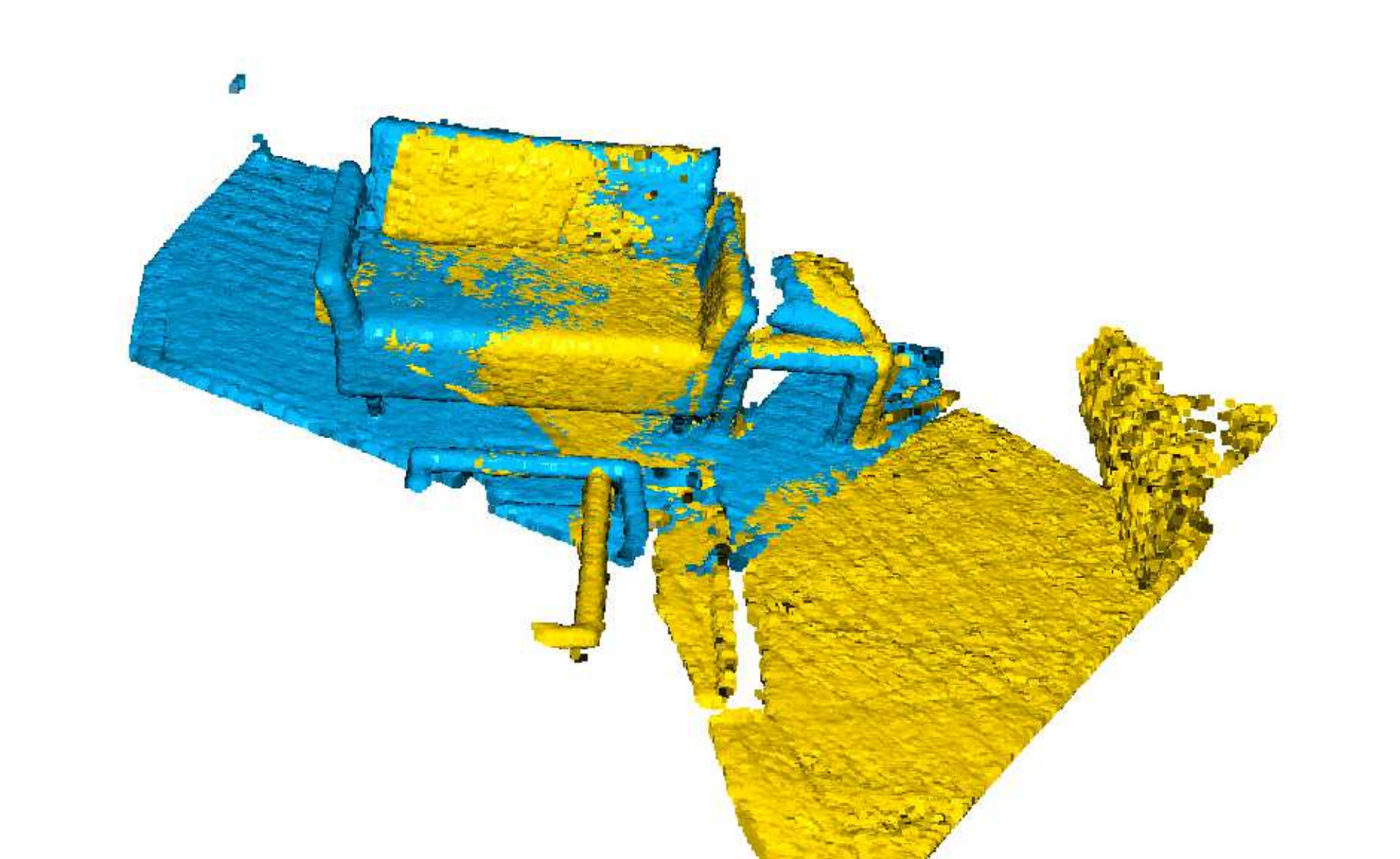}}\hfill
    \subfloat{\includegraphics[trim={500pt 200pt 400pt 250pt}, clip, width=0.16\linewidth]{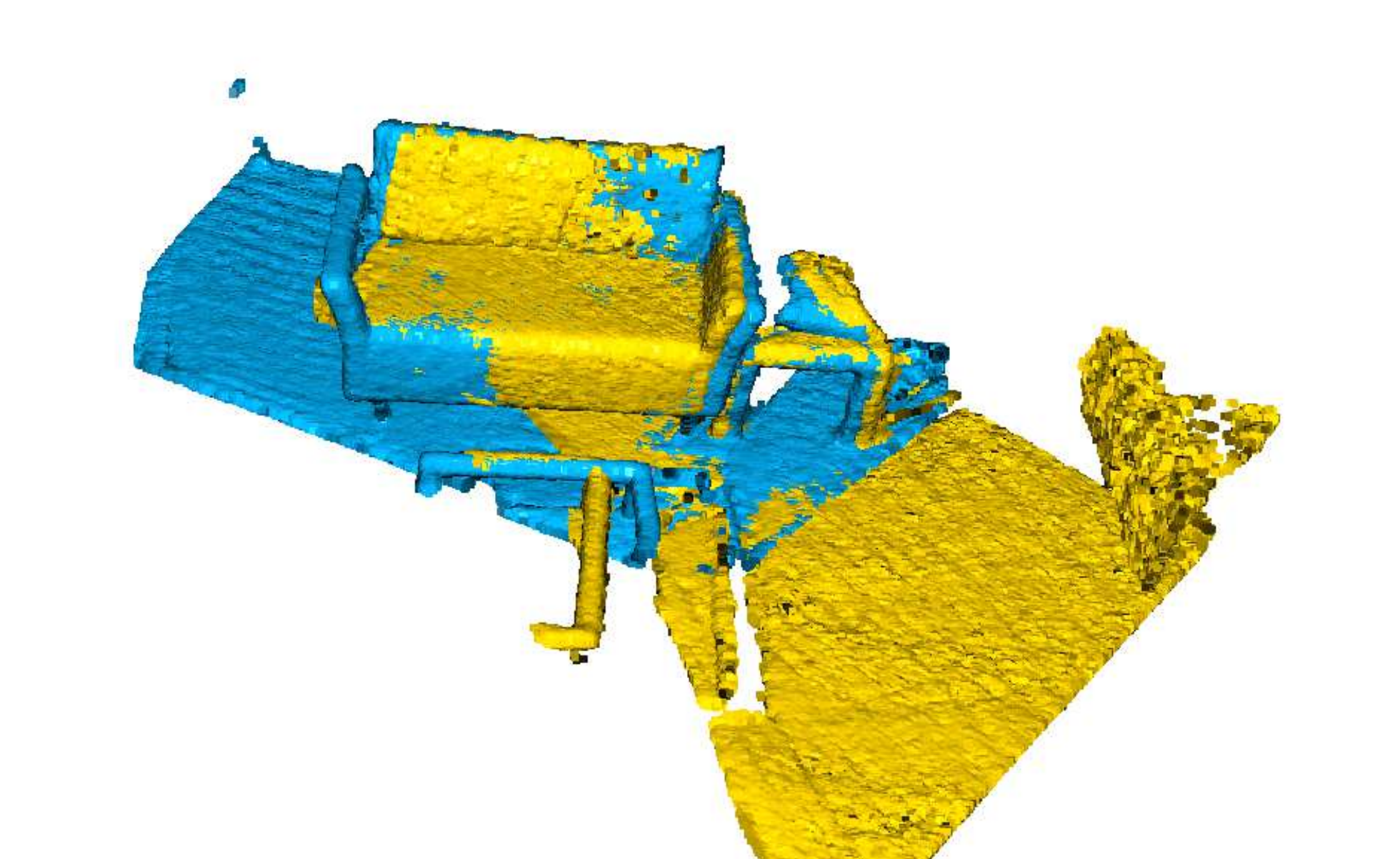}}\hfill

  \subfloat{\includegraphics[trim={550pt 250pt 400pt 200pt}, clip, width=0.16\linewidth]{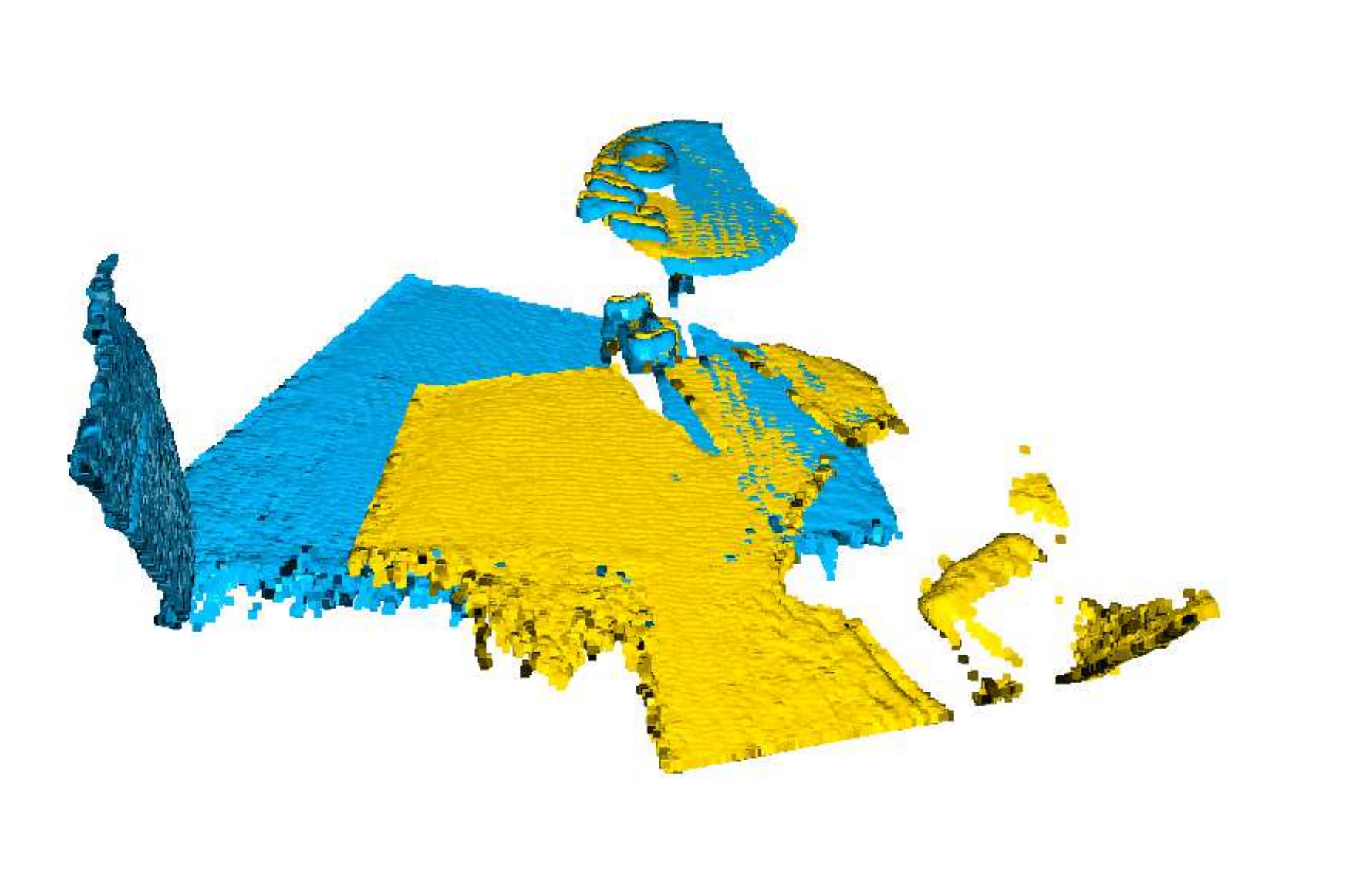}}\hfill
    \subfloat{\includegraphics[trim={550pt 250pt 400pt 200pt}, clip, width=0.16\linewidth]{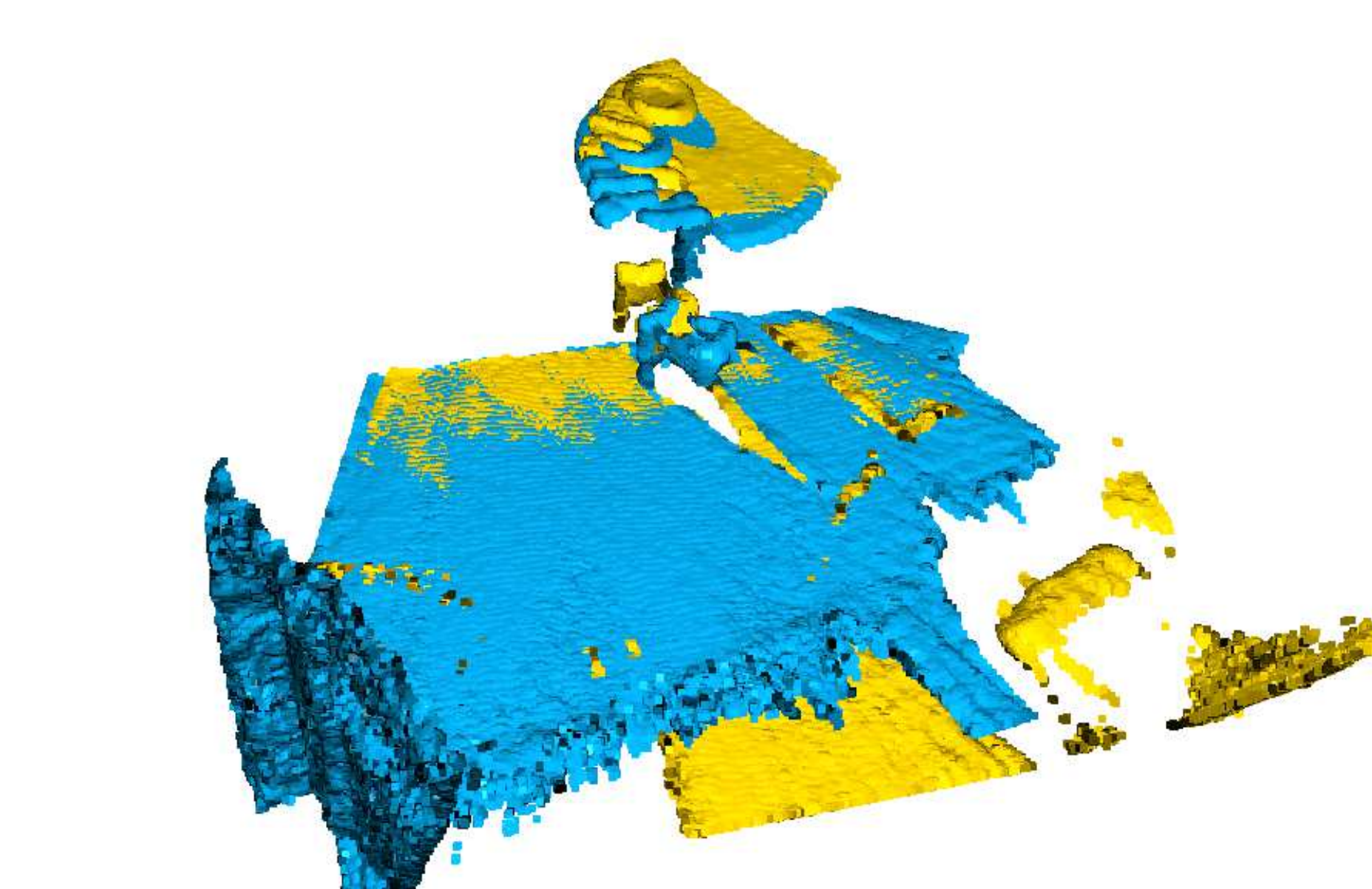}}\hfill
    \subfloat{\includegraphics[trim={550pt 250pt 400pt 200pt}, clip, width=0.16\linewidth]{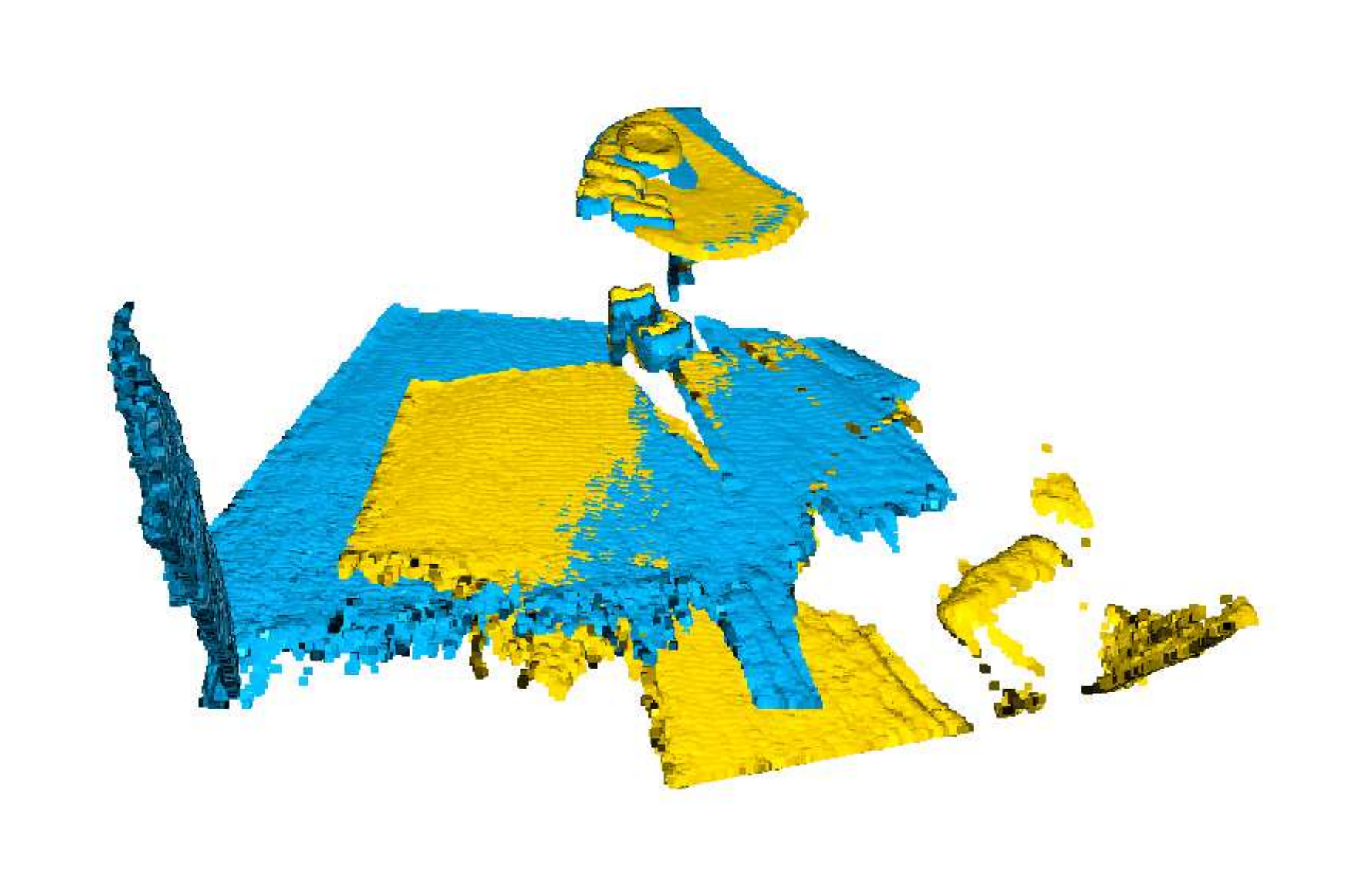}}\hfill
    \subfloat{\includegraphics[trim={550pt 250pt 400pt 200pt}, clip, width=0.16\linewidth]{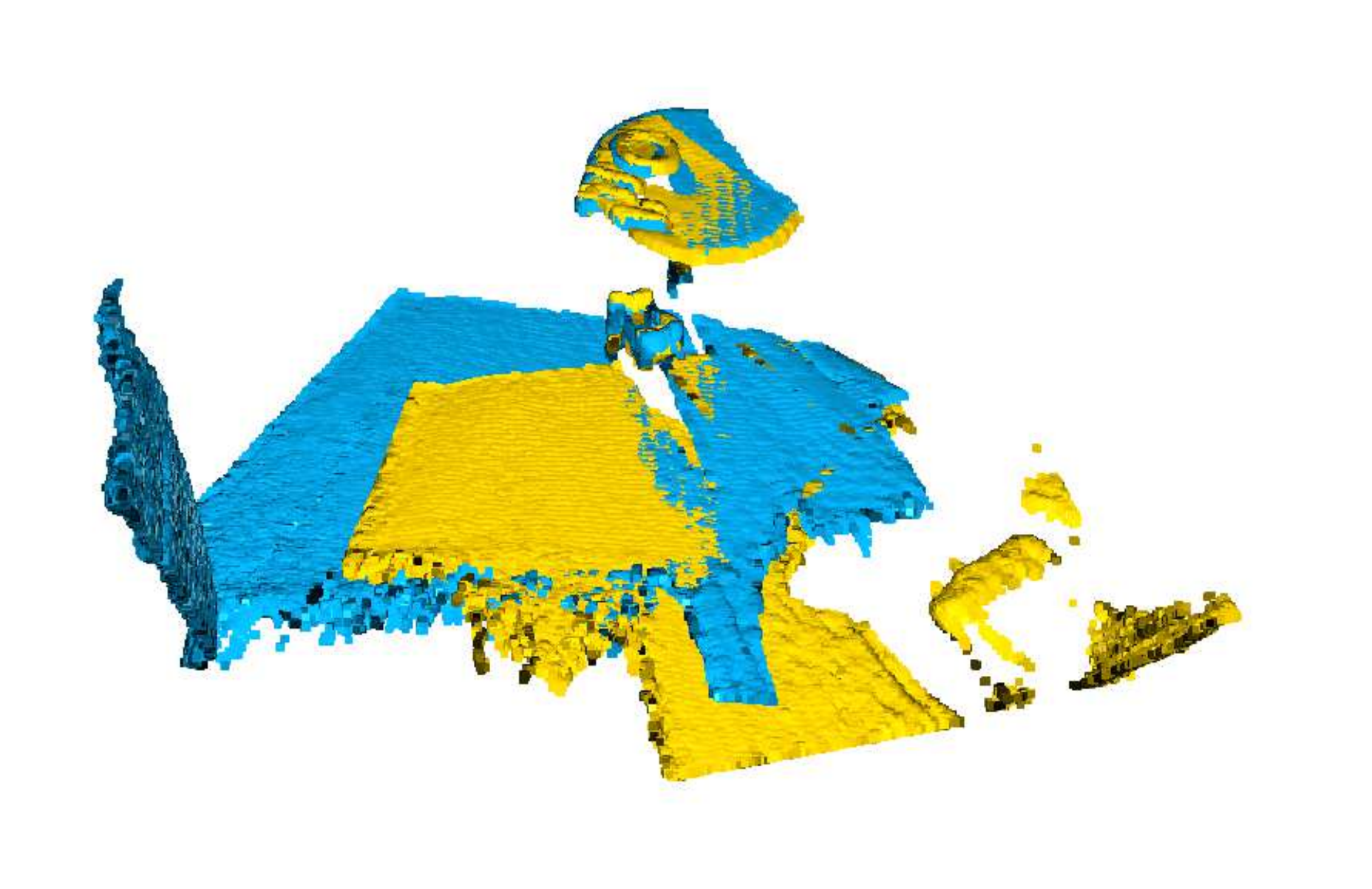}}\hfill
    \subfloat{\includegraphics[trim={550pt 250pt 400pt 200pt}, clip, width=0.16\linewidth]{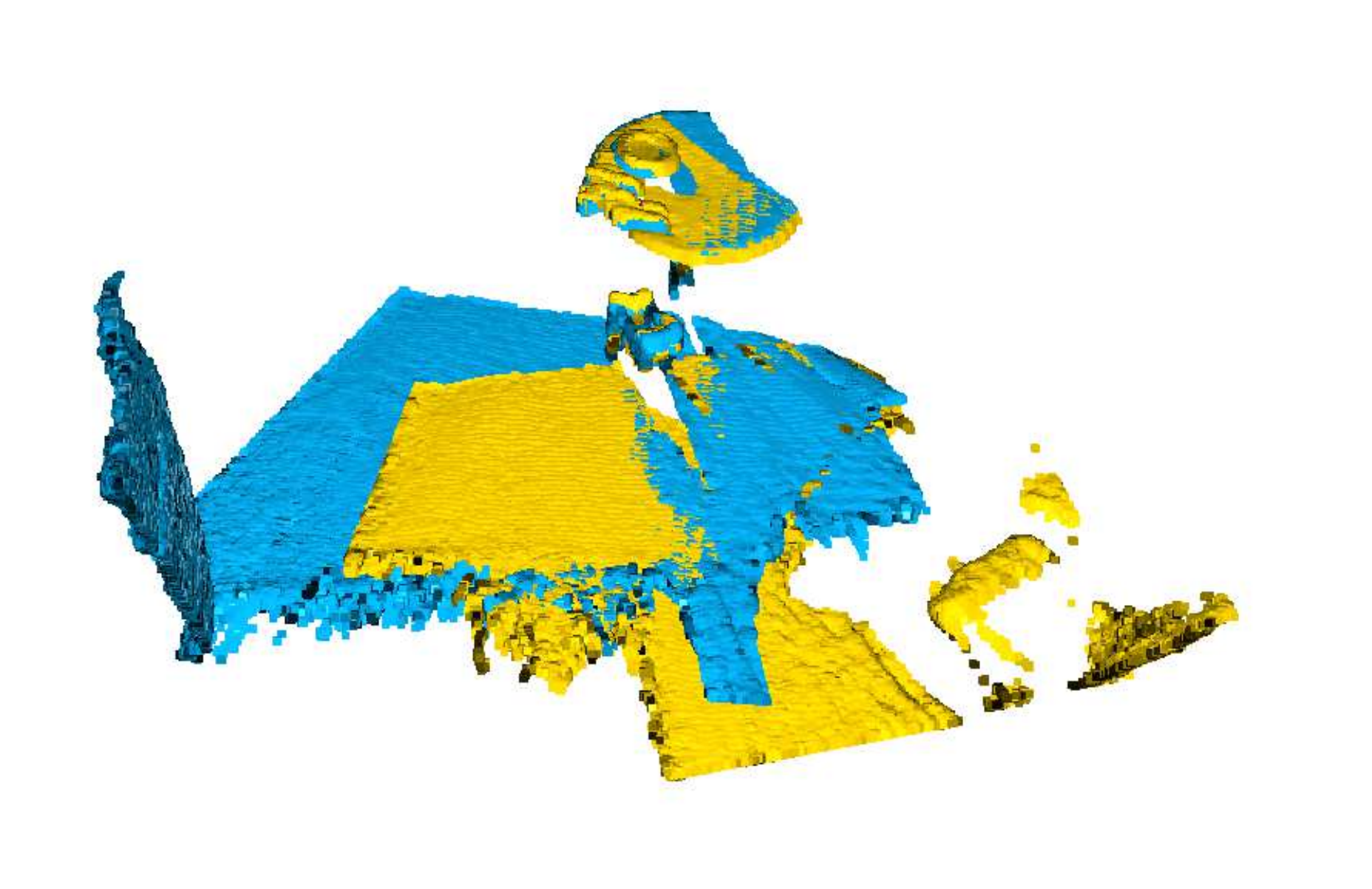}}\hfill
    \subfloat{\includegraphics[trim={550pt 250pt 400pt 200pt}, clip, width=0.16\linewidth]{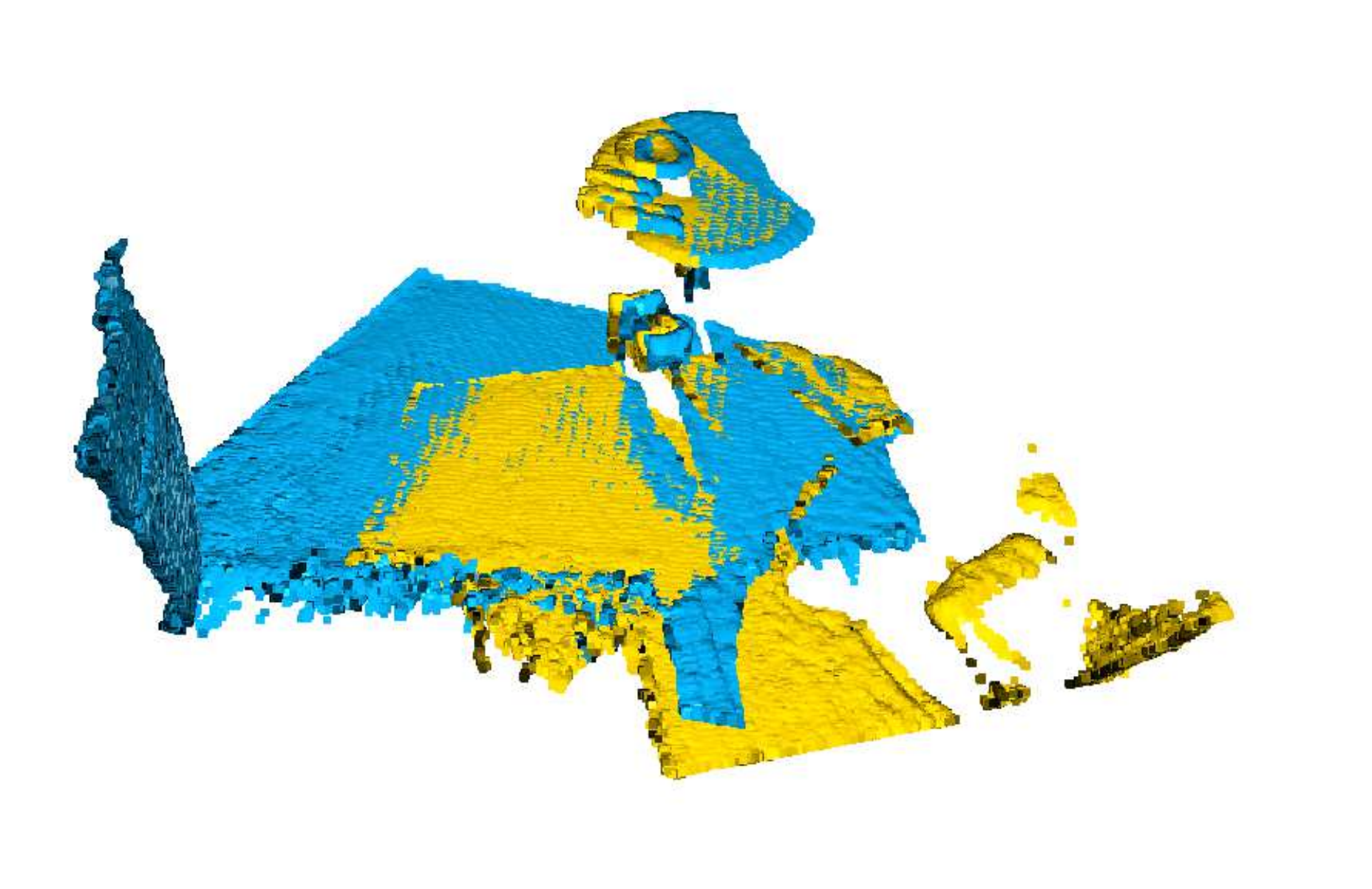}}\hfill
\caption{Qualitative comparison on 3DMatch. The scenes are listed in order from top to bottom as follows: 7-scenes-redkitchen, sun3d-home\_at-home\_at\_scan1\_2013\_jan\_1, sun3d-home\_md-home\_md\_scan9\_2012\_sep\_30, sun3d-hotel\_uc-scan3, sun3d-hotel\_umd-maryland\_hotel1, sun3d-hotel\_umd-maryland\_hotel3, sun3d-mit\_76\_studyroom-76-1studyroom2, and sun3d-mit\_lab\_hj-lab\_hj\_tea\_nov\_2\_2012\_scan1\_erika.}
\label{fig:vis_3DMatch}
\end{figure*}

Figures \ref{fig:vis_ETH} and \ref{fig:vis_3DMatch} show the visualization results of some of the more effective methods. The experiments were conducted on the first pair in each scene across both datasets. The MAC and MAC+ICP algorithms failed to register the facade scenario, whereas our method successfully registered in all cases. The results of our method closely resemble the ground truth (GT), demonstrating its accuracy. This further proves that our coarse-to-fine registration method outperforms others at each stage.

\subsection{Comparison With Learning Baseline\label{sec:Comparison_With_Learning_Baseline}}

To further assess our approach, we have also compared it with several state-of-the-art deep learning methods, using SpinNet \cite{ao2021spinnet}, Predator \cite{huang2021predator}, CoFiNet \cite{yu2021cofinet},  and GeoTransformer \cite{qin2022geometric} as baselines. Our experiments on the 3DMatch dataset were conducted using the deep learning descriptor FCGF. Each method was tested with different sample sizes, where sample size refers to the number of sampled points or correspondences. The results are presented in table \ref{tab:performance_comparison_networks}, referring to the results \cite{zhang20233d}.

In general, our method achieves a higher registration rate than SpinNet, Predator, and CoFiNet and remains competitive compared to GeoTransformer. Nevertheless, an inherent drawback of deep learning methods is their requirement for additional training and expensive GPU memory, which our method does not require.
\begin{table}[t]
    \centering
	 \caption{Performance comparison of various registration networks}
    \label{tab:performance_comparison_networks}
	\resizebox{\linewidth}{!}{
    \begin{tabular}{lccccc}
        \toprule
			\multirow{2}{*}{\# Samples} & \multicolumn{5}{c}{3DMatch RR (\%)} \\
			\cmidrule(lr){2-6} 
			 & 5000 & 2500 & 1000 & 500 & 250 \\
        \midrule
        SpinNet & 88.6 & 86.6 & 85.5 & 83.5 & 70.2 \\
        Predator & 89.0 & 89.9 & 90.6 & 88.5 & 86.6 \\
        CoFiNet & 89.3 & 88.9 & 88.4 & 87.4 & 87.0 \\
        GeoTransformer & 92.0 & 91.8 & 91.8 & {\bf91.4} & {\bf91.2} \\
        Proposed & {\bf93.5} & {\bf93.5} & {\bf92.9} & 88.9 & 87.5 \\
        \bottomrule
    \end{tabular}
}
\end{table}

\subsection{Analysis Experiments\label{sec:Analysis_Experiments}}
In this section, we analyze the role of each component in our proposed coarse-to-fine registration method. To ensure fairness, for the ETH dataset, we select the first pair from each scene for experimentation. For the 3DMatch dataset, given the similar characteristics across its scenes, we select only the first pair from the first scene.

\noindent{\bf Contribution of Graph-Based Hierarchical Outlier Removal Strategy.} We sequentially calculate the outlier ratio within the correspondence set by outlier removal strategies based on the reliability of graph nodes (RGN) and edges (RGE). This procedure entails applying the ground truth transformation to the source correspondences and computing Euclidean distances to the target points. Points are classified as outliers if their distances exceed a defined threshold $2\ell$. 

\begin{figure}[htp]
\centering
\includegraphics[width=\linewidth]{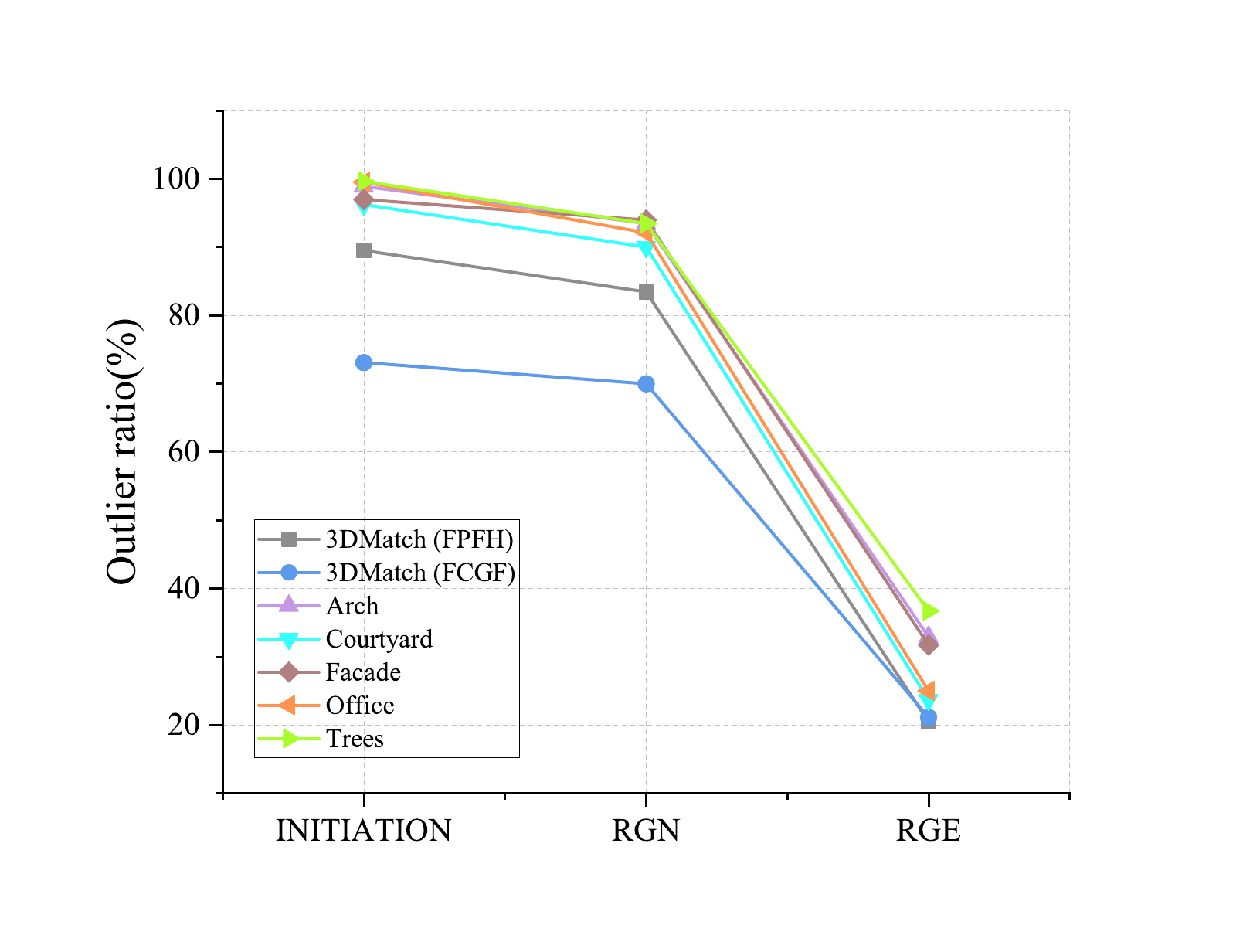}
\caption{Outlier removal results on different datasets.}
\label{fig:result_outlier_removal}
\end{figure}

As shown in figure \ref{fig:result_outlier_removal}, we can observe that, similar to our previous experimental results \cite{yan2022new}, outlier removal primarily relies on the graph-based edge strategy. The graph-based node weight strategy effectively reduces the correspondences' scale, significantly improving efficiency. The final results are minimally affected by the ${{K}_{opt}}$ parameter. After node-based outlier removal, the outlier rate drops to between 20\% and 40\%, effectively reducing the outlier rate of correspondences. These strategies demonstrated a synergistic effect: The graph-based node weight strategy selects relatively reliable small-scale correspondences, while the edge removal strategy contributes to accurately screening out inliers.
\begin{figure*}[t]
    \centering
    \subfloat{\includegraphics[width=.33\linewidth]{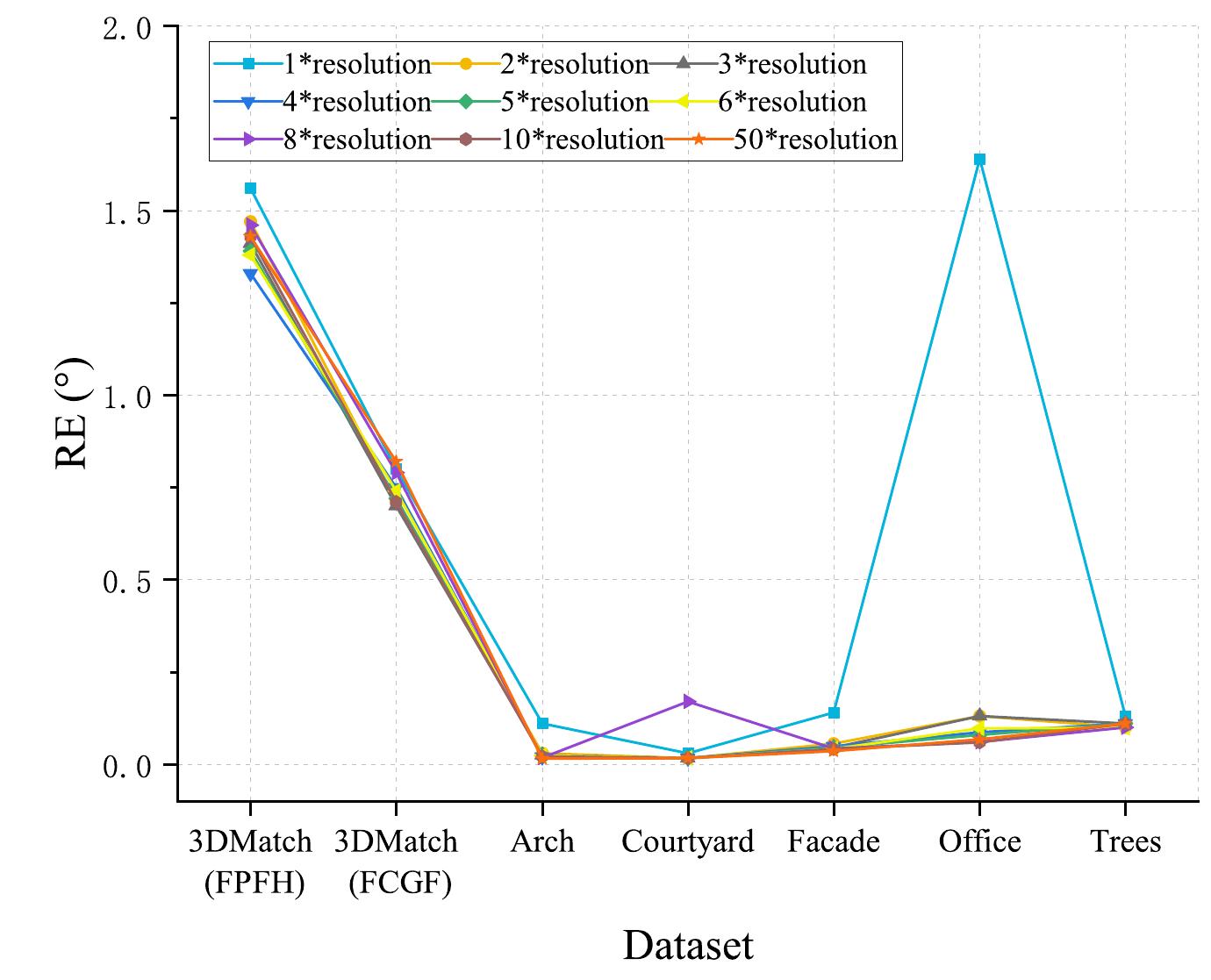}}
    \subfloat{\includegraphics[width=.33\linewidth]{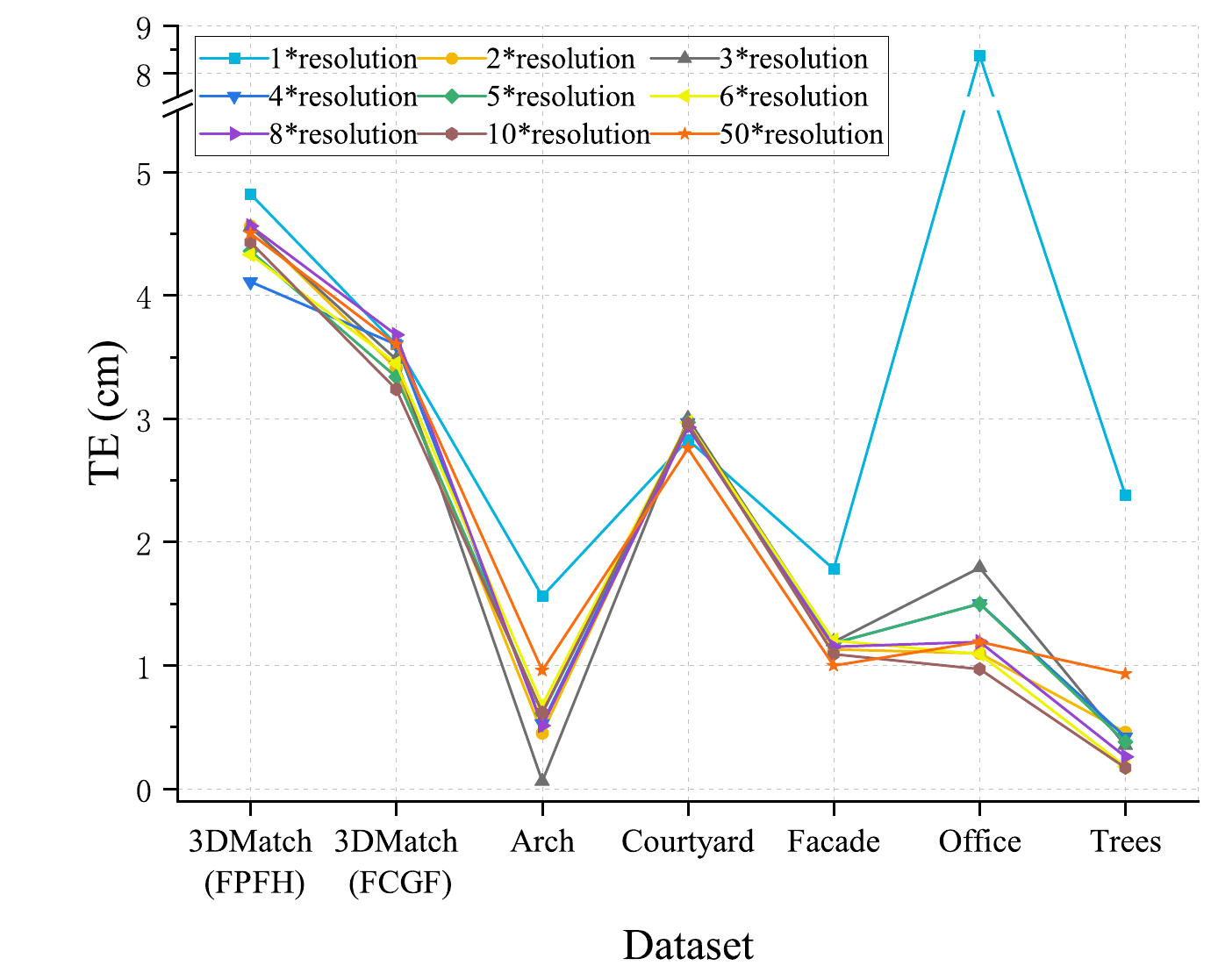}}
\subfloat{\includegraphics[width=.33\linewidth]{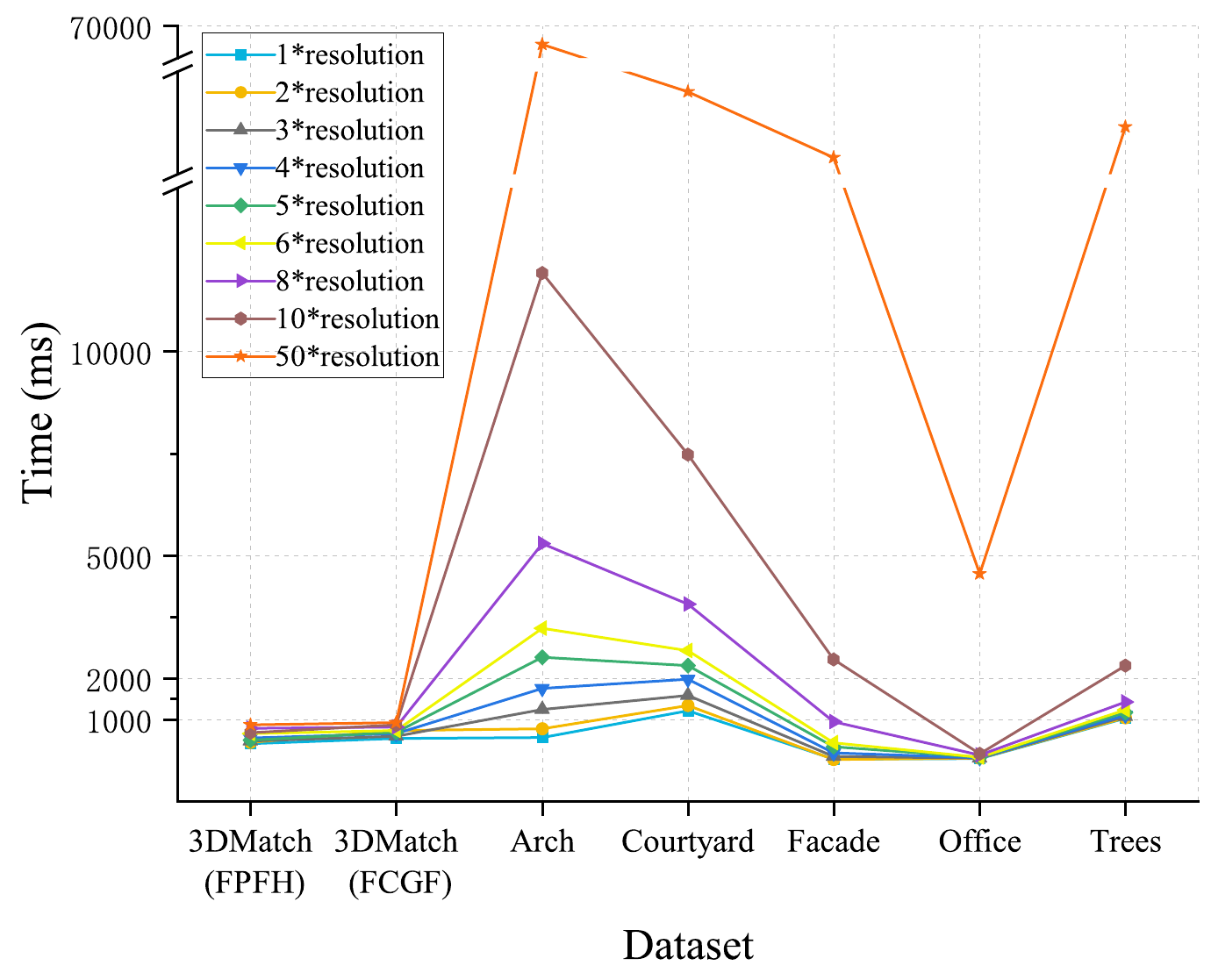}}
    \caption{Index evaluation of different $L$ values}
    \label{fig:Lsize}
\end{figure*}
\begin{figure}[htp]
    \centering
    \subfloat{\includegraphics[width=.5\linewidth]{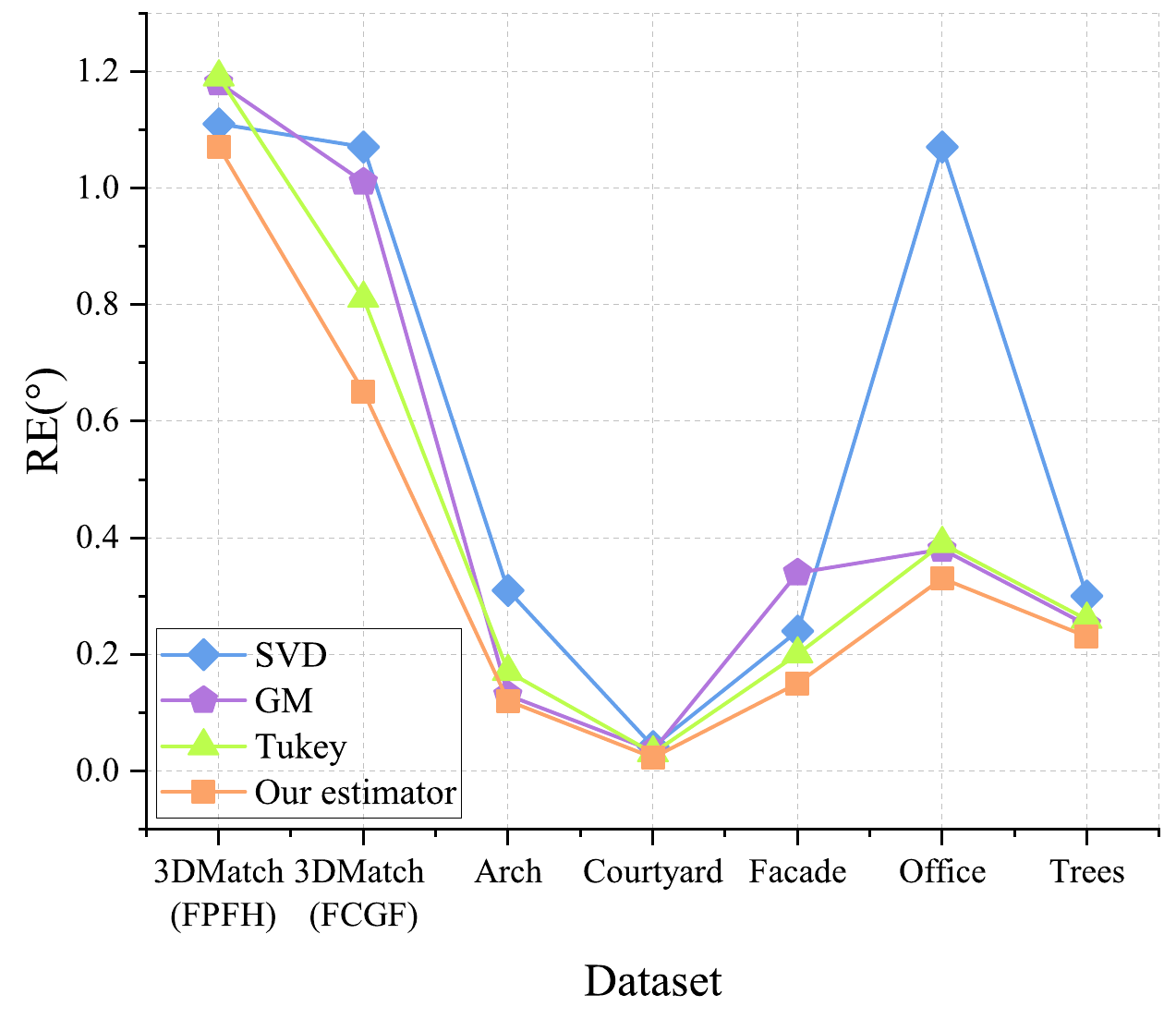}}
    \subfloat{\includegraphics[width=.5\linewidth]{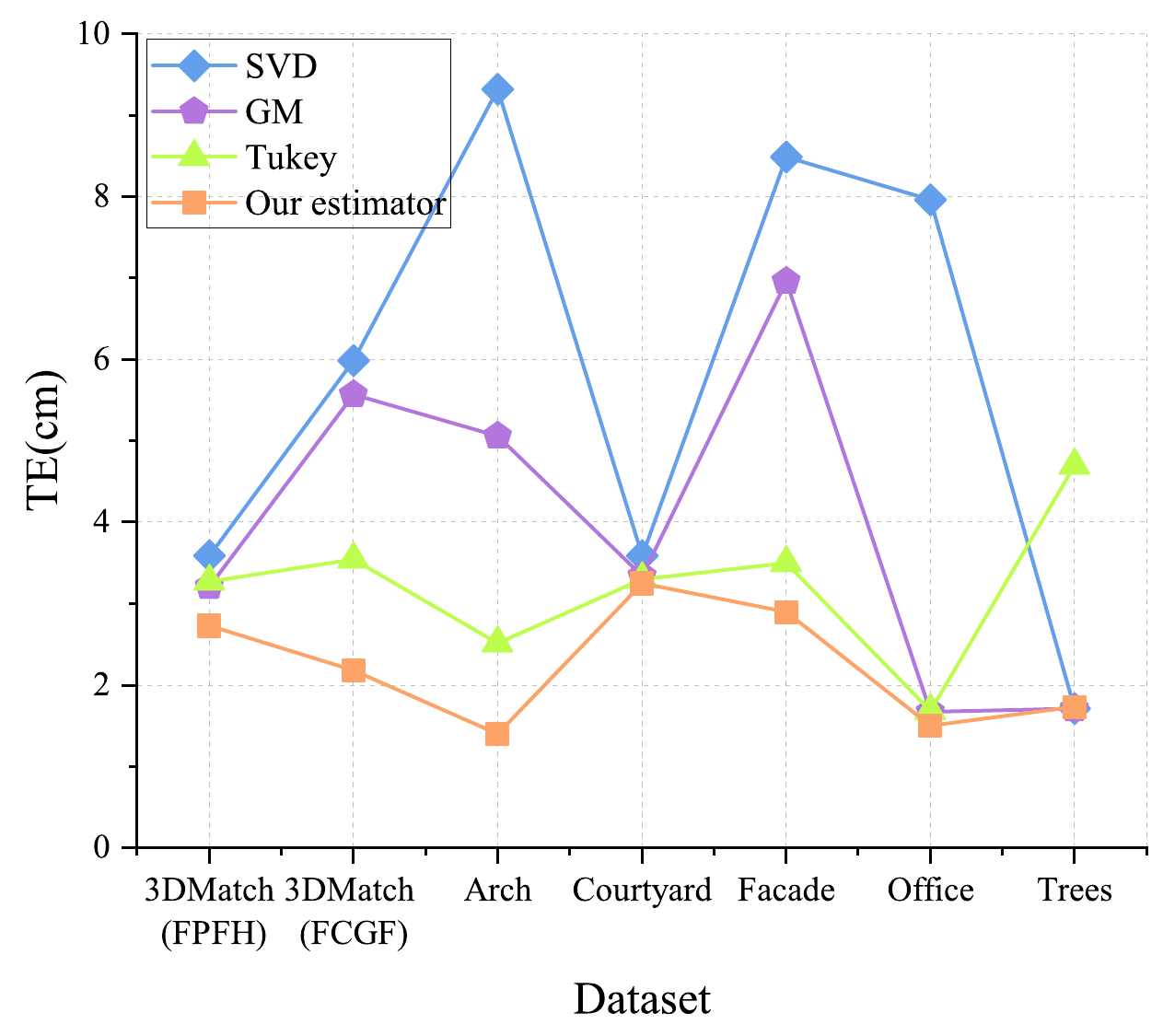}}\\
    \caption{Comparison results of different estimators}
    \label{fig:estimator_RETE}
\end{figure}

\begin{figure}[htp]
\centering
\includegraphics[width=.8\linewidth]{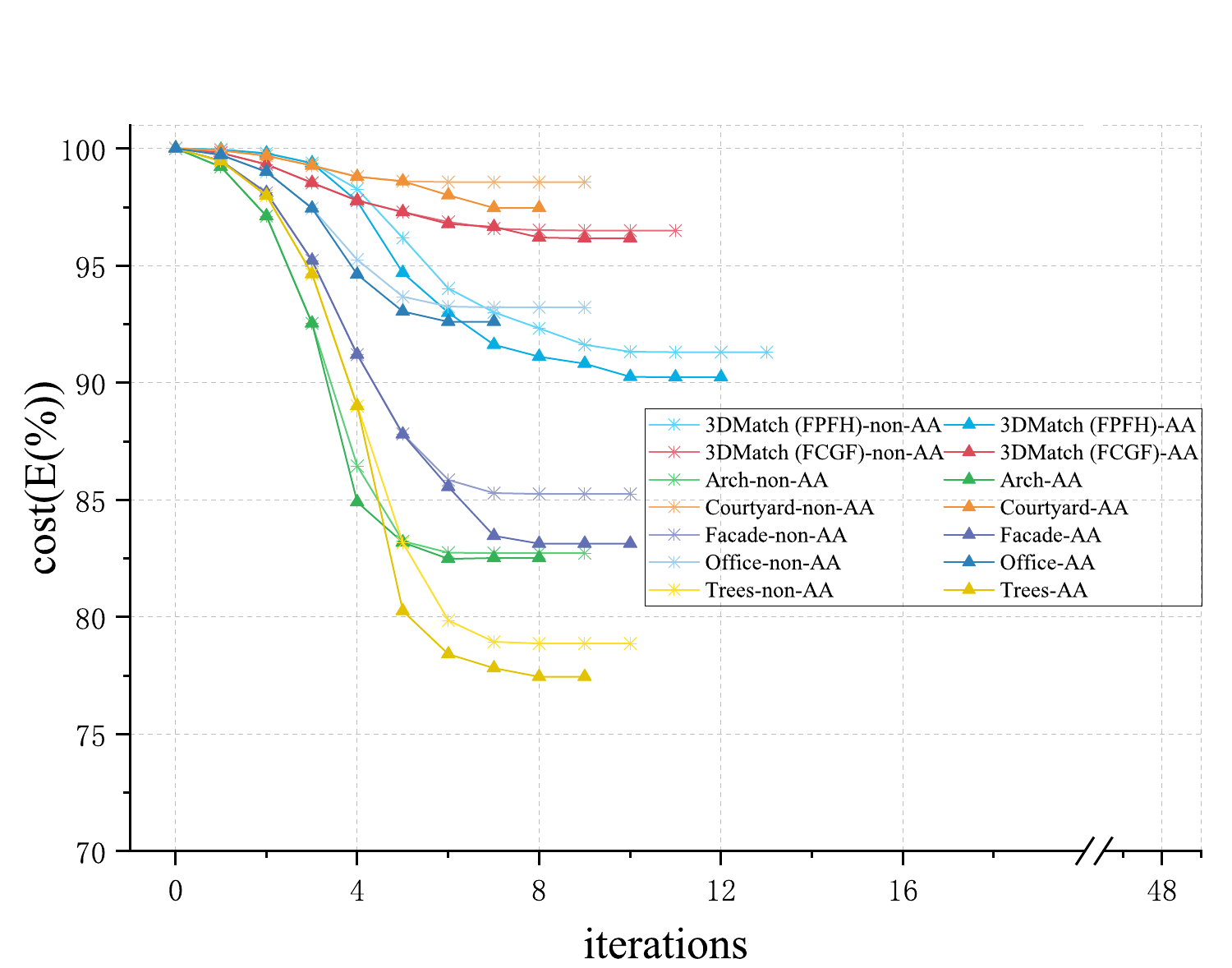}
\caption{The cost variation before and after applying Anderson Acceleration.}
\label{fig:AA_nonAA}
\end{figure}

\noindent{\bf The Advantages of the GNC-Welsch Estimator Based on the Equivalent Outlier Process.}
After outlier removal, the outlier rate in the set of correspondences is already relatively low, posing little challenge for many estimation methods. However, we aim to obtain an even higher accuracy estimator, which is meaningful for improving the precision of our coarse-to-fine registration method. We will conduct experiments by replacing our estimator with SVD, Geman-McClure estimator, and Tukey estimator. The Geman-McClure and Tukey estimators' parameters will be set to the same values as the $\sigma$ parameter in our estimator, following a gradually decreasing sequence.

Due to a considerable number of outliers in the set of correspondences, using SVD directly is still affected by these outliers. Thus, employing robust functions to reduce the influence of outliers is a better choice. However, the Geman-McClure weight function changes more dramatically for small residuals, but smaller residuals do not necessarily indicate inliers at the beginning of the iterative process. The cutoff-based weighting of the Tukey estimator may cause it to ignore some potentially helpful information and may not effectively suppress the influence of outliers. Our estimator weight function changes smoothly for small residuals, making it more likely to converge to the optimal result. As shown in figure \ref{fig:estimator_RETE}, our estimator achieved the lowest error across all tested datasets. Moreover, $TE$ performed the stable ability to maintain a low error level in our estimator.

\noindent{\bf The Effectiveness of Anderson Acceleration in Fine Registration.} 
We will compare the cost changes before and after applying Anderson acceleration in our coarse-to-fine registration method, with the cost calculated using (\ref{eq:13}). To facilitate comparison in a single plot, the cost values for each dataset are normalized by their initial cost and then rescaled. The results are shown in figure \ref{fig:AA_nonAA}.

Anderson acceleration analyzes historical information to extrapolate a more accurate approximate solution, which is then refined by comparison with the current result. Effective Anderson acceleration results typically appear in our experiments after 4 or 5 iterations. Hence, $h$ can be set to 5. Across all our datasets, Anderson acceleration consistently reduces the cost value to varying degrees, with a maximum reduction of 2.13\%. Given its relatively simple implementation, incorporating Anderson acceleration into the Levenberg-Marquardt (LM) iteration is a straightforward and effective method for optimizing results.

\subsection{Sensitivity of parameters\label{sec:Sensitivity of parameters}}

We need three parameters in our registration method: the optimal selection parameter ${{K}_{opt}}$, the downsampling parameter $\ell$, and the size of the captured micro-structure $L$. The parameter ${{K}_{opt}}$ is not sensitive, but it does impact the method's efficiency. The smaller the ${{K}_{opt}}$ value, the more significant the improvement in efficiency. The ${{K}_{opt}}$ parameter usually has a minimum value that stabilizes its influence on accuracy. The resolution parameter $\ell$  is related to the scale of the scenario; larger datasets require larger $\ell$ values. Generally, a smaller $\ell$ value results in higher accuracy, but with a significant drop in efficiency.

In practice, although we use the octree approach for adaptive search of plane features, this remains challenging for scenes like trees, where planar features are less abundant. When the resolution $\ell$ is very small, we only use the micro-structures of one resolution size $L$ to the correspondence. This will lead to a tiny spatial structure, making it challenging to detect geometric features. Furthermore, it is evident that when there are few correspondences, the number of planar features obtained is also limited, making it difficult to achieve good optimization results. Therefore, we also capture the nearby micro-structures for simultaneous detection. Centered on each micro-structure, we will conduct experiments by considering the surrounding micro-structures with $L$ set to $1\ell$, $2\ell$, $3\ell$, $4\ell$, $5\ell$, $6\ell$, $8\ell$, $10\ell$ and $50\ell$.
As shown in figure \ref{fig:Lsize}, except for when $L=\ell$, changes in $L$ have a minor effect on accuracy and tend to stabilize. With the combined benefits of adaptive search and the enlarged $L$, our method also performs well in the trees scene. Moreover, a larger $L$ is not always better. For large-scale scenarios, increasing $L$ extends the computation time significantly. This is mainly because the number of points increases, leading to more segmentation layers and detections. Overall, selecting $L$ as $2\ell$ generally provides high accuracy and efficiency. Additionally, while maintaining efficiency, $L$ can be appropriately increased depending on the scale of the scene to achieve optimal accuracy.

\section{Conclusion}
\label{sec:conclusion}

This paper introduces a micro-structures graph-based coarse-to-fine global point cloud registration method. This method employs a hierarchical outlier removal strategy based on graph nodes and edges, combined with the GNC-Welsch estimator, to ensure robustness during coarse registration. At finer scales, PA-AA optimization is utilized to further exploit the geometric features of corresponding micro-structures, enhancing accuracy with minimal additional computational cost. It works well even in tree scenes, where planar features are limited.
The entire process incrementally leverages adjacency and geometric details within micro-structures graph, making our method non-redundant and highly effective. Through real-world data experiments and a series of analytical tests, our coarse-to-fine registration method demonstrates advanced performance at each stage, maintaining high accuracy and efficiency. Moreover, our method exhibits significant efficiency advantages on large-scale datasets, making it well-suited for practical applications.

\section{acknowledgment}
\label{sec:acknowledgment}

This work was supported by the National Natural Science Foundation of China (Grant numbers 42371451, 42394061), the Open Fund of Hubei Luojia Laboratory (Grant number 220100053). The authors are grateful for the support, and they would also like to thank the editors and anonymous reviewers for their in-depth reading and valuable comments and suggestions.

{
    
    \bibliographystyle{IEEEtran}
    \bibliography{references1}
	
}
\section{Biography}
\label{sec:biog}

\begin{IEEEbiography}[{\includegraphics[width=1in,height=1.25in,clip,keepaspectratio]{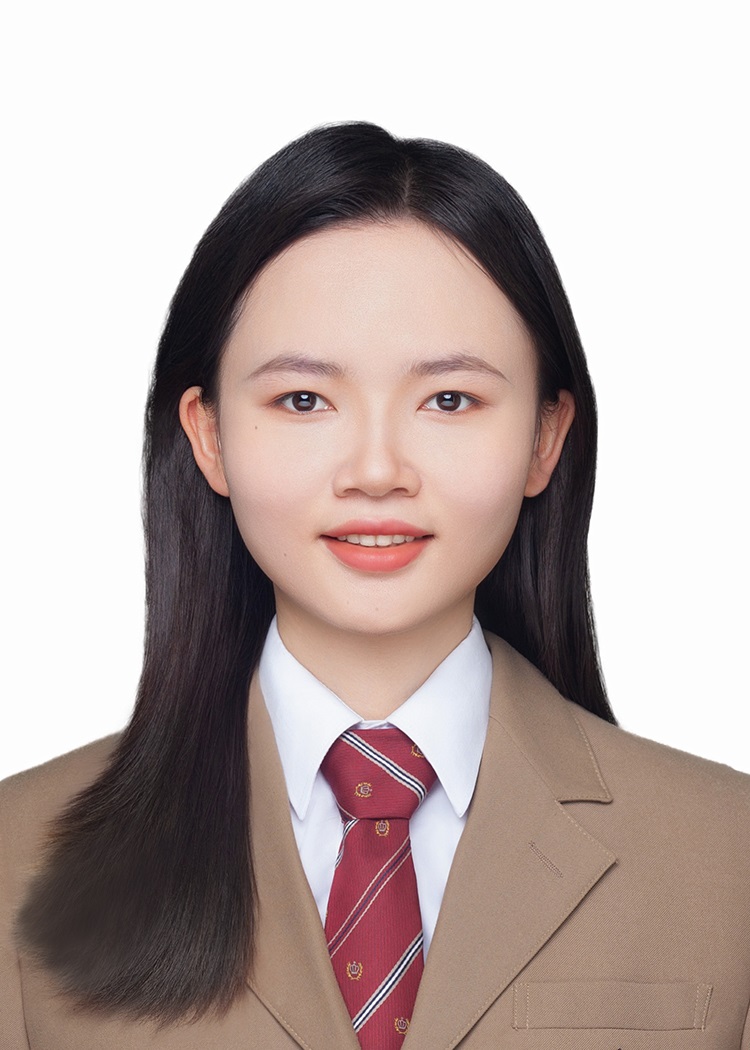}}]{Rongling Zhang}
received the B.S. degree in remote sensing science and technology in 2022 from Harbin Institute of Technology, Harbin, China, where she is currently pursuing the M.S. degree under the supervision of Prof. Li Yan. Her research interests include 3D data processing and 3D reconstruction.
\end{IEEEbiography}

\begin{IEEEbiography}[{\includegraphics[width=1in,height=1.25in,clip,keepaspectratio]{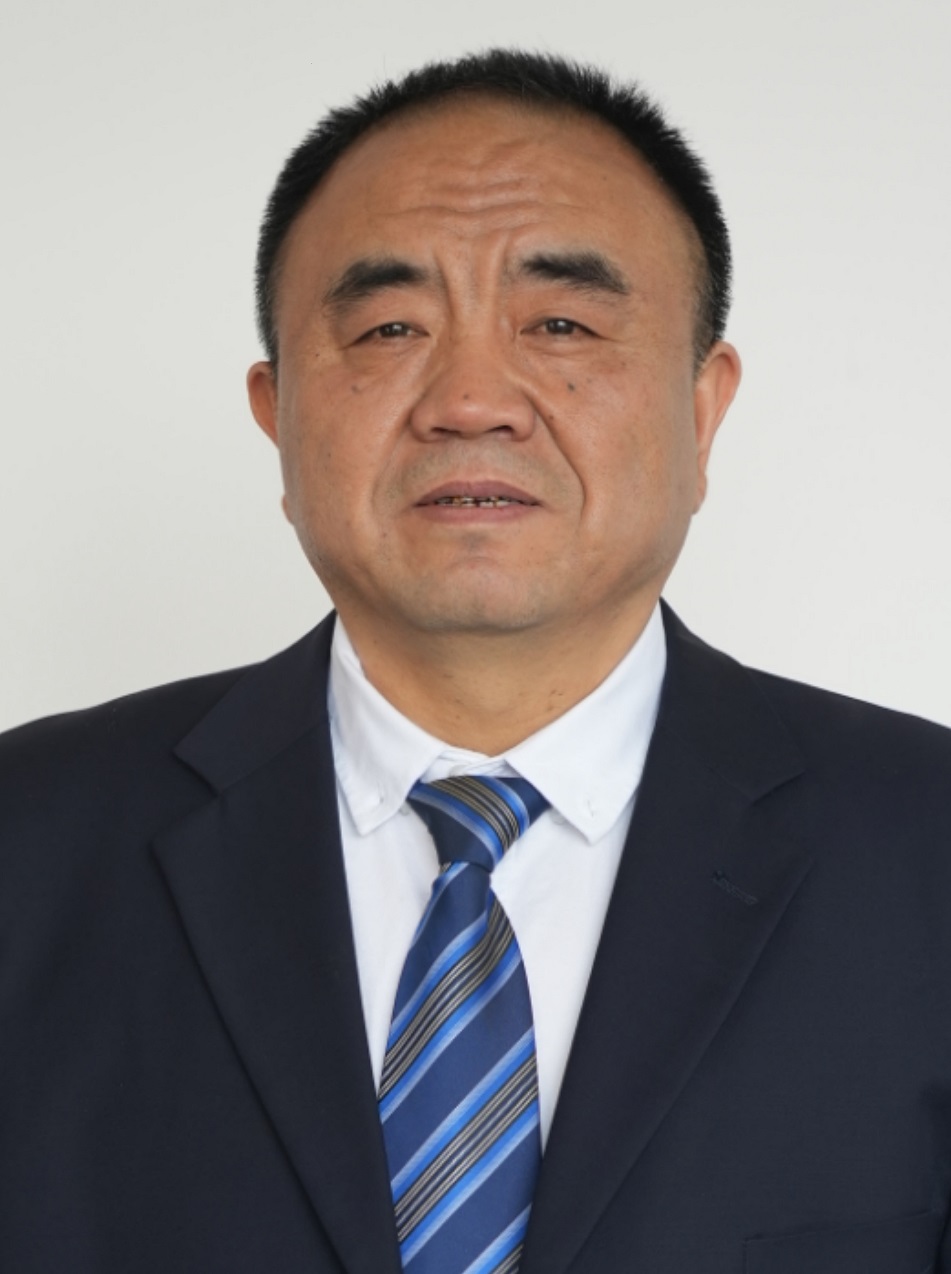}}]{Li Yan}
received the B.S., M.S., and Ph.D. degrees in photogrammetry and remote sensing from Wuhan University, Wuhan, China, in 1989, 1992, and 1999, respectively. He is currently a Luojia Distinguished Professor with the School of Geodesy and Geomatics, Wuhan University. His research interests include 3D reconstruction and measurement, real-time mobile mapping and surveying, intelligent remote sensing, and precise image measurement.
\end{IEEEbiography}

\begin{IEEEbiography}[{\includegraphics[width=1in,height=1.25in,clip,keepaspectratio]{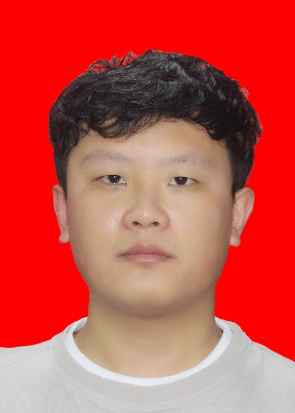}}]{Pengcheng Wei}
received the M.S. degree in photogrammetry and remote sensing from Beijing University of Civil Engineering and Architecture, Beijing, China, in 2020. He is currently pursuing the Eng.D. degree at the School of Geodesy and Geomatics, Wuhan University, Wuhan, China. His research interests include point cloud registration, segmentation, and classification.
\end{IEEEbiography}

\begin{IEEEbiography}[{\includegraphics[width=1in,height=1.25in,clip,keepaspectratio]{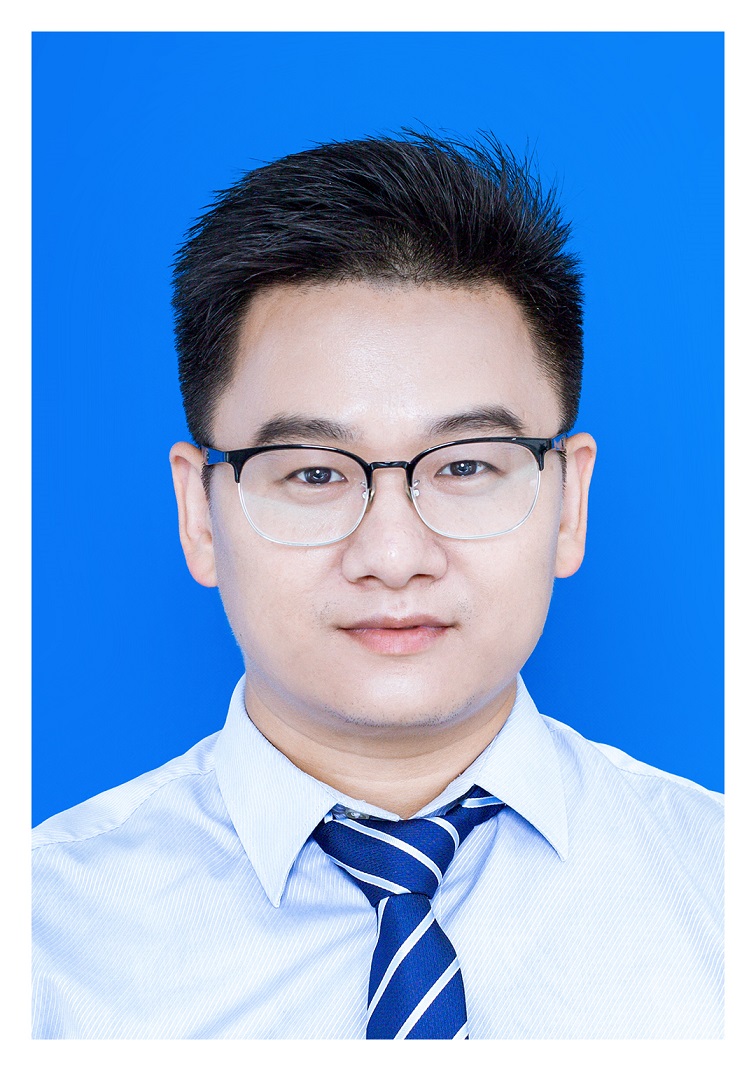}}]{Hong Xie}
received the B.S., M.S., and Ph.D. degrees in photogrammetry and remote sensing from Wuhan University, Wuhan, China, in 2007, 2009, and 2013, respectively. He is currently an associate professor with the School of Geodesy and Geomatics, Wuhan University. His research interests include target detection based on image deep learning, point cloud data quality improvement, point cloud information extraction and model reconstruction, mobile mapping, and surveying.
\end{IEEEbiography}

\begin{IEEEbiography}[{\includegraphics[width=1in,height=1.25in,clip,keepaspectratio]{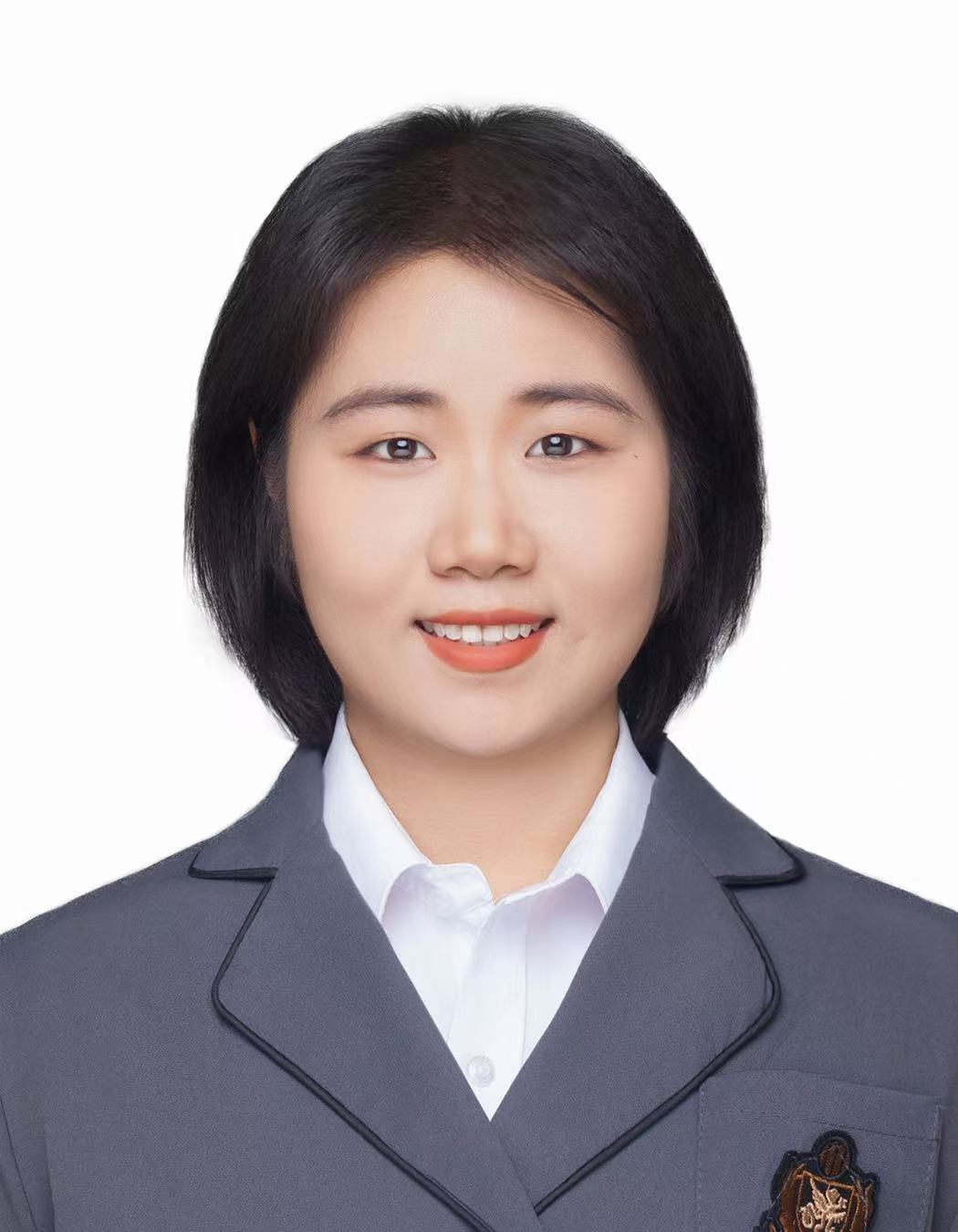}}]{Pinzhuo Wang}
received the B.S. degree in Surveying and Mapping Engineering in 2018 from Wuhan University, Wuhan, China, where she is currently pursuing the M.S. degree under the supervision of Prof. Li Yan.  Her research interests include photogrammetry and light detection and ranging (LiDAR).
\end{IEEEbiography}

\begin{IEEEbiography}[{\includegraphics[width=1in,height=1.25in,clip,keepaspectratio]{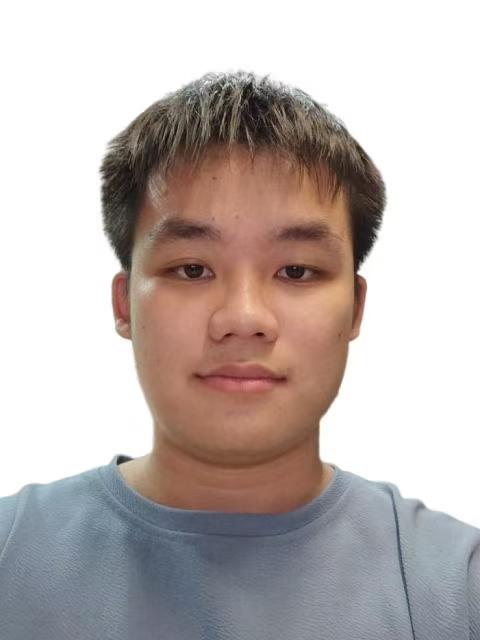}}]{Binbing Wang} received the B.S. degree in geodesy and surveying engineering in 2023 from Wuhan University, where he is currently pursuing the M.S. degree under the guidance of Prof. Hong Xie. His current research focuses on the registration and fusion of multi-source point clouds.
\end{IEEEbiography}


\end{document}